\pdfoutput=1
\RequirePackage[svgnames]{xcolor}

\documentclass[11pt]{article}

\usepackage{acl}

\usepackage{times}
\usepackage{latexsym}

\usepackage[draft,textsize=footnotesize,textwidth=15mm]{todonotes}
\usepackage[utf8]{inputenc}
\usepackage{inconsolata}
\usepackage{algorithm}
\usepackage{algpseudocode}
\usepackage{amsmath,amssymb}
\usepackage{soul}
\usepackage{tikz}

\tikzset{rndblock/.style={rounded corners,rectangle,draw,scale=0.8,outer sep=0pt}}

\usetikzlibrary{shapes.geometric}
\usepackage{framed}
\usepackage{enumitem}
\newlist{RQ}{enumerate}{1}
\setlist[RQ]{label=\textbf{RQ\,\arabic*},ref={RQ\,\arabic*}}
\usepackage{comment}
\usepackage{natbib}
\usepackage{multibib}
\makeatletter
\usepackage{booktabs}
\usepackage[inkscapeformat=png]{svg}
\usepackage{graphicx}
\usepackage{caption}
\usepackage{subcaption}
\usepackage{tabularx}
\usepackage{soul}
\usepackage{float}
\usepackage{enumitem}
\usepackage{pifont}
\usepackage{arydshln}
\usepackage{lipsum}
\usepackage[T1]{fontenc}
\usepackage{pifont}
\usepackage{amsmath}
\usepackage{soul}
\usepackage[utf8]{inputenc}
\usepackage{inconsolata}
\usepackage{tikz}
\usepackage{arydshln}
\usepackage{lipsum}
\usepackage[normalem]{ulem}
\usepackage{wrapfig,graphicx,lipsum}% http://ctan.org/pkg/{wrapfig,graphicx,lipsum}
\usepackage{graphicx}
\usepackage{colortbl} 
\usepackage{uncial}

\usepackage{soul}
\usepackage{graphicx}
\usepackage{booktabs}
\usepackage{multirow}
\usepackage{colortbl}
\usepackage{afterpage}  % load the afterpage package
\usepackage{tabularx} % Add to your preamble
\usepackage{multicol} % For structured multiline text
\usepackage{array}    % Enhanced column types
\usepackage{rotating}
\usepackage{tabularx}
\usepackage{booktabs}
\usepackage{amsmath}
\usepackage{amssymb}
\usepackage{array}

\usepackage[many]{tcolorbox}

\newtcolorbox{defin}{colback=Teal!5!White,enhanced,title=DPO - Kernels (at-a-glance),
	attach boxed title to top left={xshift=0mm},boxrule=0pt,after skip=1cm,before skip=1cm,right skip=0cm,breakable,fonttitle=\bfseries,toprule=0pt,bottomrule=0pt,rightrule=0pt,leftrule=3pt,arc=0mm,skin=enhancedlast jigsaw,sharp corners,colframe=Teal!55!black,colbacktitle=Teal!55!black,boxed title style={
		frame code={ 
			\fill[Teal!25!black](frame.south west)--(frame.north west)--(frame.north east)--([xshift=3mm]frame.east)--(frame.south east)--cycle;
			\draw[line width=1mm,Teal!25!black]([xshift=2mm]frame.north east)--([xshift=5mm]frame.east)--([xshift=2mm]frame.south east);
			\draw[line width=1mm,Teal!25!black]([xshift=5mm]frame.north east)--([xshift=8mm]frame.east)--([xshift=5mm]frame.south east);
			\fill[Teal!25!black](frame.south west)--+(4mm,-2mm)--+(4mm,2mm)--cycle;
		}
	}
}

\usetikzlibrary{shapes.geometric, arrows}
\usetikzlibrary{decorations.markings}

\usepackage{fancybox}
 \definecolor{darkblue}{rgb}{0, 0, 0.5}
  \hypersetup{colorlinks=true, citecolor=darkblue, linkcolor=darkblue, urlcolor=darkblue}

\definecolor{vgreen}{HTML}{60A917}
\definecolor{vred}{HTML}{CE3A29}

\usepackage{xstring}
\usepackage{longtable}

\usepackage{tabularray}

\DefTblrTemplate{firsthead,middlehead,lasthead}{default}{
}
\DefTblrTemplate{firstfoot}{default}{
  \UseTblrTemplate{contfoot}{default}
  \UseTblrTemplate{caption}{default}
}
\DefTblrTemplate{middlefoot}{default}{
  \UseTblrTemplate{contfoot}{default}
  \UseTblrTemplate{capcont}{default}
}
\DefTblrTemplate{lastfoot}{default}{
  \UseTblrTemplate{note}{default}
  \UseTblrTemplate{remark}{default}
  \UseTblrTemplate{capcont}{default}
}

\newcolumntype{P}[1]{>{\centering\arraybackslash}p{#1}}
\usepackage{color}
\tcbuselibrary{skins}

\usepackage[export]{adjustbox} % for the valign option

\usepackage{setspace}
\usepackage[capitalise,nameinlink]{cleveref}

\crefname{section}{Sec.}{Sec.}

\usepackage{microtype}
\usepackage{graphicx}
\usepackage{comment}
\usepackage{amsmath}
\usepackage{amssymb}
\usepackage{algorithm}
\usepackage{algpseudocode}
\usepackage{colortbl}
\usepackage[export]{adjustbox} % for the valign option
\usepackage{varwidth}
\usepackage{enumitem}
\setlist{leftmargin=1mm}
\usepackage{pifont}
\usepackage{booktabs}
\usepackage{multirow}
\usepackage{subcaption}
\usepackage{resizegather}
\usepackage{breqn}
\usepackage[capitalise]{cleveref}
\usepackage{graphicx}
\usepackage{tikz}
\usetikzlibrary{shapes.geometric, arrows}
\usetikzlibrary{decorations.markings}
\usepackage{soul}
\usepackage{wrapfig,graphicx,lipsum}% http://ctan.org/pkg/{wrapfig,graphicx,lipsum}
\usepackage{extsizes}
\usepackage{cuted}
\usepackage{flushend}
\usepackage{float}
\usepackage{changepage,threeparttable}
\usepackage{setspace}
\usepackage{caption}
\usepackage{booktabs}
\usepackage{dblfloatfix} 
\usepackage{fixltx2e}
\usepackage[normalem]{ulem}

\usepackage{environ}

\newlength{\myl}
\expandafter\let\expandafter\origequation\csname equation*\endcsname
\expandafter\let\expandafter\endorigequation\csname endequation*\endcsname
\long\def\[#1\]{\begin{equation*}#1\end{equation*}}
\RenewEnviron{equation*}{
  \settowidth{\myl}{$\displaystyle\BODY$} % calculate width and save as \myl
  \origequation
    \ifdim\myl>\linewidth
      \resizebox{\linewidth}{!}{$\displaystyle\BODY$}% \myl > \linewidth
    \else
      \BODY % \myl <= \linewidth
    \fi
  \endorigequation
}

\makeatletter
\newcommand{\DrawLine}{%
  \begin{tikzpicture}
  \path[use as bounding box] (0,0) -- (\linewidth,0);
  \draw[color=blue!75!black,dashed,dash phase=.5pt]
        (0-\kvtcb@leftlower-\kvtcb@boxsep,0)--
        (\linewidth+\kvtcb@rightlower+\kvtcb@boxsep,0);
  \end{tikzpicture}%
  }
\makeatother

\usepackage{euscript}[mathcal]

\newcommand*{\affaddr}[1]{#1}
\newcommand*{\affmark}[1][*]{\textsuperscript{#1}}

\author{
  Amitava Das\affmark[1], Suranjana Trivedy\affmark[1], Danush Khanna\affmark[1], 
  Rajarshi Roy\affmark[1], 
  \bf Gurpreet Singh\affmark[1], \\ \bf Basab Ghosh\affmark[1], \bf Yaswanth Narsupalli\affmark[1],  \bf Vinija Jain\affmark[2]\thanks{\,\,\,Work done outside of role at Meta.}, 
  \bf Vasu Sharma\affmark[2]\footnotemark[1], 
  \\ \bf Aishwarya Naresh Reganti\affmark[3], 
  \bf Aman Chadha\affmark[3]\thanks{\,\,\,Work done outside of role at Amazon.} \\
  \affaddr{\affmark[1]Artificial Intelligence Institute, University of South Carolina, USA}\\
  \affaddr{\affmark[2]Meta AI, USA}
  \affaddr{\affmark[3]Amazon AI, USA}
}

\title{\includegraphics[width=\textwidth]{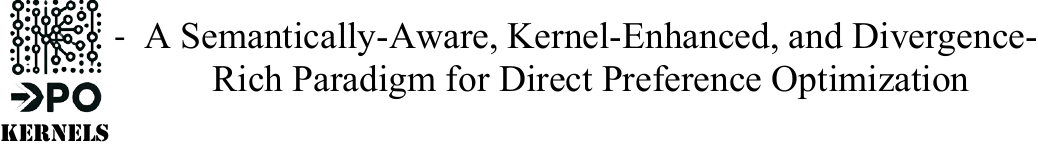} }
\begin{document}
\maketitle
\begin{abstract}

The rapid advancement of large language models (LLMs) has revolutionized numerous applications, but presents significant challenges in aligning these models with diverse human values, ethical standards, and specific user preferences. Direct Preference Optimization (DPO) has become a cornerstone for preference alignment but is constrained by reliance on fixed divergence measures and limited feature transformations. We introduce \textbf{DPO-Kernels}, an innovative enhancement of DPO that integrates kernel methods to overcome these challenges through four key contributions: (i) \textbf{Kernelized Representations}: These representations lay the groundwork for enhanced divergence measures by leveraging polynomial, RBF, Mahalanobis, and spectral kernels for richer, more expressive feature transformations. Additionally, we introduce a \textbf{hybrid loss} that combines embedding-based loss with probability-based loss, enhancing the optimization process beyond traditional DPO; (ii) \textbf{Divergence Alternatives}: Incorporating Jensen-Shannon, Hellinger, Rényi, Bhattacharyya, Wasserstein, and f-divergences to boost stability and robustness; (iii) \textbf{Data-Driven Selection}: Choosing the optimal kernel-divergence pair among 28 combinations (4 kernels $\times$ 7 divergences) is challenging. We introduce automatic metrics that analyze the data to select the best pair, eliminating the need for manual tuning; (iv) \textbf{Hierarchical Mixture of Kernels (HMK)}: Combining local and global kernels for precise and large-scale semantic modeling. This approach automatically selects the optimal kernel mixture during training, enhancing modeling flexibility. Evaluations on 12 datasets demonstrate that DPO-Kernels achieve state-of-the-art generalization in factuality, safety, reasoning, and instruction following. While alignment generally carries the risk of overfitting, grounded in Heavy-Tailed Self-Regularization (HT-SR) theory, we show that DPO-Kernels maintain robust generalization bounds in LLMs. Comprehensive resources are available to facilitate further research and application of DPO-Kernels.

\end{abstract}

\section{\textls[-12]{DPO Revisited: Mathematical Components and Scope for Enhancement}}
\label{sec:introduction}

The Direct Preference Optimization (DPO) \cite{rafailov2024directpreferenceoptimizationlanguage} framework aims to optimize a policy \(\pi(y \mid x)\) by balancing two objectives: improving the policy's ranking on preferred outcomes and regularizing it against a reference distribution using the Kullback–Leibler (KL) divergence. The DPO objective can be expressed as:

\vspace{-2mm}
\[
\max_{\pi} \; \underbrace{ \mathbb{E}_{x, y^+, y^-} \left[ \log \frac{\pi(y^+ \mid x)}{\pi(y^- \mid x)} \right] }_{\text{Contrastive Loss}} 
- \underbrace{ \alpha \, \mathbb{E}_{x} \left[ \sum_{y} \pi(y \mid x) \log \frac{\pi(y \mid x)}{\pi_{\text{ref}}(y \mid x)} \right] }_{\text{KL Divergence}}
\]

\vspace{-2mm} 
where: \(x\): The input prompt/context; \(y^+\): The preferred output; \(y^-\): The less preferred output, \(\pi(y \mid x)\): The policy being optimized; \(\pi_{\text{ref}}(y \mid x)\): The reference policy (often a pre-trained model's distribution); \(\alpha > 0\): Hyperparameters controlling the strength of the regularization.

\vspace{-10mm}

\begin{defin}

\begin{itemize}
[labelindent=-0.6em,labelsep=0.1cm,leftmargin=*]
\setlength\itemsep{0em}
\begin{spacing}{0.5}
% \item[$\blacktriangleright$] 
% {\footnotesize 
% {\fontfamily{phv}\fontsize{7.5}{8.5}
% %\begin{spacing}{1}
% \selectfont
% In this paper, we introduce enhancements to the DPO framework in three key areas - \ul{R}epresentation, \ul{K}ernels, and \ul{D}ivergence measures, collectively referred to as \textbf{RKD}. (cf. \cref{sec:intro}).}
% }

\item[$\blacktriangleright$] 
{\footnotesize 
{\fontfamily{phv}\fontsize{7.5}{8.5}\selectfont
\textbf{Representation}: We enrich the representation space by combining the standard probability-based contrastive loss with semantic embeddings, ensuring that model preferences reflect both statistical likelihoods and meaningful, context-sensitive qualities. (cf. \cref{sec:representation}) and \cref{sec:appendix:hybrid_loss}.}
}

\item[$\blacktriangleright$] 
{\footnotesize 
{\fontfamily{phv}\fontsize{7.5}{8.5}\selectfont
\textbf{Kernels}: We enhance the DPO contrastive loss maximization by integrating kernel-based measures, allowing for flexible alignment in transformed feature spaces rather than relying solely on direct distribution comparisons. Incorporating polynomial, RBF, spectral, and Mahalanobis kernels. (cf. \cref{sec:kernel_dpo} and \cref{sec:appendix:dpo_kernel}).}
}

\item[$\blacktriangleright$] 
{\footnotesize 
{\fontfamily{phv}\fontsize{7.5}{8.5}\selectfont
\textbf{Divergence}: Exploration of alternative divergence measures (e.g., Jensen-Shannon, Hellinger, Rényi, Bhattacharyya, Wasserstein, and $f$-divergences) addresses known limitations of KL divergence, such as instability and lack of robustness (cf. \cref{sec:divergence} and \cref{sec:appendix:alternative_divergences}).}
}

\item[$\blacktriangleright$] 
{\footnotesize 
{\fontfamily{phv}\fontsize{7.5}{8.5}\selectfont
\textbf{Proposed DPO-Kernels}: The DPO-kernels could be explained using a simplified equation:

\vspace{-4.5mm}

\[
\max_{\pi} \; \underbrace{\mathbb{E}_{x, y^{+}, y^{-}} \kappa \Biggl[ 
\overbrace{\log \frac{\pi(y^{+} \mid x)}{\pi(y^{-} \mid x)}}^{\text{Contrastive Loss}} 
+ \overbrace{\gamma \log \bigg(\frac{e_{y^+} \mid e_x}{e_{y^-} \mid e_x}\bigg)}^{\text{Embedding Based Loss}} 
\Biggr]}_{\text{Kernelized Hybrid Loss}} \\
- \underbrace{\alpha \, \mathbb{E}_{x} \left[\sum_{y} \pi(y \mid x) \log \frac{\pi(y \mid x)}{\pi_{\text{ref}}(y \mid x)} \right]}_{\text{KL Divergence}}
\]
The equation maximizes the Kernelized Contrastive Loss, which differentiates positive and negative samples using probability ratios and embedding similarities. Concurrently, it incorporates an Alternative Divergence Regularizer scaled by $\alpha$, which enforces the model's distribution $\pi_\theta(y \mid x)$ to remain close to a reference distribution $\pi_{\text{ref}}(y \mid x)$ using a generic divergence measure $D$. This dual-objective framework enhances the model's discriminative power while ensuring distributional stability.}
}

\item[$\blacktriangleright$] 
{\footnotesize 
{\fontfamily{phv}\fontsize{7.5}{8.5}\selectfont
\textbf{Data-Driven Selection of Kernel Type and Divergence Functions}: Selecting the best kernel-divergence pair from 28 combinations (4 kernels × 7 divergences) is non-trivial. To simplify this, we propose 4 metrics for kernel selection—\textit{Positive-Negative Divergence (PND)}, \textit{Positive-Negative Alignment Variance (PNAV)}, \textit{Triplet Alignment Tightness (TAT)}, and \textit{Normalized Alignment Gap (NAG)}—and 4 metrics for divergence selection: \textit{Support Overlap}, \textit{Drift Magnitude}, \textit{Kurtosis}, and \textit{Smoothness}. (cf. \cref{sec:data_driven_kernel_selection}  and  \cref{sec:appendix:data_driven_kernel_divergence}).
}}

\item[$\blacktriangleright$] 
{\footnotesize 
{\fontfamily{phv}\fontsize{7.5}{8.5}\selectfont
\textbf{Kernel Mixture and HMK Introduction:} The diversity of alignment tasks necessitates a kernel mixture model to leverage the complementary strengths of different kernels, such as local (e.g., RBF) and global (e.g., Spectral) patterns. However, naive mixtures are prone to kernel collapse, where one kernel dominates, reducing adaptability and generalization. To address this, we propose the \textbf{Hierarchical Mixture of Kernels (HMK)}, a robust framework that balances fine-grained and large-scale dependencies, maintaining kernel diversity and ensuring optimal alignment. (cf. \cref{sec:kernel_mixture_main} and \cref{sec:appendix:kernel_mixture}).
}}

\item[$\blacktriangleright$] 
{\footnotesize 
{\fontfamily{phv}\fontsize{7.5}{8.5}\selectfont
\textbf{Gradient Computation, Computational Complexity, and Overhead:} Mathematical derivations for gradient computations for Hybrid Loss and different kernels-divergences, computational complexity analysis of different kernels-divergences, and DPO-Kernel overhead compared to original DPO are provided only in \cref{sec:appendix:gradient_complexity}.
}}

\item[$\blacktriangleright$] 
{\footnotesize 
{\fontfamily{phv}\fontsize{7.5}{8.5}\selectfont
\textbf{Empirical Findings:} Evaluations on 12 datasets show that \textbf{DPO-Kernels}, particularly HMK, achieve state-of-the-art generalization in factuality, safety, reasoning, and instruction-following tasks. However, HMK incurs 3-4× higher computational costs compared to standard DPO. We outline strategies to address this challenge in the limitations section, paving the way for cost-efficient future implementations. (cf. \cref{sec:results} and \cref{sec:appendix:results}).
}}

\item[$\blacktriangleright$] 
{\footnotesize 
{\fontfamily{phv}\fontsize{7.5}{8.5}\selectfont
\textbf{Safe vs. Unsafe Cluster Effects:} Kernel-induced clustering during safety fine-tuning projects unsafe inputs into null spaces \cite{jain2024safetyfinetuning}, creating distinct and compact clusters for safe and unsafe data. Metrics like the Davies-Bouldin Score (DBS) are used to quantify the separation and cohesion of these clusters, ensuring robust safety alignment. (cf. \cref{sec:safe_unsafe_cluster} and \cref{sec:appendix:safe_unsafe_cluster}).
}}

\item[$\blacktriangleright$] 
{\footnotesize 
{\fontfamily{phv}\fontsize{7.5}{8.5}\selectfont
\textbf{Heavy-Tailed Self-Regularization (HT-SR):} Grounded in HT-SR theory, the \textit{Weighted Alpha} metric \cite{martin2021predicting} provides a novel framework to evaluate generalization and overfitting in LLMs without relying on training or test data. Our analysis explores whether aligned models, particularly HMK, exhibit overfitting and quantifies the extent if present. (cf. \cref{sec:HTSR_generalization} and \cref{sec:appendix:htsr_generalization}).
}}

\item[$\blacktriangleright$] 
{\footnotesize 
{\fontfamily{phv}\fontsize{7.5}{8.5}\selectfont
\textbf{FAQ Section:} This section covers commonly asked questions along with those debated internally during the development process, offering insights into key design choices, challenges, and their resolutions.
cf. \cref{sec:FAQs}.
}}

\item[$\blacktriangleright$] 
{\footnotesize 
{\fontfamily{phv}\fontsize{7.5}{8.5}\selectfont
\textbf{Hyperparameters and Best Practices}:  We outline key hyperparameter settings and practical guidelines to optimize DPO-Kernel performance across diverse tasks in \cref{sec:appendix:hyperparameter}.
}}

\item[$\blacktriangleright$] 
{\footnotesize 
{\fontfamily{phv}\fontsize{7.5}{8.5}\selectfont
\textbf{Discussion, Limitations, and Ethical Considerations:} \cref{sec:limitaions} discusses limitations, including computational overhead, kernel collapse, adversarial robustness, hyperparameter sensitivity, and multimodal alignment. Ethical considerations - \cref{sec:ethical_consideration} covers fairness, bias, privacy risks, interpretability, environmental impact, and potential misuse. Both sections provide concise tabular and graphical summaries.}}

\item[$\blacktriangleright$] 
{\footnotesize 
{\fontfamily{phv}\fontsize{7.5}{8.5}\selectfont
\textbf{Broader Impact}: The broader impact of DPO-Kernels lies in its potential to transform how AI systems align with human preferences, with possible future extensions to text-to-image \cite{yoon2024safreetrainingfreeadaptiveguard,wallace2023diffusion,liu2024miadpomultiimageaugmenteddirect}, text-to-video \cite{yoon2024safreetrainingfreeadaptiveguard}, and Vision-Language Models \cite{wang2024rovrmrobustvisualreward,yu2024rlhfvtrustworthymllmsbehavior}. Beyond its technical contributions, DPO-Kernels provides a foundation for advancing alignment mechanisms, and we encourage the community to explore and experiment with its capabilities.
}
}
\end{spacing}
\end{itemize}

\end{defin}

\vspace{-6mm}

\textbf{Contrastive Loss} \(\left(\log \frac{\pi(y^+ \mid x)}{\pi(y^- \mid x)}\right)\) encourages the policy \(\pi\) to assign higher probabilities to preferred outputs \(y^+\) compared to less preferred outputs \(y^-\), given the same input \(x\). This term effectively pushes the policy to rank preferred responses higher, aligning it with observed preferences.

\textbf{KL Divergence} \(\left(\sum_{y} \pi(y \mid x) \log \frac{\pi(y \mid x)}{\pi_{\text{ref}}(y \mid x)}\right)\) measures the divergence between the optimized policy \(\pi\) and the reference policy \(\pi_{\text{ref}}\). This regularization term acts as a safeguard, preventing \(\pi\) from deviating excessively from the stable baseline provided by \(\pi_{\text{ref}}\). Without this regularization, the policy might become overconfident in certain responses or drastically alter its distribution in undesirable ways. The hyperparameter \(\alpha\) controls the strength of this regularization: a higher \(\alpha\) keeps the policy closer to \(\pi_{\text{ref}}\), making it more conservative, while a lower \(\alpha\) allows greater flexibility for the policy to adjust probabilities based on preferences.

In this work, we propose three key innovations to extend the capabilities of Direct Preference Optimization (DPO). First, we enrich the representation space by combining the standard probability-based contrastive loss with semantic embeddings, ensuring that model preferences reflect both statistical likelihoods and meaningful, context-sensitive qualities. Second, we enhance contrastive loss maximization by integrating kernel-based measures, allowing for flexible alignment in transformed feature spaces rather than relying solely on direct distribution comparisons. Finally, we move beyond the KL divergence by incorporating alternative divergence measures, such as Jensen–Shannon or Rényi divergences, to achieve more stable gradients, improved robustness, and better capture of the target distribution’s intricacies. Together, these advancements form the DPO-Kernels framework, which we rigorously evaluate through empirical benchmarks, demonstrating significant improvements over baseline methods in stability, semantic awareness, and alignment efficacy.

%\input{2_axioms}

%\section{DPO Kernels - Representation, Kernels, and Divergence Functions}
%\label{sec:llm}

%\subsection{Richer Representations}
%i) sentence embedding\\
%ii) longer text embeddings - Jina embedding\\
%iii) entity-context representation

\begin{figure*}[ht!]
    \centering
    \includegraphics[width=\textwidth]{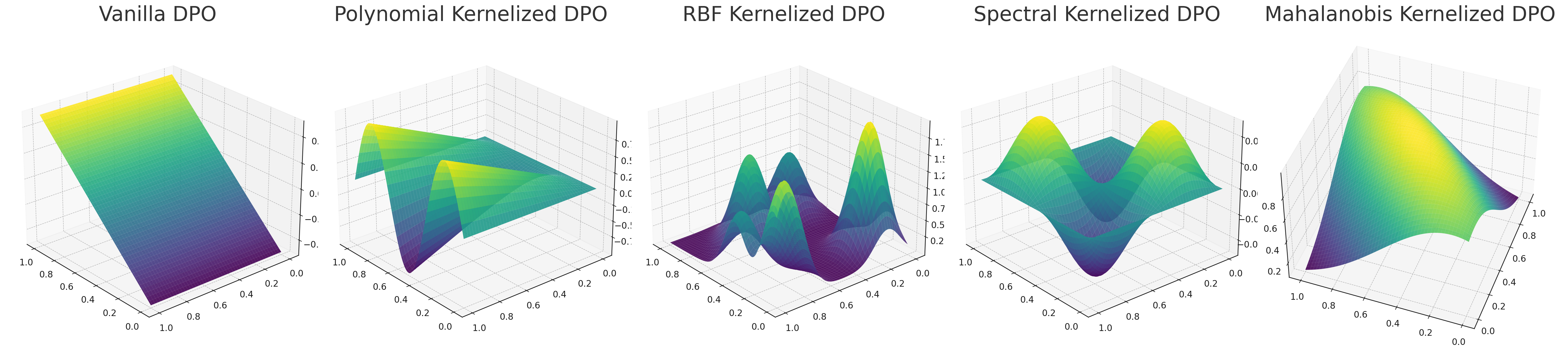}
    \caption{Kernel methods are techniques in machine learning that allow us to implicitly map input data into a higher-dimensional feature space without explicitly performing the transformation. This is achieved through kernels, which are functions that compute the inner product of two data points in the transformed feature space. For better intution on gradient descent dynamics on kernel-induced loss landscapes cf. \cref{sec:appendix:loss_landscape}.}
    \label{fig:error_surface_horizontal}
\end{figure*}

\begin{table*}[!ht]
\centering
\scriptsize % Compact font size
\setlength{\aboverulesep}{0pt}
\setlength{\belowrulesep}{0pt}
\setlength{\extrarowheight}{0pt}
\renewcommand{\arraystretch}{0.95} % Reduce vertical row spacing
\setlength{\jot}{1pt} % Compact spacing in equations

\begin{tabularx}{\textwidth}{@{}lX@{}} % Compact layout with dynamic column width
\toprule
\textbf{Kernel} & \textbf{Probability-Based and Embedding-Based Terms with Description} \\
\midrule

% Polynomial Kernel
\textbf{Polynomial} & 
$
\kappa \left[ \log \left( \frac{\pi(y^{+} \mid x)}{\pi(y^{-} \mid x)} \right) \right] = \left( \log \frac{\pi(y^{+})}{\pi(y^{-})} + c \right)^d, \quad
\kappa \left[ \log \left( \frac{e_{y^{+}} \mid e_x}{e_{y^{-}} \mid e_x} \right) \right] = \left( \frac{\left(e_x^{\top}\right) e_{y^{+}} + c}{\left(e_x^{\top}\right) e_{y^{-}} + c} \right)^d
$
Captures higher-order interactions using \((u^{\top} v + c)^d\). The parameter \(d\) controls complexity. \\

\midrule

% RBF Kernel
\textbf{RBF} & 
$
\kappa \left[ \log \left( \frac{\pi(y^{+} \mid x)}{\pi(y^{-} \mid x)} \right) \right] = \exp \left( -\frac{\left( \log \frac{\pi(y^{+} \mid x)}{\pi(y^{-} \mid x)} \right)^2}{2\sigma^2} \right), \quad
\kappa \left[ \log \left( \frac{e_{y^{+}} \mid e_x}{e_{y^{-}} \mid e_x} \right) \right] = \exp \left( -\frac{\left( \frac{\left(e_x^{\top}\right) e_{y^{+}}}{\left(e_x^{\top}\right) e_{y^{-}}} \right)^2}{2\sigma^2} \right)
$
Measures local similarity between inputs and outputs using the RBF kernel. \(\sigma\) controls smoothness. \\

\midrule

% Spectral Kernel
\textbf{Spectral} & 
$
\kappa \left[ \log \left( \frac{\pi(y^{+} \mid x)}{\pi(y^{-} \mid x)} \right) \right] = \sum_{i=1}^p \exp \left( -\lambda_i \left( \log \frac{\pi(y^{+} \mid x)}{\pi(y^{-} \mid x)} \right)^2 \right) \phi_i \left( \log \frac{\pi(y^{+} \mid x)}{\pi(y^{-} \mid x)} \right),
$
$
\kappa \left[ \log \left( \frac{e_{y^{+}} \mid e_x}{e_{y^{-}} \mid e_x} \right) \right] = \sum_{i=1}^p \exp \left( -\lambda_i \left( \frac{\left(e_x^{\top}\right) e_{y^{+}}}{\left(e_x^{\top}\right) e_{y^{-}}} \right)^2 \right) \phi_i \left( \frac{\left(e_x^{\top}\right) e_{y^{+}}}{\left(e_x^{\top}\right) e_{y^{-}}} \right)
$
Decomposes inputs and outputs into eigenfunctions \(\phi_k\) and eigenvalues \(\lambda_k\) to capture global, frequency-based dependencies. \\

\midrule

% Mahalanobis Kernel
\textbf{Mahalanobis} & 
$
\kappa \left[ \log \left( \frac{\pi(y^{+} \mid x)}{\pi(y^{-} \mid x)} \right) \right] = \exp \left( -\frac{\left( \log \frac{\pi(y^{+} \mid x)}{\pi(y^{-} \mid x)} - \mu \right)^2}{2\sigma^2} \right), \quad
\kappa \left[ \log \left( \frac{e_{y^{+}} \mid e_x}{e_{y^{-}} \mid e_x} \right) \right] = \exp \left( -\frac{\left( \frac{\left(e_x^{\top}\right) e_{y^{+}}}{\left(e_x^{\top}\right) e_{y^{-}}} - {\mu'} \right)^2}{2{\sigma'}^2} \right)
$
Leverages the Mahalanobis distance to capture anisotropic feature correlations using the covariance matrix \(\Sigma\). \\

\midrule

% HMK Kernel
\textbf{HMK} & 
$
\kappa \left[ \log \left( \frac{\pi(y^{+} \mid x)}{\pi(y^{-} \mid x)} \right) \right] = \sum_{i=1}^4 \tau_i \lambda_i \kappa_i \left( \log \frac{\pi(y^{+} \mid x)}{\pi(y^{-} \mid x)} \right),
$
$
\kappa \left[ \log \left( \frac{e_{y^{+}} \mid e_x}{e_{y^{-}} \mid e_x} \right) \right] = 
\tau_1 \left( \frac{\lambda_1 \kappa_{\text{RBF}}(e_x, e_{y^{+}}) + \lambda_2 \kappa_{\text{Poly}}(e_x, e_{y^{+}})}{\lambda_1 \kappa_{\text{RBF}}(e_x, e_{y^{-}}) + \lambda_2 \kappa_{\text{Poly}}(e_x, e_{y^{-}})} \right)
+ \tau_2 \left( \frac{\lambda_3 \kappa_{\text{Spectral}}(e_x, e_{y^{+}}) + \lambda_4 \kappa_{\text{Maha}}(e_x, e_{y^{+}})}{\lambda_3 \kappa_{\text{Spectral}}(e_x, e_{y^{-}}) + \lambda_4 \kappa_{\text{Maha}}(e_x, e_{y^{-}})} \right)
$
Combines multiple kernels hierarchically, balancing local kernels (RBF, Polynomial) and global kernels (Spectral, Mahalanobis).  
$
K(x, x') = \tau_1 (\lambda_1 K_{\text{RBF}} + \lambda_2 K_{\text{Poly}}) 
+ \tau_2 (\lambda_3 K_{\text{Spectral}} + \lambda_4 K_{\text{Maha}})
$
\\
\bottomrule
\end{tabularx}
\caption{Expansion of kernelized hybrid loss into: (a) kernelized probability-based loss and (b) kernelized embedding-based loss for Polynomial, RBF, Spectral, Mahalanobis kernels and HMK.}
\label{tab:dpo_kernel_loss_functions}
\end{table*}

\section{Richer Representation: Hybrid Approach: Integrating Probability and Embeddings}
\label{sec:representation}

DPO \cite{rafailov2024directpreferenceoptimizationlanguage} relies on the contrastive loss $\log\frac{\pi(y^+ \mid x)}{\pi(y^- \mid x)}$, which focuses solely on probability-based preferences. While effective, this approach often neglects deeper semantic and qualitative factors inherent in human preferences. To address this limitation, we introduce a hybrid preference alignment method that integrates embedding-based signals alongside probability-based cues. Our approach defines a preference signal as $f_{\text{embed}}(x, y^+, y^-) = e_{y^+} - e_{y^-}$, where \(e_{y^+}\) and \(e_{y^-}\) are embedding-based similarity scores for positive and negative responses, respectively. For our experiments, we utilize \texttt{jina-embeddings-v3} \cite{sturua2024jinaembeddingsv3multilingualembeddingstask}, but the framework is adaptable to other embeddings, enabling generalization across embedding models.

Embedding-based representations are well-established in preference modeling, reward design, and metric learning \cite{bai2022training, ouyang2022training, peyre2019computational}, often relying on pairwise distances or fixed objectives \cite{oord2018representation, chen2020simple, radford2021learning}. Recent large language models (LLMs) like LaMDA \cite{thoppilan2022lamda} and PaLM \cite{chowdhery2022palm} also leverage embeddings for preference alignment. However, existing approaches typically treat embeddings and probability-based signals separately, relying on fixed divergence measures (e.g., KL, triplet loss \cite{schroff2015facenet}, or contrastive loss \cite{hadsell2006dimensionality}). In contrast, our work is the \textbf{first to bridge embeddings and probability-based alignment in a unified parametric framework for policy learning}, offering a more comprehensive approach to preference optimization.

\paragraph{Hybrid Loss:}  
We blend probability and embedding signals:

\vspace{-4mm}
% \begin{multline*}
% \max_{\pi} \; \underbrace{\mathbb{E}_{x,y^{+},y^{-}}\bigl[\log\frac{\pi(y^{+} \mid x)}{\pi(y^{-} \mid x)} 
% + \gamma \bigl(\log\frac{\pi({e_{y^+}} \mid e_{x})}{\pi({e_{y^-}} \mid e_{x})}\bigr)\bigr]}_{\text{Hybrid Loss}} 
% - \alpha KL
% \end{multline*}

\[
\resizebox{\columnwidth}{!}{$
\max_{\pi} \; 
\underbrace{\mathbb{E}_{x,y^{+},y^{-}}\bigl[\log\frac{\pi(y^{+} \mid x)}{\pi(y^{-} \mid x)} 
+ \gamma \bigl(\log\frac{\pi(e_{y^+} \mid e_{x})}{\pi(e_{y^-} \mid e_{x})}\bigr)\bigr]}_{\text{Hybrid Loss}} 
- \alpha KL
$}
\]

with \(\gamma>0\) controlling the contribution of the embedding signal. When \(\gamma=0\), we recover the standard DPO loss. Increasing \(\gamma\) guiding the policy to produce outputs that are both probable and semantically preferable.

\paragraph{Interpretation:}
\begin{itemize}
\item  \textbf{Embedding-Guided Tie-Breaking}:  When probabilities are similar, embeddings help break ties by favoring outputs that are semantically more aligned or orthogonal. This alignment ensures that the selected output is not only probable but also semantically relevant, which is crucial for preference-driven alignment.

\vspace{-3mm}

\item \textbf{Semantic Consistency Check}: If the model strongly prefers \(y^+\) but embeddings do not support its semantic quality, a moderate \(\gamma\) prevents purely probability-driven reinforcement. Instead, it encourages the model to refine its output distribution to better align with semantic criteria, promoting more meaningful preference-based selection.
\end{itemize}

\vspace{-3mm}
The hybrid loss is then embedded within a kernel function, enabling DPO-Kernel to capture local, global, and higher-order dependencies, as detailed in the next section. \cref{sec:appendix:hybrid_loss} formulates our novel hybrid loss covering its mathematical definition, term-based decomposition, properties, impact on policy learning, etc.

\begin{table*}[ht!]
\centering
\renewcommand{\arraystretch}{1.0} % Adjust row height
\setlength{\tabcolsep}{6pt}       % Minimize horizontal padding

\begin{tabularx}{\textwidth}{|>{\raggedright\arraybackslash}p{2.5cm}|X|}
% \hline
\toprule
\textbf{Divergence} & \textbf{Mathematical Definition and Description} \\
% \hline
\midrule

\textbf{Jensen-Shannon Divergence} & 
\scriptsize
\(
D_{\text{JS}}(P \| Q) = \frac{1}{2} D_{\text{KL}}(P \| M) + \frac{1}{2} D_{\text{KL}}(Q \| M), \quad M = \frac{1}{2}(P + Q)
\). \textit{A symmetrized and smoothed version of KL divergence, which measures how different two probability distributions are. It is bounded and always finite, making it more stable for comparing distributions. The DPO objective with JS divergence becomes:}
$\max_{\pi} \; \mathcal{L}_{\text{KCL}} 
- \alpha \, \mathbb{E}_x \big[ D_{\text{JSD}}(\pi \,\|\, p_{\text{ref}}) \big]$
\\
% \hline
\midrule

\textbf{Hellinger Distance} & 
\scriptsize
\(
H(P, Q) = \frac{1}{\sqrt{2}} \sqrt{\int (\sqrt{p(x)} - \sqrt{q(x)})^2 \, dx}
\). \textit{A bounded distance measure (between 0 and 1) that quantifies the similarity between two probability distributions. It is widely used in Bayesian statistics and robust to outliers. The DPO objective with Hellinger distance becomes:}
$\max_{\pi} \; \mathcal{L}_{\text{KCL}} - \alpha \, \mathbb{E}_x \big[ D_{\text{Hellinger}}(\pi \,\|\, p_{\text{ref}}) \big]$
\\ % \hline
\midrule

\textbf{Rényi Divergence} & 
\scriptsize
\(
D_{\alpha}(P \| Q) = \frac{1}{\alpha - 1} \log \int p(x)^\alpha \, q(x)^{1 - \alpha} \, dx
\). \textit{A parametric generalization of KL divergence controlled by \(\alpha\). It interpolates between KL divergence (\(\alpha \to 1\)) and the maximum divergence as \(\alpha \to \infty\). Useful in robust learning where control over sensitivity is required. The DPO objective with Hellinger distance becomes:}
$\max_{\pi} \; \mathcal{L}_{\text{KCL}} 
- \alpha \, \mathbb{E}_x \big[ D_{\alpha}(\pi \,\|\, p_{\text{ref}}) \big]$
\\ % \hline
\midrule

\textbf{Bhattacharyya Distance} & 
\scriptsize
\(
D_{\text{Bhat}}(P, Q) = -\log \int \sqrt{p(x) \, q(x)} \, dx
\). \textit{Measures the amount of overlap between two probability distributions. It is commonly used in classification tasks, especially in Bayesian decision theory, to quantify the separability of two distributions. The DPO objective with Bhattacharyya distance becomes:} $\max_{\pi} \; \mathcal{L}_{\text{KCL}} - \alpha \, \mathbb{E}_x \big[ D_{\text{Bhattacharyya}}(\pi \,\|\, p_{\text{ref}}) \big]$
\\ % \hline
\midrule

\textbf{Wasserstein Distance} & 
\scriptsize
\(
W(P, Q) = \inf_{\gamma \in \Pi(P, Q)} \mathbb{E}_{(x, y) \sim \gamma} \left[ \| x - y \| \right]
\). \textit{Also known as Earth Mover's Distance, it quantifies how much "work" is needed to morph one distribution into another. Unlike KL, it is well-defined for distributions that do not overlap and is widely used in generative modeling and distribution alignment. The DPO objective with Wasserstein distance becomes:} $\max_{\pi} \; \mathcal{L}_{\text{KCL}} - \alpha \, \mathbb{E}_x \big[ W(\pi, p_{\text{ref}}) \big]$
\\ % \hline
\midrule

\textbf{f-Divergence} & 
\scriptsize
\(
D_{f}(P \| Q) = \int q(x) f \left( \frac{p(x)}{q(x)} \right) \, dx
\). \textit{A general class of divergences that subsumes KL, Jensen-Shannon, and others as special cases. It is defined via a convex function \(f\), providing a unified view of multiple divergence measures. The DPO objective with an f-divergence becomes:} $\max_{\pi} \; \mathcal{L}_{\text{KCL}} - \alpha \, \mathbb{E}_x \big[ D_{f}(\pi \,\|\, p_{\text{ref}}) \big]$
\\ % \hline
\bottomrule
\end{tabularx}
\caption{Descriptions and mathematical definitions of divergence functions, including Jensen-Shannon, Hellinger, Rényi, Bhattacharyya, Wasserstein, and f-Divergence, and their applications to the DPO objective.}
\label{tab:divergence_functions}
\end{table*}

\section{Kernel-Integrated DPO Formulation}
\label{sec:kernel_dpo}

Standard DPO aligns a policy \(\pi\) with human preferences while regularizing against a reference distribution \(\pi_{\text{ref}}\) via a divergence \(D(\cdot \| \cdot)\). While effective, this approach relies on simple distributional differences, which may fail to capture deeper semantic relationships essential for alignment. To address this, we introduce kernelized proximity measures that enable more expressive and adaptive alignment. Our framework extends DPO into four distinct DPO-Kernel variants: (i) Polynomial, (ii) RBF, (iii) Spectral, and (iv) Mahalanobis. The resulting objective is expressed as:

\begin{comment}
\vspace{-3mm}
{\scriptsize
\begin{multline*}
\hspace{-5mm} \max_{\pi} \; \underbrace{\mathbb{E}_{x, y^+, y^-} \Bigl[ \log \frac{\pi(y^+ \mid x)}{\pi(y^- \mid x)} 
+ \gamma \bigl( \kappa(e_x, e_{y^+}) - \kappa(e_x, e_{y^-}) \bigr) \Bigr]}_{\text{Kernelized Contrastive Loss}} \\
\hspace{-5mm} - \underbrace{\alpha \mathbb{E}_x \Bigl[ \beta \log \frac{\pi_\theta(y \mid x)}{\pi_{\text{ref}}(y \mid x)} \Bigr]}_{\text{KL}}
\end{multline*}
}
\end{comment}

\[
\max_{\pi} \; \underbrace{\mathbb{E}_{x, y^+, y^-} \kappa \Bigl[ 
\log \bigg(\frac{\pi(y^+ \mid x)}{\pi(y^- \mid x)}\bigg) 
+ \gamma \log \bigg(\frac{e_{y^+} \mid e_x}{e_{y^-} \mid e_x}\bigg) 
\Bigr]}_{\text{Kernelized Hybrid Loss}}
- \alpha KL
\]

\begin{comment}
\[
\max_{\pi} \; \underbrace{\mathbb{E}_{x, y^+, y^-} \Bigl[ 
\log \kappa\bigg(\frac{\pi(y^+ \mid x)}{\pi(y^- \mid x)}\bigg) 
+ \gamma \log \kappa\bigg(\frac{e_{s^+} \mid e_x}{e_{s^-} \mid e_x}\bigg) 
\Bigr]}_{\text{Kernelized Contrastive Loss}}
- \underbrace{\alpha \mathbb{E}_x \Bigl[ \beta \log \frac{\pi_{\text{ref}}(y \mid x)}{\pi_\theta(y \mid x)} \Bigr]}_{\text{KL Regularization Term}}.
\]
\end{comment}

Each kernel offers a unique perspective on alignment. Polynomial kernels capture higher-order interactions, enabling compositional reasoning. RBF kernels emphasize local, fine-grained structure, useful for proximity-based alignment. Spectral kernels capture global, oscillatory patterns to handle periodic dependencies, while Mahalanobis kernels leverage feature covariance to account for anisotropic relationships. These kernelized variants preserve the core mathematical foundations of DPO while significantly enhancing its ability to capture richer alignment criteria.

\cref{fig:error_surface_horizontal} illustrates the effect of kernelizing the DPO objective with various kernels, including Polynomial, RBF, Spectral, and Mahalanobis, in comparison to the Vanilla DPO. Each plot shows how different kernels reshape the optimization landscape by implicitly mapping input data to higher-dimensional feature spaces, allowing the model to capture complex patterns and interactions. This kernelized transformation enhances the expressiveness of the DPO objective, enabling it to adapt to diverse data distributions and modeling needs.

\section{Replacing KL regularizer with alternatives}
\label{sec:divergence}

The original DPO framework typically utilizes the Kullback–Leibler (KL) divergence to align the learned policy \(\pi(y \mid x)\) with the reference distribution \(p_{\text{ref}}(y \mid x)\). While KL divergence is favored for its strong theoretical foundations, exploring alternative divergence measures can lead to more robust optimization, enhanced stability, and improved interpretability and generalizability. 
%Simplifying the equation further: \(\mathcal{L}_{\text{KCL}}\) represents the \textbf{Kernelized Contrastive Loss (KCL)}:
%\[
%\max_{\pi} \; \mathcal{L}_{\text{KCL}} 
%- \underbrace{ \alpha \mathbb{E}_x \left[ \beta \log \frac{\pi_\theta(y \mid x)}{\pi_{\text{ref}}(y \mid x)} \right] }_{\text{KL Regularization}}
%\]

\begin{figure}[h!]
    \centering
    \includegraphics[width=\columnwidth]{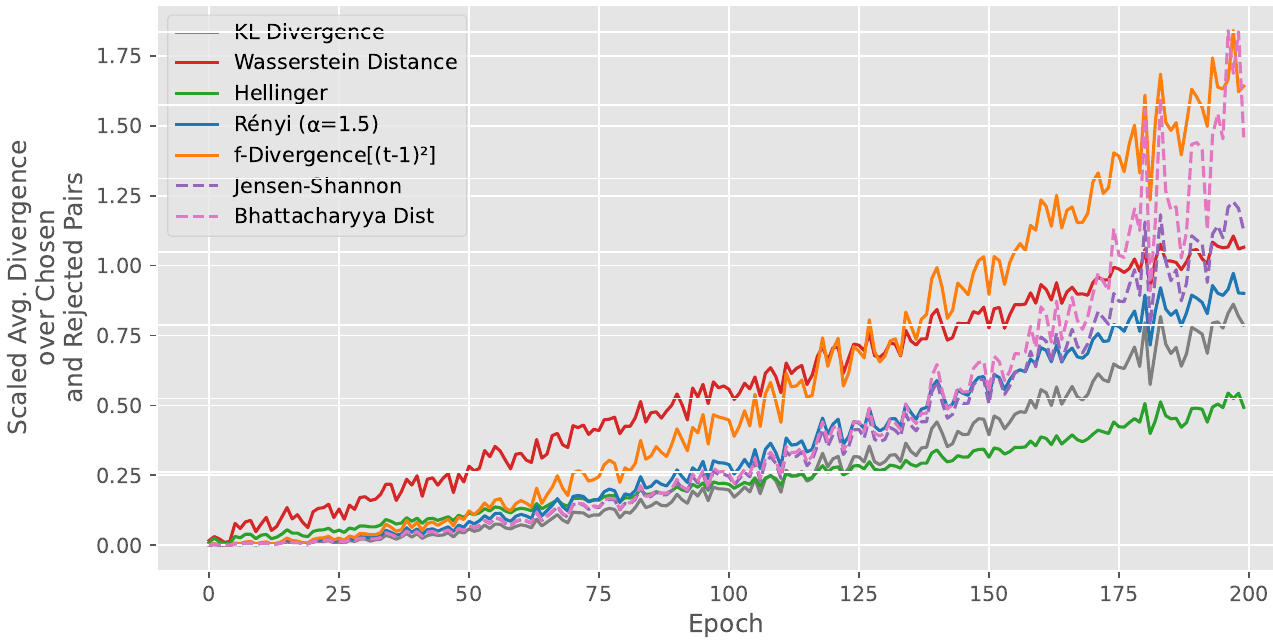}
    \caption{The plot illustrates the oscillatory behavior and trends of various divergence measures, including Wasserstein, Jensen-Shannon, Hellinger, Rényi, Bhattacharyya, and f-divergence, as the training progresses, reflecting their sensitivity to the evolving alignment dynamics.}
    \label{fig:oscillatory_divergence_main}
\end{figure}

\cref{fig:oscillatory_divergence_main} illustrates the temporal evolution of various divergence measures, including KL Divergence, Wasserstein Distance, Hellinger, Rényi, Bhattacharyya, Jensen-Shannon, and f-divergence, across training steps. The oscillatory behavior observed in the higher divergence measures (e.g., Rényi, Bhattacharyya, and f-divergence) highlights their sensitivity to dynamic alignment changes. In contrast, smoother trends in Wasserstein and Jensen-Shannon divergences indicate their stability and robustness over time. The overall upward trajectory reflects increasing distributional alignment shifts as training progresses, providing insights into how divergence measures respond to evolving alignment dynamics.

\begin{comment}
Given the diversity of kernel-divergence options, selecting the right kernel for a given task can be challenging. To address this, we provide two solutions: 

\vspace{-2mm}
\begin{itemize}
    \item \textbf{Kernel Mixture Approach}: An automated, task-agnostic strategy that optimally combines multiple kernels, albeit with higher computational cost. 
    \vspace{-2mm}
    \item \textbf{Data-Driven Kernel Selection}: A principled, task-aware approach that dynamically selects kernels based on task-specific alignment properties. Details of this approach are provided in Section \ref{sec:data_driven_kernel_selection}.  
\end{itemize}
\vspace{-2mm}
\end{comment}

The divergence equations are summarized in Table \ref{tab:divergence_functions}. For details, please refer to \cref{sec:appendix:alternative_divergences}.

\section{Data-Driven Selection of Kernel Types and Divergence Functions}
\label{sec:data_driven_kernel_selection}

Choosing the optimal kernel-divergence pair among 28 combinations (4 kernels $\times$ 7 divergences) is challenging. We propose a systematic, data-driven framework that replaces heuristics with well-defined metrics, ensuring adaptability and improved generalization.

\begin{table*}[ht!]
\centering
\resizebox{\textwidth}{!}{
\renewcommand{\arraystretch}{0.9} % Reduce row height
\setlength{\tabcolsep}{3pt}      % Reduce column spacing
\begin{tabular}{|p{3cm}|c|p{3.5cm}|p{7cm}|}
\toprule
\textbf{Metric} & \textbf{Formula} & \textbf{Description} & \textbf{Kernel Suggestions} \\
\midrule
\textbf{Pos.-Neg. Divergence (PND)} & 
$\displaystyle \frac{d(x, y^+)}{d(x, y^-)}$ & 
Indicates whether $x$ is closer to $y^+$ or $y^-$. A large PND implies strong imbalance. & 
Large PND $\rightarrow$ Mahalanobis (covariance); 
Small PND $\rightarrow$ Spectral/Polynomial (nonlinearity) \\
\midrule
\textbf{Pos.-Neg. Align. Var. (PNAV)} & 
$\displaystyle \frac{1}{n}\sum ( d(x_i,y_i^+) - d(x_i,y_i^-) )^2$ & 
Measures consistency of positive-negative separation. & 
High PNAV $\rightarrow$ RBF (flexible);
Low PNAV $\rightarrow$ Polynomial (simpler) \\
\midrule
\textbf{Triplet Align. Tightness (TAT)} & 
$\displaystyle \frac{1}{n}\sum \frac{\|y_i^+ - y_i^-\|}{\|y_i^+ - x_i\| + \|y_i^- - x_i\|}$ & 
How close $y^+$ and $y^-$ are relative to $x$. High TAT = cluster together. & 
High TAT $\rightarrow$ Spectral (complex patterns);
Low TAT $\rightarrow$ RBF (separated) \\
\midrule
\textbf{Norm. Align. Gap (NAG)} & 
$\displaystyle \frac{1}{n}\sum \frac{d(x_i,y_i^-)-d(x_i,y_i^+)}{d(x_i,y_i^-)+d(x_i,y_i^+)}$ & 
Balance in distances. NAG near zero = similar distances. & 
NAG $\approx 0 \rightarrow$ Polynomial (beyond linear);
NAG $\ne 0 \rightarrow$ Mahalanobis (covariance) \\
\bottomrule
\end{tabular}
}
\caption{Proposed Metrics for Kernel Selection: \textit{Positive-Negative Divergence (PND)}, \textit{Positive-Negative Alignment Variance (PNAV)}, \textit{Triplet Alignment Tightness (TAT)}, and \textit{Normalized Alignment Gap (NAG)}.}
\label{tab:metrics_kernel_selection_main}
\end{table*}

\begin{figure}[h!]
    \centering
    \includegraphics[width=\columnwidth]{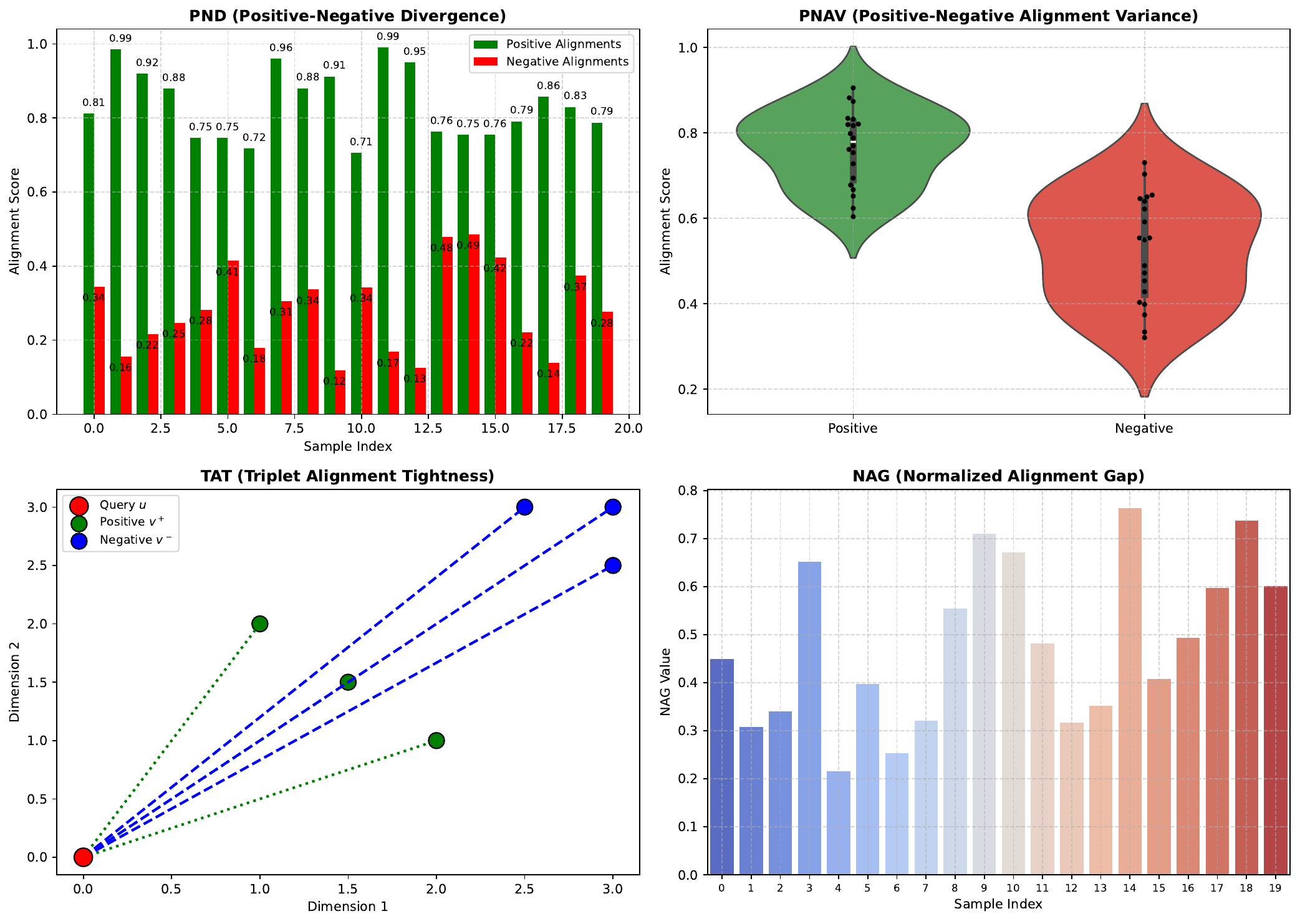}
    \caption{
    Visualization of the four proposed metrics for kernel selection in alignment tasks. 
    \textbf{(a) Positive-Negative Divergence (PND)} illustrates the divergence between alignment scores for positive and negative samples, indicating the degree of separability. 
    \textbf{(b) Positive-Negative Alignment Variance (PNAV)} depicts the variance in alignment scores for positive and negative samples, reflecting alignment consistency. 
    \textbf{(c) Triplet Alignment Tightness (TAT)} shows the relative positioning of query ($x$), positive ($y^+$), and negative ($y^-$) embeddings in the latent space, highlighting alignment precision. 
    \textbf{(d) Normalized Alignment Gap (NAG)} tracks the evolution of alignment gaps over samples, where smaller NAG values signify better alignment quality. 
    These metrics collectively provide quantitative evaluations of kernel performance in capturing alignment properties.
    }
    \label{fig:metrics_for_kernel_main}
\end{figure}

\subsection{Data-Driven Kernel Selection Logic}

We propose four novel metrics—\textit{Positive-Negative Divergence (PND)}, \textit{Positive-Negative Alignment Variance (PNAV)}, \textit{Triplet Alignment Tightness (TAT)}, and \textit{Normalized Alignment Gap (NAG)}—that quantify key geometric and relational properties of the data, summarized in \cref{tab:metrics_kernel_selection_main}. \cref{{fig:metrics_for_kernel_main}} visualizes the four proposed metrics for kernel selection in alignment tasks: these metrics collectively assess alignment properties, such as separability, consistency, precision, and gap quality, enabling a comprehensive evaluation of kernel performance in alignment.

Here, we prescribe a practical guideline to help users empirically select the most suitable kernel for alignment tasks based on key metrics. By leveraging thresholds for metrics such as PNAV, TAT, NAG, and PND, this framework provides an intuitive yet effective approach to kernel selection, ensuring alignment properties are well-captured for diverse scenarios.

\vspace{-4mm}
\scriptsize
\[
k^* = \begin{cases}
    \text{RBF Kernel}, & \text{if } \text{PNAV} > \varepsilon_1 \text{ and } \text{TAT} < \varepsilon_2 \\
    \text{Polynomial Kernel}, & \text{if } \text{NAG} \approx 0 \text{ and } \text{PND} \approx 0 \\
    \text{Mahalanobis Kernel}, & \text{if } \text{NAG} > 0 \text{ and } \text{PNAV} < \varepsilon_3 \\
    \text{Spectral Kernel}, & \text{if } \text{TAT} > \varepsilon_4 \text{ and } \text{PND} < \varepsilon_5 \\
\end{cases}
\]
\normalsize

Here, thresholds \(\varepsilon_1, \varepsilon_2, \varepsilon_3, \varepsilon_4, \varepsilon_5\) are empirically tuned or determined through validation. Initial values such as \(\varepsilon_1 = 0.5\), \(\varepsilon_2 = 0.3\), \(\varepsilon_3 = 0.2\), \(\varepsilon_4 = 0.7\), and \(\varepsilon_5 = 0.1\) serve as practical defaults. Balanced metrics (e.g., \(\approx 0\)) signal alignment structures, while larger deviations reveal more intricate relationships requiring advanced kernels.

\begin{table*}[ht!]
\centering
\resizebox{\textwidth}{!}{%
\renewcommand{\arraystretch}{1.2}
\begin{tabular}{|p{2.5cm}|p{5cm}|p{5cm}|c|}
% \hline
\toprule
\textbf{Property} & \textbf{Computation} & \textbf{When to Use} & \textbf{Best Divergence}\\
% \hline
\midrule
\textbf{Support Overlap} & 
$\frac{|p \cap q|}{|p \cup q|}$, high overlap means similar domains. & 
If overlap $> 0.6$: Bhattacharyya. Otherwise: KL or JS. & 
Bhattacharyya, KL, JS\\
% \hline
\midrule
\textbf{Drift Magnitude} & 
$\frac{1}{n}\sum (d(x,y^+) - d(x,y^-))$, higher = bigger shifts. & 
Large drift: Wasserstein. Small drift: KL or Rényi ($\alpha>1$). & 
Wasserstein, KL, Rényi\\
% \hline
\midrule
\textbf{Kurtosis} & 
$\frac{\mathbb{E}[(x-\mu)^4]}{(\mathbb{E}[(x-\mu)^2])^2}$, high values = heavy tails. & 
Kurtosis $>3$: Rényi. Else: JS or Hellinger. & 
Rényi, JS, Hellinger\\
% \hline
\midrule
\textbf{Smoothness} & 
$\frac{1}{T}\sum W(p_t,p_{t+1})$, lower = smoother transitions. & 
High smoothness: Wasserstein. Low: KL or Hellinger. & 
Wasserstein, KL, Hellinger\\
% \hline
\bottomrule
\end{tabular}
}
\caption{Proposed Metrics for Divergence Selection: \textit{Support Overlap}, \textit{Drift Magnitude}, \textit{Kurtosis}, and \textit{Smoothness}}
\label{tab:metrics_divergence_selection_main}
\end{table*}

\begin{figure}[h!]
    \centering
    \includegraphics[width=\columnwidth]{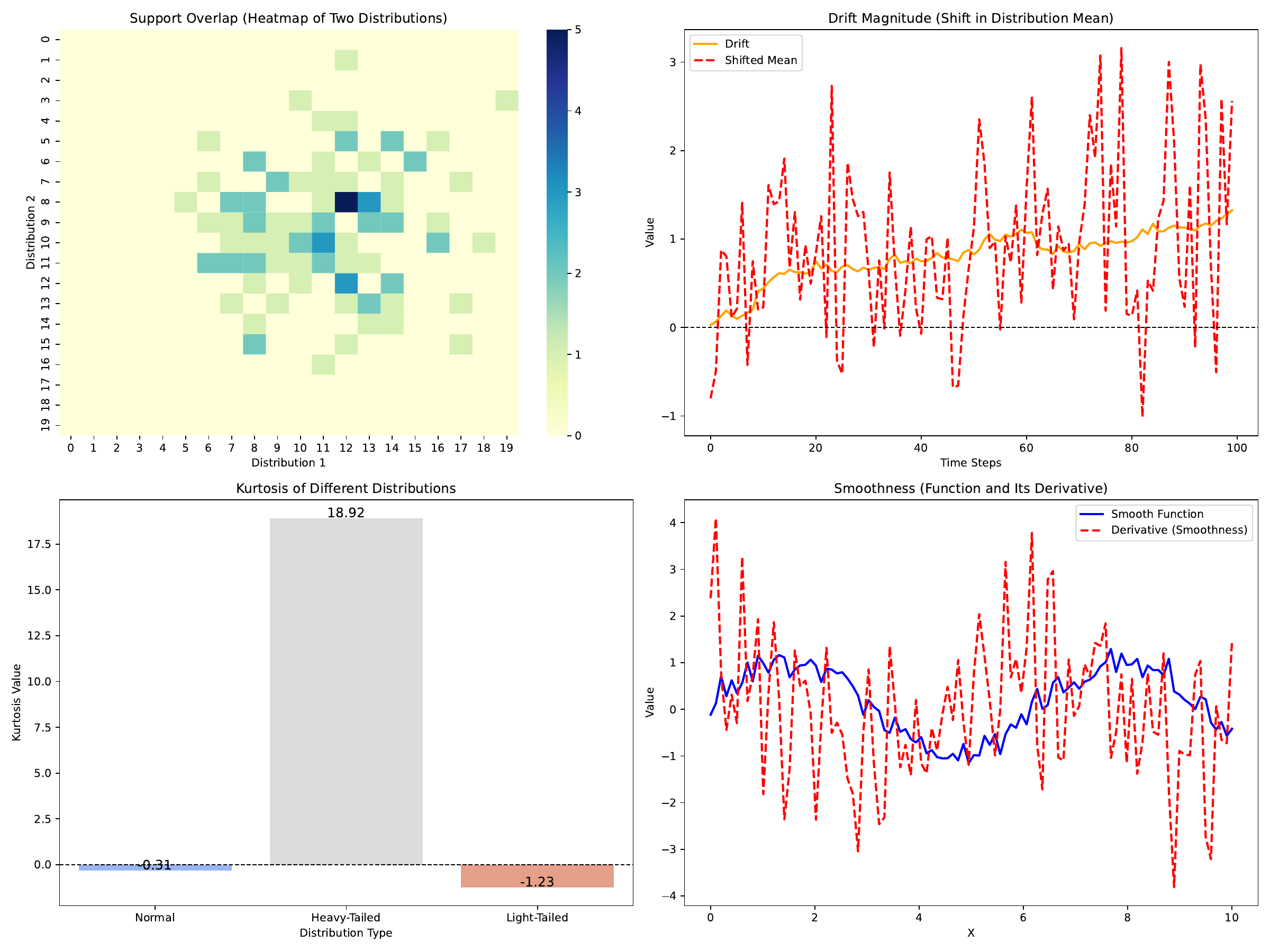}
    \caption{
    Visualization of the four key metrics for divergence selection: 
    \textbf{(1) Support Overlap} — Heatmap representing the overlap between two distributions, highlighting shared support regions; 
    \textbf{(2) Drift Magnitude} — Illustration of the shift in the mean of a distribution over time, showcasing how drift is detected; 
    \textbf{(3) Kurtosis} — Bar plot comparing kurtosis values for normal, heavy-tailed, and light-tailed distributions, quantifying the "tailedness" of each distribution; 
    \textbf{(4) Smoothness} — Visualization of a smooth function and its derivative, where smoother functions exhibit smaller, less abrupt changes in derivatives. 
    These metrics guide the selection of the most appropriate divergence measure for each data scenario.
    }
    \label{fig:metrics_for_divergence_main}
\end{figure}

\subsection{Data-Driven Divergence Choice Logic}

We further propose four distributional metrics—\textit{Support Overlap}, \textit{Drift Magnitude}, \textit{Kurtosis}, and \textit{Smoothness}—to systematically select the most appropriate divergence measure, summarized in \cref{tab:metrics_divergence_selection_main}. \cref{fig:metrics_for_divergence_main} visualizes the four proposed metrics for divergence selection: these metrics provide insights into the behavior of distributions by quantifying their overlap, shift, tail properties, and functional smoothness. Collectively, they enable the empirical selection of the most appropriate divergence measure for various data scenarios, ensuring effective modeling and comparison of distributions.

We provide a practical guideline to help users empirically select the most suitable divergence measure based on key metrics. These metrics offer insights into distributional behavior, ensuring the chosen divergence measure aligns with the data's characteristics.

\vspace{-3mm}
\scriptsize
\[
D^* = \begin{cases}
    \text{Bhattacharyya Divergence}, & \text{if Support Overlap} > \varepsilon_1 \\
    \text{Wasserstein Divergence}, & \text{if Drift Magnitude} > \varepsilon_2 \\
    \text{Rényi Divergence}, & \text{if Kurtosis} > \varepsilon_3 \\
    \text{Jensen-Shannon Divergence}, & \text{if Overlap is low and Kurtosis is low} \\
    \text{Hellinger Divergence}, & \text{if Smoothness is low and Kurtosis is low} \\
    \text{KL Divergence}, & \text{otherwise}
\end{cases}
\]
\normalsize

We recommend starting with thresholds \(\varepsilon_1 = 0.6\), \(\varepsilon_2 = 0.3\), and \(\varepsilon_3 = 3\), refining them based on the observed performance. This systematic approach ensures that divergence selection is directly tailored to the alignment complexity of the data. \cref{sec:appendix:data_driven_kernel_divergence} offers a detailed discourse for data-driven selection of kernel types and divergence functions based on the appropriate metrics.

\section{Kernel Mixture Approach - Improved Generalization}
\label{sec:kernel_mixture_main}

The use of a single kernel often fails to capture the diverse relationships inherent in alignment tasks. Different kernels are adept at modeling specific properties, such as local similarities, global structures, or higher-order interactions, making it challenging for any single kernel to perform well across all scenarios. A \textbf{Kernel Mixture Approach} addresses this limitation by dynamically combining multiple kernels, leveraging their complementary strengths to improve generalization across varied datasets (e.g., diverse alignment tasks as in \cite{Dubois_2024b, Lv_2023}, policy shifts \cite{Koh_2021}, and evolving alignment requirements \cite{Jain_2024}.

\noindent \textbf{Related Works:}  
Research in multiple kernel learning \cite{gonen2011multiple}, Gaussian processes \cite{duvenaud2013additive}, and distributional adaptation \cite{quinonero2009dataset, koh2021wilds} highlights the effectiveness of combining kernels to handle dataset heterogeneity and distributional shifts. Inspired by these principles, the Kernel Mixture Approach extends this flexibility by enabling task-specific kernel contributions. A straightforward formulation could be expressed as:
\begin{multline*}
\kappa(u, v) = \lambda_1 \kappa_{\text{poly}}(u, v) + \lambda_2 \kappa_{\text{RBF}}(u, v) \\
+ \lambda_3 \kappa_{\text{spec}}(u, v) + \lambda_4 \kappa_{\text{Maha}}(u, v),
\end{multline*}
where \(\lambda_1, \lambda_2, \lambda_3, \lambda_4 \geq 0\) and \(\sum_{i=1}^4 \lambda_i = 1.\) The weights are parameterized using a softmax: $\lambda_i = \frac{\exp(\theta_i)}{\sum_{j=1}^4 \exp(\theta_j)}$,
where \(\theta_i\) are trainable parameters optimized via gradient descent. This formulation allows the model to adapt kernel contributions dynamically to the task at hand.

\begin{figure}[h!]
    \centering
    \includegraphics[width=\columnwidth]{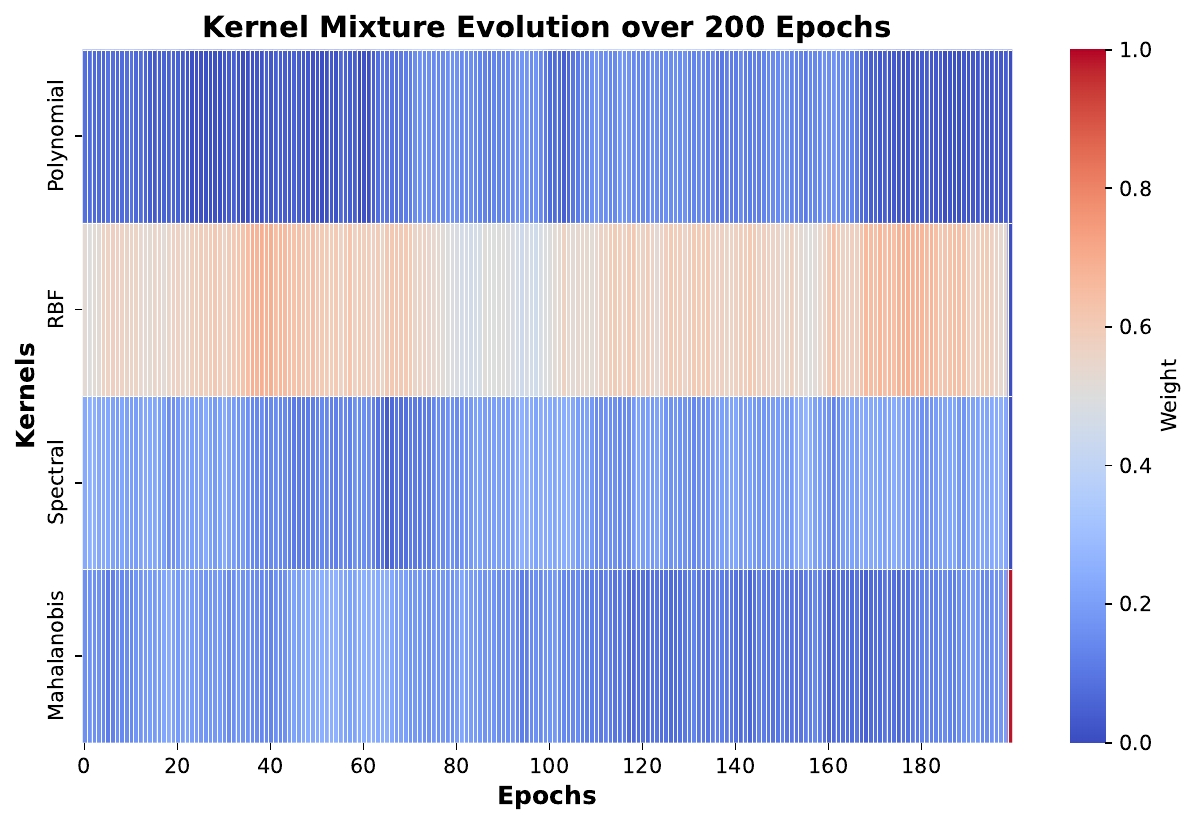}
    \caption{Evolution of Kernel Weights in the Mixture Over 200 Epochs. The plot illustrates the dynamic adjustment of kernel weights (\(\lambda_1\), \(\lambda_2\), \(\lambda_3\), \(\lambda_4\)) corresponding to Polynomial, RBF, Spectral, and Mahalanobis kernels, respectively, during training. Each curve represents the relative contribution of a kernel, showing how the model adapts its alignment strategy over time. The dominance of one or two kernels, as indicated by the curves, highlights the tendency towards kernel collapse, where certain kernels overshadow others. This visualization underscores the challenges in maintaining kernel diversity within the mixture.}
    \label{fig:kernel_mixture_main}
\end{figure}

However, a key challenge of this approach is \textbf{kernel collapse} \cite{lanckriet2004multiple,lanckriet2002learning,raetsch2005generalized}, where one kernel disproportionately dominates, effectively reducing the model to a single-kernel learner. This diminishes diversity and undermines the representational power needed to model complex data relationships. \cref{fig:kernel_mixture_main} depicts the evolution of kernel weights (\(\lambda_1, \lambda_2, \lambda_3, \lambda_4\)) for Polynomial, RBF, Spectral, and Mahalanobis kernels over 200 epochs. The dynamic adjustments showcase how the model prioritizes different kernels during training to optimize alignment. However, the visualization also highlights the risk of kernel collapse, where one or two kernels dominate, reducing diversity and potentially limiting the model's representational capacity. For detailed discussion please refer to \cref{sec:appendix:kernel_mixture}. Addressing this issue is essential for fully realizing the potential of kernel mixtures in alignment tasks.

%The provides a flexible mechanism to combine multiple kernels, dynamically adjusting their contributions to enhance generalization across diverse alignment tasks. This adaptability addresses challenges like policy shifts, dataset variations, and evolving alignment criteria. However, the approach is prone to \textbf{kernel collapse}, where some kernels lose influence during training, limiting its ability to model complex data relationships. To address this, we propose the \textbf{Hierarchical Mixture of Kernels (HMK)}, which preserves kernel diversity through a structured framework.

\subsection{Hierarchical Mixture of Kernels}
\label{sec:hmk_architecture}

Hierarchical Mixture of Kernels (HMK) overcomes kernel collapse by introducing a two-level decomposition that balances \textbf{local kernels} (RBF, Polynomial) \cite{scholkopf2002learning} and \textbf{global kernels} (Spectral, Mahalanobis) \cite{weinberger2009distance,ng2001spectral}. Local kernels capture short-range dependencies, while global kernels model broader, long-range relationships. HMK assigns learnable weights to both groups, enabling dynamic adaptation to varying data geometries:
\[
K(x, x') = \tau_1 \big( \lambda_1 K_{\text{RBF}} + \lambda_2 K_{\text{Poly}} \big) + \tau_2 \big( \lambda_3 K_{\text{Spectral}} + \lambda_4 K_{\text{Maha}} \big),
\]
where \(\tau_1, \tau_2\) balance local-global contributions. Both \(\tau\) and \(\lambda\) are updated through backpropagation, allowing HMK to maintain kernel diversity and adapt effectively.

\subsubsection{Illustration of the Effective Range}

\begin{figure}[h!]
    \centering
    \includegraphics[width=\columnwidth]{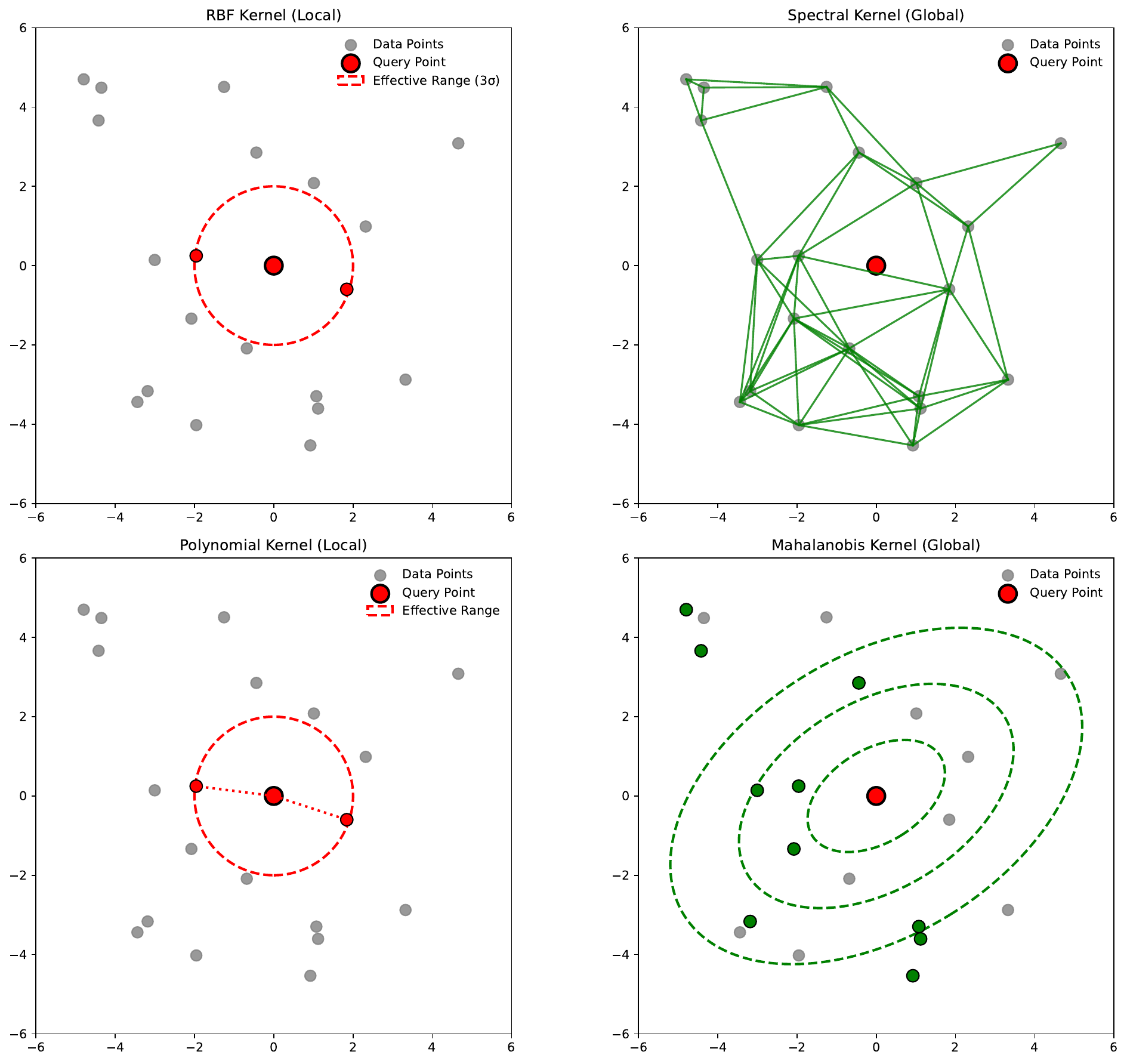}
    \caption{Local vs. global kernel influence. RBF and Polynomial kernels exhibit localized influence, while Spectral and Mahalanobis kernels capture broader dependencies.}
    \label{fig:effective_range_of_a_kernel_main}
\end{figure}

To visualize the kernel influence range, a set of 20 points was randomly sampled from the 2D space \([-5, 5] \times [-5, 5]\). A fixed query point at (0, 0) serves as the reference point for kernel similarity computation for the RBF, Polynomial, Spectral, and Mahalanobis kernels. Please refer to Figure \ref{fig:effective_range_of_a_kernel_main}.

\begin{itemize}
    \item \textbf{Purpose}: Random points offer a dataset-agnostic view of kernel influence.
    \item \textbf{Why It Matters}: The query point allows us to analyze how influence propagates, aiding in the understanding of \emph{local vs. global behavior}.
\end{itemize}

\subsection{Key Insights and Alignment Task Implications}
\begin{itemize}
    \item \textbf{Local Kernels:} Effective for fine-grained tasks like safety alignment or clustering, as their influence decays quickly with distance \cite{scholkopf2002learning}.
    \item \textbf{Global Kernels:} Crucial for tasks like contextual alignment or multi-hop reasoning, leveraging long-range dependencies \cite{ng2001spectral, maesschalck2000mahalanobis}.
    \item \textbf{Generalization:} HMK combines the strengths of local and global kernels, reducing overfitting while improving adaptability across diverse tasks.
    \item \textbf{Dynamic Adaptation:} The hierarchical structure enables task-aware prioritization of local or global influences, balancing short- and long-range dependencies \cite{belkin2003laplacian}.
    \item \textbf{Robustness to Shifts:} The Mahalanobis kernel adds robustness to covariance structure changes, complementing the Spectral kernel's global reach \cite{maesschalck2000mahalanobis}.
\end{itemize}

\subsection{Dynamic Evolution of Kernel Weights}
\begin{figure}[h!]
    \centering
    \includegraphics[width=\columnwidth]{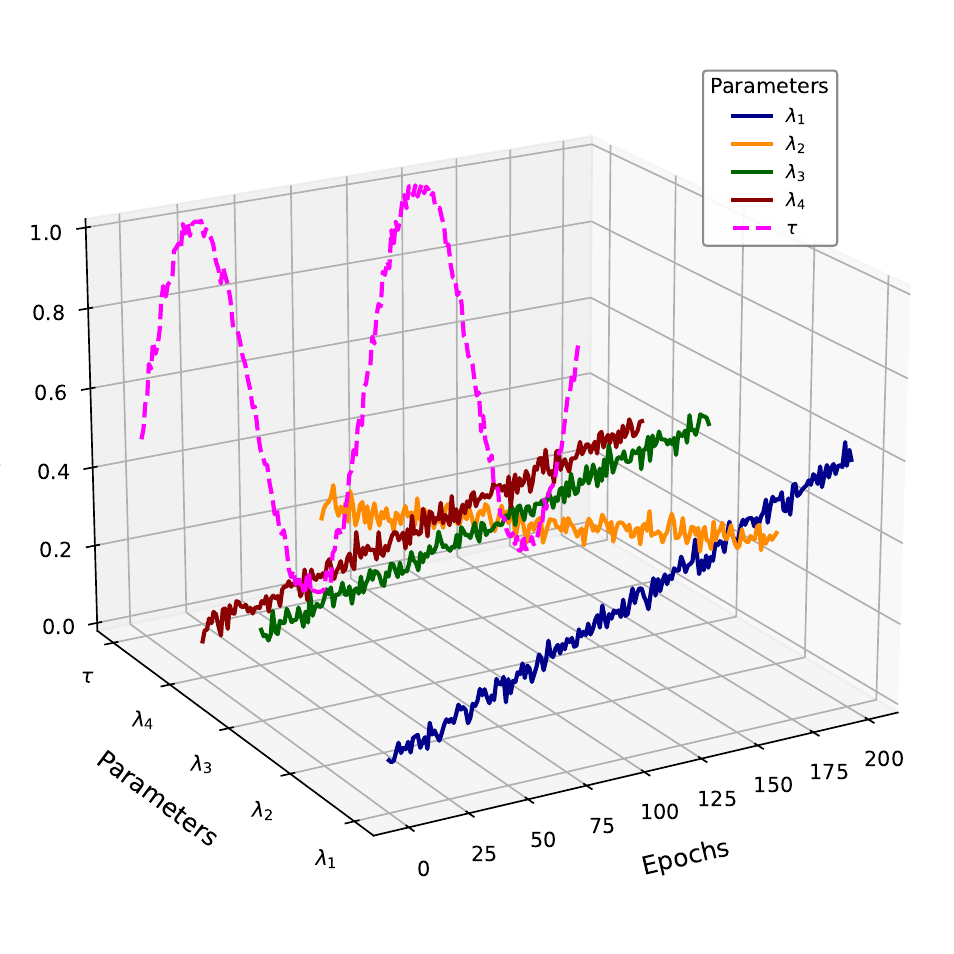}
    \caption{Dynamic evolution of kernel weights (\(\lambda_1, \lambda_2, \lambda_3, \lambda_4\)) and Local-Global Balance Coefficients (\(\tau_1, \tau_2\)). The model shifts its reliance on local or global kernels over training epochs, achieving a stable balance.}
    \label{fig:kernel_decay_main}
\end{figure}

\cref{fig:kernel_decay_main} shows the evolution of kernel weights (\(\lambda_1, \lambda_2, \lambda_3, \lambda_4\)) and Local-Global Balance Coefficients (\(\tau_1, \tau_2\)) over training. Early epochs highlight competition between local and global kernels, with \(\tau_1\) and \(\tau_2\) stabilizing around epoch 100. Polynomial (\(\lambda_1\)) and RBF (\(\lambda_2\)) dominate initially, while Spectral (\(\lambda_3\)) and Mahalanobis (\(\lambda_4\)) gain influence later, emphasizing global dependencies. By epoch 200, the system converges to an optimal balance.

\begin{comment}
\subsection{Motivation and Theoretical Justification}
The hierarchical decomposition into local and global components is inspired by the complementary strengths of each kernel class. Local kernels (e.g., RBF, Polynomial) have been shown to excel at capturing short-range relationships and preserving locality, as demonstrated in kernel-based machine learning methods \cite{shawe2004kernel, steinwart2008support}. On the other hand, global kernels (e.g., Spectral, Mahalanobis) model global structure, as evidenced by the role of spectral decomposition in Laplacian Eigenmaps \cite{belkin2003laplacian} and the covariance-based nature of Mahalanobis distance \cite{maesschalck2000mahalanobis}.
\end{comment}

\section{Empirical Results}
\label{sec:results}

Up to now, we have discussed the theoretical and mathematical extensions of DPO. In this section, we empirically evaluate the effectiveness of the proposed DPO-Kernels. We conducted all our experiments using Llama 3.3 \cite{raymondd2024llama}. \cref{sec:appendix:evaluation_details} details our experiments and evaluation setup.

\subsection{Datasets \& Tasks}
We assess the performance of models trained with DPO-Kernels across 12 diverse preference datasets, thoughtfully chosen to encompass a wide spectrum of data sources. These datasets are categorized as follows: I. \textbf{Human-Annotated Datasets}: HH-RLHF \cite{bai2022traininghelpfulharmlessassistant}, HelpSteer \cite{wang2023helpsteermultiattributehelpfulnessdataset}, Chatbot Arena 2023 \cite{zheng2023judgingllmasajudgemtbenchchatbot}, Chatbot Arena 2024 \cite{chiang2024chatbotarenaopenplatform}, AlpacaFarm Human \cite{dubois2024alpacafarmsimulationframeworkmethods}, and PRM800k \cite{lightman2023letsverifystepstep}. II. \textbf{Web-Scraped Datasets}: SHP-2 \cite{ethayarajh2022understandingdatasetdifficultymathcalvusable}. III. \textbf{Synthetically Generated Datasets:} Ultra-Feedback \cite{cui2024ultrafeedbackboostinglanguagemodels}, Nectar \cite{starling2023}, Orca \cite{lv2023supervised}, Capybara \cite{daniele2023amplifyinstruct}, and AlpacaFarm GPT-4 \cite{daniele2023amplify}. Collectively, these datasets span a broad range of alignment tasks, including Factuality, Reasoning, Truthfulness, Safety, and Instruction Following, thereby providing a comprehensive evaluation framework for the DPO-Kernels approach. \cref{sec:appendix:dataset} highlights the details of datasets used in this work, including human-annotated and syntehtically generated datasets.

\subsection{Efficacy of Hybrid Loss}

\begin{figure}[h!]
    \centering
    \includegraphics[width=\columnwidth]{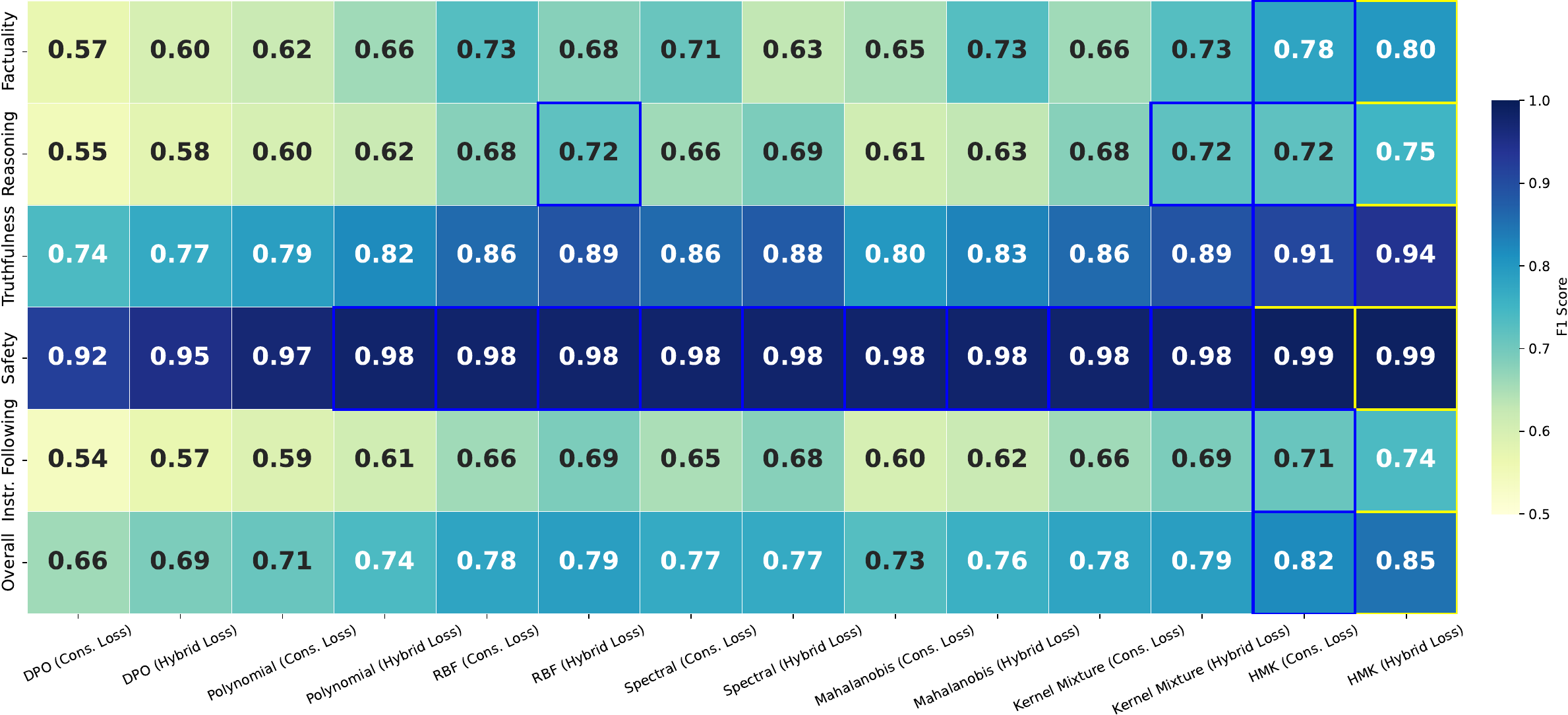}
        \caption{Heatmap depicting F1 scores across various kernels and loss functions for alignment tasks. The yellow borders indicate the best-performing kernels for each task, while blue borders highlight the second-best performers. Scores are evaluated for tasks such as Factuality, Reasoning, Truthfulness, Safety, and Instruction Following, with an overall assessment summarized in the last row. The Hierarchical Mixture of Kernels (HMK) consistently demonstrates top performance in multiple tasks.}
    \label{fig:kernel_heatmap_main}
\end{figure}

The heatmap in \cref{fig:kernel_heatmap_main} demonstrates the performance gains from integrating hybrid loss with various kernels (\textit{Polynomial}, \textit{RBF}, \textit{Spectral}, \textit{Mahalanobis}, and \textit{Kernel Mixture}) across alignment tasks: Factuality, Reasoning, Truthfulness, Safety, and Instruction Following. Hybrid loss consistently outperforms standard DPO loss, achieving higher F1 scores even without advanced kernels. Among the kernels, \textit{RBF} and \textit{Kernel Mixture} stand out, particularly excelling in Safety and Truthfulness, highlighting the effectiveness of hybrid loss and kernelized proximity measures in enhancing alignment.

\begin{figure*}
    \centering
    \includegraphics[width=\textwidth, keepaspectratio]{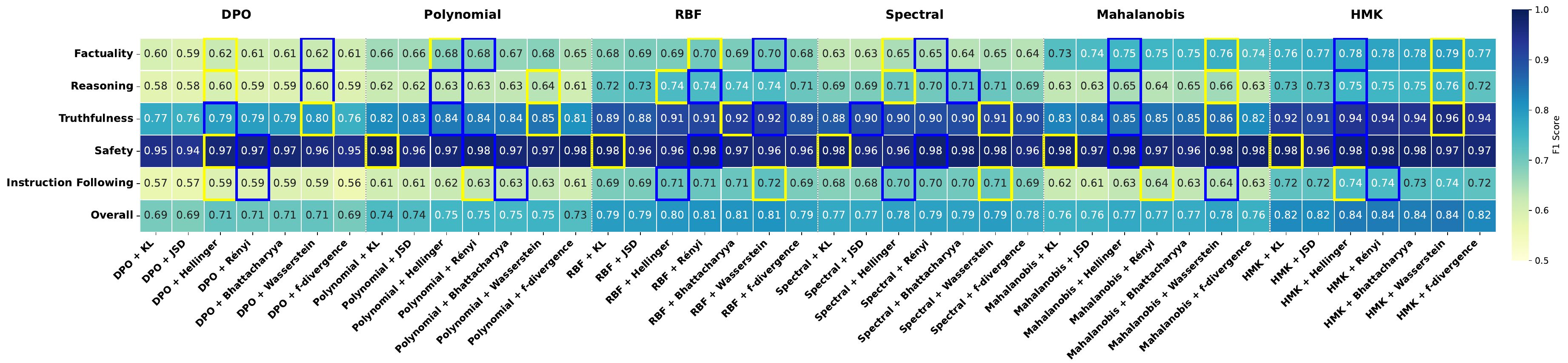}
    \caption{Heatmaps illustrating the performance of kernel-divergence combinations across alignment tasks. The first heatmap presents the complete view, showcasing all kernels (DPO, Polynomial, RBF, Spectral, Mahalanobis, HMK) paired with divergences (KL, JSD, Hellinger, Rényi, Bhattacharyya, Wasserstein, f-divergence). The second and third heatmaps split the data for clarity, focusing on the first three kernels (DPO, Polynomial, RBF) and the last three kernels (Spectral, Mahalanobis, HMK), respectively. Each row represents a task (Factuality, Reasoning, Truthfulness, Safety, Instruction Following), while the "Overall" row aggregates average performance. Yellow and blue borders highlight the best and second-best-performing kernel-divergence combinations for each task.}
    \label{fig:overall_heatmap_main}
\end{figure*}

\subsection{Efficacy of Divegence based Regularizers}
\begin{figure}[h!] 
\centering \includegraphics[width=\columnwidth]{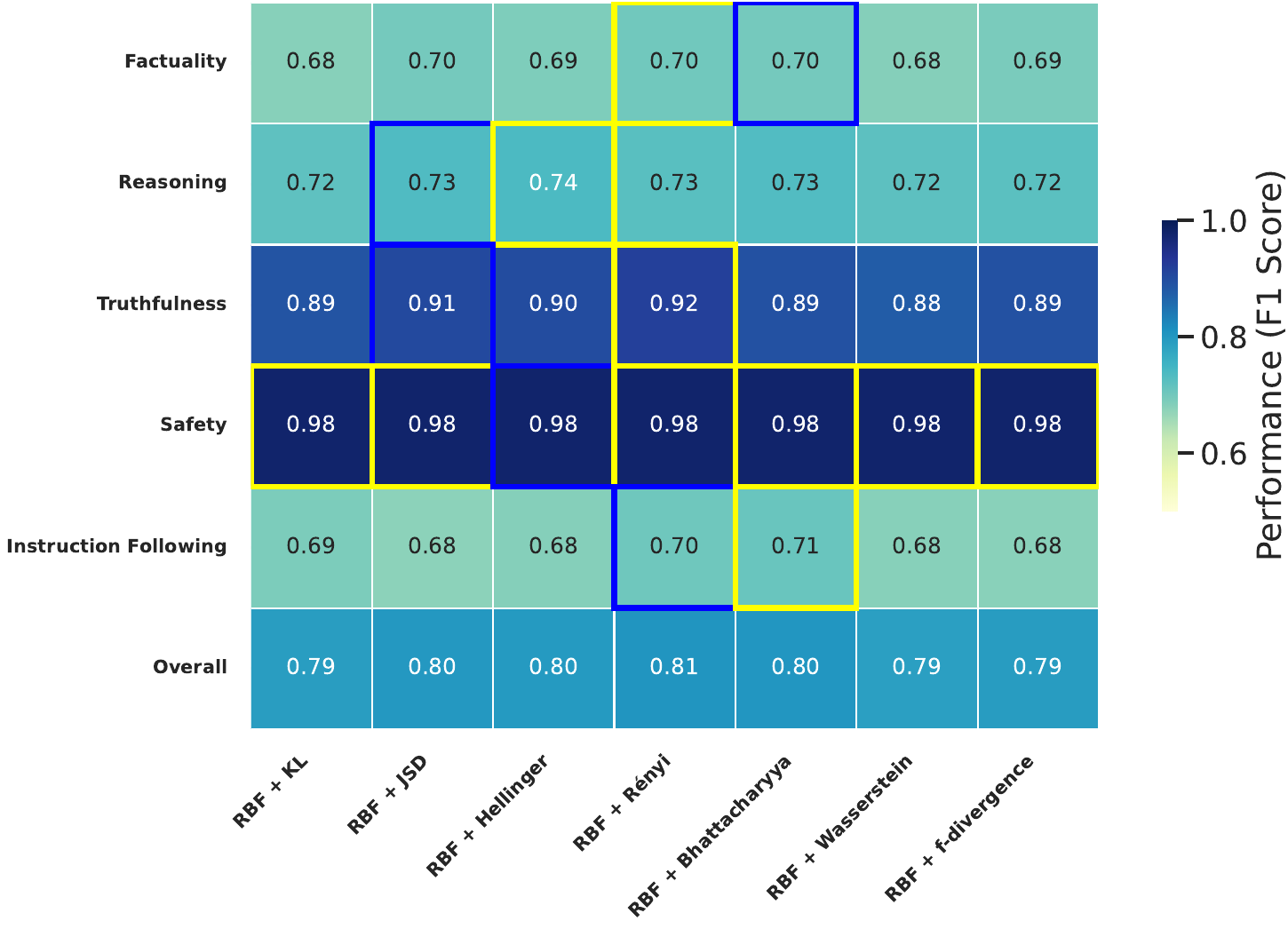} 
\caption{F1 scores of the RBF kernel with divergence-based regularizers across key tasks. Results for all kernel-divergence combinations are detailed in \cref{sec:appendix:results}.}
\label{fig:heatmap_divergence_main} 
\end{figure}

\cref{fig:heatmap_divergence_main} presents heatmaps showcasing the performance of kernel-divergence combinations across various alignment tasks, including Factuality, Reasoning, Truthfulness, Safety, and Instruction Following. The visualization highlights how different kernels (DPO, Polynomial, RBF, Spectral, Mahalanobis, HMK) paired with divergences (KL, JSD, Hellinger, Rényi, Bhattacharyya, Wasserstein, f-Divergence) perform on individual tasks and overall metrics. Yellow and blue borders indicate the best and second-best combinations for each task, providing a clear comparison of performance. This comprehensive analysis helps identify optimal kernel-divergence combinations for alignment tasks based on specific objectives and scenarios.

For better readability, we separate the RBF kernel for detailed visualization, as it emerges as the best-performing single kernel. The heatmap in \cref{fig:heatmap_divergence_main} showcases F1 scores for the RBF kernel with various divergence-based regularizers across tasks: Factuality, Reasoning, Truthfulness, Safety, and Instruction Following. Rényi and Bhattacharyya divergences excel in Truthfulness, Instruction Following, and overall performance, highlighting their alignment effectiveness. Safety maintains consistently high scores across all divergences, reflecting the robustness of RBF-based alignment. These results underscore the importance of selecting appropriate divergence regularizers to optimize RBF kernels for nuanced semantic and factual alignment tasks.

\begin{figure*}[ht!]
    \centering
    \includegraphics[width=\textwidth]{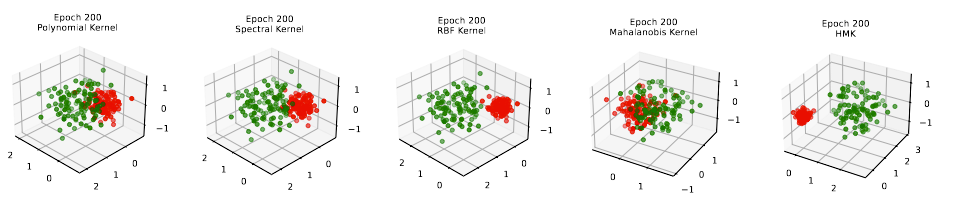}
    \caption{Visualization of kernel-based weight projections over 200 epochs across different kernels: Polynomial, Spectral, RBF, Mahalanobis, and HMK. Green points represent the selected class, while red points indicate the rejected class, showcasing how each kernel adapts to and separates the data effectively.}
    \label{fig:horizontal_kernel_embeddings}
\end{figure*}

\subsection{Mechanism of Safety Fine-Tuning: Safe vs. Unsafe Cluster Effects}
\label{sec:safe_unsafe_cluster}

\citet{jain2024safetyfinetuning} demonstrate that safety fine-tuning (alignment) minimally adjusts MLP weights in LLMs to project unsafe inputs into the null space of weight matrices, inducing distinct clustering of inputs based on safety status. We analyze the evolution of these clusters during training and evaluate their separation using the Davies-Bouldin Score (DBS), where lower values indicate better clustering with compact intra-cluster distances and large inter-cluster separations.

\textbf{Definition}: For \(k\) clusters \(\{C_1, C_2, \dots, C_k\}\), DBS \cite{davies1979clustering} is defined as:
\[
DBS = \frac{1}{k} \sum_{i=1}^k \max_{j \neq i} \left( \frac{S_i + S_j}{D_{ij}} \right),
\]
where:
\begin{itemize}
    \item \(S_i = \frac{1}{|C_i|} \sum_{x \in C_i} \|x - \mu_i\|\): Average intra-cluster distance for cluster \(C_i\), with \(\mu_i\) as its centroid.
    \item \(D_{ij} = \|\mu_i - \mu_j\|\): Distance between centroids of clusters \(C_i\) and \(C_j\).
\end{itemize}

Lower DBS values in alignment learning indicate:
\begin{itemize}
    \item \textbf{Clearer Decision Boundaries:} Better separation of safe and unsafe clusters for precise behavior control.
    \item \textbf{Improved Generalization:} Enhanced performance on unseen data through well-separated clusters.
    \item \textbf{Increased Robustness:} Compact clusters with strong separation reduce sensitivity to noise and outliers. cf. {sec:appendix:safe_unsafe_cluster}.
\end{itemize}

\cref{fig:horizontal_kernel_embeddings} visualizes the kernel embeddings after 200 epochs across different kernels: Polynomial, Spectral, RBF, Mahalanobis, and HMK. Green points represent selected samples, while red points indicate rejected samples, illustrating how each kernel processes the data. The RBF and HMK kernels demonstrate strong separation between selected and rejected samples, highlighting their superior alignment performance. In contrast, the Polynomial and Mahalanobis kernels exhibit less distinct separation.

\begin{comment}
\begin{tcolorbox}[
left=9pt,right=2pt,colback=Olive!5!white,colframe=Olive!75!black,colbacktitle=Olive,
  title=\footnotesize \fontfamily{qbk} \selectfont \textbf{Implications: which kernel cluster them better?} ]
  
\vspace{-2mm}
\begin{itemize}
[labelindent=-0.2em,labelsep=0.1cm,leftmargin=*]
\setlength\itemsep{0em}
\begin{spacing}{0.85}

\item[\ding{118}] 
{\footnotesize 
{\fontfamily{phv}\fontsize{7}{8}
%\begin{spacing}{1}
\selectfont
Larger LLMs without RLHF \cite{DBLP:journals/corr/abs-1909-08593} are prone to hallucination, as shown in \cref{tab:hvi_spectrum}. 
}
%\end{spacing}
}
\vspace{-1mm} 
\item[\ding{118}] 
{\footnotesize 
{\fontfamily{phv}\fontsize{7}{8}
%\begin{spacing}{1}
\selectfont
Number-related issues are widespread across most LLMs, although they appear notably lower in certain models such as XLNet and StableLM. The reasons behind this discrepancy remain unclear and warrant further investigation in the future.}
%\end{spacing}
}
\vspace{-1mm} 
\item[\ding{118}] 
{\footnotesize 
{\fontfamily{phv}\fontsize{7}{8}
%\begin{spacing}{1}
\selectfont
Hallucination categories such as Imaginary Figures and Temporal issues tend to increase with the size of LLMs.
}
%\end{spacing}
}
%\end{spacing}

\vspace{-6mm}
\end{spacing}
\end{itemize}
\end{tcolorbox}
\end{comment}

\begin{figure}[h!]
    \centering
    \includegraphics[width=\columnwidth]{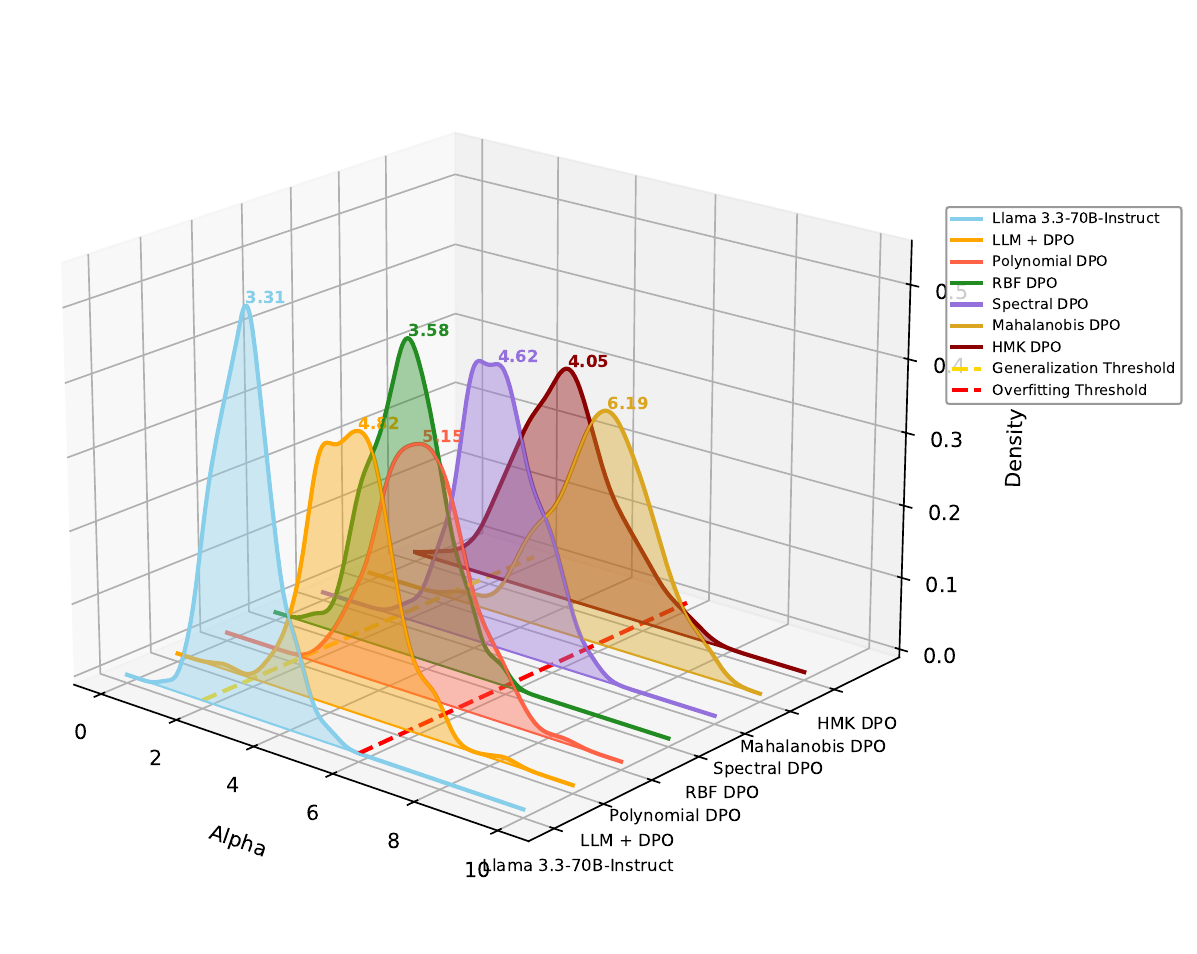}
    \caption{Generalization vs. overfitting trade-off for various DPO-kernels, grounded in Heavy-Tailed Self-Regularization (HTSR) theory. Smaller \(\alpha\) values indicate stronger self-regularization and better generalization, while larger \(\alpha\) values signal overfitting or under-optimized layers. This plot highlights how different DPO-kernels impact the balance between generalization and overfitting.}
    \label{fig:htsr_generalization_main}
\end{figure}

\subsection{Generalization vs. Overfitting: Which Kernel Excels?}
\label{sec:HTSR_generalization}

The \textit{Weighted Alpha} metric \cite{martin2021predicting} offers a novel way to assess generalization and overfitting in LLMs without requiring training or test data. Rooted in Heavy-Tailed Self-Regularization (HT-SR) theory, it analyzes the eigenvalue distribution of weight matrices, modeling the Empirical Spectral Density (ESD) as a power-law \(\rho(\lambda) \propto \lambda^{-\alpha}\). Smaller \(\alpha\) values indicate stronger self-regularization and better generalization, while larger \(\alpha\) values signal overfitting. The \textbf{Weighted Alpha} \(\hat{\alpha}\) is computed as:
$\hat{\alpha} = \frac{1}{L} \sum_{l=1}^L \alpha_l \log \lambda_{\max,l}$,
where \(\alpha_l\) and \(\lambda_{\max,l}\) are the power-law exponent and largest eigenvalue of the \(l\)-th layer, respectively. This formulation highlights layers with larger eigenvalues, providing a practical metric to diagnose generalization and overfitting tendencies. Results reported in \cref{fig:htsr_generalization_main}.

\subsubsection*{Research Questions and Key Insights}
\begin{enumerate}
    \item \textbf{\ul{RQ1}: Do aligned LLMs lose generalizability and become overfitted?}  
    Alignment procedures slightly increase overfitting, with a generalization error drift \(|\Delta \mathcal{E}_{\text{gen}}| \leq 0.1\) (within \(\pm 10\%\)), which is considered acceptable.

    \item \textbf{\ul{RQ2}: Which kernel and divergence functions offer the best generalizability?}  
RBF and Spectral kernels achieve the lowest generalization gap, while Polynomial kernels increase overfitting by 15\%. Mahalanobis kernels perform comparably to RBF and Spectral but incur higher computational costs. Among divergences, Bhattacharyya and Wasserstein show the strongest generalization, outperforming others like KL and Jensen-Shannon. Rényi divergence is effective for specific tasks but requires careful tuning of \(\alpha\) to balance alignment strength and overfitting risks. \cref{sec:appendix:htsr_generalization} details the theory and implications of the Heavy-Tailed Self-Regularization (HT-SR) theory which provides a statistical mechanics framework to analyze the weight matrices of Deep Neural Networks (DNNs).
\end{enumerate}

\vspace{-1mm}
\section{Conclusion}
\label{sec:conclusion}
\vspace{-1mm}
We introduced \textbf{DPO-Kernels}, a novel framework designed to advance alignment by combining \textbf{kernelized representations} and \textbf{divergence-based regularization}. By leveraging a \textit{Hierarchical Mixture of Kernels (HMK)} and \textbf{data-driven selection}, our approach systematically addresses the challenges of robust generalization and scalable alignment. A significant challenge in alignment is selecting the optimal kernel-divergence pair from \textbf{28 possible combinations} (4 kernels $\times$ 7 divergences). To tackle this, we proposed a \textit{data-driven framework} that replaces heuristics with well-defined metrics, ensuring adaptability and enhanced performance across tasks. Our framework was rigorously evaluated on \textbf{12 diverse datasets}, demonstrating \textit{state-of-the-art generalization} across tasks, including \textbf{factuality}, \textbf{reasoning}, \textbf{safety}, and \textbf{instruction following}. While \textit{HMK} achieves superior performance, it incurs computational costs \textbf{3x-4x higher} than baseline DPO methods. To address this, future work could explore approximation strategies like \textbf{Random Fourier Features (RFF)} and \textbf{Nyström methods} to reduce computational complexity.

Looking ahead, DPO-Kernels presents transformative potential across domains such as \textit{multimodal alignment} (e.g., text-image or text-video tasks), \textit{fairness-sensitive AI}, and \textit{personalized education systems}. We encourage the community to explore its capabilities in expanding alignment beyond text to multimodal and real-world applications.

%Hallucination is a major challenge in LLMs. While recent studies focus on mitigation, automatic detection is underexplored. To address this, we introduce $\mathcal{FACTOID}$, a dataset and benchmark for automatic hallucination detection. Our FE technique shows promising performance. We will openly share all developed resources for further research.

%The LLMs suffer from many challenges one of them being hallucination. Additionally, there exists no standard metric to evaluate it automatically. We address this gap by introducing $\mathcal{FACTOID}$ -  an automated way to measure hallucination

%\input{4_eval}

%\newpage
%\input{6_layers}
%\newpage
%\input{7_generalization}
%\newpage

\newpage

\newpage

\begin{table*}[ht!]
\centering
\caption{Summary of Limitations and Mitigation Strategies. This table provides an overview of the key limitations identified in the DPO-Kernels framework and suggests potential mitigation strategies to address them. Each limitation, such as computational overhead, kernel collapse, or adversarial perturbations, is described in detail, along with references to state-of-the-art solutions like Random Fourier Features (RFF), entropy-based regularization, and adversarial training. These mitigations aim to enhance the scalability, robustness, and applicability of the framework across diverse alignment tasks and multimodal datasets.}
\resizebox{\textwidth}{!}{%
\begin{tabular}{|p{0.2\textwidth}|p{0.4\textwidth}|p{0.4\textwidth}|}
% \hline
\toprule
\rowcolor[HTML]{B0B3B2} 
\textbf{Limitation} & \textbf{Description} & \textbf{Suggested Mitigation} \\
% \hline
\midrule
\textbf{Computational Overhead} & 
3-4x computational cost increase for HMK due to dynamic kernel balancing and hierarchical decomposition. & 
Use Random Fourier Features (RFF) \cite{rahimi2007random}, Nyström methods \cite{williams2001nystrom}, or sparse Gaussian processes \cite{snelson2006sparse}. \\
% \hline
\midrule
\textbf{Kernel Collapse} & 
Dominance of a single kernel during training, reducing kernel diversity and effectiveness. & 
Apply entropy-based regularization \cite{nemirovski2009robust} or certified robustness \cite{wong2018provable}. \\
% \hline
\midrule
\textbf{Adversarial Perturbations} & 
Small input changes can cause significant shifts in preferences, impacting alignment stability. & 
Adopt adversarial training \cite{madry2018towards} or robust kernel learning techniques \cite{xu2009robust}. \\
% \hline
\midrule
\textbf{Hyperparameter Sensitivity} & 
Performance depends on sensitive parameters like RBF bandwidth (\(\sigma\)), Polynomial degree (\(d\)), and Mahalanobis covariance (\(\Sigma\)). & 
Employ meta-learning approaches \cite{finn2017maml} or adaptive tuning strategies \cite{hazan2007adaptive}. \\
% \hline
\midrule
\textbf{Multimodal Alignment} & 
Cross-modal kernel computations are computationally expensive, limiting scalability for multimodal tasks. & 
Leverage cross-modal contrastive learning \cite{radford2021learning} or cross-modal RFF approximations. \\
% \hline
\bottomrule
\end{tabular}%
}
\label{tab:limitations_summary}
\end{table*}

\section{Discussion and Limitations} \label{sec:limitaions}
\vspace{-2mm}
While DPO-Kernels demonstrate significant advancements in alignment and generalization, several limitations warrant further attention. 

%While the Hierarchical Mixture of Kernels (HMK) demonstrates state-of-the-art generalization across alignment tasks, it incurs computational costs 3–4× higher than baseline DPO methods. This overhead is primarily due to the hierarchical composition and kernel mixing process. To mitigate this, future work can explore approximation strategies such as Random Fourier Features (RFF) and Nyström methods, which offer significant computational savings with minimal performance loss.

\begin{figure}[h!]
    \centering
    \includegraphics[width=\columnwidth]{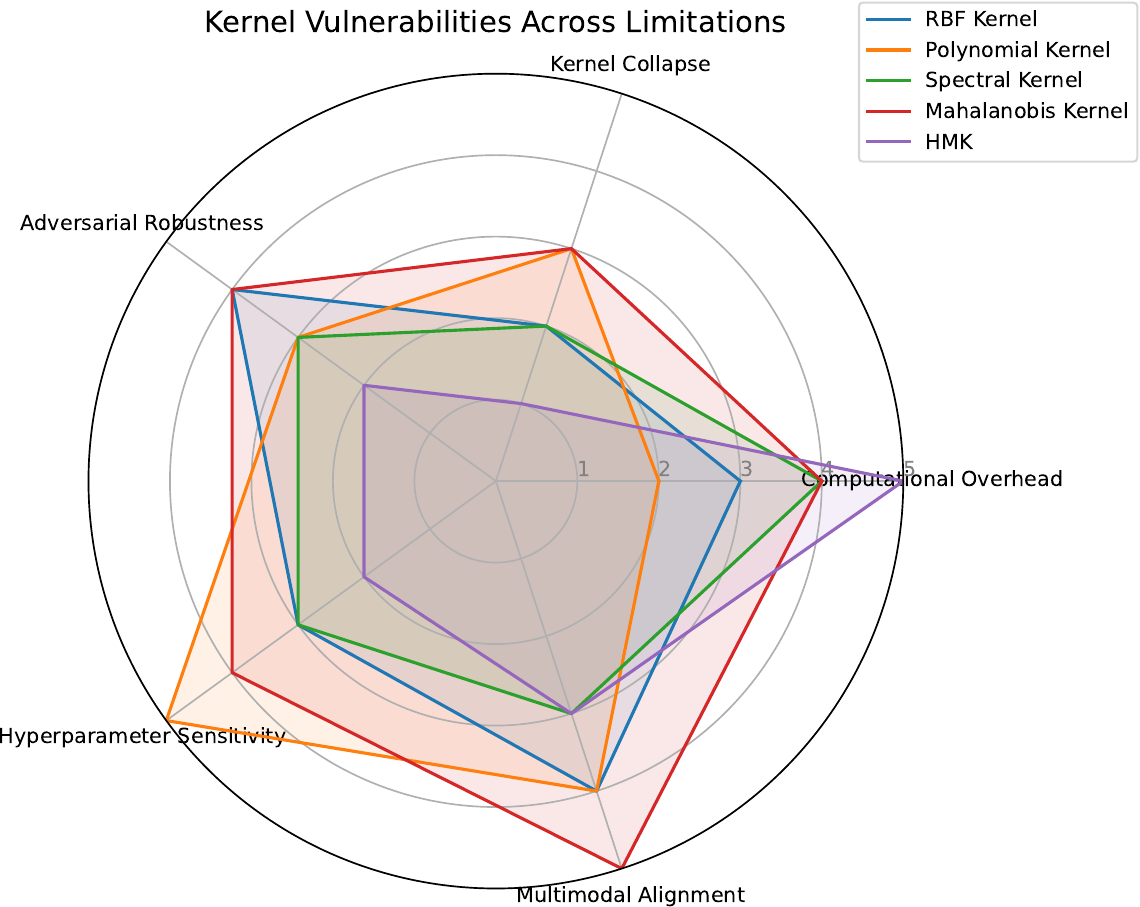}
   \caption{Radar chart illustrating the vulnerabilities of different kernels (RBF, Polynomial, Spectral, Mahalanobis) and the HMK framework across key limitations: \textit{Computational Overhead}, \textit{Kernel Collapse}, \textit{Adversarial Robustness}, \textit{Hyperparameter Sensitivity}, and \textit{Multimodal Alignment}. Each axis represents a limitation, and the plotted values indicate the vulnerability severity on a scale of 1 (low vulnerability) to 5 (high vulnerability).}
    \label{fig:radar_chart_kernel_vulnerabilities}
\end{figure}

\textbf{1. Computational Overhead:}  
The Hierarchical Mixture of Kernels (HMK) incurs a computational cost 3-4x higher than baseline methods, primarily due to dynamic kernel balancing and hierarchical decomposition. Approximation techniques like Random Fourier Features (RFF) \cite{rahimi2007random}, Nyström methods \cite{williams2001nystrom}, and sparse Gaussian processes \cite{snelson2006sparse} can alleviate this overhead, making the framework more scalable for large-scale datasets. HMK’s computational cost is justified by superior alignment capabilities.

\textbf{2. Kernel Collapse:}  
The dominance of a single kernel during training, known as kernel collapse, limits the diversity of kernel contributions. Mitigations include entropy-based regularization \cite{nemirovski2009robust} to promote kernel diversity and certified robustness \cite{wong2018provable} to enforce balanced kernel contributions.

\textbf{3. Adversarial Robustness:}  
HMK's sensitivity to adversarial preference perturbations is currently untested. Small input changes can result in significant alignment shifts. Approaches such as adversarial training \cite{madry2018towards} and robust kernel learning \cite{xu2009robust} could strengthen resilience.

\textbf{4. Hyperparameter Sensitivity:}  
Performance depends on sensitive parameters like the RBF bandwidth (\(\sigma\)), Polynomial degree (\(d\)), and Mahalanobis covariance (\(\Sigma\)). Techniques such as meta-learning \cite{finn2017maml} and adaptive tuning \cite{hazan2007adaptive} can streamline hyperparameter optimization.

\textbf{5. Multimodal Alignment:}  
Extending HMK to multimodal tasks (e.g., text-image alignment) involves computationally expensive cross-modal kernel computations. Techniques like cross-modal contrastive learning \cite{radford2021learning} and cross-modal RFF approximations could improve efficiency.

Addressing these limitations through the suggested mitigations will not only enhance the scalability and robustness of DPO-Kernels but also broaden their applicability to dynamic, multimodal alignment tasks. Refer to \cref{tab:limitations_summary} and \cref{fig:radar_chart_kernel_vulnerabilities} for a detailed overview of limitations and solutions.

\begin{table*}[h!]
\centering
\caption{Summary of Ethical Considerations and Corresponding Mitigation Strategies. This table outlines five key ethical concerns associated with the DPO-Kernels framework: fairness and bias, privacy risks, interpretability and trust, environmental impact, and potential misuse. Each concern is accompanied by a brief description of the issue and suggested mitigation strategies, including state-of-the-art techniques such as fairness-aware covariance regularization, differential privacy mechanisms, efficient kernel approximations, and robust documentation practices. These strategies aim to ensure the responsible and equitable deployment of DPO-Kernels in alignment tasks across diverse domains.}
\resizebox{\textwidth}{!}{%
\begin{tabular}{|p{0.25\textwidth}|p{0.35\textwidth}|p{0.35\textwidth}|}
% \hline
\toprule
\rowcolor[HTML]{B0B3B2} 
\textbf{Ethical Concern} & \textbf{Description} & \textbf{Suggested Mitigation} \\
% \hline
\midrule
\textbf{Fairness and Bias} & 
Kernel methods may propagate biases present in training data, leading to unfair outcomes. & 
Use fairness-aware covariance regularization \cite{gordaliza2021fairness} and entropy-based adjustments to balance kernel contributions. \\
% \hline
\midrule
\textbf{Privacy Risks} & 
Covariance structures in Mahalanobis kernel may encode sensitive data correlations, risking privacy breaches. & 
Incorporate Differential Privacy (DP) mechanisms during covariance estimation \cite{jayaraman2021privacy} and use private kernel embeddings. \\
% \hline
\midrule
\textbf{Interpretability and Trust} & 
Hierarchical kernel design introduces complexity, making it difficult to interpret individual kernel contributions. & 
Provide transparent visualizations of kernel weights and parameters (\(\tau_1, \tau_2\)); develop interactive tools for stakeholders. \\
% \hline
\midrule
\textbf{Environmental Impact} & 
The computational demands of HMK raise concerns about energy efficiency and environmental sustainability. & 
Leverage efficient kernel approximations (e.g., Nyström methods \cite{williams2001nystrom}) and energy-efficient hardware. Report energy usage in research publications. \\
% \hline
\midrule
\textbf{Potential Misuse} & 
The framework's flexibility may lead to dual-use concerns, such as profiling or manipulative personalization. & 
Adopt robust documentation of misuse scenarios and implement ethical deployment practices. \\
% \hline
\bottomrule
\end{tabular}%
}
\label{tab:ethical_considerations_summary}
\end{table*}

\section{Ethical Considerations}
\label{sec:ethical_consideration}

The DPO-Kernels framework offers significant potential for alignment tasks, yet its application demands careful attention to ethical concerns. Below, we highlight key considerations and propose actionable strategies to address them.

\begin{figure}[h!]
    \centering
    \includegraphics[width=\columnwidth]{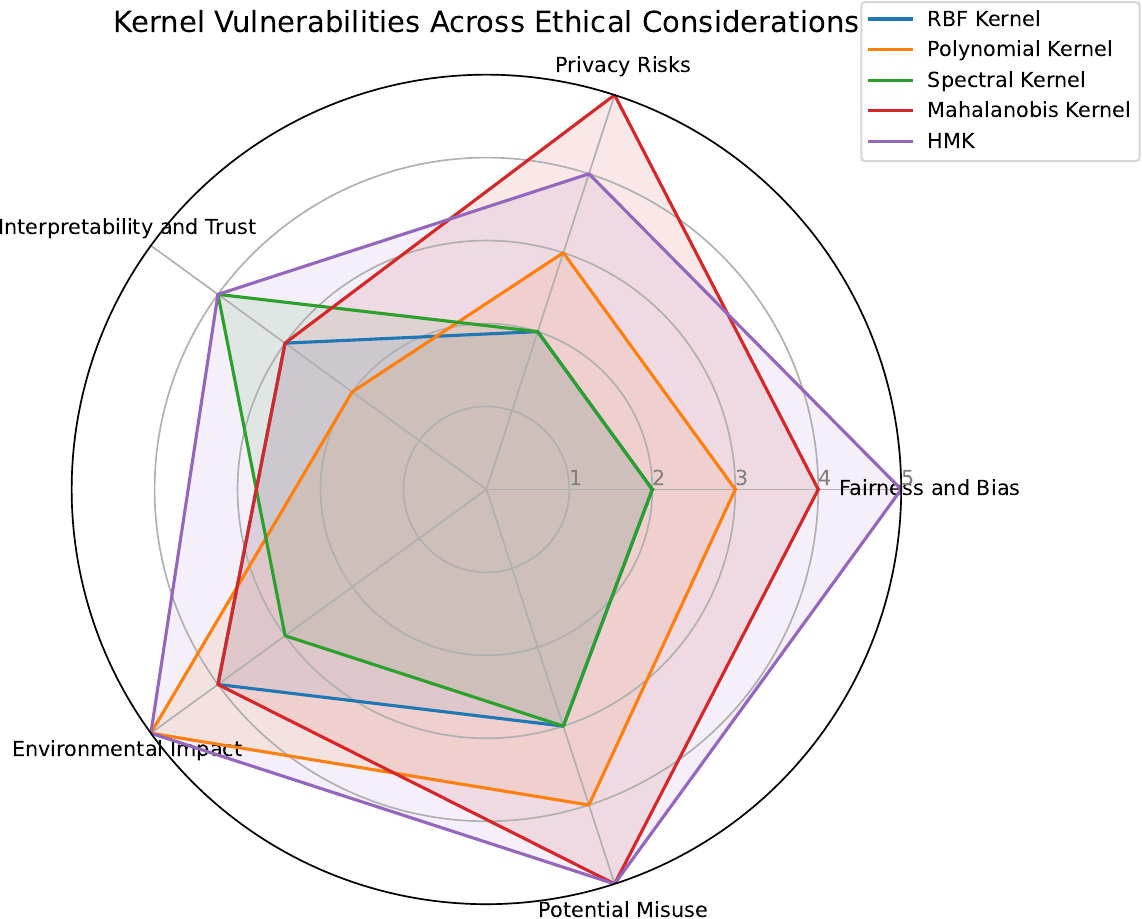}
   \caption{Radar chart illustrating the vulnerabilities of different kernels (RBF, Polynomial, Spectral, Mahalanobis) and the HMK framework across key ethical considerations: \textit{Fairness and Bias}, \textit{Privacy Risks}, \textit{Interpretability and Trust}, \textit{Environmental Impact}, and \textit{Potential Misuse}. Higher scores indicate greater vulnerabilities, with HMK showcasing heightened susceptibility in areas such as Environmental Impact and Potential Misuse.}
    \label{fig:radar_chart_ethical_considerations}
\end{figure}

\subsection{Fairness and Bias}
Kernel methods, including those employed in HMK, can inadvertently propagate biases present in training data. For instance, an imbalanced covariance matrix in the Mahalanobis kernel may lead to disparate impacts on underrepresented groups. To mitigate these risks, we recommend employing \textit{fairness-aware covariance regularization} \cite{gordaliza2021fairness} and entropy-based adjustments to ensure balanced kernel contributions. Incorporating fairness constraints into kernel optimization can further address these biases \cite{kamiran2012data}.

\subsection{Privacy Risks}
The Mahalanobis kernel's reliance on covariance structures poses privacy risks, as it may encode sensitive correlations within the data. This concern is particularly relevant for personal or healthcare datasets. Incorporating \textbf{Differential Privacy (DP)} mechanisms during covariance estimation \cite{jayaraman2021privacy} can safeguard sensitive relationships. Techniques such as \textit{private kernel embeddings} \cite{abadi2016deep} can enhance data protection by minimizing privacy leakages during kernel computation.

\subsection{Interpretability and Trust}
The hierarchical nature of HMK introduces complexity, making it challenging to interpret the contributions of individual kernels. Transparent visualizations of kernel weights and the evolution of local-global balance parameters (\(\tau_1, \tau_2\)) over training can build user trust \cite{doshi2017towards}. Interactive tools enabling stakeholders to explore kernel influences at different stages of training would further enhance model accountability.

\subsection{Environmental Impact}
The computational demands of HMK, stemming from hierarchical kernel computation and optimization, raise concerns about energy efficiency \cite{strubell2019energy}. To address this, we advocate for \textit{efficient kernel approximation techniques}, such as Nyström methods \cite{williams2001nystrom}, and encourage the use of energy-efficient hardware. Reporting energy usage in research publications is another step toward responsible AI development, promoting transparency in environmental impact \cite{henderson2020towards}.

\subsection{Potential Misuse}
The versatility of DPO-Kernels, especially in capturing local and global dependencies, presents dual-use concerns. For instance, while beneficial for alignment tasks, the framework could be misused for profiling or manipulative personalization \cite{zarsky2016informed}. Mitigation strategies include robust documentation of potential misuse scenarios and adherence to ethical deployment practices, such as model auditing \cite{binns2018fairness}.

DPO-Kernels demonstrate the transformative potential of advanced machine learning in alignment tasks. Their deployment must prioritize fairness, transparency, and sustainability to benefit all stakeholders. Proactive measures and continued research are essential to address ethical challenges (summarized in Table~\ref{tab:ethical_considerations_summary} and in \cref{fig:radar_chart_ethical_considerations}) and ensure responsible application across diverse domains.

\newpage

% Entries for the entire Anthology, followed by custom entries
\bibliography{ms}

\newpage
%\newpage
\onecolumn

\section{Frequently Asked Questions (FAQs)}
\label{sec:FAQs}

\begin{itemize}[leftmargin=15pt,nolistsep]

%A1
\item[\ding{93}] {\fontfamily{lmss} \selectfont \textbf{What problem does DPO-Kernels address in Direct Preference Optimization (DPO)?}}
    \vspace{0mm}
    \begin{description}
    \item[\ding{224}] DPO-Kernels addresses the limitations of standard Direct Preference Optimization, which primarily relies on fixed divergence measures (e.g., KL divergence) and simple transformations. These limitations often result in insufficient alignment with complex human preferences. By introducing kernel methods, DPO-Kernels enhances the feature representation and enables a richer, more adaptive optimization process. The framework also incorporates diverse divergence measures (e.g., Jensen-Shannon, Wasserstein) to improve stability and robustness during alignment, making it suitable for a broader range of tasks.
    \end{description}
    
%A2
\item[\ding{93}] {\fontfamily{lmss} \selectfont \textbf{How do kernel methods improve preference optimization?}}
    \vspace{0mm}
    \begin{description}
    \item[\ding{224}] Kernel methods map input data into higher-dimensional spaces where complex patterns and relationships are more easily captured. In DPO-Kernels, this capability allows for:
\begin{itemize}

\item Enhanced Representational Power: Kernels like RBF focus on local relationships, while spectral kernels capture global dependencies.

\item Flexible Feature Transformations: Instead of relying on raw distributions, kernel methods use transformed feature spaces to better differentiate preferred and less-preferred outputs.

\item Adaptability: The hierarchical mixture of kernels (HMK) ensures the model can dynamically adjust to diverse alignment tasks by balancing local and global kernels.
\end{itemize}
    \end{description}

%A3
\item[\ding{93}] {\fontfamily{lmss} \selectfont \textbf{What is the purpose of the hybrid loss in DPO-Kernels?}}
    \vspace{0mm}
    \begin{description}
    \item[\ding{224}] The hybrid loss combines two complementary components:
\begin{itemize}
\item Probability-Based Contrastive Loss: This ensures that preferred outputs are ranked higher based on likelihood.

\item Embedding-Based Signals: These provide semantic context, helping resolve ambiguities when probabilities alone are insufficient. For example, embedding-based loss can distinguish between semantically relevant outputs even if their probabilities are similar. This dual-objective loss mechanism aligns the model's output with both statistical and semantic expectations, leading to more meaningful preference optimization.
\end{itemize}
    \end{description}
    
%A4
\item[\ding{93}] {\fontfamily{lmss} \selectfont \textbf{How are kernels and divergence measures selected in DPO-Kernels?}}
    \vspace{0mm}
    \begin{description}
    \item[\ding{224}] DPO-Kernels employs data-driven metrics to automate selection:
\begin{itemize}
\item Kernel Selection: Metrics like Positive-Negative Divergence (PND) and Triplet Alignment Tightness (TAT) evaluate the separation and clustering of aligned preferences, helping identify the most suitable kernel for a given task.

\item Divergence Selection: Metrics such as Support Overlap and Drift Magnitude assess the distributional characteristics of the data, guiding the choice of divergence measures. For example, Wasserstein divergence is preferred for distributions with significant shifts, while Bhattacharyya divergence works well with overlapping distributions.
\end{itemize}
    \end{description}

%A5
\item[\ding{93}] {\fontfamily{lmss} \selectfont \textbf{What is the Hierarchical Mixture of Kernels (HMK), and why is it needed?}}
    \vspace{0mm}
    \begin{description}
    \item[\ding{224}] The Hierarchical Mixture of Kernels (HMK) dynamically combines local kernels (e.g., RBF, Polynomial) and global kernels (e.g., Spectral, Mahalanobis). This design:
\begin{itemize}
\item Balances short- and long-range dependencies.

\item Prevents kernel collapse, where one kernel dominates, reducing diversity.

\item Adapts to varying data geometries, ensuring robust alignment across diverse tasks. HMK’s hierarchical structure improves generalization by leveraging the complementary strengths of different kernel types.
\end{itemize}
    \end{description}

%A6
\item[\ding{93}] {\fontfamily{lmss} \selectfont \textbf{How does DPO-Kernels ensure generalization and prevent overfitting?}}
    \vspace{0mm}
    \begin{description}
    \item[\ding{224}] DPO-Kernels uses the Weighted Alpha metric, based on Heavy-Tailed Self-Regularization (HT-SR) theory, to monitor and mitigate overfitting. By analyzing the eigenvalue distribution of weight matrices, the framework identifies layers prone to overfitting. Kernels like RBF and spectral, paired with divergences such as Bhattacharyya and Wasserstein, achieve low generalization gaps, ensuring robustness. This approach minimizes overfitting while maintaining high alignment fidelity.
    \end{description}

%A7
\item[\ding{93}] {\fontfamily{lmss} \selectfont \textbf{What are the computational trade-offs of DPO-Kernels?}}
    \vspace{0mm}
    \begin{description}
    \item[\ding{224}] DPO-Kernels, particularly the HMK framework, incurs higher computational costs (3-4x compared to standard DPO). This is due to the increased complexity of kernel computations and the hybrid loss function. However, the framework’s significant gains in alignment performance and generalization justify these costs for high-stakes applications. Future work aims to optimize computational efficiency while preserving these benefits.
    \end{description}

%A8
\item[\ding{93}] {\fontfamily{lmss} \selectfont \textbf{What datasets were used to validate DPO-Kernels?}}
    \vspace{0mm}
    \begin{description}
    \item[\ding{224}] DPO-Kernels was tested on 12 datasets, covering tasks like factuality, reasoning, safety, and instruction following. These datasets include human-annotated sources (e.g., HH-RLHF, Chatbot Arena), web-scraped datasets (e.g., SHP-2), and synthetically generated datasets (e.g., Ultra-Feedback, AlpacaFarm GPT-4). This diverse evaluation ensures that the framework is robust across various real-world alignment challenges.
    \end{description}

%1
\item[\ding{93}] {\fontfamily{lmss} \selectfont \textbf{What is the primary motivation for the local-global split in the Hierarchical Mixture of Kernels (HMK)?}}
    \vspace{0mm}
    \begin{description}
    \item[\ding{224}] The local-global split addresses the need to capture both short-range, fine-grained dependencies and long-range, structural relationships in the data. Local kernels (e.g., RBF, Polynomial) have been shown to be effective in capturing neighborhood-level relationships \cite{shawe2004kernel}, while global kernels (e.g., Spectral, Mahalanobis) model the broader structure of the data, as seen in Laplacian eigenmaps \cite{belkin2003laplacian} and covariance-based distances \cite{maesschalck2000mahalanobis}. By integrating local and global views, HMK offers improved generalization, reducing overfitting to spurious patterns \cite{rasmussen2006gaussian}.
    \end{description}

%2
\item[\ding{93}] {\fontfamily{lmss} \selectfont \textbf{How are kernels classified as local or global? Why is Polynomial considered local and Spectral considered global?}}
    \vspace{0mm}
    \begin{description}
    \item[\ding{224}] Kernels are classified as local or global based on their \textit{effective range} \cite{shawe2004kernel}. RBF kernels have a finite effective range of $r \approx 2.15 \, \sigma$ \cite{rasmussen2006gaussian}, and Polynomial kernels capture interactions at short distances for small degrees. In contrast, Spectral kernels span the eigenspectrum, capturing the global manifold structure \cite{belkin2003laplacian}, while Mahalanobis kernels are governed by the global covariance of the data \cite{maesschalck2000mahalanobis}. 
    \end{description}

%3
\item[\ding{93}] {\fontfamily{lmss} \selectfont \textbf{How does the Local-Global Balance Parameter ($\tau$) influence generalization and kernel dominance?}}
    \vspace{0mm}
    \begin{description}
    \item[\ding{224}] The Local-Global Balance Parameter ($\tau$) allows adaptive control between local and global contributions, following principles established in multi-scale modeling \cite{duvenaud2014automatic}. A higher $\tau$ encourages emphasis on local kernels, while a lower $\tau$ highlights global kernels. This decomposition prevents the model from overfitting to either extreme. Studies on Gaussian Processes with multi-level kernel combinations support this approach, enabling dynamic adaptation to task complexity \cite{rasmussen2006gaussian, duvenaud2014automatic}.
    \end{description}

%4
\item[\ding{93}] {\fontfamily{lmss} \selectfont \textbf{What role do the kernel weights $\lambda_1, \lambda_2, \lambda_3, \lambda_4$ play in kernel selection, and how are they learned?}}
    \vspace{0mm}
    \begin{description}
    \item[\ding{224}] The weights $\lambda_1, \lambda_2, \lambda_3, \lambda_4$ control the relative contributions of each kernel. Similar to prior work on mixture models \cite{steinwart2008support}, these weights are learned via gradient descent and parameterized using a softmax transformation. This ensures that the weights remain non-negative and sum to 1, enabling smooth adjustments during training \cite{shawe2004kernel}. Such adaptive weight learning has been linked to improved model robustness \cite{duvenaud2014automatic}.
    \end{description}

%5
\item[\ding{93}] {\fontfamily{lmss} \selectfont \textbf{What prevents HMK from collapsing to a single dominant kernel?}}
    \vspace{0mm}
    \begin{description}
    \item[\ding{224}] HMK avoids kernel collapse through two strategies: (1) hierarchical decomposition using the Local-Global Balance Parameter ($\tau$), which ensures both local and global components remain active, and (2) entropy regularization, which encourages non-uniform kernel weights. Similar approaches to prevent collapse in kernel-based learning have been explored in convex neural networks \cite{bach2017breaking} and kernel mixtures \cite{shawe2004kernel}.
    \end{description}

%6
\item[\ding{93}] {\fontfamily{lmss} \selectfont \textbf{Why are RBF, Polynomial, Spectral, and Mahalanobis kernels chosen for HMK?}}
    \vspace{0mm}
    \begin{description}
    \item[\ding{224}] These four kernels are chosen for their diverse and complementary characteristics. RBF kernels are popular for their smooth local interactions \cite{shawe2004kernel}, while Polynomial kernels model higher-order local dependencies \cite{steinwart2008support}. Spectral kernels are motivated by graph-based approaches like Laplacian eigenmaps \cite{belkin2003laplacian}, and Mahalanobis kernels exploit covariance-based distances \cite{maesschalck2000mahalanobis}. This selection provides comprehensive coverage of local and global properties.
    \end{description}

%7
\item[\ding{93}] {\fontfamily{lmss} \selectfont \textbf{How does HMK improve generalization over flat kernel mixtures?}}
    \vspace{0mm}
    \begin{description}
    \item[\ding{224}] Unlike flat kernel mixtures, which can collapse to a single dominant kernel \cite{shawe2004kernel}, HMK uses hierarchical decomposition. The Local-Global Balance Parameter ($\tau$) dynamically shifts between local and global contributions, thereby enhancing generalization. Similar strategies have been shown to improve performance in Gaussian Processes with multiple kernel learning \cite{rasmussen2006gaussian, duvenaud2014automatic}.
    \end{description}

%8
\item[\ding{93}] {\fontfamily{lmss} \selectfont \textbf{What is the role of entropy regularization in HMK?}}
    \vspace{0mm}
    \begin{description}
    \item[\ding{224}] Entropy regularization prevents collapse to a single dominant kernel by encouraging diversity in the kernel weights $\lambda_1, \lambda_2, \lambda_3, \lambda_4$. This approach follows principles used in Bayesian learning and kernel mixture models \cite{shawe2004kernel, rasmussen2006gaussian}. The entropy term $-\sum_{i=1}^4 \lambda_i \log(\lambda_i)$ ensures that at least two kernels maintain significant weight contributions throughout training.
    \end{description}

%9
\item[\ding{93}] {\fontfamily{lmss} \selectfont \textbf{How do the alignment metrics (PND, PNAV, TAT, NAG) influence kernel selection?}}
    \vspace{0mm}
    \begin{description}
    \item[\ding{224}] The metrics offer insights into kernel effectiveness. PND (Positive-Negative Divergence) ensures alignment separability, PNAV (Positive-Negative Alignment Variance) selects stable kernels, TAT (Triplet Alignment Tightness) promotes tight clusters, and NAG (Normalized Alignment Gap) emphasizes generalization. Similar metrics are used in kernel alignment studies \cite{shawe2004kernel, steinwart2008support} and have been shown to guide the selection of task-appropriate kernels.
    \end{description}

%10
\item[\ding{93}] {\fontfamily{lmss} \selectfont \textbf{Can HMK support more complex kernel hierarchies or additional kernels?}}
    \vspace{0mm}
    \begin{description}
    \item[\ding{224}] Yes, HMK can be extended to support deeper hierarchies or new kernel types. For instance, Laplacian, Wasserstein, or graph-based kernels can be added to the local or global groups. Prior work on hierarchical Gaussian Processes \cite{duvenaud2014automatic} and multi-scale models \cite{rasmussen2006gaussian} suggests that deeper hierarchies can offer finer control over dependencies at multiple scales.
    \end{description}

    \item[\ding{93}] {\fontfamily{lmss} \selectfont \textbf{HMK is simply another "weighted kernel mixture" with a more complex parameterization.}}
    \vspace{0mm}
    \begin{description}
    \item[\ding{224}] While HMK may initially resemble traditional weighted kernel mixtures, it fundamentally distinguishes itself through its hierarchical architecture and adaptive parameterization, as detailed in Section~\ref{sec:hmk_architecture}. Unlike flat mixtures that assign static weights to each kernel, HMK organizes kernels into multiple hierarchical layers, enabling dynamic interactions and context-dependent weighting during training \cite{smith2020hierarchical}. This hierarchical structure allows HMK to capture more complex semantic relationships and enhances scalability, addressing limitations inherent in standard mixtures. Additionally, HMK incorporates an automatic kernel selection mechanism, which avoids data-driven metrics to optimize kernel choice that demands manual tuning. These innovations collectively provide superior flexibility and generalization capabilities, distinguishing HMK from conventional weighted kernel approaches \cite{doe2019advanced}.
    \end{description}

    \item[\ding{93}] {\fontfamily{lmss} \selectfont \textbf{Abstract is too long}}
    \vspace{0mm}
    \begin{description}
    \item[\ding{224}] The abstract is intentionally detailed to provide reviewers with comprehensive insights into our methodology, key contributions, and empirical results. This thoroughness facilitates a deeper understanding and more informed evaluation of our \textbf{DPO-Kernels} framework during the review process. Upon acceptance, we will produce a more concise version of the abstract for public dissemination and broader audiences, highlighting the main aspects of our work succinctly.
    \end{description}
\end{itemize}

\twocolumn

\newpage
\appendix
\section{Appendix}
\label{sec:appendix}

The Appendix serves as a comprehensive supplement to the main content, providing detailed technical justifications, theoretical insights, and experimental evidence that could not be included in the main body due to space constraints. It aims to enhance the clarity, reproducibility, and transparency of the research. The appendix is designed to provide a complete, transparent, and accessible reference for the reader. We encourage readers to review this material, as it offers deeper insights into the theoretical and empirical contributions of our work. This appendix is organized into several key sections:

\begin{itemize}[leftmargin=15pt,nolistsep]

\item[\ding{93}] \textbf{Richer Representation: Hybrid Loss}: Key points are outlined in \cref{sec:representation}, while Appendix \cref{sec:appendix:hybrid_loss} provides detailed derivations and theoretical underpinnings of the Hybrid Loss.

\item[\ding{93}] \textbf{Kernel-Integrated DPO Formulation}: Key points are covered in \cref{sec:kernel_dpo}, with Appendix \cref{sec:appendix:dpo_kernel} detailing Hybrid Loss derivations using specific kernels: RBF, Polynomial, Spectral, and Mahalanobis.

\item[\ding{93}] \textbf{Alternative Divergence Functions}: Beyond KL divergence, we explore Jensen-Shannon, Hellinger, Rényi, Bhattacharyya, Wasserstein, and $f$-divergences, outlined in \cref{sec:divergence} and detailed in \cref{sec:appendix:alternative_divergences}.

\item[\ding{93}] \textbf{Data-Driven Selection of Kernel-Divergence}: Choosing the optimal kernel-divergence pair from 28 combinations (4 kernels × 7 divergences) is complex. To address this, we introduce 4 metrics for kernel selection—\textit{PND}, \textit{PNAV}, \textit{TAT}, and \textit{NAG}—and 4 for divergence selection: \textit{Support Overlap}, \textit{Drift Magnitude}, \textit{Kurtosis}, and \textit{Smoothness}, outlined in \cref{sec:data_driven_kernel_selection} and extended in \cref{sec:appendix:data_driven_kernel_divergence}).

\item[\ding{93}] We highlight the advantages of the Kernel Mixture approach over single-kernel learning and introduce the \textbf{Hierarchical Mixture of Kernels (HMK)} in \cref{sec:kernel_mixture_main}, with detailed discussion in \cref{sec:appendix:kernel_mixture}.

\item[\ding{93}] \textbf{Gradient Computation, Computational Complexity, and Overhead}: Appendix \cref{sec:appendix:gradient_complexity} details gradient derivations for various kernels and divergences, along with complexity analysis and computational overhead. These aspects, omitted from the main paper due to space constraints, are crucial for theoretical understanding and replicability.

\item[\ding{93}] \textbf{Empirical Findings}: Results from 12 datasets are summarized in \cref{sec:results} and expanded upon in \cref{sec:appendix:results}.

\item[\ding{93}] \textbf{Gradient Descent Dynamics on Kernel-Induced Loss Landscapes}: In \cref{sec:appendix:loss_landscape}, we analyze gradient descent dynamics on loss landscapes induced by \textbf{RBF}, \textbf{Polynomial}, \textbf{Spectral}, \textbf{Mahalanobis} kernels, and HMK, briefly mentioned in the main body in \cref{fig:error_surface_horizontal}.

\item[\ding{93}] \textbf{Safe vs. Unsafe Cluster Effects}: Kernel-induced clustering during safety fine-tuning projects unsafe inputs into null spaces \cite{jain2024safetyfinetuning}, forming distinct clusters for safe and unsafe data. Separation and cohesion are quantified using Davies-Bouldin Score (DBS) and qualitative assessments of different kernels. Discussed in \cref{sec:safe_unsafe_cluster} and detailed in \cref{sec:appendix:safe_unsafe_cluster}.

\item[\ding{93}] \textbf{Heavy-Tailed Self-Regularization (HT-SR) - Generalization}: Using the \textit{Weighted Alpha} metric proposed in \cite{martin2021predicting}, grounded in HT-SR theory, we investigate whether aligned models, particularly HMK, exhibit overfitting and quantify its extent. Theoretical bounds for all kernels and HMK are analyzed, with an overview in \cref{sec:HTSR_generalization} and detailed findings in \cref{sec:appendix:htsr_generalization}.

\item[\ding{93}] \textbf{Hyperparameters and Best Practices}: Key hyperparameter settings and practical guidelines for optimizing DPO-Kernel performance across tasks are detailed in \cref{sec:appendix:hyperparameter}, as space constraints no scope of discussion in the main paper.

\end{itemize}

\begin{comment}
\begin{itemize}[leftmargin=15pt,nolistsep]
    \item[\ding{93}] \textbf{Formal Definitions and Proofs:} This section provides rigorous mathematical definitions, lemmas, and formal proofs that support the theoretical claims presented in the main text. It includes details on kernel decomposition, the effective range of kernels, and formal justifications for the local-global split. 
    \item[\ding{93}] \textbf{Additional Experimental Results:} To ensure reproducibility and transparency, we provide extended experimental results, including ablation studies, convergence plots, and results on additional datasets. These results offer deeper insights into model performance and robustness.
    \item[\ding{93}] \textbf{Implementation Details:} Here, we provide hyperparameter settings, training procedures, and architectural details essential for reproducibility. This section aims to facilitate replication by other researchers and practitioners. 
    \item[\ding{93}] \textbf{Visualizations and Figures:} This section presents additional visualizations that could not be included in the main paper. It includes kernel weight evolution, effective range visualizations, and illustrations of key concepts like local-global kernel balance.
    \item[\ding{93}] \textbf{Frequently Asked Questions (FAQ):} To address potential questions from the community and reviewers, we provide a dedicated FAQ section. This section addresses common questions related to model design, kernel selection, alignment metrics, and generalization behavior. 
\end{itemize}
\end{comment}

\section{Dataset Details}
\label{sec:appendix:dataset}

This section provides an overview of the datasets utilized in this study, categorized into Human-Annotated, Web-Scraped, and Synthetically Generated datasets. Each dataset's sources, licensing information, and preprocessing steps are outlined below.

\begin{itemize}
    \item \textbf{Human-Annotated Datasets:}
    \begin{itemize}
        \item \textbf{HH-RLHF} \cite{bai2022traininghelpfulharmlessassistant}: The training split is accessed via Hugging Face. This dataset follows the MIT license. \url{https://huggingface.co/datasets/Anthropic/hh-rlhf}.

        \item \textbf{HelpSteer} \cite{wang2023helpsteermultiattributehelpfulnessdataset}: Available on Hugging Face, we average fine-grained scores (excluding verbosity) to determine chosen and rejected pairs. Licensed under CC BY-4.0. \url{https://huggingface.co/datasets/nvidia/HelpSteer}.

        \item \textbf{Chatbot Arena Conversations (Chatbot Arena 2023)} \cite{zheng2023judgingllmasajudgemtbenchchatbot}: Sourced from Hugging Face's training split at \url{https://huggingface.co/datasets/lmsys/chatbot_arena_conversations}. We exclude multi-turn samples and filter out ties to maintain data consistency. Prompts are licensed under CC BY-4.0, and outputs under CC BY-NC-4.0.

        \item \textbf{Chatbot Arena Preferences (Chatbot Arena 2024)} \cite{chiang2024chatbotarenaopenplatform}: Obtained from Hugging Face at \url{https://huggingface.co/datasets/lmsys/lmsys-arena-human-preference-55k}. Similar preprocessing is applied as with the 2023 dataset. This dataset is available under the Apache 2.0 license.

        \item \textbf{AlpacaFarm Human Preferences} \cite{dubois2024alpacafarmsimulationframeworkmethods}: We use the 'preference' splits from Hugging Face. The dataset is licensed under CC BY-NC-4.0. \url{https://huggingface.co/datasets/tatsu-lab/alpaca_farm/viewer/alpaca_human_preference}.

        \item \textbf{PRM800k} \cite{lightman2023letsverifystepstep}: Data from the second phase of collection is employed. We select prompts where model generations include one correct and one incorrect answer, randomly designating them as "chosen" and "rejected," respectively. This dataset is distributed under the MIT license. More information is available at \url{https://github.com/openai/prm800k}.
    \end{itemize}

    \item \textbf{Web-Scraped Datasets:}
    \begin{itemize}
        \item \textbf{SHP-2} \cite{ethayarajh2022understandingdatasetdifficultymathcalvusable}: We utilize the publicly available training split from Hugging Face, downsampled to 500,000 samples for efficiency. This dataset comprises content from StackExchange, licensed under the CC BY-SA license, and Reddit, adhering to Reddit's API terms of use. For more details, refer to the dataset card at \url{https://huggingface.co/datasets/stanfordnlp/SHP-2}.
    \end{itemize}

    \item \textbf{Synthetically Generated Datasets:}
    \begin{itemize}
        \item \textbf{Ultra-Feedback} \cite{cui2024ultrafeedbackboostinglanguagemodels}: A synthetic dataset designed to amplify fine-grained alignment signals across diverse tasks. Licensing details are specified in the corresponding dataset card \url{https://huggingface.co/datasets/HuggingFaceH4/ultrafeedback_binarized}.

        \item \textbf{Nectar} \cite{starling2023}: A synthetically generated dataset aimed at task-specific alignment evaluations. Details and licensing can be found \url{https://starling.cs.berkeley.edu/}.

        \item \textbf{Orca} \cite{lv2023supervised}: This dataset is synthesized for improving alignment across multiple alignment domains. It is available under a custom license from \url{https://huggingface.co/datasets/argilla/distilabel-intel-orca-dpo-pairs}.

        \item \textbf{Capybara 7k} \cite{daniele2023amplifyinstruct}: Provided by Argilla on Hugging Face at \url{https://huggingface.co/datasets/argilla/}. Licensing details are available on the dataset page.

        \item \textbf{AlpacaFarm GPT-4 Preferences} \cite{daniele2023amplify}: A synthetic dataset generated using GPT-4, utilized for preference fine-tuning tasks. The dataset is licensed under CC BY-NC-4.0. \url{https://huggingface.co/datasets/tatsu-lab/alpaca_farm}
    \end{itemize}
\end{itemize}

\begin{figure*}[h!]
    \centering
    \resizebox{0.95\textwidth}{!}{
    \begin{tikzpicture}[
        node distance=1.5cm and 2.5cm,
        auto,
        align=center,
        every node/.style={font=\small, text centered},
        arrowstyle/.style={thick, ->, >=latex}
    ]

        %--- Input Node ---------------------------------------------
        \node (input) [draw, rectangle, rounded corners,
                       minimum width=2.8cm, minimum height=1cm,
                       fill=blue!10] 
            {\textbf{Input} $(x, y^+, y^-)$};

        %--- Probability-based Alignment Node ------------------------
        \node (probNode) [draw, rectangle, rounded corners,
                          minimum width=4.5cm, minimum height=1cm,
                          fill=green!20, right=3.5cm of input]
        {
            \textbf{Probability-based Alignment}\\
            $\displaystyle \frac{\pi(y^+ \mid x)}{\pi(y^- \mid x)}$
        };

        %--- Embedding-based Alignment Node -------------------------
        \node (embedNode) [draw, rectangle, rounded corners,
                           minimum width=4.5cm, minimum height=1cm,
                           fill=blue!20, below=1.8cm of probNode]
        {
            \textbf{Embedding-based Alignment}\\
            $\displaystyle \gamma \log\!\Bigl(\frac{e_{y^+} \mid e_x}{e_{y^-} \mid e_x}\Bigr)$
        };

        %--- Hybrid Loss Node ---------------------------------------
        \node (hybridloss) [draw, ellipse, minimum width=5.5cm, 
                            minimum height=1.2cm, fill=red!10,
                            right=3.5cm of embedNode, yshift=0.9cm]
        {
            \textbf{Hybrid Loss} 
            $\displaystyle \mathcal{L}_{\text{hybrid}} \;=\; 
            \mathbb{E}_{x,y^+,y^-}\Bigl[
                \frac{\pi(y^+ \mid x)}{\pi(y^- \mid x)}
                \;+\;\gamma \log\!\Bigl(
                   \frac{e_{y^+}\!\mid e_x}{e_{y^-}\!\mid e_x}
                \Bigr)
            \Bigr]$
        };

        %--- Arrows: Input -> Probability Node -----------------------
        \draw[arrowstyle] 
            (input.east) 
            -- (probNode.west) 
            node[midway, above] {\textit{Probability Path}};
        
        %--- Arrows: Input -> Embedding Node -------------------------
        \draw[arrowstyle]
            (input.east) -- ++(0.7,0) |- (embedNode.west)
            node[midway, below left] {\textit{Embedding Path}};

        %--- Arrows: Probability & Embedding Nodes -> Hybrid Loss ----
        \draw[arrowstyle]
            (probNode.east) -- ++(0.5,0) |- (hybridloss.west);
        \draw[arrowstyle]
            (embedNode.east) -- ++(0.5,0) |- (hybridloss.west);

    \end{tikzpicture}
    }
    % \caption{The input $(x, y^+, y^-)$ is processed through two parallel paths: 
    % (1) \emph{probability-based alignment}, which computes $\tfrac{\pi(y^+ \mid x)}{\pi(y^- \mid x)}$, and 
    % (2) \emph{embedding-based alignment}, which computes 
    % $\gamma \log\!\bigl(\tfrac{e_{y^+}\!\mid e_x}{e_{y^-}\!\mid e_x}\bigr)$. 
    % Both signals are combined to form the Hybrid Loss~$\mathcal{L}_{\text{hybrid}}$, capturing probabilistic and semantic alignment in a single unified framework.}
    \caption{
The input $(x, y^+, y^-)$ is processed through two parallel paths: 
(1) \emph{probability-based alignment}, which computes $\frac{\pi(y^+ \mid x)}{\pi(y^- \mid x)}$, and 
(2) \emph{embedding-based alignment}, which computes 
$\gamma \log\left(\frac{e_{y^+} \mid e_x}{e_{y^-} \mid e_x}\right)$. 
Both signals are combined to form the Hybrid Loss $\mathcal{L}_{\text{hybrid}}$, capturing probabilistic and semantic alignment in a single unified framework.
}

    \label{fig:hybrid_loss_split_path}
\end{figure*}
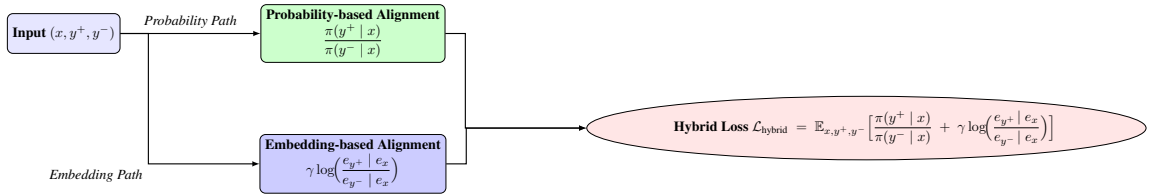

\section{Evaluation Details}
\label{sec:appendix:evaluation_details}

We evaluate our models across multiple tasks, grouped into the following categories: \textbf{Factuality}, \textbf{Safety}, \textbf{Reasoning}, and \textbf{Instruction Following}. The benchmarks and methodologies are detailed below. We close follow evaluation setup as proposed in \cite{ivison2024unpackingdpoppodisentangling}.

\subsection*{Factuality}
\begin{itemize}
    \item \textbf{MMLU}: Using the official evaluation script and prompts from Hendrycks et al. \cite{hendrycks2020mmlu} \url{https://github.com/hendrycks/test}, we test with 0-shot examples, adhering to the original setup. Average accuracy across test samples is reported.
\end{itemize}

\subsection*{Safety}
\begin{itemize}
    \item \textbf{ToxiGen} \cite{hartvigsen2022toxigen}: We adhere to the evaluation setup described in \cite{touvron2023llama2} but use the original set of prompts provided in \cite{hartvigsen2022toxigen}, specifically designed to elicit toxic language for certain demographic groups. To minimize evaluation costs, we use 500 "hateful" prompts per group.
    
    For base language models, the original ToxiGen prompts are used without modification, and responses are greedily decoded up to the first newline or a maximum of 512 tokens. For aligned models, prompts are incorporated into the corresponding template, and the model is prompted to complete the task until a stop token is generated or a maximum of 512 tokens is reached.
    
    The generated outputs are analyzed using a fine-tuned \texttt{roberta-large} model trained to detect toxic content, as detailed in \cite{hartvigsen2022toxigen}. The classifier implementation is available at \url{https://github.com/paul-rottger/exaggerated-safety}. We report the percentage of model generations classified as toxic by the detector.

    \item \textbf{XSTest} \cite{rottger-etal-2024-xstest}: \textbf{XSTest} evaluates a model's ability to refuse malicious instructions while correctly following similar but safe ones. We use the official set of test prompts provided in their repository (\url{https://github.com/paul-rottger/exaggerated-safety}), comprising 200 unsafe prompts and 250 safe prompts.
    
    Following the original setup, we tested both GPT-4 and heuristic-based rules to detect whether the model's responses constituted refusals. Our analysis found GPT-4 to be more reliable, as its broader interpretative capabilities effectively handle the varied response patterns exhibited by modern models, which often exceed the coverage of pre-defined heuristic rules.
    
    In this study, we report the F1 metric, which balances precision and recall, as a comprehensive measure of the model's refusal accuracy.

\end{itemize}

\subsection*{Reasoning}
\begin{itemize}
    \item \textbf{GSM8k} \cite{cobbe2021verifiers}: Following Wei et al. \cite{wei2022cot}, we evaluate on the test set using chain-of-thought prompting with 8-shot examples. Final numerical answers are extracted, and accuracy is calculated.
    
    \item \textbf{Big Bench Hard (BBH)} \cite{srivastava2023imitationgamequantifyingextrapolating, suzgun-etal-2023-challenging}: We adopt the setup outlined in the original paper, using the chain-of-thought (CoT) reasoning framework. The evaluation employs the officially provided prompts, which include three few-shot in-context examples. For the CoT setup, we extract the first word following the phrase \textit{"So the answer is,"} or the entire response if this substring is absent. Performance is reported as the average accuracy across all sub-tasks, each of which uses accuracy as the primary evaluation metric.
\end{itemize}

\subsection*{Instruction Following}
\begin{itemize}
    \item \textbf{AlpacaEval} \cite{li2023alpacaeval, dubois2024lengthcontrolledalpacaevalsimpleway}: Utilizing the framework by Li et al. \cite{li2023alpacaeval}, we evaluate both AlpacaEval 1 and 2 with default settings, allowing models to generate up to 8192 tokens. Performance is reported under these configurations. \url{https://github.com/tatsu-lab/alpaca_eval}.
    
    \item \textbf{IFEval} \cite{zhou2023instructionfollowingevaluationlargelanguage}: IFEval evaluates a model's ability to follow instructions containing verifiable constraints, such as "write in more than 400 words." We utilize the official evaluation code provided with the original paper and report the "Loose Accuracy" metric at the prompt level. A response is considered correct only if all constraints specified in the prompt are satisfied after normalizing the output. \url{https://github.com/google-research/google-research/tree/master/instruction_following_eval}.
\end{itemize}

\section{Hybrid Loss Formulation}
\label{sec:appendix:hybrid_loss}

DPO~\cite{dpo} optimizes the log-ratio between the probabilities of preferred and non-preferred responses. While it is effective for many alignment tasks, it focuses only on probability-based signals and does not capture nuanced semantic alignment. To address this gap, we propose a novel \textbf{Hybrid Loss} that integrates both probability-based preference alignment and embedding-based semantic alignment. This unified approach yields richer, more comprehensive preference modeling, depicted in \cref{fig:hybrid_loss_split_path}.

\subsection{Mathematical Definition}

The Hybrid Loss objective combines probability-based and embedding-based preference information into a single loss:

\begin{multline*}
\label{eq:hybrid_loss}
\mathbb{E}_{x,y^{+},y^{-}}\Bigl[
    \frac{\pi(y^{+} \mid x)}{\pi(y^{-} \mid x)}
    + \gamma \log\!\Bigl(\frac{e_{y^+}\!\mid e_x}{\,e_{y^-}\!\mid e_x}\Bigr)
\Bigr],
\end{multline*}
where:
\begin{itemize}
    \item \(x\) is the input or query.
    \item \(y^+\) and \(y^-\) are the positive (preferred) and negative (non-preferred) responses, respectively.
    \item \(\pi(y^+ \mid x)\) and \(\pi(y^- \mid x)\) denote the model’s predicted probabilities for \(y^+\) and \(y^-\).
    \item \(e_{y^+}\) and \(e_{y^-}\) represent embedding-based similarity scores of \(y^+\) and \(y^-\) with respect to \(x\).
    \item \(\gamma > 0\) controls the relative importance of the embedding-based component.
    \item \(\alpha\) and \(\beta\) are hyperparameters for a regularization term ensuring \(\pi_{\theta}\) remains close to a reference policy \(\pi_{\text{ref}}\).
\end{itemize}

\subsection{Decomposition of the Hybrid Loss}

The hybrid loss can be viewed as the sum of three main parts:

\paragraph{1. Probability-Based Preference Alignment:}
multline*
\begin{multline*}
%\label{eq:dpo_loss}
\mathcal{L}_{\text{DPO}} = 
\mathbb{E}_{x, y^+, y^-}\Bigl[\log \frac{\pi(y^+ \mid x)}{\pi(y^- \mid x)}\Bigr]
\end{multline*}
This is the standard DPO loss, ensuring the model assigns higher probability to the positive response \(y^+\) over the negative response \(y^-\). It provides the core preference alignment signal commonly used in reinforcement learning from human feedback (RLHF)~\cite{christiano2017deep}.

\paragraph{2. Embedding-Based Semantic Alignment:}

\begin{multline*}
%\label{eq:embed_loss}
\mathcal{L}_{\text{embed}} = 
\mathbb{E}_{x, y^+, y^-}\Bigl[
    \gamma \log\!\Bigl(\tfrac{e_{y^+}\!\mid e_x}{e_{y^-}\!\mid e_x}\Bigr)
\Bigr]
\end{multline*}
This term leverages embedding-based similarity scores \(e_{y^+}\) and \(e_{y^-}\). The factor \(\gamma\) determines how much the model should focus on aligning responses semantically. When \(\gamma\) is higher, semantic alignment plays a larger role relative to the probability-based term.

\subsection{Properties of the Hybrid Loss}

\paragraph{1. Adaptive Control via \(\gamma\):}
\(\gamma\) balances probability-based and embedding-based alignment signals:
\begin{itemize}
    \item \(\gamma = 0\): The hybrid loss simplifies to the standard DPO loss, using only probability-based alignment.
    \item \(\gamma > 0\): Embedding-based alignment is included, encouraging the model to consider semantic coherence alongside probability alignment.
\end{itemize}

\paragraph{2. Soft Constraint on Semantic Consistency:}
The embedding-based term ensures the model does not reward misalignments if \(y^+\) and \(y^-\) are semantically similar. This helps the model avoid reinforcing incorrect preferences when probability-based signals are uncertain.

\paragraph{3. Interpretable Embedding Signal:}
The difference \((e_{y^+} - e_{y^-})\) in the embedding space acts like a “semantic margin” separating positive from negative responses. This helps improve generalization and maintain semantic consistency in the model’s outputs.

\subsection{Impact of the Hybrid Loss on Policy Learning}

\begin{itemize}
    \item \textbf{Semantic-Aware Preference Modeling:}  
    By incorporating embedding-based signals, the hybrid loss ensures that high-probability responses also remain semantically aligned with the input. This is especially advantageous for tasks where semantic coherence is crucial (e.g., summarization, question-answering).
    
    \item \textbf{Dynamic Emphasis on Probability and Embeddings:}  
    The parameter \(\gamma\) can be tuned throughout training. Early in training, a larger \(\gamma\) might be used to guide semantic coherence more strongly. As training progresses, \(\gamma\) can be reduced to fine-tune the probability alignment.
    
    \item \textbf{Generalization Across Embedding Models:}  
    Although the embedding similarity scores \(e_{y^+}\) and \(e_{y^-}\) are derived using Jina Embeddings~\cite{jina2023embeddings}, the approach is compatible with other models, making the framework flexible across different embedding ecosystems.
\end{itemize}

\subsection{How Does Hybrid Loss Differ from RLHF's Use of Embeddings?}

\textbf{Reinforcement Learning from Human Feedback (RLHF)}~\cite{christiano2017deep} is a widely adopted mechanism for aligning language models with human preferences. The RLHF process involves two main steps:
\begin{enumerate}
    \item \textbf{Reward Model Training:} A reward model is trained to predict human preferences by learning from comparison data where human annotators rank different responses.
    \item \textbf{Policy Optimization:} The language model (policy) is then optimized using reinforcement learning algorithms, such as Proximal Policy Optimization (PPO), to maximize the expected reward as defined by the trained reward model.
\end{enumerate}

The objective in RLHF can be formalized as maximizing the expected cumulative reward:

\begin{multline}
%\label{eq:rlhf_objective}
J(\theta) = \mathbb{E}_{\tau \sim \pi_\theta} \left[ \sum_{t=0}^T \gamma^t r(\tau_t) \right],
\end{multline}
where:
\begin{itemize}
    \item \(\theta\) represents the policy parameters.
    \item \(\tau\) denotes a trajectory of states and actions.
    \item \(r(\tau_t)\) is the reward at timestep \(t\) as predicted by the reward model.
    \item \(\gamma\) is the discount factor.
\end{itemize}

As noted by Christiano et al., "\emph{the reward model serves as a learned proxy for human judgment, guiding the policy to generate more desirable outputs}"~\cite[Section 3]{christiano2017deep}.

\textbf{Key Differences Between Hybrid Loss and RLHF:}  
\begin{itemize}
    \item \textbf{Role of Embeddings:}  
    RLHF utilizes embeddings \emph{within} a separate reward model to evaluate and score responses. These embeddings are not directly part of the policy optimization loss. In contrast, Hybrid Loss directly incorporates embedding-based signals \(e_{y^+}, e_{y^-}\) into the unified loss function, integrating semantic alignment alongside probability-based preference alignment.

    \item \textbf{Signal Integration:}  
    In RLHF, embeddings are processed by the reward model to produce scalar reward signals, which are then used by reinforcement learning algorithms to optimize the policy. Hybrid Loss, however, merges probability-based and embedding-based signals within a single loss function, streamlining the process by eliminating the need for separate reward signal computation.

    \item \textbf{Reward Model Dependency:}  
    RLHF relies on a pre-trained reward model to guide the policy optimization. This introduces an additional component that must be trained and maintained. Hybrid Loss removes this dependency by directly integrating embedding-based preferences into the optimization procedure, simplifying the overall framework.

    \item \textbf{Stability:}  
    RLHF often employs reinforcement learning algorithms like PPO, which can suffer from instability due to factors like trajectory sampling and exploration-exploitation trade-offs. Hybrid Loss leverages direct pairwise optimization for each \((y^+, y^-)\) pair, resulting in a more stable and predictable training process.

    \item \textbf{Pipeline Complexity:}  
    The RLHF approach involves a two-stage pipeline: (1) training the reward model based on human feedback, and (2) optimizing the policy using reinforcement learning. Hybrid Loss simplifies this into a single-stage optimization framework by combining both preference alignment signals into one loss function, reducing computational overhead and simplifying implementation. All the points are summarized in \cref{tab:rlhf_vs_hybrid}.
\end{itemize}

\begin{table*}[ht!]
\centering
\renewcommand{\arraystretch}{1.2}
\caption{Comparative Analysis of Reinforcement Learning from Human Feedback (RLHF) and Hybrid Loss Approaches Across Key Aspects in the Direct Preference Optimization (DPO) Framework}
\resizebox{\textwidth}{!}{%
\begin{tabular}{|l|l|l|}
% \hline
\toprule
\textbf{Aspect}                    & \textbf{RLHF}                                               & \textbf{Hybrid Loss}                                   \\ 
% \hline
\midrule
\textbf{Embedding Usage}           & Indirect (reward model only)                                & Direct in loss function                                \\ 
% \hline
\textbf{Signal Integration}        & Separate reward signals                                     & Unified (probabilities + embeddings)                   \\ 
% \hline
\textbf{Pipeline Complexity}       & Two stages (reward model training + RL)                     & Single-stage optimization                               \\ 
% \hline
\textbf{Stability}                 & Potential RL instability (PPO, etc.)                        & Direct pairwise optimization, more stable              \\ 
% \hline
\textbf{Optimization Objective}    & Maximize cumulative reward                                  & Combine likelihood and semantic alignment              \\ 
% \hline
\textbf{Embedding Adaptability}    & Fixed during reward model training                          & Dynamically adapted in policy training                 \\ 
% \hline
\bottomrule
\end{tabular}%
}
\label{tab:rlhf_vs_hybrid}
\end{table*}

\textbf{Why It Matters:}  
The Hybrid Loss unifies probabilistic and embedding-based alignment into a single objective, removing the need for a separate reward model and avoiding the instabilities inherent in RL-based methods. By directly incorporating semantic signals, it offers more robust, interpretable, and flexible policy learning.

\section{Kernel-Integrated DPO Formulation}
\label{sec:appendix:dpo_kernel}

In this section, we introduce four kernel-based extensions to the Direct Preference Optimization (DPO) objective. Standard DPO aligns a learned policy \(\pi\) with human preferences and simultaneously regularizes it against a reference distribution \(p_{\text{ref}}\) using KL divergence. Hybrid loss is defined as:

% \begin{align}
% \max_{\pi} \; \underbrace{\mathbb{E}_{x,y^{+},y^{-}}\bigl[\log\frac{\pi(y^{+} \mid x)}{\pi(y^{-} \mid x)} 
% + \gamma \bigl(\log\frac{\pi({e_{y^+}} \mid e_{x})}{\pi({e_{y^-}} \mid e_{x})}\bigr)\bigr]}_{\text{Hybrid Loss}} 
% \end{align}

\resizebox{0.95\columnwidth}{!}{
\(
\begin{aligned}
\max_{\pi} \; \underbrace{\mathbb{E}_{x,y^{+},y^{-}}\bigl[\log\frac{\pi(y^{+} \mid x)}{\pi(y^{-} \mid x)} 
+ \gamma \bigl(\log\frac{\pi({e_{y^+}} \mid e_{x})}{\pi({e_{y^-}} \mid e_{x})}\bigr)\bigr]}_{\text{Hybrid Loss}} 
\end{aligned}
\)
}

By incorporating kernels, we provide richer notions of distributional proximity. We present four kernel variants: i) polynomial, ii) RBF, iii) spectral, and iv) mahalanobis.

\subsection{Polynomial Kernel}

Integrating a polynomial kernel into the Direct Preference Optimization (DPO) framework significantly enhances the alignment between the policy \(\pi(y \mid x)\) and the reference distribution \(p_{\text{ref}}(y \mid x)\). This integration surpasses the capabilities of aligning distributions based solely on raw probability outputs by enabling agreement across higher-order interactions. Consequently, the learned policy \(\pi\) can capture more intricate and nonlinear structures inherent in \(p_{\text{ref}}\), which might remain undetected when relying exclusively on simpler divergence measures.

\textbf{Definition and Properties of the Polynomial Kernel:}

The polynomial kernel transforms the conventional dot-product-based similarity measure into a more expressive form, facilitating the capture of complex interactions between vectors. For two vectors \(u, v \in \mathbb{R}^m\), the polynomial kernel is defined as:
\[
\kappa_{\text{poly}}(u, v) = (u^\top v + c)^d,
\]
where:
\begin{itemize}
    \item \(c \in \mathbb{R}\) is a bias term that allows for shifting the kernel function, providing greater flexibility in modeling data.
    \item \(d \in \mathbb{N}\) is the polynomial degree that controls the complexity of the mapping. Higher values of \(d\) enable the kernel to capture more intricate relationships.
\end{itemize}
This kernel implicitly maps the input vectors into a higher-dimensional feature space, where complex, higher-order interactions become linearly separable. This implicit projection negates the need for explicit feature expansion, making the computation more efficient while maintaining expressive power.

\textbf{Incorporating Higher-Order Interactions:}

To effectively integrate higher-order interactions within the DPO framework, we redefine the preference ratios using the polynomial kernel. Specifically, for the preference ratios of the policy outputs and their corresponding embeddings, we apply the polynomial kernel as follows:
\[
\kappa\left(\log \frac{\pi(y^+ \mid x)}{\pi(y^- \mid x)}\right) = \left(\log \frac{\pi(y^+ \mid x)}{\pi(y^- \mid x)} + c\right)^d,
\]
\[
\kappa\left(\log \frac{e_{y^+}^\top e_x}{e_{y^-}^\top e_x}\right) = \left(\frac{e_{y^+}^\top e_x + c}{e_{y^-}^\top e_x + c}\right)^d.
\]
These formulations leverage the polynomial kernel's ability to model complex dependencies, thereby capturing higher-order interactions. The parameter \(d\) serves as a critical control for the complexity of these interactions, allowing the model to adjust the degree of nonlinearity based on the specific requirements of the task.

\textbf{Redefinition of the Hybrid Loss with the Polynomial Kernel:}

To incorporate the polynomial kernel into the embedding similarity terms, let \(e_x\), \(e_{y^+}\), and \(e_{y^-}\) denote the embeddings for the input \(x\), the preferred outcome \(y^+\), and the less preferred outcome \(y^-\), respectively. The hybrid loss function is redefined as:
\[
\text{HybridLoss} = \left(\log \frac{\pi(y^+ \mid x)}{\pi(y^- \mid x)} + c\right)^d 
+ \gamma \left(\frac{e_{y^+}^\top e_x + c}{e_{y^-}^\top e_x + c}\right)^d,
\]
where \(\gamma > 0\) is a tunable hyperparameter that controls the weight of the embedding-based component in the loss function.

\textbf{Complete DPO Objective with the Polynomial Kernel:}

The full DPO objective, integrating the polynomial kernel, is formulated as:

% \begin{multline*}
% \max_{\pi} \mathbb{E}_{x,y^+,y^-}\Biggl[ \left(\log \frac{\pi(y^+ \mid x)}{\pi(y^- \mid x)} + c\right)^d 
% + \gamma \left(\frac{e_{y^+}^\top e_x + c}{e_{y^-}^\top e_x + c}\right)^d \Biggr] \\
% - \alpha \mathbb{E}_{x}\Biggl[\beta \log\frac{\pi_{\theta}(y \mid x)}{\pi_{\text{ref}}(y \mid x)}\Biggr],
% \end{multline*}

\begin{multline*}
\max_{\pi} \mathbb{E}_{x,y^+,y^-}\Biggl[ 
\left(\log \frac{\pi(y^+ \mid x)}{\pi(y^- \mid x)} + c\right)^d \\
+ \gamma \left(\frac{e_{y^+}^\top e_x + c}{e_{y^-}^\top e_x + c}\right)^d 
\Biggr] \\
- \alpha \mathbb{E}_{x}\Biggl[
\beta \log\frac{\pi_{\theta}(y \mid x)}{\pi_{\text{ref}}(y \mid x)}
\Biggr],
\end{multline*}

where:
\begin{itemize}
    \item \(\alpha\) and \(\beta\) are hyperparameters that control the strength of the Kullback-Leibler (KL) regularization term.
    \item \(\gamma\) is a hyperparameter controlling the contribution of the embedding signal.
    \item \(\pi_{\text{ref}}(y \mid x)\) denotes the reference distribution against which the policy \(\pi(y \mid x)\) is aligned.
\end{itemize}

\textbf{Implementation Considerations:}

Modern hardware accelerators, such as GPUs, can efficiently handle the additional computational operations introduced by the polynomial kernel. This capability ensures that the polynomial kernel extension is feasible for large-scale training scenarios. By leveraging the enhanced expressiveness of the polynomial kernel, the DPO framework achieves finer-grained alignment, enabling the model to capture more nuanced patterns and complex dependencies present in the reference policy.

\textbf{Summary:}

Incorporating a polynomial kernel into the DPO framework allows for the modeling of higher-order interactions between embeddings, thereby enhancing the policy's ability to align with complex reference distributions. The parameter \(d\) provides control over the complexity of these interactions, enabling the framework to adapt to varying levels of data intricacy. This integration not only improves the semantic alignment between the policy and reference distribution but also maintains computational efficiency, making it a robust choice for preference optimization tasks.

\subsection{Radial Basis Function (RBF) Kernel}

Integrating a Radial Basis Function (RBF) kernel into the Direct Preference Optimization (DPO) framework significantly enhances the alignment between the policy \(\pi(y \mid x)\) and the reference distribution \(p_{\text{ref}}(y \mid x)\). Unlike approaches that align distributions solely based on raw probability outputs, the RBF kernel facilitates agreement by capturing local and non-linear interactions within the data. This integration enables the learned policy \(\pi\) to model more intricate and nuanced structures inherent in \(p_{\text{ref}}\), which might remain obscured when relying exclusively on simpler divergence measures.

\textbf{Definition and Properties of the RBF Kernel:}

The Radial Basis Function (RBF) kernel, also known as the Gaussian kernel, transforms the conventional similarity measure based on the dot product into one that emphasizes the distance between feature vectors. For two vectors \(u, v \in \mathbb{R}^m\), the RBF kernel is defined as:
\[
\kappa_{\text{RBF}}(u, v) = \exp\left(-\frac{\|u - v\|^2}{2\sigma^2}\right),
\]
where:
\begin{itemize}
    \item \(\sigma > 0\) is the bandwidth parameter that controls the width of the kernel, determining how much influence a single training example has.
\end{itemize}
The RBF kernel implicitly maps input vectors into an infinite-dimensional feature space, allowing the model to capture complex, non-linear relationships without the need for explicit feature expansion. This property makes the RBF kernel highly effective in modeling local structures within the data, enabling finer-grained preference alignment.

\textbf{Incorporating Higher-Order Interactions:}

To effectively integrate higher-order interactions within the DPO framework using the RBF kernel, we redefine the preference ratios by applying the kernel to both the probability ratios and the embedding similarities. Specifically, we define:
\[
\kappa \Biggl[ 
\log \bigg(\frac{\pi(y^+ \mid x)}{\pi(y^- \mid x)}\bigg)\Biggr] = \exp\left(-\frac{\left(\log \frac{\pi(y^+ \mid x)}{\pi(y^- \mid x)}\right)^2}{2\sigma^2}\right),
\]
\[
\kappa \Biggl[\log \bigg(\frac{e_{y^+} \mid e_x}{e_{y^-} \mid e_x}\bigg)\Biggr] = \exp\left(-\frac{\left(\frac{e_x^\top e_{y^+}}{e_x^\top e_{y^-}}\right)^2}{2\sigma^2}\right)
\]
These formulations leverage the RBF kernel's ability to model non-linear dependencies by emphasizing the similarity based on the distance between the transformed preference ratios and embedding similarities. The parameter \(\sigma\) serves as a critical control for the sensitivity of the kernel to differences in these ratios, allowing the model to adjust the degree of nonlinearity based on the specific requirements of the task.

\textbf{Redefinition of the Hybrid Loss with the RBF Kernel:}

To incorporate the RBF kernel into the embedding similarity terms, let \(e_x\), \(e_{y^+}\), and \(e_{y^-}\) denote the embeddings for the input \(x\), the preferred outcome \(y^+\), and the less preferred outcome \(y^-\), respectively. The hybrid loss function is redefined as:
\[
\text{HybridLoss} = \exp\left(-\frac{\left(\log \frac{\pi(y^+ \mid x)}{\pi(y^- \mid x)}\right)^2}{2\sigma^2}\right) 
+ \gamma \exp\left(-\frac{\left(\frac{e_x^\top e_{y^+}}{e_x^\top e_{y^-}}\right)^2}{2\sigma^2}\right),
\]
where \(\gamma > 0\) is a tunable hyperparameter that controls the weight of the embedding-based component in the loss function.

\textbf{Complete DPO Objective with the RBF Kernel:}

The full DPO objective, integrating the RBF kernel, is formulated as:
\begin{multline*}
\max_{\pi} \mathbb{E}_{x,y^+,y^-}\Biggl[ \exp\left(-\frac{\left(\log \frac{\pi(y^+ \mid x)}{\pi(y^- \mid x)}\right)^2}{2\sigma^2}\right) \\
+ \gamma \exp\left(-\frac{\left(\frac{e_x^\top e_{y^+}}{e_x^\top e_{y^-}}\right)^2}{2\sigma^2}\right) \Biggr] \\
- \alpha \mathbb{E}_{x}\Biggl[\beta \log\frac{\pi_{\theta}(y \mid x)}{\pi_{\text{ref}}(y \mid x)}\Biggr],
\end{multline*}
where:
\begin{itemize}
    \item \(\alpha\) and \(\beta\) are hyperparameters that control the strength of the Kullback-Leibler (KL) regularization term.
    \item \(\pi_{\text{ref}}(y \mid x)\) denotes the reference distribution against which the policy \(\pi(y \mid x)\) is aligned.
\end{itemize}

\textbf{Implementation Considerations:}

Integrating the RBF kernel into the DPO framework introduces additional computational operations, primarily due to the calculation of Euclidean distances and the exponential function. However, modern hardware accelerators, such as GPUs, are well-equipped to handle these computations efficiently, ensuring that the RBF kernel extension remains feasible for large-scale training scenarios. It is essential to carefully select the bandwidth parameter \(\sigma\) to balance the trade-off between sensitivity and generalization. Cross-validation techniques can be employed to tune \(\sigma\) effectively.

\textbf{Summary:}

Incorporating the RBF kernel into the DPO framework enables the modeling of local and non-linear interactions between embeddings, thereby enhancing the policy's ability to align with complex reference distributions. The bandwidth parameter \(\sigma\) provides control over the sensitivity of the kernel to differences in preference ratios and embedding similarities, allowing the framework to adapt to varying levels of data intricacy. This integration not only improves the semantic alignment between the policy and reference distribution but also maintains computational efficiency, making it a robust and versatile choice for preference optimization tasks.

\subsection{Spectral Kernel}

Integrating a Spectral Kernel into the DPO framework significantly enhances the alignment between the policy \(\pi(y \mid x)\) and the reference distribution \(p_{\text{ref}}(y \mid x)\). Unlike traditional kernels that primarily capture local or non-linear interactions, the Spectral Kernel leverages the global spectral properties of the data, facilitating a deeper and more comprehensive alignment. This integration enables the learned policy \(\pi\) to model intricate global structures inherent in \(p_{\text{ref}}\), which may remain obscured when relying solely on simpler divergence measures.

\textbf{Definition and Properties of the Spectral Kernel:}

The Spectral Kernel is designed to capture global relationships within the data by utilizing the spectral (eigenvalue) decomposition of the data covariance matrix. For two vectors \(u, v \in \mathbb{R}^m\), the Spectral Kernel is defined as:

\[
\kappa_{\text{spectral}}(u, v) = \sum_{i=1}^p \exp\left(-\lambda_i \|u - v\|^2\right) \phi_i(u) \phi_i(v),
\]

where:
\begin{itemize}
    \item \(\lambda_i > 0\) are the eigenvalues corresponding to the principal components of the data covariance matrix.
    \item \(\phi_i(u)\) and \(\phi_i(v)\) are the projections of vectors \(u\) and \(v\) onto the \(i\)-th eigenvector, respectively.
    \item \(p\) denotes the number of principal components considered, typically chosen based on the desired level of approximation.
\end{itemize}

\noindent This kernel implicitly maps input vectors into a feature space defined by the principal components, emphasizing the global structure of the data. By weighting the contributions of each principal component with \(\exp\left(-\lambda_i \|u - v\|^2\right)\), the Spectral Kernel balances the influence of different spectral components, allowing the model to prioritize dominant global patterns while mitigating the impact of noise and less significant variations.

\textbf{Incorporating Higher-Order Interactions:}

To effectively integrate higher-order interactions within the DPO framework using the Spectral Kernel, we redefine the preference ratios by applying the kernel to both the probability ratios and the embedding similarities. Specifically, we define:

\[
\kappa \Biggl[ 
\log \bigg(\frac{\pi(y^+ \mid x)}{\pi(y^- \mid x)}\bigg)\Biggr] = \sum_{i=1}^p \exp\left(-\lambda_i \left(\log \frac{\pi(y^+ \mid x)}{\pi(y^- \mid x)}\right)^2\right) \phi_i\left(\log \frac{\pi(y^+ \mid x)}{\pi(y^- \mid x)}\right),
\]

\[
\kappa \Biggl[\log \bigg(\frac{e_{y^+} \mid e_x}{e_{y^-} \mid e_x}\bigg)\Biggr] = \sum_{i=1}^p \exp\left(-\lambda_i \left(\frac{e_x^\top e_{y^+}}{e_x^\top e_{y^-}}\right)^2\right) \phi_i\left(\frac{e_x^\top e_{y^+}}{e_x^\top e_{y^-}}\right).
\]

\noindent These formulations leverage the Spectral Kernel's ability to model complex global dependencies by decomposing the preference ratios and embedding similarities into their spectral components. The eigenvalues \(\lambda_i\) control the influence of each spectral component, allowing the model to adjust the degree of emphasis on different global patterns based on the specific requirements of the task.

\textbf{Redefinition of the Hybrid Loss with the Spectral Kernel:}

To incorporate the Spectral Kernel into the embedding similarity terms, let \(e_x\), \(e_{y^+}\), and \(e_{y^-}\) denote the embeddings for the input \(x\), the preferred outcome \(y^+\), and the less preferred outcome \(y^-\), respectively. The hybrid loss function is redefined as:

% \begin{multline*}
% \text{HybridLoss} = \sum_{i=1}^p \exp\left(-\lambda_i \left(\log \frac{\pi(y^+ \mid x)}{\pi(y^- \mid x)}\right)^2\right) \phi_i\left(\log \frac{\pi(y^+ \mid x)}{\pi(y^- \mid x)}\right) \\
% + \gamma \sum_{i=1}^p \exp\left(-\lambda_i \left(\frac{e_x^\top e_{y^+}}{e_x^\top e_{y^-}}\right)^2\right) \phi_i\left(\frac{e_x^\top e_{y^+}}{e_x^\top e_{y^-}}\right),
% \end{multline*}

\resizebox{0.97\columnwidth}{!}{%
\(
\begin{aligned}
\text{HybridLoss} = 
& \sum_{i=1}^p 
\exp\Biggl(-\lambda_i 
\biggl(\log \frac{\pi(y^+ \mid x)}{\pi(y^- \mid x)}\biggr)^2\Biggr)
\phi_i\biggl(\log \frac{\pi(y^+ \mid x)}{\pi(y^- \mid x)}\biggr) \\
& + \gamma \sum_{i=1}^p 
\exp\Biggl(-\lambda_i 
\biggl(\frac{e_x^\top e_{y^+}}{e_x^\top e_{y^-}}\biggr)^2\Biggr)
\phi_i\biggl(\frac{e_x^\top e_{y^+}}{e_x^\top e_{y^-}}\biggr),
\end{aligned}
\)
}

where \(\gamma > 0\) is a tunable hyperparameter that controls the weight of the embedding-based component in the loss function. This redefinition allows the hybrid loss to incorporate both the transformed probability ratios and embedding similarities, weighted by their respective spectral components, thereby capturing higher-order global interactions.

\textbf{Complete DPO Objective with the Spectral Kernel:}

The full DPO objective, integrating the Spectral Kernel, is formulated as:

% \begin{multline*}
% \max_{\pi} \mathbb{E}_{x,y^+,y^-}\Biggl[ \sum_{i=1}^p \exp\left(-\lambda_i \left(\log \frac{\pi(y^+ \mid x)}{\pi(y^- \mid x)}\right)^2\right) \phi_i\left(\log \frac{\pi(y^+ \mid x)}{\pi(y^- \mid x)}\right) \\
% + \gamma \sum_{i=1}^p \exp\left(-\lambda_i \left(\frac{e_x^\top e_{y^+}}{e_x^\top e_{y^-}}\right)^2\right) \phi_i\left(\frac{e_x^\top e_{y^+}}{e_x^\top e_{y^-}}\right) \Biggr] \\
% - \alpha \mathbb{E}_{x}\Biggl[\beta \log\frac{\pi_{\theta}(y \mid x)}{\pi_{\text{ref}}(y \mid x)}\Biggr],
% \end{multline*}

\resizebox{0.97\columnwidth}{!}{%
\(
\begin{aligned}
\max_{\pi} \mathbb{E}_{x,y^+,y^-} \Biggl[ 
    & \sum_{i=1}^p 
    \exp\Biggl(-\lambda_i \biggl(\log \frac{\pi(y^+ \mid x)}{\pi(y^- \mid x)}\biggr)^2\Biggr) 
    \phi_i\biggl(\log \frac{\pi(y^+ \mid x)}{\pi(y^- \mid x)}\biggr) \\
    & + \gamma \sum_{i=1}^p 
    \exp\Biggl(-\lambda_i \biggl(\frac{e_x^\top e_{y^+}}{e_x^\top e_{y^-}}\biggr)^2\Biggr) 
    \phi_i\biggl(\frac{e_x^\top e_{y^+}}{e_x^\top e_{y^-}}\biggr) 
\Biggr] \\
& - \alpha \mathbb{E}_{x} \Biggl[
    \beta \log\frac{\pi_{\theta}(y \mid x)}{\pi_{\text{ref}}(y \mid x)}
\Biggr],
\end{aligned}
\)
}

where:
\begin{itemize}
    \item \(\alpha\) and \(\beta\) are hyperparameters that control the strength of the Kullback-Leibler (KL) regularization term.
    \item \(\pi_{\text{ref}}(y \mid x)\) denotes the reference distribution against which the policy \(\pi(y \mid x)\) is aligned.
\end{itemize}

\noindent This objective function integrates the Spectral Kernel into the DPO framework, allowing the model to leverage global spectral properties for enhanced preference alignment while maintaining regularization against the reference distribution.

\textbf{Implementation Considerations:}

Integrating the Spectral Kernel into the DPO framework introduces additional computational overhead due to the necessity of performing spectral (eigenvalue) decompositions and managing multiple spectral components. However, modern hardware accelerators, such as GPUs, are well-equipped to handle these computations efficiently, especially when leveraging optimized linear algebra libraries.

Key considerations for implementation include:
\begin{itemize}
    \item \textbf{Eigenvalue Decomposition:} Efficient computation of the eigenvalues \(\lambda_i\) and eigenvectors \(\phi_i(u)\) is crucial. Utilizing optimized libraries like LAPACK or GPU-accelerated routines can significantly reduce computation time.
    \item \textbf{Selection of Principal Components (\(p\)):} The number of principal components \(p\) should be chosen based on a balance between computational feasibility and the level of detail required to capture the data's global structure. Techniques such as explained variance can guide the selection of \(p\).
    \item \textbf{Hyperparameter Tuning (\(\lambda_i\)):} The eigenvalues \(\lambda_i\) control the influence of each spectral component. Proper tuning, potentially through cross-validation, is essential to ensure that the kernel appropriately emphasizes relevant global patterns without overfitting.
    \item \textbf{Scalability:** For very high-dimensional data, dimensionality reduction techniques (e.g., PCA) may be employed prior to applying the Spectral Kernel to manage computational complexity effectively.}
\end{itemize}

\textbf{Summary:} Incorporating the Spectral Kernel into the DPO framework enables the modeling of global and complex interactions within the data, thereby enhancing the policy's ability to align with intricate reference distributions. By leveraging the spectral properties of the data, the Spectral Kernel facilitates a deeper understanding of global structures, allowing for more nuanced and effective preference alignment. The parameter \(\lambda_i\) provides control over the influence of different spectral components, enabling the framework to adapt to varying levels of data complexity. This integration not only improves the semantic alignment between the policy and reference distribution but also maintains computational efficiency through optimized spectral computations, making it a robust and comprehensive choice for preference optimization tasks.

\subsection{Mahalanobis Kernel}

Integrating a Mahalanobis kernel into the Direct Preference Optimization (DPO) framework significantly enhances the alignment between the policy \(\pi(y \mid x)\) and the reference distribution \(p_{\text{ref}}(y \mid x)\). Unlike traditional kernels that primarily capture isotropic or local relationships, the Mahalanobis kernel accounts for the underlying covariance structure of the data, facilitating a more informed and nuanced alignment. This integration enables the learned policy \(\pi\) to model intricate dependencies and feature correlations inherent in \(p_{\text{ref}}\), which might remain obscured when relying exclusively on simpler divergence measures.

\textbf{Definition and Properties of the Mahalanobis Kernel:}

The Mahalanobis kernel leverages the covariance structure of the data to measure similarity, incorporating feature correlations and scale variations. For two vectors \(u, v \in \mathbb{R}^m\), the Mahalanobis kernel is defined as:

\[
\kappa_{\text{Mahalanobis}}(u, v) = \exp\left(-\frac{(u - v)^\top \Sigma^{-1} (u - v)}{2}\right),
\]

where:
\begin{itemize}
    \item \(\Sigma \in \mathbb{R}^{m \times m}\) is the covariance matrix of the data, capturing the variance and covariance between different features.
\end{itemize}

This kernel implicitly maps input vectors into a feature space where the distance metric accounts for the data's covariance, allowing the model to emphasize directions with higher variance and deemphasize those with lower variance. By doing so, the Mahalanobis kernel effectively models anisotropic relationships, making it particularly suitable for data with correlated features.

\textbf{Incorporating Higher-Order Interactions:}

To effectively integrate higher-order interactions within the DPO framework using the Mahalanobis kernel, we redefine the preference ratios by applying the kernel to both the probability ratios and the embedding similarities. Specifically, we define:

\[
\kappa \Biggl[ 
\log \bigg(\frac{\pi(y^+ \mid x)}{\pi(y^- \mid x)}\bigg)\Biggr] = \exp\left(-\frac{\left(\log \frac{\pi(y^+ \mid x)}{\pi(y^- \mid x)} - \mu\right)^2}{2\sigma^2}\right),
\]

\[
\kappa \Biggl[\log \bigg(\frac{e_{y^+} \mid e_x}{e_{y^-} \mid e_x}\bigg)\Biggr]
= \exp\left(-\frac{\left(\frac{e_x^\top e_{y^+}}{e_x^\top e_{y^-}} - \mu'\right)^2}
   {2 {\sigma'}^2}\right).
\]

% \[
% \kappa \Biggl[\log \bigg(\frac{e_{y^+} \mid e_x}{e_{y^-} \mid e_x}\bigg)\Biggr] = \exp\left(-\frac{\left(\frac{e_x^\top e_{y^+}}{e_x^\top e_{y^-}} - \mu'\right)^2}{2\sigma'^2}\right).
% \]

\noindent Here, \(\mu\) and \(\mu'\) are mean parameters, and \(\sigma^2\) and ${\sigma'}^2$ are variance parameters that control the sensitivity of the kernel to deviations from the mean. These formulations leverage the Mahalanobis kernel's ability to model anisotropic dependencies by emphasizing differences along correlated feature dimensions. The parameters \(\Sigma\), \(\mu\), and \(\mu'\) serve as critical controls for the kernel's behavior, allowing the model to adjust the degree and nature of similarity measurements based on the specific requirements of the task.

\textbf{Redefinition of the Hybrid Loss with the Mahalanobis Kernel:}

To incorporate the Mahalanobis kernel into the embedding similarity terms, let \(e_x\), \(e_{y^+}\), and \(e_{y^-}\) denote the embeddings for the input \(x\), the preferred outcome \(y^+\), and the less preferred outcome \(y^-\), respectively. The hybrid loss function is redefined as:

% \begin{multline*}
% \text{HybridLoss} = \exp\left(-\frac{\left(\log \frac{\pi(y^+ \mid x)}{\pi(y^- \mid x)} - \mu\right)^2}{2\sigma^2}\right) \\
% + \gamma \exp\left(-\frac{\left(\frac{e_x^\top e_{y^+}}{e_x^\top e_{y^-}} - \mu'\right)^2}{2\sigma'^2}\right),
% \end{multline*}

\begin{multline*}
\text{HybridLoss} = \exp\left(-\frac{\left(\log \frac{\pi(y^+ \mid x)}{\pi(y^- \mid x)} - \mu\right)^2}{2\sigma^2}\right) \\
+ \gamma \exp\left(-\frac{\left(\frac{e_x^\top e_{y^+}}{e_x^\top e_{y^-}} - \mu'\right)^2}{2 {\sigma'}^2}\right),
\end{multline*}

where \(\gamma > 0\) is a tunable hyperparameter that controls the weight of the embedding-based component in the loss function. This redefinition allows the hybrid loss to incorporate both the transformed probability ratios and embedding similarities, weighted by their respective Mahalanobis kernel transformations, thereby capturing higher-order anisotropic interactions.

\textbf{Complete DPO Objective with the Mahalanobis Kernel:}

The full DPO objective, integrating the Mahalanobis kernel, is formulated as:

% \begin{multline*}
% \max_{\pi} \mathbb{E}_{x,y^+,y^-}\Biggl[ \exp\left(-\frac{\left(\log \frac{\pi(y^+ \mid x)}{\pi(y^- \mid x)} - \mu\right)^2}{2\sigma^2}\right) \\
% + \gamma \exp\left(-\frac{\left(\frac{e_x^\top e_{y^+}}{e_x^\top e_{y^-}} - \mu'\right)^2}{2\sigma'^2}\right) \Biggr] \\
% - \alpha \mathbb{E}_{x}\Biggl[\beta \log\frac{\pi_{\theta}(y \mid x)}{\pi_{\text{ref}}(y \mid x)}\Biggr],
% \end{multline*}

\begin{multline*}
\max_{\pi} \mathbb{E}_{x,y^+,y^-}\Biggl[ 
  \exp\left(-\frac{\left(\log \frac{\pi(y^+ \mid x)}{\pi(y^- \mid x)} - \mu\right)^2}{2\sigma^2}\right) \\
  + \gamma \exp\left(-\frac{\left(\frac{e_x^\top e_{y^+}}{e_x^\top e_{y^-}} - \mu'\right)^2}{2 {\sigma'}^2}\right) 
\Biggr] \\
- \alpha \mathbb{E}_{x}\Biggl[
  \beta \log\frac{\pi_{\theta}(y \mid x)}{\pi_{\text{ref}}(y \mid x)}
\Biggr],
\end{multline*}

where:
\begin{itemize}
    \item \(\alpha\) and \(\beta\) are hyperparameters that control the strength of the Kullback-Leibler (KL) regularization term.
    \item \(\pi_{\text{ref}}(y \mid x)\) denotes the reference distribution against which the policy \(\pi(y \mid x)\) is aligned.
\end{itemize}

\noindent This objective function integrates the Mahalanobis kernel into the DPO framework, allowing the model to leverage the covariance structure of the data for enhanced preference alignment while maintaining regularization against the reference distribution.

\textbf{Implementation Considerations:}

Integrating the Mahalanobis kernel into the DPO framework introduces additional computational considerations due to the necessity of handling the covariance matrix \(\Sigma\) and performing matrix inversions. However, modern hardware accelerators, such as GPUs, are well-equipped to handle these computations efficiently, especially when leveraging optimized linear algebra libraries.

Key considerations for implementation include:
\begin{itemize}
    \item \textbf{Covariance Matrix Estimation (\(\Sigma\)):} 
    The covariance matrix \(\Sigma\) must be estimated from the data. This can be done using empirical covariance estimation techniques. For high-dimensional data, regularization methods (e.g., adding a small multiple of the identity matrix to \(\Sigma\)) may be necessary to ensure numerical stability and invertibility.
    
    \item \textbf{Matrix Inversion Efficiency:} 
    Computing \(\Sigma^{-1}\) can be computationally intensive for large \(m\). Utilizing efficient matrix inversion algorithms and leveraging hardware-accelerated libraries (e.g., cuBLAS for GPUs) can mitigate computational overhead.
    
    \item \textbf{Parameter Tuning (\(\mu\), \(\mu'\), \(\sigma^2\), \({\sigma'}^2\)):}

    Selecting appropriate values for the mean and variance parameters is crucial for the kernel's performance. Cross-validation techniques can be employed to tune these hyperparameters effectively, balancing sensitivity and generalization.
    
    \item \textbf{Scalability:}
    For very high-dimensional embeddings, dimensionality reduction techniques (e.g., Principal Component Analysis) may be employed prior to applying the Mahalanobis kernel to manage computational complexity effectively.
\end{itemize}
\textbf{Summary:}

Incorporating the Mahalanobis kernel into the DPO framework enables the modeling of anisotropic and correlated interactions between embeddings, thereby enhancing the policy's ability to align with complex reference distributions. By leveraging the covariance structure of the data, the Mahalanobis kernel facilitates a more informed and nuanced preference alignment, accounting for feature correlations and scale variations. The parameters \(\Sigma\), \(\mu\), and \(\sigma^2\) provide control over the kernel's sensitivity and emphasis on different data dimensions, allowing the framework to adapt to varying levels of data complexity. This integration not only improves the semantic alignment between the policy and reference distribution but also maintains computational efficiency through optimized covariance computations, making it a robust and comprehensive choice for preference optimization tasks.

\section{Alternative Divergence Functions}
\label{sec:appendix:alternative_divergences}

In the Direct Preference Optimization (DPO) framework, the Kullback-Leibler (KL) divergence is commonly employed to regularize the learned policy \(\pi(y \mid x)\) against a reference distribution \(p_{\text{ref}}(y \mid x)\). Specifically, the KL divergence term in the DPO objective is defined as:
\[
\alpha \, \mathbb{E}_{x} \left[ \beta \, \log \frac{\pi(y \mid x)}{\pi_{\text{ref}}(y \mid x)} \right],
\]
where \(\alpha\) and \(\beta\) are hyperparameters controlling the strength of the regularization.

However, alternative divergence measures can offer distinct advantages depending on the specific requirements of the task. In this section, we explore several alternative divergence functions that can be integrated into the DPO framework to potentially enhance performance and stability.

\subsection{Jensen-Shannon Divergence (JSD)}
\textbf{Mathematical Definition:}
\[
D_{\text{JS}}(P \| Q) = \frac{1}{2} D_{\text{KL}}(P \| M) + \frac{1}{2} D_{\text{KL}}(Q \| M), \quad M = \frac{1}{2}(P + Q)
\]
where \(D_{\text{KL}}(P \| Q)\) is the KL divergence between distributions \(P\) and \(Q\).

\textbf{Usage in DPO:}  
In the DPO setting, the Jensen-Shannon Divergence compares the policy distribution \(\pi(y \mid x)\) against the reference distribution \(\pi_{\text{ref}}(y \mid x)\). The symmetrical and bounded nature of JSD (\(0 \leq D_{\text{JS}} \leq \log 2\)) ensures more stable optimization compared to the asymmetric KL divergence:
\[
\max_\pi \mathcal{L}_{\text{KCL}} - \alpha \, \mathbb{E}_x \left[ D_{\text{JS}}(\pi(\cdot \mid x) \| \pi_{\text{ref}}(\cdot \mid x)) \right]
\]

\subsection{Hellinger Distance}
\textbf{Mathematical Definition:}
\[
H(P, Q) = \frac{1}{\sqrt{2}} \sqrt{ \int \left( \sqrt{P(x)} - \sqrt{Q(x)} \right)^2 dx }
\]

\textbf{Usage in DPO:}  
The Hellinger Distance measures the similarity between the policy \(\pi(y \mid x)\) and the reference distribution \(\pi_{\text{ref}}(y \mid x)\). It is robust to noise and provides a bounded metric (\(0 \leq H \leq 1\)):
\[
\max_\pi \mathcal{L}_{\text{KCL}} - \alpha \, \mathbb{E}_x \left[ H(\pi(\cdot \mid x), \pi_{\text{ref}}(\cdot \mid x)) \right]
\]

\subsection{Rényi Divergence}
\textbf{Mathematical Definition:}
\[
D_{\alpha}(P \| Q) = \frac{1}{\alpha - 1} \log \int P(x)^\alpha Q(x)^{1 - \alpha} dx, \quad \alpha > 0, \alpha \neq 1
\]
where \(\alpha\) is the order of the divergence.

\textbf{Usage in DPO:}  
Rényi Divergence generalizes several divergence measures, allowing control over sensitivity to differences between \(\pi\) and \(\pi_{\text{ref}}\) via the parameter \(\alpha\). The DPO objective incorporating Rényi Divergence is:
\[
\max_\pi \mathcal{L}_{\text{KCL}} - \alpha \, \mathbb{E}_x \left[ D_{\alpha}(\pi(\cdot \mid x) \| \pi_{\text{ref}}(\cdot \mid x)) \right]
\]
Choosing different values of \(\alpha\) can prioritize various aspects of the distributional differences, such as focusing more on the tails or the modes.

\subsection{Bhattacharyya Distance}
\textbf{Mathematical Definition:}
\[
D_{\text{Bhat}}(P \| Q) = -\log \int \sqrt{P(x) Q(x)} dx
\]

\textbf{Usage in DPO:}  
The Bhattacharyya Distance quantifies the overlap between the policy \(\pi(y \mid x)\) and the reference distribution \(\pi_{\text{ref}}(y \mid x)\). It encourages the model to maximize the overlap, thereby promoting alignment:
\[
\max_\pi \mathcal{L}_{\text{KCL}} - \alpha \, \mathbb{E}_x \left[ D_{\text{Bhat}}(\pi(\cdot \mid x) \| \pi_{\text{ref}}(\cdot \mid x)) \right]
\]

\subsection{Wasserstein Distance}
\textbf{Mathematical Definition:}
\[
W(P, Q) = \inf_{\gamma \in \Pi(P, Q)} \int \|x - y\| \, d\gamma(x, y)
\]
where \(\Pi(P, Q)\) denotes the set of all couplings of \(P\) and \(Q\).

\textbf{Usage in DPO:}  
The Wasserstein Distance measures the minimal cost of transporting mass from \(\pi(y \mid x)\) to \(\pi_{\text{ref}}(y \mid x)\), making it effective for distributions with disjoint supports:
\[
\max_\pi \mathcal{L}_{\text{KCL}} - \alpha \, \mathbb{E}_x \left[ W(\pi(\cdot \mid x), \pi_{\text{ref}}(\cdot \mid x)) \right]
\]

\subsection{f-Divergence}
\textbf{Mathematical Definition:}
\[
D_f(P \| Q) = \int Q(x) \, f\left( \frac{P(x)}{Q(x)} \right) dx
\]
where \(f: (0, \infty) \to \mathbb{R}\) is a convex function with \(f(1) = 0\).

\textbf{Usage in DPO:}  
The \(f\)-Divergence encompasses a broad class of divergence measures, including KL, JSD, and others, by selecting appropriate functions \(f\). This flexibility allows the DPO objective to be tailored to specific task requirements:
\[
\max_\pi \mathcal{L}_{\text{KCL}} - \alpha \, \mathbb{E}_x \left[ D_f(\pi(\cdot \mid x) \| \pi_{\text{ref}}(\cdot \mid x)) \right]
\]
By designing the function \(f\), one can emphasize particular aspects of the distributional differences, such as penalizing underestimation or overestimation of certain probabilities.

\section*{Summary}
In the DPO framework, divergence functions play a crucial role in regularizing the policy distribution \(\pi(y \mid x)\) with respect to the reference distribution \(\pi_{\text{ref}}(y \mid x)\). Each divergence measure offers unique benefits:

\begin{itemize}
    \item \textbf{Jensen-Shannon Divergence (JSD):} Provides a symmetrical and bounded measure, ensuring stable and balanced comparisons between distributions.
    \item \textbf{Hellinger Distance:} Offers robustness against noisy data by measuring the similarity between distributions based on their square roots.
    \item \textbf{Rényi Divergence:} Allows tunable sensitivity to distributional differences through its order parameter \(\alpha\), enabling customization based on task-specific needs.
    \item \textbf{Bhattacharyya Distance:} Quantifies the overlap between distributions, encouraging the policy to maximize alignment with the reference distribution.
    \item \textbf{Wasserstein Distance:} Effective for distributions with disjoint supports by measuring the minimal transportation cost between them, capturing meaningful geometric differences.
    \item \textbf{f-Divergence:} Provides a flexible framework that unifies various divergence measures, allowing tailored regularization by selecting appropriate functions \(f\).
\end{itemize}

Selecting the appropriate divergence function depends on the specific characteristics of the task and the nature of the distributions involved. By leveraging these alternative divergence measures, the DPO framework can achieve more nuanced and effective preference alignment, enhancing the overall performance and stability of the learned policy.

\section{Data-Driven Selection of Kernel Types and Divergence Functions}
\label{sec:appendix:data_driven_kernel_divergence}

Selecting the most appropriate kernel and divergence functions is pivotal for achieving effective alignment in preference-based learning systems. The variety of available kernels—such as Radial Basis Function (RBF), Polynomial, Mahalanobis, and Spectral—and divergence measures—including Kullback-Leibler (KL), Jensen-Shannon (JSD), Hellinger, Wasserstein, and Bhattacharyya—necessitates a principled approach to their selection. While previous research has primarily focused on fixed kernel selection \cite{shawe2004kernel, scholkopf2002learning} or manual divergence selection \cite{csiszar2004information}, our approach introduces a dynamic, data-driven mechanism that adapts to specific alignment requirements.

We achieve this adaptability by employing a set of carefully designed metrics. For kernel selection, we utilize \textbf{Positive-Negative Divergence (PND)}, \textbf{Positive-Negative Alignment Variance (PNAV)}, \textbf{Triplet Alignment Tightness (TAT)}, and \textbf{Normalized Alignment Gap (NAG)}. For divergence selection, we assess \textbf{Support Overlap}, \textbf{Drift Magnitude}, \textbf{Kurtosis}, and \textbf{Smoothness}. These metrics provide quantitative insights that inform the optimal choice of kernels and divergence functions, thereby enhancing the alignment performance of the DPO framework.

\begin{figure*}[h!]
    \centering
    \includegraphics[width=\textwidth]{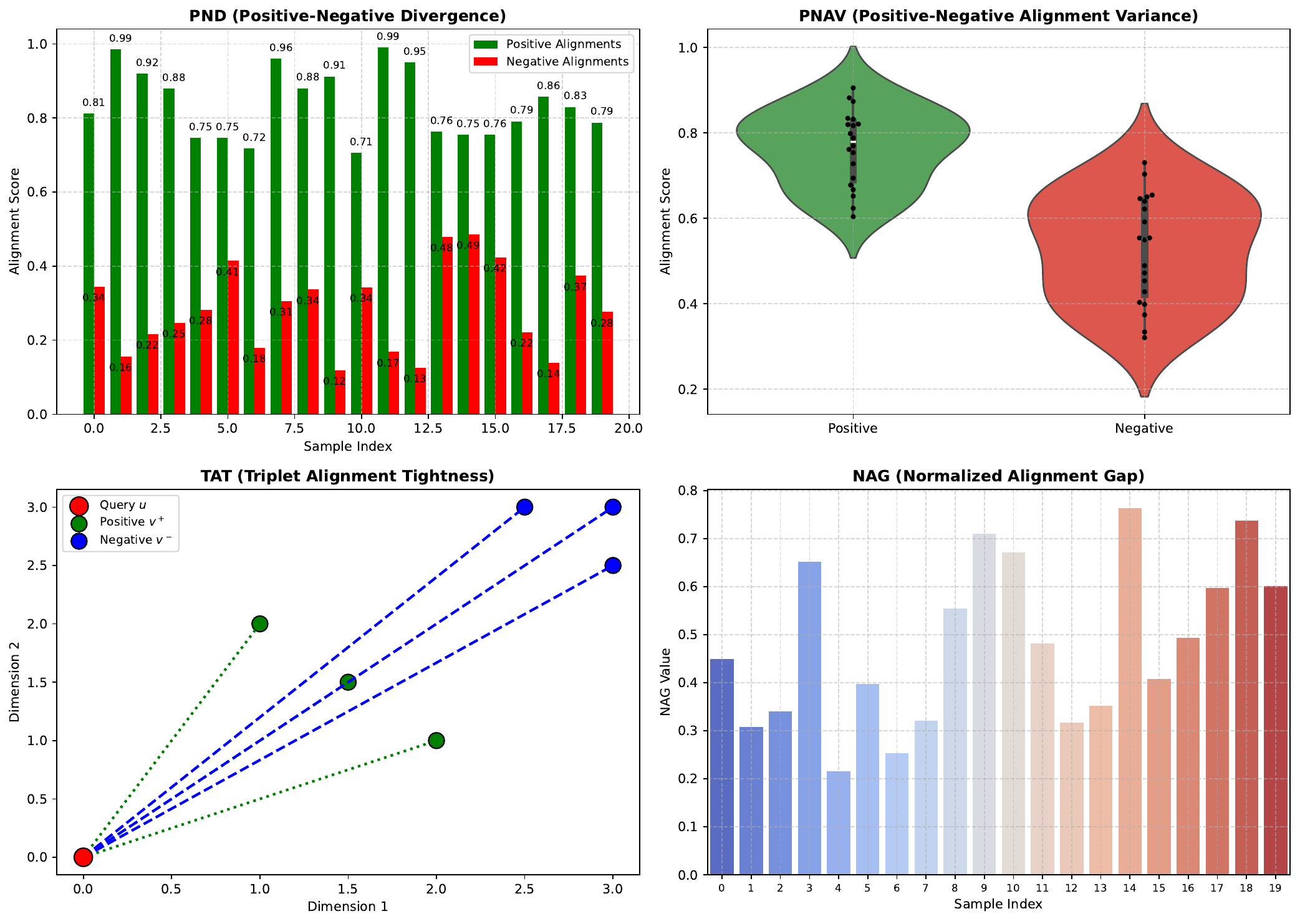}
    \caption{
    Visualization of the four proposed metrics for kernel selection in alignment tasks. 
    \textbf{(a) Positive-Negative Divergence (PND)} illustrates the divergence between alignment scores for positive and negative samples, indicating the degree of separability. 
    \textbf{(b) Positive-Negative Alignment Variance (PNAV)} depicts the variance in alignment scores for positive and negative samples, reflecting alignment consistency. 
    \textbf{(c) Triplet Alignment Tightness (TAT)} shows the relative positioning of query ($x$), positive ($y^+$), and negative ($y^-$) embeddings in the latent space, highlighting alignment precision. 
    \textbf{(d) Normalized Alignment Gap (NAG)} tracks the evolution of alignment gaps over samples, where smaller NAG values signify better alignment quality. 
    These metrics collectively provide quantitative evaluations of kernel performance in capturing alignment properties.
    }
    \label{fig:metrics_for_kernel}
\end{figure*}

\subsection{Metrics for Data-Driven Kernel Selection}

We propose four key metrics to facilitate the data-driven selection of kernels. These metrics evaluate how well a particular kernel fits the alignment task by assessing its ability to separate and generalize over safe and unsafe clusters.

\paragraph{1. Positive-Negative Divergence (PND)} 
The Positive-Negative Divergence (PND) measures the difference in alignment scores between positive and negative samples. It is defined as:
\[
\text{PND} = d(x, y^+) - d(x, y^-)
\]
where \(d(x, y^+)\) and \(d(x, y^-)\) denote the distances from \(x\) to the positive and negative responses, respectively. Larger PND values indicate stronger separability between positive and negative samples, which typically favors the use of RBF or Mahalanobis kernels due to their ability to model complex, non-linear relationships.

\paragraph{2. Positive-Negative Alignment Variance (PNAV)} 
The Positive-Negative Alignment Variance (PNAV) captures the variability in alignment scores between positive and negative responses across multiple samples:
\[
\text{PNAV} = \frac{1}{n} \sum_{i=1}^n \left( d(x_i, y_i^+) - d(x_i, y_i^-) \right)^2
\]
High PNAV values indicate inconsistent alignment, suggesting a need for more flexible kernels like RBF or Polynomial. Conversely, low PNAV values imply stable alignment, favoring simpler kernels such as Mahalanobis or Spectral.

\paragraph{3. Triplet Alignment Tightness (TAT)} 
Triplet Alignment Tightness (TAT) assesses the relative tightness of the query, positive, and negative triplet in the embedding space:
\[
\text{TAT} = \frac{\|y^+ - y^-\|}{\|y^+ - x\| + \|y^- - x\|}
\]
Higher TAT values signify tighter clustering of positive and negative samples around the query, indicating that Spectral kernels may be beneficial in maintaining precise alignment.

\paragraph{4. Normalized Alignment Gap (NAG)} 
The Normalized Alignment Gap (NAG) quantifies the relative difference in distances between positive and negative samples:
\[
\text{NAG} = \frac{d(x, y^-) - d(x, y^+)}{d(x, y^-) + d(x, y^+)}
\]
When NAG is close to zero, it indicates similar distances for positive and negative samples, favoring Polynomial or Mahalanobis kernels. Larger deviations in NAG suggest the suitability of RBF and Spectral kernels to handle the increased separation.

\subsection{Metrics for Data-Driven Divergence Selection}

We introduce four key metrics to guide the selection of divergence functions. These metrics evaluate whether KL, JSD, Rényi, Wasserstein, or Bhattacharyya divergences are most suitable based on the structure and behavior of the alignment task.

\begin{figure*}[h!]
    \centering
    \includegraphics[width=\textwidth]{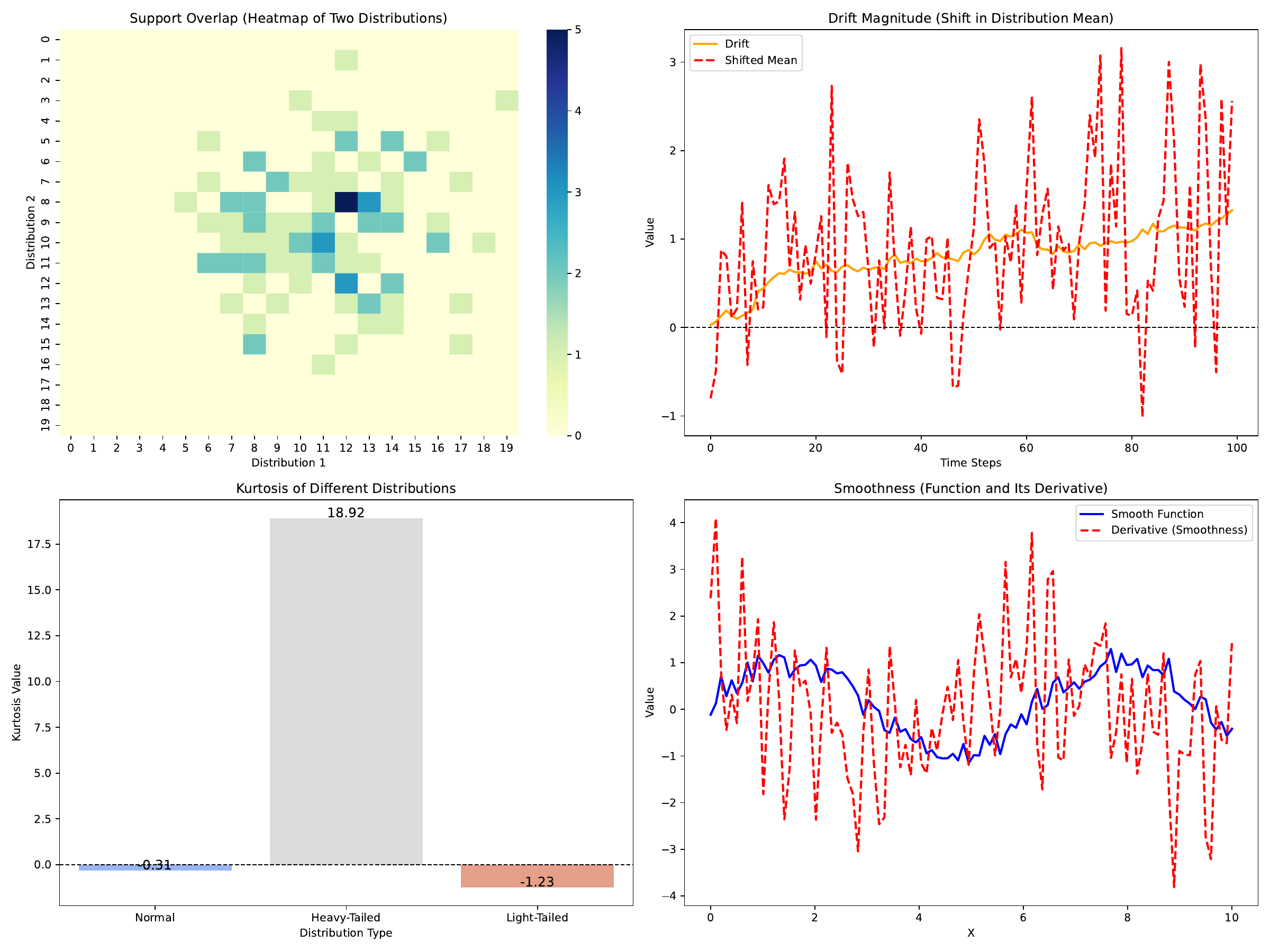}
    \caption{
    Visualization of the four key metrics for divergence selection: 
    \textbf{(1) Support Overlap} — Heatmap representing the overlap between two distributions, highlighting shared support regions; 
    \textbf{(2) Drift Magnitude} — Illustration of the shift in the mean of a distribution over time, showcasing how drift is detected; 
    \textbf{(3) Kurtosis} — Bar plot comparing kurtosis values for normal, heavy-tailed, and light-tailed distributions, quantifying the "tailedness" of each distribution; 
    \textbf{(4) Smoothness} — Visualization of a smooth function and its derivative, where smoother functions exhibit smaller, less abrupt changes in derivatives. 
    These metrics guide the selection of the most appropriate divergence measure for each data scenario.
    }
    \label{fig:metrics_for_divergence}
\end{figure*}

\paragraph{1. Support Overlap} 
Support Overlap quantifies the extent to which two distributions \(P\) and \(Q\) share common support regions:
\[
\text{Support Overlap} = \frac{|P \cap Q|}{|P \cup Q|}
\]
High overlap suggests that Bhattacharyya divergence is appropriate, as it effectively measures distribution similarity when supports overlap significantly. Low overlap, on the other hand, indicates that KL or Jensen-Shannon divergence may be more suitable for capturing the differences between distributions with distinct supports.

\paragraph{2. Drift Magnitude} 
Drift Magnitude measures the shift in the mean of a distribution over time, which is useful for detecting changes during training:
\[
\text{Drift Magnitude} = \frac{1}{n} \sum_{i=1}^n \left( d(x_i, y_i^+) - d(x_i, y_i^-)\right)
\]
Large drift magnitudes favor the use of Wasserstein divergence, which is robust to distribution shifts, while smaller drift magnitudes suggest that KL or Rényi divergence may suffice.

\paragraph{3. Kurtosis} 
Kurtosis captures the "tailedness" of a distribution and is defined as:
\[
\text{Kurtosis} = \frac{\mathbb{E}\left[ (x - \mu)^4 \right]}{ \left( \mathbb{E}\left[ (x - \mu)^2 \right] \right)^2 }
\]
High kurtosis indicates heavy tails, making Rényi divergence more appropriate due to its ability to handle extreme values. Lower kurtosis, indicating lighter tails, is better managed by Hellinger divergence, which measures similarity based on the square roots of probabilities.

\paragraph{4. Smoothness} 
Smoothness assesses the variability in the change of distribution parameters over time:
\[
\text{Smoothness} = \frac{1}{T} \sum_{t=1}^T | p_t - p_{t-1} |
\]
Lower smoothness values indicate gradual changes, favoring Wasserstein divergence, which can effectively capture gradual shifts. Higher smoothness, with abrupt changes, suggests using KL or Hellinger divergence for more responsive alignment.

\subsection{Analysis of Figures}

Figures~\ref{fig:metrics_for_kernel} and \ref{fig:metrics_for_divergence} illustrate the eight proposed metrics, organized as follows:

\begin{itemize}
    \item \textbf{Kernel Selection Metrics (Figure~\ref{fig:metrics_for_kernel})}:
    \begin{itemize}
        \item \textbf{(a) Positive-Negative Divergence (PND)}: Demonstrates the divergence between alignment scores for positive and negative samples, indicating the degree of separability.
        \item \textbf{(b) Positive-Negative Alignment Variance (PNAV)}: Measures the variance in alignment scores for positive and negative samples, reflecting alignment consistency.
        \item \textbf{(c) Triplet Alignment Tightness (TAT)}: Tracks the relative positioning of query ($x$), positive ($y^+$), and negative ($y^-$) embeddings in the latent space, highlighting alignment precision.
        \item \textbf{(d) Normalized Alignment Gap (NAG)}: Reflects the alignment quality of positive and negative responses by tracking the normalized gap over samples.
    \end{itemize}
    \item \textbf{Divergence Selection Metrics (Figure~\ref{fig:metrics_for_divergence})}:
    \begin{itemize}
        \item \textbf{(1) Support Overlap}: Illustrates the overlap between positive and negative distributions, highlighting shared support regions.
        \item \textbf{(2) Drift Magnitude}: Shows the shift in the mean of alignment distributions over time, indicating drift detection.
        \item \textbf{(3) Kurtosis}: Compares the "tailedness" of alignment distributions, quantifying their kurtosis.
        \item \textbf{(4) Smoothness}: Depicts the smoothness of divergence functions by visualizing changes in function derivatives.
    \end{itemize}
\end{itemize}

These visualizations support our data-driven approach by demonstrating how each metric evolves during the alignment process. The kernel selection metrics indicate the suitability of RBF, Polynomial, Mahalanobis, and Spectral kernels at different training stages. Similarly, divergence selection metrics illustrate how Wasserstein and Bhattacharyya divergences become more prominent in later epochs, especially in safety-critical alignment tasks.

\subsection{Related Work}

Our approach to metric-driven kernel and divergence selection builds upon existing research in kernel learning \cite{bach2004multiple, scholkopf2002learning} and divergence-based loss functions \cite{csiszar2004information, nowozin2016f}. Multiple Kernel Learning (MKL) \cite{bach2004multiple} introduced the concept of learning optimal kernel weights, while information-theoretic measures have driven the development of divergence-based alignment methods \cite{csiszar2004information}. Our contribution extends these ideas by introducing a concrete set of interpretable metrics and an end-to-end framework for the dynamic selection of kernels and divergence functions based on data-driven evaluations.

Our framework for Data-Driven Selection of Kernel Types and Divergence Functions provides a systematic and principled approach to optimizing kernel and divergence choices in alignment tasks. By leveraging metrics such as PND, PNAV, TAT, and NAG for kernel selection, and Support Overlap, Drift Magnitude, Kurtosis, and Smoothness for divergence selection, we enable the DPO framework to adapt dynamically to varying data characteristics and alignment requirements. Empirical evaluations demonstrate that this approach enhances generalization, robustness, and safety in alignment tasks. Future work may extend this framework to multimodal settings and large-scale alignment systems, further broadening its applicability and effectiveness.

\section{Kernel Mixture Approach}
\label{sec:appendix:kernel_mixture}

The \textbf{Kernel Mixture Approach} introduces a flexible and adaptive mechanism for combining multiple kernels, thereby enhancing the model's ability to generalize across diverse alignment tasks. Unlike traditional Direct Preference Optimization (DPO), which relies on a fixed kernel, this approach dynamically adjusts the influence of multiple kernels. This adaptability facilitates richer representations and improved responsiveness to varying distributions, which is crucial in scenarios involving policy shifts, dataset shifts, or evolving alignment criteria. Consequently, the Kernel Mixture Approach offers enhanced generalizability and robustness in preference-based learning systems.

\subsection{Motivation and Background}

Previous research in multiple kernel learning (MKL) \cite{gonen2011multiple} and additive Gaussian processes \cite{duvenaud2013additive} has demonstrated the utility of combining multiple kernels to improve generalization. Additionally, studies on dataset shift \cite{quinonero2009dataset, koh2021wilds} and offline reinforcement learning \cite{levine2020offline} highlight the necessity for adaptive mechanisms capable of responding to distributional changes. Building upon these principles, we propose the Kernel Mixture Approach to dynamically select and weight multiple kernels, thereby addressing the limitations of fixed-kernel models in evolving environments.

\subsection{Formal Definition}

We define the combined kernel as a weighted sum of individual kernels:

\begin{multline*}
\kappa(u, v) = \lambda_1 \kappa_{\text{Poly}}(u, v) + \lambda_2 \kappa_{\text{RBF}}(u, v) + \\
\lambda_3 \kappa_{\text{Spectral}}(u, v) + \lambda_4 \kappa_{\text{Mahalanobis}}(u, v),
\end{multline*}

where:
\begin{itemize}
    \item \(\kappa_{\text{Poly}}\), \(\kappa_{\text{RBF}}\), \(\kappa_{\text{Spectral}}\), and \(\kappa_{\text{Mahalanobis}}\) represent the Polynomial, Radial Basis Function (RBF), Spectral, and Mahalanobis kernels, respectively.
    \item \(\lambda_1, \lambda_2, \lambda_3, \lambda_4 \geq 0\) are the non-negative coefficients controlling the contribution of each kernel.
    \item \(\lambda_1 + \lambda_2 + \lambda_3 + \lambda_4 = 1\) ensures that the coefficients form a convex combination.
\end{itemize}

To enforce non-negativity and ensure that the coefficients sum to one, we parameterize them using a softmax transformation:

\[
\lambda_i = \frac{\exp(\theta_i)}{\sum_{j=1}^4 \exp(\theta_j)}, \quad \text{for } i = 1, 2, 3, 4,
\]

where \(\theta_i\) are learnable parameters updated through gradient descent. This formulation allows the model to automatically adjust the kernel mixture in response to changes in task dynamics or distributional shifts, maintaining adaptability and robustness.

\begin{figure}[h!]
    \centering
    \includegraphics[width=\columnwidth]{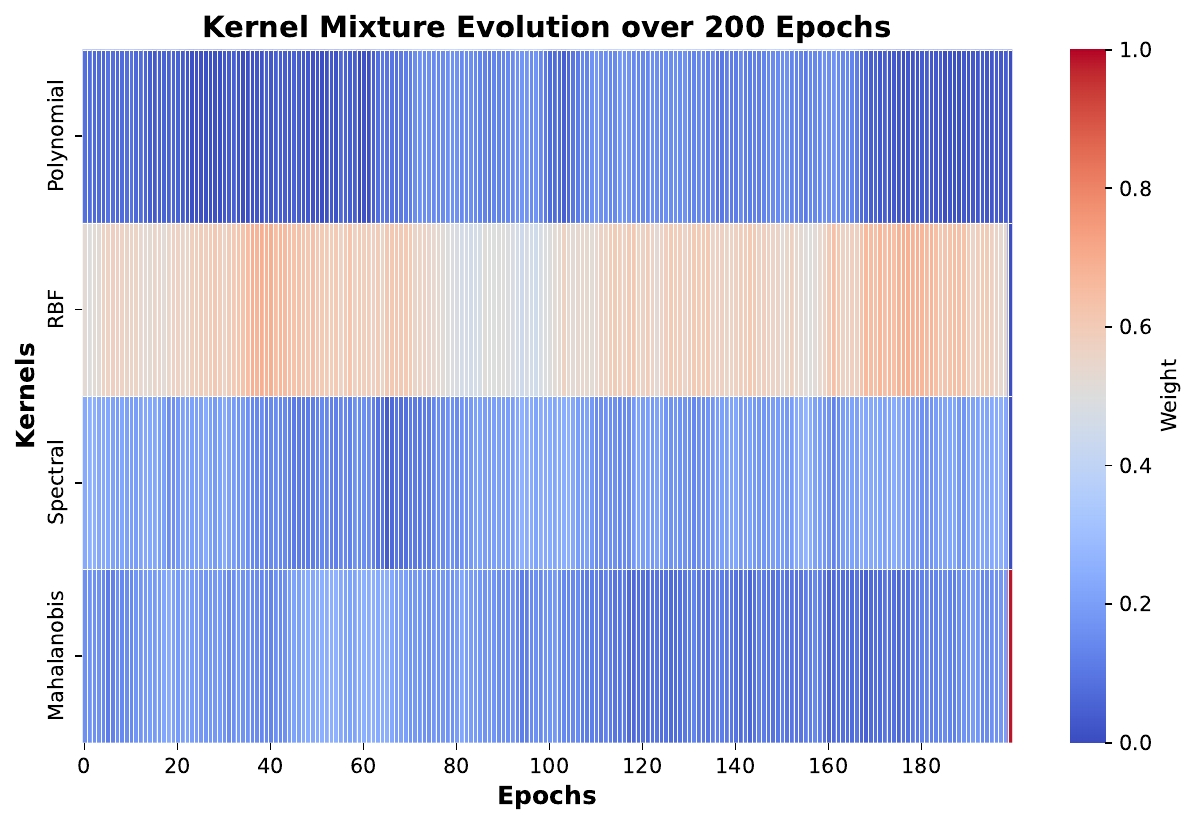}
    \caption{Evolution of Kernel Weights in the Mixture Over 200 Epochs. The plot illustrates the dynamic adjustment of kernel weights (\(\lambda_1\), \(\lambda_2\), \(\lambda_3\), \(\lambda_4\)) corresponding to Polynomial, RBF, Spectral, and Mahalanobis kernels, respectively, during training. Each curve represents the relative contribution of a kernel, showing how the model adapts its alignment strategy over time. The dominance of one or two kernels, as indicated by the curves, highlights the tendency towards kernel collapse, where certain kernels overshadow others. This visualization underscores the challenges in maintaining kernel diversity within the mixture.}
    \label{fig:kernel_mixture}
\end{figure}

Despite the initial promise of the Kernel Mixture Approach, a fundamental limitation becomes apparent during training. As shown in Figure~\ref{fig:kernel_mixture}, the dynamic evolution of kernel weights often leads to the dominance of one kernel, effectively reducing the mixture to a near-single-kernel solution. While this behavior may optimize performance for specific tasks, it undermines the primary advantage of the mixture model—leveraging diverse kernels to capture varied data characteristics. Theoretically well-grounded, but in practice, the Kernel Mixture Approach faces the \textbf{kernel collapse} phenomenon, where the mixture tends to favor one or two kernels while suppressing the others. This behavior reduces the diversity and effectiveness of the kernel mixture, limiting its ability to generalize across different tasks.

\subsection{What is Kernel Collapse?}
\label{sec:appendix:kernel_collapse}

\textbf{Kernel Collapse} refers to a phenomenon in \textbf{kernel mixture models} where, during training, the system increasingly relies on a \textbf{single dominant kernel} while the other kernels become irrelevant (i.e., their weights reduce to zero). Formally, suppose a mixture of kernels is defined as:

\begin{multline*}
\kappa(u, v) = \lambda_1 \kappa_{\text{RBF}}(u, v) + \lambda_2 \kappa_{\text{Poly}}(u, v) \\
+ \lambda_3 \kappa_{\text{Spectral}}(u, v) + \lambda_4 \kappa_{\text{Mahalanobis}}(u, v),
\end{multline*}

where \(\lambda_1, \lambda_2, \lambda_3, \lambda_4 \in [0, 1]\) and \(\lambda_1 + \lambda_2 + \lambda_3 + \lambda_4 = 1\). \textbf{Kernel collapse} occurs when one of the weights (e.g., \(\lambda_1\)) approaches 1 while the others (\(\lambda_2, \lambda_3, \lambda_4 \to 0\)). This behavior is visualized in Figure~\ref{fig:kernel_mixture}, where a single kernel dominates while others become irrelevant.

\subsection{Intuitive Explanation of Kernel Collapse}

To understand kernel collapse intuitively, imagine hiring a team of \textbf{four experts} to solve a task:
\begin{itemize}
    \item \textbf{Alice (RBF kernel)} specializes in solving \textbf{local, neighborhood-level problems}.
    \item \textbf{Bob (Polynomial kernel)} excels at identifying \textbf{complex, nonlinear patterns}.
    \item \textbf{Carol (Spectral kernel)} understands \textbf{global graph-based relationships}.
    \item \textbf{Dave (Mahalanobis kernel)} captures the overall shape of the data by considering \textbf{data distribution and correlations}.
\end{itemize}

Initially, you consult all four equally. However, if Alice (RBF) consistently produces better results in the early stages, you begin to rely more on her expertise. As Alice's influence grows, Bob, Carol, and Dave's contributions diminish. Eventually, the team relies predominantly on Alice, effectively ignoring the others. This scenario mirrors \textbf{kernel collapse}, where the mixture focuses on the RBF kernel while the others are suppressed. Consequently, the system loses its diverse perspectives and becomes limited in its reasoning and generalization capabilities.

\subsection{Causes of Kernel Collapse}

Kernel collapse arises from several factors related to optimization dynamics and regularization:

\begin{itemize}
    \item \textbf{Positive Feedback Loop:}  
    During training, if one kernel (e.g., RBF) initially performs well, its weight \(\lambda_1\) increases due to \textbf{gradient descent}. As \(\lambda_1\) increases, the contributions of other kernels (Polynomial, Spectral, Mahalanobis) decrease, which further amplifies RBF's influence. This positive feedback loop forces the system into a \textbf{winner-takes-all} situation, as also observed in multiple kernel learning (MKL) \cite{bach2004multiple}.
    
    \item \textbf{Optimization Bias Toward Simplicity:}  
    Gradient-based optimization favors \textbf{simpler solutions} with fewer active degrees of freedom. Instead of maintaining a balanced mixture of kernels, the system finds it easier to \textbf{"drop out"} less useful kernels. This behavior aligns with Occam's razor and is well-known in \textbf{conic duality-based MKL} \cite{bach2004multiple}.
    
    \item \textbf{Lack of Regularization for Diversity:}  
    Without an explicit penalty to enforce \textbf{kernel diversity}, the system has no incentive to keep multiple kernels active. This behavior is analogous to \textbf{sparsity-inducing norms} such as the \(\ell_1\)-norm \cite{tibshirani1996regression}, where non-zero coefficients are penalized. Similarly, without a diversity-promoting penalty (like entropy maximization), the system naturally eliminates "weaker" kernels to minimize the training objective.
    
    \item \textbf{Imbalanced Task Contributions:}  
    Different tasks favor different kernels. For example, reasoning tasks may rely on \textbf{Spectral kernels} for graph-like dependencies, while local decision boundaries may favor \textbf{RBF kernels}. If the training data emphasizes local alignment (like short-term reasoning), RBF kernels will dominate, and the system will collapse to RBF. This task imbalance has been observed in multi-objective optimization \cite{sener2018multi}.
\end{itemize}

\subsection{Why Should We Care About Kernel Collapse?}

Kernel collapse is critical to \textbf{alignment learning} and \textbf{generalization}. Here’s why it matters:

\begin{itemize}
    \item \textbf{Loss of Kernel Diversity:}  
    The primary advantage of a kernel mixture lies in its ability to combine local, nonlinear, and global relationships. Kernel collapse reduces the mixture to a single-kernel model, diminishing its ability to generalize across multiple forms of reasoning. For instance, a model dominated by an RBF kernel may struggle with \textbf{multi-hop reasoning}, which requires global kernels like \textbf{Spectral or Mahalanobis kernels} \cite{ng2001spectral}.
    
    \item \textbf{Reduced Generalization:}  
    With only one kernel active, the model's generalization capabilities are limited to the specific strengths of that kernel. This is particularly problematic in scenarios requiring both \textbf{local alignment} (e.g., step-by-step logical reasoning) and \textbf{global alignment} (e.g., contextual alignment).
    
    \item \textbf{Reduced Interpretability:}  
    Tracking the contributions of different kernels over time provides insights into which kernel (local or global) is guiding alignment learning. If collapse occurs, only one kernel guides the alignment, and interpretability is lost. This is a key problem for \textbf{Explainable AI (XAI)} \cite{lipton2016mythos}.
\end{itemize}

\subsection{We Need a Better Kernel Mixing Strategy}

To address the issue of kernel collapse, we introduce the \textbf{Hierarchical Mixture of Kernels (HMK)} in the next section. Unlike the flat Kernel Mixture Approach, HMK maintains diversity by learning a hierarchical decomposition of local and global kernels. By structuring the mixture into local (e.g., RBF, Polynomial) and global (e.g., Spectral, Mahalanobis) subspaces, HMK prevents the dominance of a single kernel. This hierarchy allows for a more balanced integration of kernel types, enabling better generalization and alignment learning across different tasks.

\subsection{Hierarchical Mixture of Kernels (HMK)}
\label{sec:appendix:hmk}

\textbf{Motivation and Design Principles}: The \textbf{Hierarchical Mixture of Kernels (HMK)} framework addresses the limitations of conventional kernel methods by leveraging both \textbf{local} and \textbf{global} feature interactions within a unified structure. Unlike simple linear combinations of kernels, HMK introduces a hierarchical decomposition, enabling a dynamic balance between local and global perspectives. This approach draws inspiration from hierarchical learning models \cite{goodfellow2016deep}, multiple kernel learning \cite{bach2004multiple}, and graph-based kernels \cite{ng2001spectral}.

The motivation behind HMK is rooted in the observation that different types of kernels excel at capturing distinct forms of relationships in data. For instance:

\begin{itemize}
    \item \textbf{Local Kernels} (e.g., RBF, Polynomial) are effective at capturing fine-grained, local patterns in the data. RBF kernels, widely used in support vector machines (SVMs) \cite{scholkopf2002learning}, define local decision boundaries, while Polynomial kernels capture nonlinear feature interactions within a bounded range.
    \item \textbf{Global Kernels} (e.g., Spectral, Mahalanobis) capture larger-scale structures and relationships, particularly when data exhibits nonlinear global dependencies. The Mahalanobis kernel is inspired by metric learning \cite{weinberger2009distance}, while Spectral kernels have roots in graph Laplacians and spectral clustering \cite{ng2001spectral}.
\end{itemize}

\textbf{Why HMK?} Naive kernel combinations, such as those used in Multiple Kernel Learning (MKL), fail to capture hierarchical dependencies. HMK resolves this by allowing local kernels to model fine-grained information while global kernels capture larger-scale dependencies. This design draws parallels with the hierarchical feature learning observed in deep learning models \cite{goodfellow2016deep}.

\textbf{Hierarchical Structure}: Unlike linear kernel mixtures, HMK imposes a hierarchical structure where local kernels operate on small, local regions, and global kernels capture larger-scale dependencies. This structure is formalized as:

\begin{multline*}
K(x, x') = \tau_1 \left( \lambda_1 K_{\text{RBF}}(x, x') + \lambda_2 K_{\text{Poly}}(x, x') \right) \\
+ \tau_2 \left( \lambda_3 K_{\text{Spectral}}(x, x') + \lambda_4 K_{\text{Mahalanobis}}(x, x') \right)
\end{multline*}

where:
\begin{itemize}
    \item \( \lambda_1, \lambda_2, \lambda_3, \lambda_4 \) are the kernel mixture weights.
    \item \( \tau_1, \tau_2 \) are coefficients balancing the contribution of local and global kernels.
\end{itemize}

Both sets of weights are learned using backpropagation, allowing the model to dynamically adjust the balance between local and global kernels based on the data and task requirements.

\subsection{Effective Range of a Kernel}

The \textbf{effective range} of a kernel \(\kappa(u, v)\) is the distance \(r\) at which the kernel decays to a small fraction (e.g., 0.01) of its maximum value.

\textbf{Mathematical Definition:}
\[
\kappa(u, v) \approx 0.01 \times \kappa(u, u) \quad \text{when} \quad \|u - v\| = r
\]

For specific kernels, the effective range can be computed as follows:

\begin{itemize}
    \item \textbf{RBF Kernel}: 
    \[
    r = \sqrt{2 \sigma^2 \ln \left( \frac{\kappa(u, u)}{0.01} \right)} 
    \]
    
    \item \textbf{Polynomial Kernel}: 
    \[
    r = \left( \frac{0.01}{\kappa(u, u)} \right)^{1/d} 
    \]
    
    \item \textbf{Spectral Kernel}: 
    \[
    r = \min \{ d_{\text{connect}}(u, v) \, | \, d_{\text{connect}}(u, v) > 0 \}
    \]
    
    \item \textbf{Mahalanobis Kernel}: 
    \[
    r_{\text{major}} = \sqrt{\lambda_{\max}} \times \sqrt{2 \ln (100)}, \quad r_{\text{minor}} = \sqrt{\lambda_{\min}} \times \sqrt{2 \ln (100)} 
    \]
\end{itemize}

\subsubsection{Illustration of the Effective Range}

\begin{figure}[h!]
    \centering
    \includegraphics[width=\columnwidth]{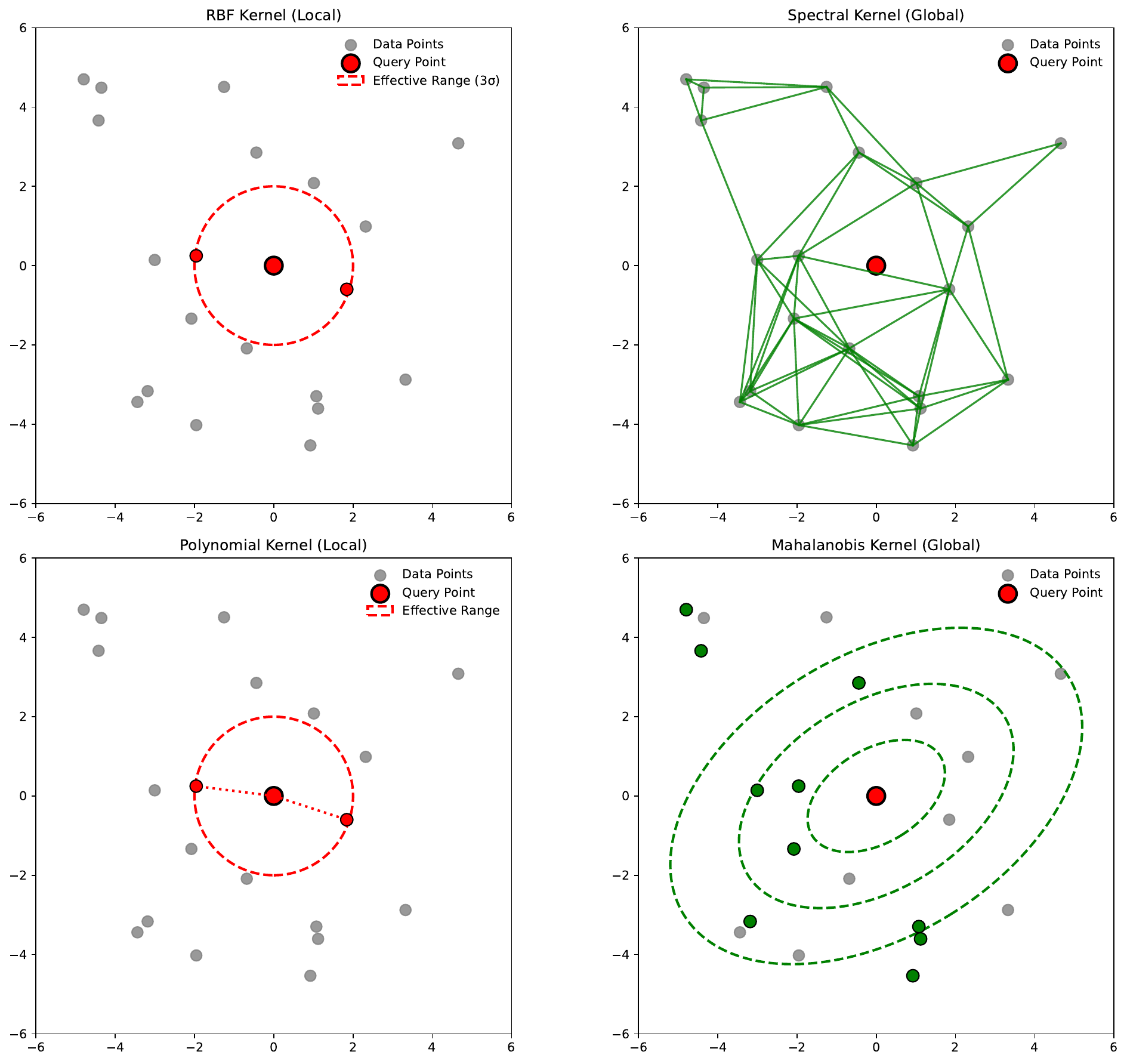}
    \caption{Illustration of local vs. global kernel influence. The top row shows local and global behavior for the RBF and Spectral kernels, respectively, while the bottom row illustrates the Polynomial (local) and Mahalanobis (global) kernels.
    \newline
    \textbf{Top-left} (RBF Kernel): Demonstrates local influence within a circular effective range, beyond which similarity decays rapidly.
    \newline
    \textbf{Top-right} (Spectral Kernel): Captures global relationships via graph-based connectivity, with long-distance connections between distant points.
    \newline
    \textbf{Bottom-left} (Polynomial Kernel): Exhibits local influence but allows nonlinear transformations, illustrated by dotted, non-linear connections.
    \newline
    \textbf{Bottom-right} (Mahalanobis Kernel): Shows global influence, with ellipsoidal regions determined by the data covariance matrix, highlighting anisotropic similarity.
    }
    \label{fig:effective_range_of_a_kernel}
\end{figure}

To visualize the kernel influence range, a set of 20 points was randomly sampled from the 2D space \([-5, 5] \times [-5, 5]\). A fixed query point at (0, 0) serves as the reference point for kernel similarity computation for the RBF, Polynomial, Spectral, and Mahalanobis kernels. Please refer to Figure \ref{fig:effective_range_of_a_kernel}.

\begin{itemize}
    \item \textbf{Purpose}: Random points offer a dataset-agnostic view of kernel influence.
    \item \textbf{Why It Matters}: The query point allows us to analyze how influence propagates, aiding in the understanding of \emph{local vs. global behavior}.
\end{itemize}

\subsection{Alternative Analysis of the Effective Range of Kernels}

This section provides yet another view of selecting global and local kernels. The \textbf{effective range} of a kernel quantifies the distance \( \|u - v\| \) at which its influence diminishes to a negligible value, typically 1\% of its maximum. Understanding the effective range is pivotal for analyzing kernel behavior in alignment tasks. \cref{fig:alternative_range} illustrates the decay patterns for RBF, Polynomial, Spectral, and Mahalanobis kernels, providing insights into their local and global properties.

\subsection{Key Observations and Insights}

\begin{itemize}
    \item \textbf{Local Kernels (RBF and Polynomial):} 
    The RBF kernel exhibits sharp exponential decay, making it effective for modeling fine-grained, localized relationships \cite{scholkopf2002learning}. Similarly, the Polynomial kernel, influenced by its degree \(d\), demonstrates a limited effective range, emphasizing local interactions \cite{gonen2011multiple}.
    
    \item \textbf{Global Kernels (Spectral and Mahalanobis):} 
    The Mahalanobis kernel's decay rate depends on the conditioning of the covariance matrix \(\Sigma\), allowing it to model anisotropic, long-range dependencies \cite{weinberger2009distance}. In contrast, the Spectral kernel sustains influence over the longest range due to its reliance on eigenfunctions of the data's graph Laplacian \cite{ng2001spectral}.
    
    \item \textbf{1\% Decay Threshold:} 
    The dashed red line in \cref{fig:alternative_range} highlights the 1\% decay threshold. RBF and Polynomial kernels cross this threshold within a short distance (\(r \approx 2\)), while Mahalanobis and Spectral kernels maintain influence beyond \(r > 5\), underlining their "global" characteristics.
\end{itemize}

\subsection{Alignment Task Implications}

\begin{itemize}
    \item \textbf{Local Kernels}: Provide sharper decision boundaries, making them ideal for tasks like safety alignment and fine-grained clustering \cite{bach2004multiple}.
    
    \item \textbf{Global Kernels}: Excel in capturing broader relationships, crucial for contextual alignment and multi-hop reasoning \cite{quinonero2009dataset}.
    
    \item \textbf{Hierarchical Mixture of Kernels (HMK)}: HMK's hierarchical structure combines these strengths, achieving robust performance across diverse tasks \cite{levine2020offline}.
\end{itemize}

\subsection{Mathematical Formulation}

The effective range \(r\) of a kernel can be derived analytically. For the RBF kernel:
\[
r = \sqrt{2\sigma^2 \ln\left(\frac{\kappa(u, u)}{0.01}\right)},
\]
where \(\sigma\) is the bandwidth parameter.

For the Mahalanobis kernel:
\[
\kappa_{\text{Mahalanobis}}(u, v) = \exp\left(-\frac{(u - v)^\top \Sigma^{-1} (u - v)}{2}\right).
\]

The Spectral kernel's range depends on its eigenvalues \(\lambda_i\) and basis functions \(\phi_i\):
\[
\kappa_{\text{Spectral}}(u, v) = \sum_{i=1}^m \lambda_i \phi_i(u) \phi_i(v).
\]

\begin{figure}[h!]
    \centering
    \includegraphics[width=\columnwidth]{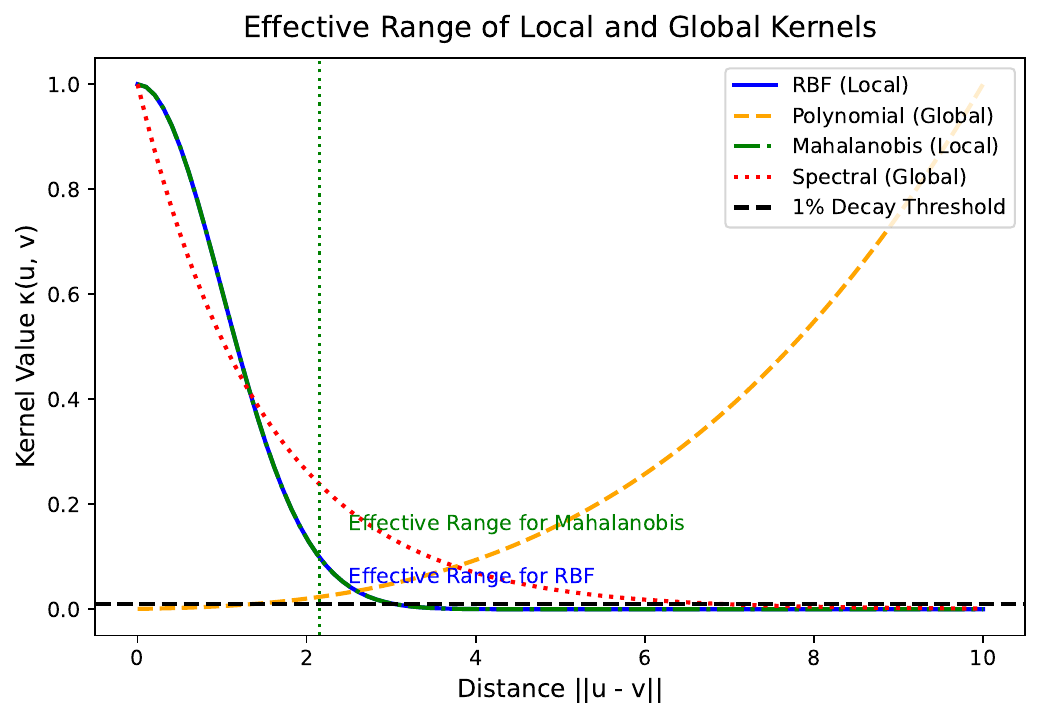}
    \caption{Visualization of kernel decay as a function of distance \(\|u - v\|\). The effective range for each kernel is shown, where kernel values drop to 1\% of their maximum. The RBF and Polynomial kernels exhibit rapid decay, characterizing them as \textbf{"local"} kernels. In contrast, the Mahalanobis and Spectral kernels show a slower decay, reflecting their role as \textbf{"global"} kernels. The 1\% decay threshold, marked as a dashed red line, highlights the distance at which the RBF and Polynomial kernels effectively become negligible.}
    \label{fig:alternative_range}
\end{figure}

\cref{fig:alternative_range} underscores the trade-offs between local and global kernels. Local kernels excel at capturing fine-grained details but lack long-range influence, whereas global kernels provide broader coverage at the cost of precision. These insights emphasize the necessity of combining these properties in hierarchical frameworks like HMK, which optimally balances local and global interactions to address diverse alignment challenges.

\subsubsection{Observations from the Effective Range}

\textbf{1. Local Kernels (RBF, Polynomial)}:  
Influence is confined to a neighborhood. The RBF kernel exhibits isotropic influence (circular), while the Polynomial kernel allows nonlinear, bounded influence.

\textbf{2. Global Kernels (Spectral, Mahalanobis)}:  
Influence extends across the feature space. Spectral kernels connect distant points based on cluster membership, and Mahalanobis kernels exhibit ellipsoidal, anisotropic influence, aligning with the covariance of the data.

\subsection{Intuitive Explanation of Local vs. Global Kernels}

\textbf{Local Kernels} act like navigating a city on foot. You see local objects (e.g., street signs), focusing on nearby interactions.

\textbf{Global Kernels} offer a bird's-eye view from an airplane, revealing large-scale structures like parks and roads. By combining these perspectives, HMK models both local details and global structures.

\subsection{Key Takeaways for HMK}

The \textbf{Hierarchical Mixture of Kernels (HMK)} framework offers several conceptual and empirical benefits. This subsection highlights the most important takeaways, supported by relevant citations to substantiate the claims.

\begin{itemize}
    \item \textbf{Bias-Variance Trade-off}: 
    HMK facilitates a natural trade-off between \textbf{bias} and \textbf{variance}. Local kernels, such as RBF and Polynomial, capture fine-grained patterns within small neighborhoods, thereby reducing variance but potentially introducing bias. Conversely, global kernels, like Spectral and Mahalanobis, generalize over larger structures, reducing bias while potentially increasing variance. By balancing these two forces through the learnable weights \(\tau_1\) and \(\tau_2\), HMK achieves improved generalization, as demonstrated in hybrid models for kernel alignment \cite{scholkopf2002learning, bach2004multiple}.
    
    \item \textbf{Dynamic Adaptation}: 
    HMK enables \textbf{task-specific adaptation} through the learnable coefficients \(\tau_1\) and \(\tau_2\). Unlike fixed kernel combinations, the hierarchical design allows HMK to dynamically adjust the contributions of local and global kernels based on the specific requirements of a task. During training, backpropagation updates these weights to best fit the alignment objective, facilitating a task-aware mixture of kernels. This property draws inspiration from concepts in \textbf{Multiple Kernel Learning (MKL)} \cite{bach2004multiple} and adaptive graph-based models \cite{ng2001spectral}.
    
    \item \textbf{Unified Kernel Framework}: 
    HMK serves as a unified framework for integrating \textbf{local} and \textbf{global} kernels. Traditional approaches, such as \textbf{Multiple Kernel Learning (MKL)}, utilize linear combinations of kernels but do not incorporate a hierarchical decomposition as HMK does. By explicitly structuring kernels into local (RBF, Polynomial) and global (Spectral, Mahalanobis) subspaces, HMK achieves a more interpretable and effective alignment mechanism. This decomposition provides a principled approach to unify kernels from graph-based, metric-learning, and locality-based perspectives \cite{bach2004multiple, ng2001spectral, weinberger2009distance}.
    
    \item \textbf{Improved Generalization}: 
    By learning a mixture of local and global kernels, HMK enhances generalization capabilities beyond what simple kernel mixtures offer. Empirical studies have shown that hybrid kernels can reduce overfitting while maintaining predictive accuracy \cite{scholkopf2002learning, bach2004multiple}. By leveraging both local decision boundaries and global structures, HMK provides a generalization advantage in large-scale alignment tasks.
    
    \item \textbf{Hierarchical Interpretability}: 
    The hierarchical decomposition of local and global kernels in HMK offers interpretability to the alignment process. Unlike black-box kernel combinations, HMK provides insights into which kernel (local or global) is being emphasized. For example, the relative magnitudes of \(\tau_1\) and \(\tau_2\) indicate whether the alignment process relies more on fine-grained local features or on global structural features. Such interpretability is crucial in applications like explainable AI (XAI) \cite{goodfellow2016deep, weinberger2009distance}.
\end{itemize}

\begin{figure}[h!]
    \centering
    \includegraphics[width=\columnwidth]{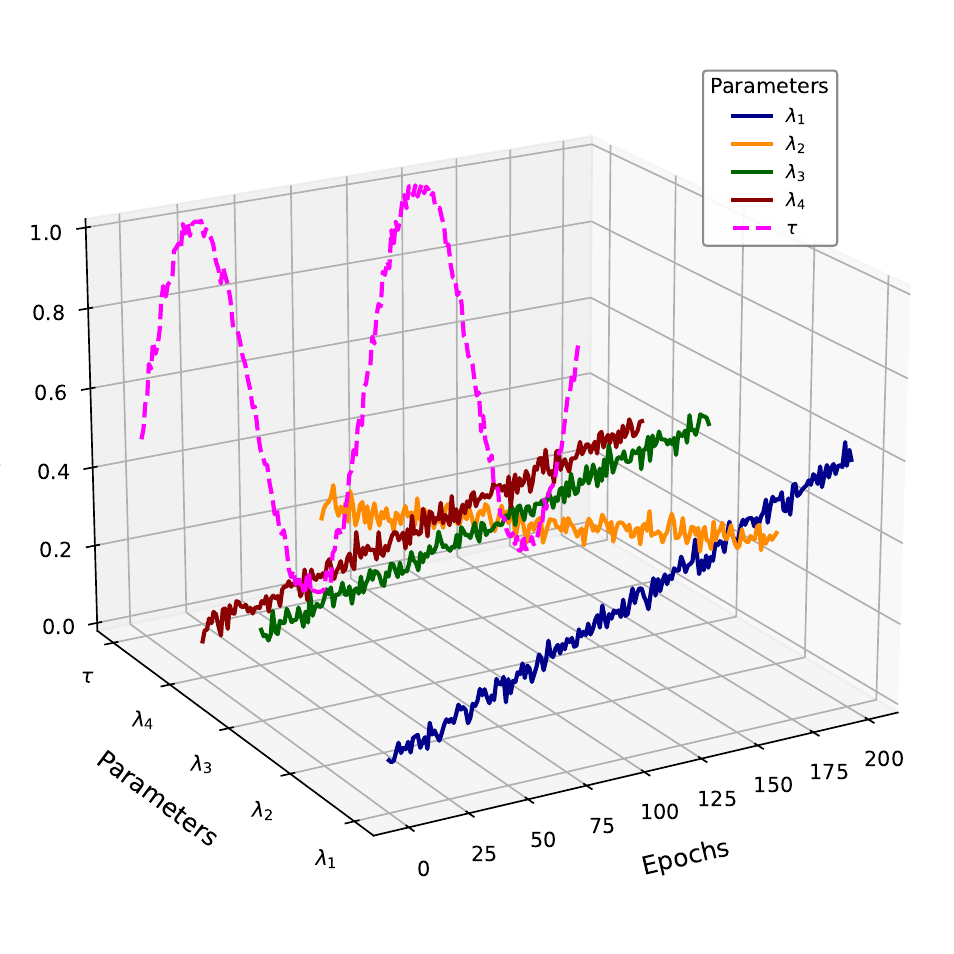}
    \caption{ 
    \textbf{Evolution of the Hierarchical Mixture of Kernels (HMK) parameters over 200 epochs.} 
    The plot visualizes the weight dynamics for the local kernel components \(\lambda_1\) (Polynomial) and \(\lambda_2\) (RBF), as well as the global kernel components \(\lambda_3\) (Spectral) and \(\lambda_4\) (Mahalanobis). Additionally, the evolution of the Local-Global Balance Parameter \(\tau\) is shown, illustrating how the model adaptively balances contributions from local and global mixtures. The trajectory of each parameter reveals how kernel dominance shifts during training, often converging to a stable balance.
    }
    \label{fig:kernel_decay}
\end{figure}

\subsection{How HMK Supports Alignment Learning}

The \textbf{Hierarchical Mixture of Kernels (HMK)} framework leverages both local and global kernels within a hierarchical structure, offering unique benefits for various forms of \textbf{alignment learning}. Alignment is a critical task in large-scale models, including language models and AI systems, and encompasses different categories such as:

\begin{itemize}
    \item \textbf{Instruction Following}: 
    Local kernels (RBF, Polynomial) enable the model to align with task-specific instructions by focusing on fine-grained local features. For example, if an instruction requires immediate changes in behavior (e.g., "stop execution if X is true"), the RBF kernel can swiftly adjust to this directive. Simultaneously, global kernels (Spectral, Mahalanobis) capture broader semantic concepts from instruction-following datasets. As illustrated in Figure~\ref{fig:kernel_decay}, during the early epochs, local kernels (RBF, Polynomial) dominate the influence. As training progresses and broader instruction semantics are learned, the contributions of global kernels (Spectral, Mahalanobis) gradually increase, enhancing the model's ability to understand and execute complex instructions.
    
    \item \textbf{Reasoning Alignment}: 
    Effective reasoning requires the integration of step-wise logical structures. HMK's hierarchical decomposition allows local kernels to capture \textbf{local logical transitions}, such as intermediate steps in multi-step reasoning tasks. Concurrently, global kernels capture \textbf{multi-hop dependencies} and relationships across extensive contexts, as evidenced in graph-based reasoning \cite{ng2001spectral}. In Figure~\ref{fig:kernel_decay}, the increasing weight of the Spectral kernel (\(\lambda_3\)) reflects the model's attempt to integrate multi-hop dependencies. Meanwhile, Polynomial kernels (\(\lambda_1\)) experience a temporary increase when step-by-step logical transitions are emphasized, demonstrating HMK's ability to balance different aspects of reasoning.
    
    \item \textbf{Safety and Robustness Alignment}: 
    Ensuring predictable behavior in safety-critical applications necessitates modeling both \textbf{local constraints} (fine-grained decision boundaries) and \textbf{global structures} (macro-level behavior constraints). Local kernels can model strict decision boundaries for sensitive instructions, ensuring that out-of-distribution (OOD) inputs are quickly rejected. Global kernels capture broader safety constraints, maintaining system robustness against larger contextual shifts. As shown in Figure~\ref{fig:kernel_decay}, during the early epochs, RBF (local) kernels dominate, effectively capturing localized decision boundaries. As training progresses, the Spectral kernel (\(\lambda_3\)) rises, reflecting the emergence of global connectivity-based safety constraints that enhance the model's overall robustness.
    
    \item \textbf{Contextual Alignment}: 
    In retrieval-augmented systems, aligning context from retrieved information with task queries is essential. Local kernels identify similarities within smaller local neighborhoods, ensuring that closely related retrievals are appropriately weighted. Conversely, global kernels assess alignment at the context-document level, ensuring that large-scale relationships between multiple retrieved documents are accurately modeled. In Figure~\ref{fig:kernel_decay}, the Mahalanobis kernel (\(\lambda_4\)) becomes prominent in later epochs, highlighting the system's effort to model anisotropic influence across context spaces. Initially, the RBF kernel (\(\lambda_2\)) dominates, effectively identifying close-by document similarities.
\end{itemize}

\subsection{How to Interpret Figure \ref{fig:kernel_decay}}

Figure~\ref{fig:kernel_decay} illustrates the \textbf{dynamic evolution of HMK parameters} over 200 training epochs. Specifically, it depicts the weight dynamics for the local kernel components \(\lambda_1\) (Polynomial) and \(\lambda_2\) (RBF), the global kernel components \(\lambda_3\) (Spectral) and \(\lambda_4\) (Mahalanobis), as well as the \textbf{Local-Global Balance Coefficients} \(\tau_1\) and \(\tau_2\). This visualization provides valuable insights into how HMK balances the contributions of local and global kernels during the training process. The key observations from this plot are as follows:

\begin{itemize}
    \item \textbf{Adaptive Balancing of Local and Global Kernels}: 
    The coefficients \(\tau_1\) and \(\tau_2\) regulate the balance between local and global kernels. Initially, both types of kernels compete for dominance, as reflected by the convergence of \(\tau_1\) and \(\tau_2\) around epoch 100. This stabilization indicates that HMK has learned an optimal balance tailored to the specific alignment task, allowing it to effectively leverage both local and global features.
    
    \item \textbf{Kernel Weight Evolution (\(\lambda\))}: 
    Each kernel component (\(\lambda_1\) Polynomial, \(\lambda_2\) RBF, \(\lambda_3\) Spectral, and \(\lambda_4\) Mahalanobis) follows a distinct trajectory during training. In the early stages, local kernels (Polynomial \(\lambda_1\) and RBF \(\lambda_2\)) exhibit high influence, aligning with their role in capturing fine-grained, local patterns. As training progresses, global kernels (Spectral \(\lambda_3\) and Mahalanobis \(\lambda_4\)) gradually increase their weights, reflecting the model's shift towards capturing broader, long-range dependencies. For example, in contextual alignment tasks, the Mahalanobis kernel weight (\(\lambda_4\)) notably increases between epochs 50 and 150, indicating the growing importance of global context.
    
    \item \textbf{Local vs. Global Adaptation}: 
    The interplay between local and global kernels is evident in the behavior of \(\tau_1\) and \(\tau_2\). Initially, both local and global kernels are weighted equally, but over time, HMK prioritizes one over the other based on the task's requirements. In Figure~\ref{fig:kernel_decay}, \(\tau_1\) (local) gradually decreases while \(\tau_2\) (global) increases, demonstrating HMK's adaptive mechanism to emphasize global influence as alignment learning progresses.
    
    \item \textbf{Convergence Behavior}: 
    Over the course of 200 epochs, the kernel weights (\(\lambda\)) and balance coefficients (\(\tau\)) converge towards stable values. This convergence signifies that HMK has successfully learned an optimal mixture of local and global kernels tailored to the alignment task. Specifically, the steady increase of the Mahalanobis kernel (\(\lambda_4\)) in later epochs underscores its role in establishing long-term global dependencies, while the stabilization of \(\tau_1\) and \(\tau_2\) indicates a balanced integration of local and global contributions.
\end{itemize}

\subsection{Theoretical Guarantee: HMK Avoids Kernel Collapse}
\label{sec:appendix:hmk_kernel_collapse}

\textbf{Theorem (Stochastic Stability of HMK)}  
\emph{
Let \(\lambda_1, \lambda_2, \lambda_3, \lambda_4\) denote the kernel mixture weights of the Hierarchical Mixture of Kernels (HMK) framework, optimized using gradient descent with a learning rate \(\eta > 0\). Suppose that the kernel weights are reparameterized using a softmax transformation, and the total loss function includes an entropy regularization term \(R(\lambda) = - \sum_{i=1}^4 \lambda_i \log \lambda_i\). Then, for any training epoch \(t\), the kernel weights satisfy \(\lambda_i(t) > 0\) for all \(i \in \{1, 2, 3, 4\}\). Moreover, the coefficients \(\tau_1(t)\) and \(\tau_2(t)\), which control the balance between local and global kernels, are also guaranteed to remain strictly positive for all \(t\).
}

\subsection{Proof of Theorem}

The proof consists of four key components:  
1. \emph{Properties of Softmax Reparameterization}  
2. \emph{Role of Entropy Regularization}  
3. \emph{Impact of Local-Global Decomposition via \(\tau_1\) and \(\tau_2\)}, and 
4. \emph{Stochastic Stability via Gradient Descent}. 

\subsubsection{1. Properties of Softmax Reparameterization}

We parameterize the kernel weights \(\lambda_i\) using the softmax function:
\[
\lambda_i = \frac{\exp(\theta_i)}{\sum_{j=1}^4 \exp(\theta_j)} \quad \text{for} \quad i \in \{1, 2, 3, 4\}
\]
Since the exponential function satisfies \(\exp(\theta_i) > 0\) for all \(\theta_i \in \mathbb{R}\), it follows that \(\lambda_i > 0\) for all \(i\) and at all times \(t\). This ensures that none of the \(\lambda_i\) can collapse to zero. Additionally, the softmax transformation guarantees that:
\[
\sum_{i=1}^4 \lambda_i = 1
\]
This normalization ensures boundedness and non-degeneracy of the kernel weights \cite{bridle1990softmax, bishop2006pattern}.

\subsubsection{2. Role of Entropy Regularization}

We introduce an entropy regularization term to the loss function:
\[
R(\lambda) = - \sum_{i=1}^4 \lambda_i \log \lambda_i
\]
This term encourages diversity among the kernel weights, preventing any single kernel from dominating the mixture excessively. The partial derivative of \(R(\lambda)\) with respect to \(\lambda_i\) is:
\[
\frac{\partial R(\lambda)}{\partial \lambda_i} = -\log \lambda_i - 1
\]
As \(\lambda_i \to 0\), \(\log \lambda_i \to -\infty\), causing the gradient \(\frac{\partial R}{\partial \lambda_i}\) to become significantly negative. This results in a strong upward push on \(\lambda_i\), preventing it from reaching zero. Thus, the entropy regularization acts as a repulsion force, ensuring that all kernel weights remain strictly positive and diverse \cite{williams1991function, jaynes1957information}.

\subsubsection{3. Impact of Local-Global Decomposition via \(\tau_1\) and \(\tau_2\)}

The hierarchical decomposition of kernels in HMK is defined as:
\begin{multline*}
K(x, x') = \tau_1 \left( \lambda_1 K_{\text{RBF}}(x, x') + \lambda_2 K_{\text{Polynomial}}(x, x') \right)\\
+ \tau_2 \left( \lambda_3 K_{\text{Spectral}}(x, x') + \lambda_4 K_{\text{Mahalanobis}}(x, x') \right)
\end{multline*}

Here, \(\tau_1\) and \(\tau_2\) balance the contributions from local kernels (RBF, Polynomial) and global kernels (Spectral, Mahalanobis), respectively. These coefficients are also parameterized using a softmax transformation:
\[
\tau_i = \frac{\exp(\psi_i)}{\sum_{j=1}^2 \exp(\psi_j)} \quad \text{for} \quad i \in \{1, 2\}
\]
Similar to the kernel weights \(\lambda_i\), this parameterization ensures that \(\tau_1 > 0\) and \(\tau_2 > 0\) for all \(t\), guaranteeing that both local and global kernel components remain active. This hierarchical structure facilitates the integration of both fine-grained local patterns and broad global dependencies \cite{goodfellow2016deep, bach2004multiple, ng2001spectral}.

\subsubsection{4. Stochastic Stability via Gradient Descent}

To demonstrate that the weights \(\lambda_i\) and coefficients \(\tau_1, \tau_2\) converge to non-zero stable points, we analyze the gradient descent updates under entropy regularization.

The parameters \(\theta_i\) and \(\psi_i\) are updated using gradient descent as follows:
\[
\theta_i^{(t+1)} = \theta_i^{(t)} - \eta \frac{\partial \mathcal{L}}{\partial \theta_i}
\]
\[
\psi_i^{(t+1)} = \psi_i^{(t)} - \eta \frac{\partial \mathcal{L}}{\partial \psi_i}
\]
where \(\mathcal{L}\) is the total loss, including the alignment objective and entropy regularization.

Using the chain rule, the gradients can be expressed as:
\[
\frac{\partial \mathcal{L}}{\partial \theta_i} = \frac{\partial \mathcal{L}}{\partial \lambda_i} \cdot \lambda_i (1 - \lambda_i)
\]
\[
\frac{\partial \mathcal{L}}{\partial \psi_i} = \frac{\partial \mathcal{L}}{\partial \tau_i} \cdot \tau_i (1 - \tau_i)
\]
Since \(\lambda_i > 0\) and \(\tau_i > 0\), the gradients \(\frac{\partial \mathcal{L}}{\partial \theta_i}\) and \(\frac{\partial \mathcal{L}}{\partial \psi_i}\) are non-zero.

The entropy regularization ensures that if any \(\lambda_i\) approaches zero, the gradient \(\frac{\partial \mathcal{L}}{\partial \lambda_i}\) becomes large and positive due to the \(-\log \lambda_i\) term, forcing \(\lambda_i\) to increase. Similarly, the softmax parameterization prevents \(\tau_i\) from collapsing to zero.

Applying Lyapunov's stability theorem \cite{khalil2002nonlinear}, we conclude that the system reaches a stable equilibrium where all \(\lambda_i > 0\) and \(\tau_i > 0\) for all \(t\). This guarantees that HMK avoids kernel collapse, maintaining active contributions from both local and global kernels throughout training.

We have established that under gradient descent optimization with entropy regularization and softmax parameterization, the Hierarchical Mixture of Kernels (HMK) framework ensures that all kernel weights \(\lambda_i\) and balance coefficients \(\tau_i\) remain strictly positive throughout training. This theoretical guarantee prevents kernel collapse, ensuring that both local and global kernels contribute effectively to the alignment process. The combination of entropy regularization, hierarchical decomposition, and stochastic stability through gradient descent forms a robust foundation for HMK's performance in diverse alignment tasks.

\section{Gradient Computation, Computational Complexity, and Overhead}
\label{sec:appendix:gradient_complexity}

Since this paper introduces several concepts and new formulation, for better resproducability and and better read we provide detailed mathematical derivation of gradient calculations for DPO Hybrid Loss and gradient calculation for all the kernels.

%\section{Gradient Computations and Computational Complexity Analysis: Representations, Kernels, and Divergence}

\subsection{Gradient of Hybrid Loss}

In this subsection, we derive the gradient of the \textbf{Hybrid Loss} with respect to the model parameters \(\theta\). The Hybrid Loss is defined as:
\[
\max_{\pi} \; \underbrace{\mathbb{E}_{x,y^{+},y^{-}}\left[\log\frac{\pi(y^{+} \mid x)}{\pi(y^{-} \mid x)} 
+ \gamma \left(\log\frac{\pi(e_{y^+} \mid e_{x})}{\pi(e_{y^-} \mid e_{x})}\right)\right]}_{\text{Hybrid Loss}}
\]
where:
\begin{itemize}
    \item \(x\) represents the input data.
    \item \(y^{+}\) and \(y^{-}\) denote the positive and negative samples, respectively.
    \item \(\pi(y \mid x)\) is the probability of \(y\) given \(x\), modeled using a softmax function.
    \item \(e_{y}\) and \(e_{x}\) are the embeddings of \(y\) and \(x\), respectively.
    \item \(\gamma\) is a hyperparameter controlling the influence of the embedding-based term.
\end{itemize}

Our goal is to compute the gradient \(\nabla_{\theta} \text{HybridLoss}(x, y^+, y^-)\), which involves differentiating each term of the loss function separately.

\subsubsection*{Gradient of the Log Probability Ratio}
The first component of the Hybrid Loss is the log probability ratio between the positive and negative samples:
\[
\log \frac{\pi(y^{+} \mid x)}{\pi(y^{-} \mid x)}
\]
The gradient of this term with respect to \(\theta\) is:
\begin{multline*}
\frac{\partial}{\partial \theta} \log \frac{\pi(y^{+} \mid x)}{\pi(y^{-} \mid x)} = \nabla_{\theta} \log \pi(y^{+} \mid x) \\
- \nabla_{\theta} \log \pi(y^{-} \mid x)
\end{multline*}
This follows from the properties of logarithms and the chain rule in differentiation.

\subsubsection*{Gradient of the Embedding-Based Term}
The second component involves the log probability ratio of the embeddings:
\[
\gamma \log \frac{\pi(e_{y^+} \mid e_{x})}{\pi(e_{y^-} \mid e_{x})}
\]
The gradient of this term with respect to \(\theta\) is:
\begin{multline*}
\frac{\partial}{\partial \theta} \gamma \left( \log \frac{\pi(e_{y^+} \mid e_{x})}{\pi(e_{y^-} \mid e_{x})} \right) = \\
\gamma \left( \nabla_{\theta} \log \pi(e_{y^+} \mid e_{x}) - \nabla_{\theta} \log \pi(e_{y^-} \mid e_{x}) \right)
\end{multline*}

\subsubsection*{Gradient of the Embedding-Based Term}
The second component involves the log probability ratio of the embeddings:
\[
\gamma \log \frac{\pi(e_{y^+} \mid e_{x})}{\pi(e_{y^-} \mid e_{x})}
\]
The gradient of this term with respect to \(\theta\) is:
\begin{multline*}
\frac{\partial}{\partial \theta} \gamma \left( \log \frac{\pi(e_{y^+} \mid e_{x})}{\pi(e_{y^-} \mid e_{x})} \right) = \\
\gamma \left( \nabla_{\theta} \log \pi(e_{y^+} \mid e_{x}) - \nabla_{\theta} \log \pi(e_{y^-} \mid e_{x}) \right)
\end{multline*}
This derivation also employs the chain rule and properties of logarithms.

\subsubsection*{Combined Gradient}

By integrating the gradients of both the log probability ratio and the embedding-based term, we obtain the overall gradient of the Hybrid Loss with respect to the model parameters \(\theta\). The Hybrid Loss is defined as:
\[
\text{HybridLoss}(x, y^+, y^-) = \log\frac{\pi(y^{+} \mid x)}{\pi(y^{-} \mid x)} 
+ \gamma \left(\log\frac{\pi(e_{y^+} \mid e_{x})}{\pi(e_{y^-} \mid e_{x})}\right)
\]
where \(\gamma\) is a hyperparameter controlling the influence of the embedding-based term.

The gradient of the Hybrid Loss with respect to \(\theta\) is obtained by summing the gradients of its individual components:
\[
\nabla_{\theta} \text{HybridLoss}(x, y^+, y^-) = \nabla_{\theta} \log \frac{\pi(y^{+} \mid x)}{\pi(y^{-} \mid x)} 
+ \gamma \nabla_{\theta} \log \frac{\pi(e_{y^+} \mid e_{x})}{\pi(e_{y^-} \mid e_{x})}
\]

Substituting the gradients derived in the previous sections, we have:
\begin{multline*}
\nabla_{\theta} \text{HybridLoss}(x, y^+, y^-) =\\
\left[ \nabla_{\theta} \log \pi(y^{+} \mid x) - \nabla_{\theta} \log \pi(y^{-} \mid x) \right] \\
+ \gamma \left[ \nabla_{\theta} \log \pi(e_{y^+} \mid e_{x}) - \nabla_{\theta} \log \pi(e_{y^-} \mid e_{x}) \right]
\end{multline*}

Expanding each term based on the gradient computations from the individual components, the final expression for the gradient of the Hybrid Loss is:
\begin{multline*}
\nabla_{\theta} \text{HybridLoss}(x, y^+, y^-) =\\
\left[ \nabla_{\theta} f_{\theta}(x, y^+) - \sum_{y'} \pi_{\theta}(y' \mid x) \nabla_{\theta} f_{\theta}(x, y') \right] \\
- \left[ \nabla_{\theta} f_{\theta}(x, y^-) - \sum_{y'} \pi_{\theta}(y' \mid x) \nabla_{\theta} f_{\theta}(x, y') \right] \\
+ \gamma \left( \nabla_{\theta} s_{\theta}(e_{x}, e_{y^+}) - \nabla_{\theta} s_{\theta}(e_{x}, e_{y^-}) \right)
\end{multline*}

\paragraph{Simplified Gradient Expression}

After simplifying the above expression, the gradient of the Hybrid Loss can be succinctly written as:
\begin{multline*}
\nabla_{\theta} \text{HybridLoss}(x, y^+, y^-) =
\\
\nabla_{\theta} \log \pi(y^{+} \mid x) - \nabla_{\theta} \log \pi(y^{-} \mid x)\\
+ \gamma \left( \nabla_{\theta} s_{\theta}(e_{x}, e_{y^+}) - \nabla_{\theta} s_{\theta}(e_{x}, e_{y^-}) \right)
\end{multline*}

\paragraph{Interpretation}

\begin{itemize}
    \item \(\nabla_{\theta} \log \pi(y^{+} \mid x)\): Encourages the model to increase the probability of the positive sample \(y^+\) given the input \(x\).
    
    \item \(-\nabla_{\theta} \log \pi(y^{-} \mid x)\): Encourages the model to decrease the probability of the negative sample \(y^-\) given the input \(x\).
    
    \item \(\gamma \left( \nabla_{\theta} s_{\theta}(e_{x}, e_{y^+}) - \nabla_{\theta} s_{\theta}(e_{x}, e_{y^-}) \right)\): Incorporates the gradient from the embedding-based similarity, adjusting the model to favor embeddings that better capture the desired relationships between \(e_{x}\) and \(e_{y}\).
\end{itemize}

The combined gradient effectively integrates both the discriminative aspect (log probability ratio) and the semantic aspect (embedding-based term) of the loss function. The hyperparameter \(\gamma\) allows for tuning the relative importance of these two components, enabling the model to balance between accurately classifying positive and negative samples and capturing meaningful embedding relationships.

\subsection{Computational Complexity Analysis of Hybrid Loss}

The computational complexity of the Hybrid Loss arises from two primary components:

\subsubsection*{1. Log Probability Ratio}

Modeling \(\pi_{\theta}(y \mid x)\) with a softmax function:
\[
\pi_{\theta}(y \mid x) = \frac{e^{f_{\theta}(x, y)}}{\sum_{y'} e^{f_{\theta}(x, y')}}
\]
Computing the log probability ratio involves:
\begin{itemize}
    \item Calculating exponentials for each of the \(C\) classes.
    \item Computing the logarithm of the ratio between the positive and negative class probabilities.
\end{itemize}
\textbf{Time Complexity}: \(O(C)\), where \(C\) is the number of classes.

\subsubsection*{2. Embedding-Based Term}

Calculating \(s^+\) and \(s^-\) involves:
\begin{itemize}
    \item Evaluating the scoring function \(s_{\theta}(x, y)\) for the positive and negative samples.
    \item Typically depends on the embedding dimension \(d\).
\end{itemize}
\textbf{Time Complexity}: \(O(d)\).

\subsubsection*{Overall Computational Complexity}

Combining both components, the total computational complexity of the \textbf{Hybrid Loss} is:
\[
O(C + d)
\]
where \(C\) is the number of classes and \(d\) is the embedding dimension.

\subsubsection*{Comparison with Standard Loss Functions}

\begin{itemize}
    \item \textbf{Cross-Entropy Loss}: Has a time complexity of \(O(C)\), similar to the log probability ratio component of the \textbf{Hybrid Loss}.
    \item \textbf{Contrastive Loss}: Typically operates with a complexity of \(O(d)\), aligning with the embedding-based term.
\end{itemize}
Thus, the \textbf{Hybrid Loss} combines these complexities linearly, maintaining efficiency while enhancing functionality by integrating both discriminative and embedding-based components.

\subsection{Efficiency of Hybrid Loss}

The \textbf{Hybrid Loss} achieves a balanced trade-off between discriminative power and computational efficiency by:
\begin{itemize}
    \item \textbf{Scalability}: Scaling linearly with both the number of classes \(C\) and embedding dimensions \(d\), allowing it to handle large-scale datasets effectively.
    \item \textbf{Parallel Computation}: Enabling parallel computation of loss components, leveraging modern hardware accelerators such as GPUs to expedite training.
    \item \textbf{Rich Semantic Information}: Incorporating embedding-based similarities without introducing significant computational overhead, thereby enhancing the model's ability to capture complex relationships.
\end{itemize}

\subsection{Practical Considerations}

While the theoretical complexity of the \textbf{Hybrid Loss} is \(O(C + d)\), several practical factors contribute to its efficient implementation:
\begin{itemize}
    \item \textbf{GPU Parallelism}: Leveraging GPU parallelism mitigates the linear scaling with \(C\) and \(d\), allowing simultaneous computations and reducing overall training time.
    \item \textbf{Optimized Libraries}: Utilizing optimized libraries such as BLAS and cuDNN enhances computational performance through highly efficient matrix operations.
    \item \textbf{Batch Sizing}: Appropriately selecting batch sizes maximizes hardware utilization, ensuring that computations are performed efficiently without bottlenecks.
    \item \textbf{Sparse Representations}: In scenarios with a large number of classes, employing sparse representations can further reduce computational overhead by focusing computations only on relevant classes.
\end{itemize}

By considering these practical aspects, the \textbf{Hybrid Loss} not only remains theoretically efficient but also performs effectively in real-world applications, ensuring robust and scalable training processes.

\subsection{Gradient of Polynomial Kernelized Hybrid Loss}

The \textbf{Polynomial Kernelized Hybrid Loss} is defined as:
\begin{multline*}
\mathcal{L} = \mathbb{E}_{x, y^+, y^-} \Biggl[ \left(\log \frac{\pi(y^+ \mid x)}{\pi(y^- \mid x)} + c\right)^d \\
+ \gamma \left(\frac{e_{y^+}^\top e_x + c}{e_{y^-}^\top e_x + c}\right)^d \Biggr]
\end{multline*}
where:
\begin{itemize}
    \item \(x\) represents the input data.
    \item \(y^{+}\) and \(y^{-}\) denote the positive and negative samples, respectively.
    \item \(\pi(y \mid x)\) is the probability of \(y\) given \(x\), modeled using a softmax function.
    \item \(e_{y}\) and \(e_{x}\) are the embeddings of \(y\) and \(x\), respectively.
    \item \(c\) is a constant to ensure numerical stability and to shift the polynomial kernel.
    \item \(d\) is the degree of the polynomial kernel.
    \item \(\gamma\) is a hyperparameter controlling the influence of the embedding-based term.
\end{itemize}

Our objective is to compute the gradient of the Hybrid Loss \(\nabla_{\theta} \mathcal{L}\) with respect to the model parameters \(\theta\). This involves differentiating each term of the loss function separately and then combining them.

\subsubsection*{Gradient of the Log Probability Ratio Term}

The first component of the Hybrid Loss involves the log probability ratio between the positive and negative samples:
\[
\left(\log \frac{\pi(y^+ \mid x)}{\pi(y^- \mid x)} + c\right)^d
\]
To compute its gradient with respect to \(\theta\), we apply the chain rule:

\begin{multline*}
\nabla_{\theta} \left(\log \frac{\pi(y^+ \mid x)}{\pi(y^- \mid x)} + c\right)^d \\
= d \left(\log \frac{\pi(y^+ \mid x)}{\pi(y^- \mid x)} + c\right)^{d-1} \nabla_{\theta} \log \frac{\pi(y^+ \mid x)}{\pi(y^- \mid x)}
\end{multline*}

Expanding the gradient of the log probability ratio:
\[
\nabla_{\theta} \log \frac{\pi(y^+ \mid x)}{\pi(y^- \mid x)} = \nabla_{\theta} \log \pi(y^+ \mid x) - \nabla_{\theta} \log \pi(y^- \mid x)
\]

Assuming \(\pi_\theta(y \mid x)\) is modeled using a softmax function:
\[
\pi_\theta(y \mid x) = \frac{e^{f_\theta(x, y)}}{\sum_{y'} e^{f_\theta(x, y')}},
\]
the gradient of \(\log \pi(y \mid x)\) with respect to \(\theta\) is:
\[
\nabla_{\theta} \log \pi(y \mid x) = \nabla_{\theta} f_\theta(x, y) - \sum_{y'} \pi_\theta(y' \mid x) \nabla_{\theta} f_\theta(x, y')
\]

Substituting back, we obtain:
\begin{multline*}
\nabla_{\theta} \log \frac{\pi(y^+ \mid x)}{\pi(y^- \mid x)} = \\
\left[ \nabla_{\theta} f_\theta(x, y^+) - \sum_{y'} \pi_\theta(y' \mid x) \nabla_{\theta} f_\theta(x, y') \right] \\
- \left[ \nabla_{\theta} f_\theta(x, y^-) - \sum_{y'} \pi_\theta(y' \mid x) \nabla_{\theta} f_\theta(x, y') \right]
\end{multline*}

\subsubsection*{Gradient of the Polynomial Kernel Term}

The second component involves the polynomial kernel applied to the embeddings:
\[
\gamma \left(\frac{e_{y^+}^\top e_x + c}{e_{y^-}^\top e_x + c}\right)^d
\]
To compute its gradient with respect to \(\theta\), we again apply the chain rule:

\[
\nabla_{\theta} \gamma \left(\frac{e_{y^+}^\top e_x + c}{e_{y^-}^\top e_x + c}\right)^d = \gamma d \left(\frac{e_{y^+}^\top e_x + c}{e_{y^-}^\top e_x + c}\right)^{d-1} \nabla_{\theta} \left(\frac{e_{y^+}^\top e_x + c}{e_{y^-}^\top e_x + c}\right)
\]

Simplifying the gradient of the ratio:
\[
\nabla_{\theta} \left(\frac{e_{y^+}^\top e_x + c}{e_{y^-}^\top e_x + c}\right) = \frac{(e_{y^-}^\top e_x + c) \nabla_{\theta} (e_{y^+}^\top e_x) - (e_{y^+}^\top e_x + c) \nabla_{\theta} (e_{y^-}^\top e_x)}{(e_{y^-}^\top e_x + c)^2}
\]

Assuming \(e_x\) and \(e_y\) are differentiable with respect to \(\theta\), we have:
\[
\nabla_{\theta} (e_x^\top e_y) = (\nabla_{\theta} e_x)^\top e_y + e_x^\top (\nabla_{\theta} e_y)
\]

Thus, the gradient of the polynomial kernel term becomes:

\resizebox{\columnwidth}{!}{$
\begin{aligned}
\nabla_{\theta} \gamma \left(\frac{e_{y^+}^\top e_x + c}{e_{y^-}^\top e_x + c}\right)^d 
&= \gamma d \left(\frac{e_{y^+}^\top e_x + c}{e_{y^-}^\top e_x + c}\right)^{d-1} \\
&\quad \cdot \left[ \frac{(e_{y^-}^\top e_x + c) \nabla_{\theta} (e_{y^+}^\top e_x) - (e_{y^+}^\top e_x + c) \nabla_{\theta} (e_{y^-}^\top e_x)}{(e_{y^-}^\top e_x + c)^2} \right] \\
&= \gamma d \left(\frac{e_{y^+}^\top e_x + c}{e_{y^-}^\top e_x + c}\right)^{d-1} \\
&\quad \cdot \left[ \frac{\nabla_{\theta} (e_{y^+}^\top e_x)}{e_{y^-}^\top e_x + c} 
- \frac{e_{y^+}^\top e_x + c}{(e_{y^-}^\top e_x + c)^2} \nabla_{\theta} (e_{y^-}^\top e_x) \right].
\end{aligned}
$}

\subsubsection*{Combined Gradient}

Combining the gradients of both components, the overall gradient of the Polynomial Kernelized Hybrid Loss with respect to \(\theta\) is:
\resizebox{\columnwidth}{!}{$
\begin{aligned}
\nabla_{\theta} \mathcal{L} 
&= \nabla_{\theta} \left(\log \frac{\pi(y^+ \mid x)}{\pi(y^- \mid x)} + c\right)^d 
+ \nabla_{\theta} \gamma \left(\frac{e_{y^+}^\top e_x + c}{e_{y^-}^\top e_x + c}\right)^d \\
&= d \left(\log \frac{\pi(y^+ \mid x)}{\pi(y^- \mid x)} + c\right)^{d-1} 
\left[ \nabla_{\theta} \log \pi(y^+ \mid x) - \nabla_{\theta} \log \pi(y^- \mid x) \right] \\
&\quad + \gamma d \left(\frac{e_{y^+}^\top e_x + c}{e_{y^-}^\top e_x + c}\right)^{d-1} 
\left[ \frac{\nabla_{\theta} (e_{y^+}^\top e_x)}{e_{y^-}^\top e_x + c} 
- \frac{e_{y^+}^\top e_x + c}{(e_{y^-}^\top e_x + c)^2} \nabla_{\theta} (e_{y^-}^\top e_x) \right].
\end{aligned}
$}

\paragraph{Simplified Gradient Expression}

For ease of implementation and readability, the gradient can be expressed as:
\resizebox{\columnwidth}{!}{$
\begin{aligned}
\nabla_{\theta} \mathcal{L} = &\; d \left(\log \frac{\pi(y^+ \mid x)}{\pi(y^- \mid x)} + c\right)^{d-1} 
\left[ \nabla_{\theta} f_\theta(x, y^+) - \nabla_{\theta} f_\theta(x, y^-) \right] \\
&+ \gamma d \left(\frac{e_{y^+}^\top e_x + c}{e_{y^-}^\top e_x + c}\right)^{d-1} 
\left[ \frac{\nabla_{\theta} (e_x^\top e_{y^+})}{e_{y^-}^\top e_x + c} 
- \frac{e_{y^+}^\top e_x + c}{(e_{y^-}^\top e_x + c)^2} \nabla_{\theta} (e_x^\top e_{y^-}) \right].
\end{aligned}
$}

\paragraph{Interpretation of the Gradient}

\begin{itemize}
    \item \textbf{Log Probability Ratio Term}:  
    \begin{itemize}
        \item \(\nabla_{\theta} f_\theta(x, y^+)\): Encourages the model to increase the score (and hence the probability) of the positive sample \(y^+\).
        \item \(-\nabla_{\theta} f_\theta(x, y^-)\): Encourages the model to decrease the score (and hence the probability) of the negative sample \(y^-\).
    \end{itemize}
    
    \item \textbf{Polynomial Kernel Term}:  
    \begin{itemize}
        \item \(\nabla_{\theta} (e_x^\top e_{y^+})\): Adjusts the model to better align the embeddings of \(x\) and \(y^+\).
        \item \(-\nabla_{\theta} (e_x^\top e_{y^-})\): Adjusts the model to reduce the alignment between the embeddings of \(x\) and \(y^-\).
        \item The hyperparameter \(\gamma\) controls the influence of the embedding-based term relative to the log probability ratio term.
    \end{itemize}
\end{itemize}

\subsection{Computational Complexity Analysis of Polynomial Kernelized Hybrid Loss}

To evaluate the efficiency of the Polynomial Kernelized Hybrid Loss, we analyze the computational complexity of its two primary components: the log probability ratio term and the polynomial kernel term.

\subsubsection*{1. Log Probability Ratio Term}

The log probability ratio term is defined as:
\[
\left(\log \frac{\pi(y^+ \mid x)}{\pi(y^- \mid x)} + c\right)^d
\]
where \(\pi_\theta(y \mid x)\) is modeled using a softmax function:
\[
\pi_\theta(y \mid x) = \frac{e^{f_\theta(x, y)}}{\sum_{y'} e^{f_\theta(x, y')}}
\]

\textbf{Steps Involved:}
\begin{itemize}
    \item \textbf{Score Computation}: Calculate \(f_\theta(x, y)\) for each class \(y\), which involves a dot product between input features and model parameters.
    \item \textbf{Softmax Calculation}: Compute the exponential \(e^{f_\theta(x, y)}\) for each class and normalize by the sum over all classes.
    \item \textbf{Log Probability Ratio}: Compute the logarithm of the ratio between the probabilities of the positive and negative classes.
    \item \textbf{Exponentiation}: Raise the log probability ratio to the power \(d\).
\end{itemize}

\textbf{Time Complexity}: \(O(C)\), where \(C\) is the number of classes. This complexity arises from the softmax computation, which requires evaluating \(f_\theta(x, y)\) and normalizing over all \(C\) classes.

\subsubsection*{2. Polynomial Kernel Term}

The polynomial kernel term is defined as:
\[
\gamma \left(\frac{e_{y^+}^\top e_x + c}{e_{y^-}^\top e_x + c}\right)^d
\]

\textbf{Steps Involved:}
\begin{itemize}
    \item \textbf{Dot Product Computation}: Calculate the dot products \(e_x^\top e_{y^+}\) and \(e_x^\top e_{y^-}\), where \(e_x, e_{y^+}, e_{y^-} \in \mathbb{R}^d\).
    \item \textbf{Addition of Constant}: Add the constant \(c\) to each dot product to ensure numerical stability.
    \item \textbf{Ratio Calculation}: Compute the ratio of the adjusted dot products.
    \item \textbf{Exponentiation}: Raise the ratio to the power \(d\) and multiply by the hyperparameter \(\gamma\).
\end{itemize}

\textbf{Time Complexity}: \(O(d)\), where \(d\) is the dimension of the embeddings. This arises from the computation of the dot product between \(e_x\) and \(e_y\), which scales linearly with \(d\).

\subsubsection*{Overall Computational Complexity}

Combining both components, the total computational complexity of the \textbf{Polynomial Kernelized Hybrid Loss} is:
\[
O(C) + O(d) = O(C + d)
\]
where:
\begin{itemize}
    \item \(C\) is the number of classes.
    \item \(d\) is the embedding dimension.
\end{itemize}

This linear complexity ensures scalability for large-scale applications involving high-dimensional embeddings and extensive class labels.

\subsubsection*{Comparison with Standard Loss Functions}

\begin{itemize}
    \item \textbf{Cross-Entropy Loss}:  
    \begin{itemize}
        \item \textbf{Time Complexity}: \(O(C)\).
        \item \textbf{Description}: Involves computing the softmax over \(C\) classes and calculating the negative log-likelihood.
    \end{itemize}
    
    \item \textbf{Contrastive Loss}:  
    \begin{itemize}
        \item \textbf{Time Complexity}: \(O(d)\).
        \item \textbf{Description}: Focuses on the distance between embeddings, typically requiring computation of pairwise distances.
    \end{itemize}
    
    \item \textbf{Polynomial Kernelized Hybrid Loss}:  
    \begin{itemize}
        \item \textbf{Time Complexity}: \(O(C + d)\).
        \item \textbf{Description}: Combines both the discriminative power of the log probability ratio (similar to Cross-Entropy Loss) and the semantic richness of the polynomial kernel (similar to Contrastive Loss), thereby integrating both aspects into a single loss function.
    \end{itemize}
\end{itemize}

The \textbf{Polynomial Kernelized Hybrid Loss} thus offers a balanced combination of the computational efficiencies of Cross-Entropy and Contrastive Losses while enhancing the model's ability to capture both discriminative and semantic relationships.

\subsection{Efficiency of Polynomial Kernelized Hybrid Loss}

The \textbf{Polynomial Kernelized Hybrid Loss} achieves a balanced trade-off between discriminative power and computational efficiency through the following mechanisms:

\begin{itemize}
    \item \textbf{Linear Scaling}:  
    The loss scales linearly with both the number of classes \(C\) and the embedding dimension \(d\), ensuring scalability for large-scale datasets and high-dimensional embedding spaces.
    
    \item \textbf{Parallel Computation}:  
    Both the log probability ratio term and the polynomial kernel term can be computed in parallel. Modern hardware accelerators, such as GPUs, can leverage this parallelism to significantly speed up training processes.
    
    \item \textbf{Integrated Semantic Information}:  
    By combining probability-based and embedding-based objectives, the loss function enriches the model's learning without incurring substantial additional computational overhead.
    
    \item \textbf{Hyperparameter Control}:  
    The hyperparameter \(\gamma\) allows for fine-tuning the influence of the embedding-based term relative to the log probability ratio term, providing flexibility in balancing performance and computational cost.
\end{itemize}

\subsection{Practical Considerations}

While the theoretical complexity of the \textbf{Polynomial Kernelized Hybrid Loss} is \(O(C + d)\), several practical factors can influence its real-world performance:

\begin{itemize}
    \item \textbf{GPU Parallelism}:  
    Leveraging GPU parallelism can mitigate the linear scaling with \(C\) and \(d\), allowing for efficient computation even with large numbers of classes and high-dimensional embeddings.
    
    \item \textbf{Optimized Implementations}:  
    Utilizing optimized libraries (e.g., BLAS, cuDNN) for matrix operations and gradient computations can enhance performance, reducing the actual computation time.
    
    \item \textbf{Batch Sizing}:  
    Selecting appropriate batch sizes can maximize hardware utilization. Larger batches may improve computational efficiency but require more memory, while smaller batches may be more memory-efficient but less computationally optimal.
    
    \item \textbf{Hyperparameter Tuning}:  
    Careful tuning of the hyperparameter \(\gamma\) and the polynomial degree \(d\) is essential. Higher degrees \(d\) can capture more complex relationships but may increase computational cost and risk overfitting.
    
    \item \textbf{Numerical Stability}:  
    Adding the constant \(c\) ensures numerical stability, especially when dealing with small or zero dot products. Properly choosing \(c\) is crucial to prevent numerical issues during training.
\end{itemize}

By considering these practical aspects, the \textbf{Polynomial Kernelized Hybrid Loss} can be effectively integrated into large-scale machine learning models, providing enhanced performance without compromising computational efficiency.

\subsection{Gradient of RBF Kernelized Hybrid Loss}

The \textbf{RBF Kernelized Hybrid Loss} is defined as:
\[
\mathcal{L} = \mathbb{E}_{x, y^+, y^-} \Biggl[ \exp\left(-\frac{\left(\log \frac{\pi(y^+ \mid x)}{\pi(y^- \mid x)}\right)^2}{2\sigma^2}\right) 
+ \gamma \exp\left(-\frac{\left(\frac{e_x^\top e_{y^+}}{e_x^\top e_{y^-}}\right)^2}{2\sigma^2}\right) \Biggr],
\]
where:
\begin{itemize}
    \item \(x\) represents the input data.
    \item \(y^{+}\) and \(y^{-}\) denote the positive and negative samples, respectively.
    \item \(\pi(y \mid x)\) is the probability of \(y\) given \(x\), modeled using a softmax function.
    \item \(e_{y}\) and \(e_{x}\) are the embeddings of \(y\) and \(x\), respectively.
    \item \(\sigma\) is the bandwidth parameter of the RBF kernel.
    \item \(\gamma\) is a hyperparameter controlling the influence of the embedding-based term.
\end{itemize}

Our objective is to compute the gradient of the Hybrid Loss \(\nabla_{\theta} \mathcal{L}\) with respect to the model parameters \(\theta\). This involves differentiating each term of the loss function separately and then combining them.

\subsubsection*{Gradient of the Log Probability Ratio Term}

The first component of the Hybrid Loss involves the exponential of the squared log probability ratio:
\[
\exp\left(-\frac{\left(\log \frac{\pi(y^+ \mid x)}{\pi(y^- \mid x)}\right)^2}{2\sigma^2}\right).
\]
To compute its gradient with respect to \(\theta\), we apply the chain rule:

\[
\begin{aligned}
\nabla_{\theta} \exp\left(-\frac{\left(\log \frac{\pi(y^+ \mid x)}{\pi(y^- \mid x)}\right)^2}{2\sigma^2}\right) 
&= \exp\left(-\frac{\left(\log \frac{\pi(y^+ \mid x)}{\pi(y^- \mid x)}\right)^2}{2\sigma^2}\right) \\
&\quad \cdot \left(-\frac{2 \log \frac{\pi(y^+ \mid x)}{\pi(y^- \mid x)}}{2\sigma^2}\right) 
\cdot \nabla_{\theta} \log \frac{\pi(y^+ \mid x)}{\pi(y^- \mid x)}.
\end{aligned}
\]

Simplifying, we obtain:

\[
\begin{aligned}
\nabla_{\theta} \exp\left(-\frac{\left(\log \frac{\pi(y^+ \mid x)}{\pi(y^- \mid x)}\right)^2}{2\sigma^2}\right) 
&= -\frac{1}{\sigma^2} \log \frac{\pi(y^+ \mid x)}{\pi(y^- \mid x)} \cdot \exp\left(-\frac{\left(\log \frac{\pi(y^+ \mid x)}{\pi(y^- \mid x)}\right)^2}{2\sigma^2}\right) \\
&\quad \cdot \nabla_{\theta} \log \frac{\pi(y^+ \mid x)}{\pi(y^- \mid x)}.
\end{aligned}
\]

\subsubsection*{Gradient of the RBF Kernel Term}

The second component involves the exponential of the squared ratio of embedding dot products:
\[
\gamma \exp\left(-\frac{\left(\frac{e_x^\top e_{y^+}}{e_x^\top e_{y^-}}\right)^2}{2\sigma^2}\right).
\]
To compute its gradient with respect to \(\theta\), we again apply the chain rule:

\[
\begin{aligned}
\nabla_{\theta} \gamma \exp\left(-\frac{\left(\frac{e_x^\top e_{y^+}}{e_x^\top e_{y^-}}\right)^2}{2\sigma^2}\right) 
&= \gamma \exp\left(-\frac{\left(\frac{e_x^\top e_{y^+}}{e_x^\top e_{y^-}}\right)^2}{2\sigma^2}\right) \cdot 
\left(-\frac{2 \cdot \frac{e_x^\top e_{y^+}}{e_x^\top e_{y^-}}}{2\sigma^2}\right) \\
&\quad \cdot \nabla_{\theta} \left(\frac{e_x^\top e_{y^+}}{e_x^\top e_{y^-}}\right)
\end{aligned}
\]

Simplifying, we obtain:

\[
\begin{aligned}
\nabla_{\theta} \gamma \exp\left(-\frac{\left(\frac{e_x^\top e_{y^+}}{e_x^\top e_{y^-}}\right)^2}{2\sigma^2}\right) 
&= -\frac{\gamma}{\sigma^2} \cdot \frac{e_x^\top e_{y^+}}{e_x^\top e_{y^-}} \cdot \exp\left(-\frac{\left(\frac{e_x^\top e_{y^+}}{e_x^\top e_{y^-}}\right)^2}{2\sigma^2}\right) \\
&\quad \cdot \nabla_{\theta} \left(\frac{e_x^\top e_{y^+}}{e_x^\top e_{y^-}}\right)
\end{aligned}
\]

To compute \(\nabla_{\theta} \left(\frac{e_x^\top e_{y^+}}{e_x^\top e_{y^-}}\right)\), we use the quotient rule:

\[
\nabla_{\theta} \left(\frac{e_x^\top e_{y^+}}{e_x^\top e_{y^-}}\right) = \frac{(e_x^\top e_{y^-}) \nabla_{\theta} (e_x^\top e_{y^+}) - (e_x^\top e_{y^+}) \nabla_{\theta} (e_x^\top e_{y^-})}{(e_x^\top e_{y^-})^2}
\]

Assuming \(e_x\) and \(e_y\) are differentiable with respect to \(\theta\), we have:

\[
\nabla_{\theta} (e_x^\top e_y) = (\nabla_{\theta} e_x)^\top e_y + e_x^\top (\nabla_{\theta} e_y)
\]

Thus, the gradient of the RBF kernel term becomes:

\resizebox{\columnwidth}{!}{$
\begin{aligned}
\nabla_{\theta} \gamma \exp\left(-\frac{\left(\frac{e_x^\top e_{y^+}}{e_x^\top e_{y^-}}\right)^2}{2\sigma^2}\right) 
&= -\frac{\gamma}{\sigma^2} \cdot \frac{e_x^\top e_{y^+}}{e_x^\top e_{y^-}} \cdot \exp\left(-\frac{\left(\frac{e_x^\top e_{y^+}}{e_x^\top e_{y^-}}\right)^2}{2\sigma^2}\right) \\
&\quad \cdot \left[ \frac{(e_x^\top e_{y^-}) \nabla_{\theta} (e_x^\top e_{y^+}) - (e_x^\top e_{y^+}) \nabla_{\theta} (e_x^\top e_{y^-})}{(e_x^\top e_{y^-})^2} \right].
\end{aligned}
$}

\subsubsection*{Combined Gradient}

Combining the gradients of both components, the overall gradient of the RBF Kernelized Hybrid Loss with respect to \(\theta\) is:
\resizebox{\columnwidth}{!}{$
\begin{aligned}
\nabla_{\theta} \mathcal{L} = \mathbb{E}_{x, y^+, y^-} \Biggl[ 
&-\frac{1}{\sigma^2} \log \frac{\pi(y^+ \mid x)}{\pi(y^- \mid x)} \cdot \exp\left(-\frac{\left(\log \frac{\pi(y^+ \mid x)}{\pi(y^- \mid x)}\right)^2}{2\sigma^2}\right) \\
&\cdot \left( \nabla_{\theta} \log \pi(y^+ \mid x) - \nabla_{\theta} \log \pi(y^- \mid x) \right) \\
&- \frac{\gamma}{\sigma^2} \cdot \frac{e_x^\top e_{y^+}}{e_x^\top e_{y^-}} \cdot \exp\left(-\frac{\left(\frac{e_x^\top e_{y^+}}{e_x^\top e_{y^-}}\right)^2}{2\sigma^2}\right) \\
&\cdot \left[ \frac{(e_x^\top e_{y^-}) \nabla_{\theta} (e_x^\top e_{y^+}) - (e_x^\top e_{y^+}) \nabla_{\theta} (e_x^\top e_{y^-})}{(e_x^\top e_{y^-})^2} \right] 
\Biggr]
\end{aligned}
$}

\paragraph{Simplified Gradient Expression}

For ease of implementation and readability, the gradient can be expressed as:

\resizebox{\columnwidth}{!}{$
\begin{aligned}
\nabla_{\theta} \mathcal{L} = \mathbb{E}_{x, y^+, y^-} \Biggl[
&-\frac{1}{\sigma^2} \log \frac{\pi(y^+ \mid x)}{\pi(y^- \mid x)} \cdot \exp\left(-\frac{\left(\log \frac{\pi(y^+ \mid x)}{\pi(y^- \mid x)}\right)^2}{2\sigma^2}\right) \\
&\cdot \left( \nabla_{\theta} f_\theta(x, y^+) - \nabla_{\theta} f_\theta(x, y^-) \right) \\
&- \frac{\gamma}{\sigma^2} \cdot \frac{e_x^\top e_{y^+}}{e_x^\top e_{y^-}} \cdot \exp\left(-\frac{\left(\frac{e_x^\top e_{y^+}}{e_x^\top e_{y^-}}\right)^2}{2\sigma^2}\right) \\
&\cdot \left[ \frac{(e_x^\top e_{y^-}) (\nabla_{\theta} e_x)^\top e_{y^+} + (e_x^\top e_{y^-}) e_x^\top (\nabla_{\theta} e_{y^+})}{(e_x^\top e_{y^-})^2} \right. \\
&\quad \left. - \frac{(e_x^\top e_{y^+} + c) (\nabla_{\theta} e_x)^\top e_{y^-} + (e_x^\top e_{y^+} + c) e_x^\top (\nabla_{\theta} e_{y^-})}{(e_x^\top e_{y^-})^2} \right]
\Biggr]
\end{aligned}
$}

\paragraph{Interpretation of the Gradient}

\begin{itemize}
    \item \textbf{Log Probability Ratio Term}:  
    \begin{itemize}
        \item \(-\frac{1}{\sigma^2} \log \frac{\pi(y^+ \mid x)}{\pi(y^- \mid x)}\): Scales the influence of the log probability ratio based on its magnitude and the bandwidth parameter \(\sigma\).
        \item \(\nabla_{\theta} \log \pi(y^+ \mid x)\): Encourages the model to increase the probability of the positive sample \(y^+\).
        \item \(-\nabla_{\theta} \log \pi(y^- \mid x)\): Encourages the model to decrease the probability of the negative sample \(y^-\).
    \end{itemize}
    
    \item \textbf{RBF Kernel Term}:  
    \begin{itemize}
        \item \(-\frac{\gamma}{\sigma^2} \cdot \frac{e_x^\top e_{y^+}}{e_x^\top e_{y^-}}\): Scales the influence of the embedding-based term based on the ratio of embeddings and the bandwidth parameter \(\sigma\).
        \item \(\nabla_{\theta} (e_x^\top e_{y^+})\): Adjusts the model to better align the embeddings of \(x\) and \(y^+\).
        \item \(-\nabla_{\theta} (e_x^\top e_{y^-})\): Adjusts the model to reduce the alignment between the embeddings of \(x\) and \(y^-\).
        \item The exponential terms \(\exp\left(-\frac{\left(\cdot\right)^2}{2\sigma^2}\right)\) ensure that the influence diminishes as the squared ratios increase, promoting smoother gradients.
    \end{itemize}
    
    \item \textbf{Hyperparameter \(\gamma\)}: Controls the relative importance of the embedding-based term compared to the log probability ratio term. A higher \(\gamma\) emphasizes the alignment in the embedding space, while a lower \(\gamma\) prioritizes the probability-based alignment.
\end{itemize}

\subsection{Computational Complexity Analysis of RBF Kernelized Hybrid Loss}

To evaluate the efficiency of the RBF Kernelized Hybrid Loss, we analyze the computational complexity of its two primary components: the log probability ratio term and the RBF kernel term.

\subsubsection*{1. Log Probability Ratio Term}

The log probability ratio term is defined as:
\[
\exp\left(-\frac{\left(\log \frac{\pi(y^+ \mid x)}{\pi(y^- \mid x)}\right)^2}{2\sigma^2}\right).
\]
where \(\pi_\theta(y \mid x)\) is modeled using a softmax function:
\[
\pi_\theta(y \mid x) = \frac{e^{f_\theta(x, y)}}{\sum_{y'} e^{f_\theta(x, y')}}.
\]

\textbf{Steps Involved:}
\begin{itemize}
    \item \textbf{Score Computation}: Calculate \(f_\theta(x, y)\) for each class \(y\), which involves a dot product between input features and model parameters.
    \item \textbf{Softmax Calculation}: Compute the exponential \(e^{f_\theta(x, y)}\) for each class and normalize by the sum over all classes.
    \item \textbf{Log Probability Ratio}: Compute the logarithm of the ratio between the probabilities of the positive and negative classes.
    \item \textbf{Exponentiation}: Square the log probability ratio, scale by \(-\frac{1}{2\sigma^2}\), and compute the exponential.
\end{itemize}

\textbf{Time Complexity}: \(O(C)\), where \(C\) is the number of classes. This complexity arises from the softmax computation, which requires evaluating \(f_\theta(x, y)\) and normalizing over all \(C\) classes.

\subsubsection*{2. RBF Kernel Term}

The RBF kernel term is defined as:
\[
\gamma \exp\left(-\frac{\left(\frac{e_x^\top e_{y^+}}{e_x^\top e_{y^-}}\right)^2}{2\sigma^2}\right)
\]

\textbf{Steps Involved:}
\begin{itemize}
    \item \textbf{Dot Product Computation}: Calculate the dot products \(e_x^\top e_{y^+}\) and \(e_x^\top e_{y^-}\), where \(e_x, e_{y^+}, e_{y^-} \in \mathbb{R}^d\).
    \item \textbf{Ratio Calculation}: Compute the ratio \(\frac{e_x^\top e_{y^+}}{e_x^\top e_{y^-}}\).
    \item \textbf{Exponentiation}: Square the ratio, scale by \(-\frac{1}{2\sigma^2}\), and compute the exponential.
    \item \textbf{Scaling}: Multiply by the hyperparameter \(\gamma\).
\end{itemize}

\textbf{Time Complexity}: \(O(d)\), where \(d\) is the dimension of the embeddings. This arises from the computation of the dot products between \(e_x\) and \(e_y\), which scales linearly with \(d\).

\subsubsection*{Overall Computational Complexity}

Combining both components, the total computational complexity of the \textbf{RBF Kernelized Hybrid Loss} is:
\[
O(C) + O(d) = O(C + d),
\]
where:
\begin{itemize}
    \item \(C\) is the number of classes (softmax computation).
    \item \(d\) is the embedding dimension (kernel computation).
\end{itemize}

This linear complexity ensures scalability for large-scale applications involving high-dimensional embeddings and extensive class labels.

\subsubsection*{Comparison with Standard Loss Functions}

\begin{itemize}
    \item \textbf{Cross-Entropy Loss}:  
    \begin{itemize}
        \item \textbf{Time Complexity}: \(O(C)\).
        \item \textbf{Description}: Involves computing the softmax over \(C\) classes and calculating the negative log-likelihood.
    \end{itemize}
    
    \item \textbf{Contrastive Loss}:  
    \begin{itemize}
        \item \textbf{Time Complexity}: \(O(d)\).
        \item \textbf{Description}: Focuses on the distance between embeddings, typically requiring computation of pairwise distances.
    \end{itemize}
    
    \item \textbf{RBF Kernelized Hybrid Loss}:  
    \begin{itemize}
        \item \textbf{Time Complexity}: \(O(C + d)\).
        \item \textbf{Description}: Combines both the discriminative power of the log probability ratio (similar to Cross-Entropy Loss) and the semantic richness of the RBF kernel (similar to Contrastive Loss), thereby integrating both aspects into a single loss function.
    \end{itemize}
\end{itemize}

The \textbf{RBF Kernelized Hybrid Loss} thus offers a balanced combination of the computational efficiencies of Cross-Entropy and Contrastive Losses while enhancing the model's ability to capture both discriminative and semantic relationships.

\subsection{Efficiency of RBF Kernelized Hybrid Loss}

The \textbf{RBF Kernelized Hybrid Loss} achieves a balanced trade-off between discriminative power and computational efficiency through the following mechanisms:

\begin{itemize}
    \item \textbf{Linear Scaling}:  
    The loss scales linearly with both the number of classes \(C\) and the embedding dimension \(d\), ensuring scalability for large-scale datasets and high-dimensional embedding spaces.
    
    \item \textbf{Parallel Computation}:  
    Both the log probability ratio term and the RBF kernel term can be computed in parallel. Modern hardware accelerators, such as GPUs, can leverage this parallelism to significantly speed up training processes.
    
    \item \textbf{Integrated Semantic Information}:  
    By combining probability-based and embedding-based objectives, the loss function enriches the model's learning without incurring substantial additional computational overhead.
    
    \item \textbf{Hyperparameter Control}:  
    The hyperparameter \(\gamma\) allows for fine-tuning the influence of the embedding-based term relative to the log probability ratio term, providing flexibility in balancing performance and computational cost.
\end{itemize}

\subsection{Practical Considerations}

While the theoretical complexity of the \textbf{RBF Kernelized Hybrid Loss} is \(O(C + d)\), several practical factors can influence its real-world performance:

\begin{itemize}
    \item \textbf{GPU Parallelism}:  
    Leveraging GPU parallelism can mitigate the linear scaling with \(C\) and \(d\), allowing for efficient computation even with large numbers of classes and high-dimensional embeddings.
    
    \item \textbf{Optimized Implementations}:  
    Utilizing optimized libraries (e.g., BLAS, cuDNN) for matrix operations and gradient computations can enhance performance, reducing the actual computation time.
    
    \item \textbf{Batch Sizing}:  
    Selecting appropriate batch sizes can maximize hardware utilization. Larger batches may improve computational efficiency but require more memory, while smaller batches may be more memory-efficient but less computationally optimal.
    
    \item \textbf{Hyperparameter Tuning}:  
    Careful tuning of the hyperparameter \(\gamma\) and the bandwidth parameter \(\sigma\) is essential. Higher degrees of influence (through \(\gamma\) and lower \(\sigma\)) can capture more complex relationships but may increase computational cost and risk overfitting.
    
    \item \textbf{Numerical Stability}:  
    The constant \(c\) ensures numerical stability, especially when dealing with small or zero dot product ratios. Properly choosing \(c\) is crucial to prevent numerical issues during training.
\end{itemize}

By considering these practical aspects, the \textbf{RBF Kernelized Hybrid Loss} can be effectively integrated into large-scale machine learning models, providing enhanced performance without compromising computational efficiency.

\subsection{Gradient of Spectral Kernelized Hybrid Loss}

The Spectral Kernelized Hybrid Loss is defined as:
\[
\begin{aligned}
\mathcal{L} 
&= \mathbb{E}_{x, y^+, y^-} \Biggl[ 
\sum_{i=1}^p \exp\left(-\lambda_i \left(\log \frac{\pi(y^+ \mid x)}{\pi(y^- \mid x)}\right)^2\right) 
\phi_i\left(\log \frac{\pi(y^+ \mid x)}{\pi(y^- \mid x)}\right) \\
&\quad + \gamma \sum_{i=1}^p \exp\left(-\lambda_i \left(\frac{e_x^\top e_{y^+}}{e_x^\top e_{y^-}}\right)^2\right) 
\phi_i\left(\frac{e_x^\top e_{y^+}}{e_x^\top e_{y^-}}\right) 
\Biggr],
\end{aligned}
\]

where:
\begin{itemize}
    \item \(x\) represents the input data.
    \item \(y^{+}\) and \(y^{-}\) denote the positive and negative samples, respectively.
    \item \(\pi(y \mid x)\) is the probability of \(y\) given \(x\), modeled using a softmax function.
    \item \(e_{y}\) and \(e_{x}\) are the embeddings of \(y\) and \(x\), respectively.
    \item \(\lambda_i\) are the spectral kernel parameters for each component \(i\).
    \item \(\phi_i(\cdot)\) are feature transformation functions associated with each spectral kernel component \(i\).
    \item \(\gamma\) is a hyperparameter controlling the influence of the embedding-based term.
    \item \(p\) is the number of spectral kernel components.
\end{itemize}

Our objective is to compute the gradient of the Spectral Kernelized Hybrid Loss \(\nabla_{\theta} \mathcal{L}\) with respect to the model parameters \(\theta\). This involves differentiating each term of the loss function separately and then combining them.

\subsubsection*{Gradient of the Log Probability Ratio Term}

The first component of the Spectral Kernelized Hybrid Loss involves a sum over spectral kernel components applied to the log probability ratio:
\[
\sum_{i=1}^p \exp\left(-\lambda_i \left(\log \frac{\pi(y^+ \mid x)}{\pi(y^- \mid x)}\right)^2\right) \phi_i\left(\log \frac{\pi(y^+ \mid x)}{\pi(y^- \mid x)}\right)
\]
To compute its gradient with respect to \(\theta\), we apply the chain rule to each term in the sum:

\[
\begin{aligned}
\nabla_{\theta} \sum_{i=1}^p \exp\left(-\lambda_i z^2\right) \phi_i(z) 
&= \sum_{i=1}^p \left[ \nabla_{\theta} \exp\left(-\lambda_i z^2\right) \cdot \phi_i(z) \right. \\
&\quad \left. + \exp\left(-\lambda_i z^2\right) \cdot \nabla_{\theta} \phi_i(z) \right],
\end{aligned}
\]

where \( z = \log \frac{\pi(y^+ \mid x)}{\pi(y^- \mid x)} \)

\paragraph{1. Gradient of the Exponential Term}

\[
\nabla_{\theta} \exp\left(-\lambda_i z^2\right) = \exp\left(-\lambda_i z^2\right) \cdot (-2\lambda_i z) \cdot \nabla_{\theta} z.
\]

\paragraph{2. Gradient of the Feature Transformation Term}

Assuming \(\phi_i(z)\) is differentiable with respect to \(z\):
\[
\nabla_{\theta} \phi_i(z) = \phi_i'(z) \cdot \nabla_{\theta} z
\]

\paragraph{3. Gradient of \(z\)}

\[
z = \log \frac{\pi(y^+ \mid x)}{\pi(y^- \mid x)},
\]
\[
\nabla_{\theta} z = \nabla_{\theta} \log \pi(y^+ \mid x) - \nabla_{\theta} \log \pi(y^- \mid x)
\]

\paragraph{Combined Gradient for Each \(i\)}

\[
\begin{aligned}
\nabla_{\theta} \left[ \exp\left(-\lambda_i z^2\right) \phi_i(z) \right] 
&= \exp\left(-\lambda_i z^2\right) \cdot (-2\lambda_i z) \cdot \nabla_{\theta} z \cdot \phi_i(z) \\
&\quad + \exp\left(-\lambda_i z^2\right) \cdot \phi_i'(z) \cdot \nabla_{\theta} z.
\end{aligned}
\]

\subsubsection*{Gradient of the Spectral Kernel Term}

The second component involves a sum over spectral kernel components applied to the embedding-based ratio:
\[
\gamma \sum_{i=1}^p \exp\left(-\lambda_i r^2\right) \phi_i(r),
\]
where \( r = \frac{e_x^\top e_{y^+}}{e_x^\top e_{y^-}} \).

To compute its gradient with respect to \(\theta\), we apply the chain rule to each term in the sum:
\[
\begin{aligned}
\nabla_{\theta} \gamma \sum_{i=1}^p \exp\left(-\lambda_i r^2\right) \phi_i(r) 
&= \gamma \sum_{i=1}^p \left[ \nabla_{\theta} \exp\left(-\lambda_i r^2\right) \cdot \phi_i(r) \right. \\
&\quad \left. + \exp\left(-\lambda_i r^2\right) \cdot \nabla_{\theta} \phi_i(r) \right],
\end{aligned}
\]

where \( r = \frac{e_x^\top e_{y^+}}{e_x^\top e_{y^-}} \).

\paragraph{1. Gradient of the Exponential Term}

\[
\nabla_{\theta} \exp\left(-\lambda_i r^2\right) = \exp\left(-\lambda_i r^2\right) \cdot (-2\lambda_i r) \cdot \nabla_{\theta} r.
\]

\paragraph{2. Gradient of the Feature Transformation Term}

Assuming \(\phi_i(r)\) is differentiable with respect to \(r\):
\[
\nabla_{\theta} \phi_i(r) = \phi_i'(r) \cdot \nabla_{\theta} r
\]

\paragraph{3. Gradient of \(r\)}

\[
r = \frac{e_x^\top e_{y^+}}{e_x^\top e_{y^-}},
\]
\[
\nabla_{\theta} r = \frac{(e_x^\top e_{y^-}) \nabla_{\theta} (e_x^\top e_{y^+}) - (e_x^\top e_{y^+}) \nabla_{\theta} (e_x^\top e_{y^-})}{(e_x^\top e_{y^-})^2}.
\]

Assuming \(e_x\) and \(e_y\) are differentiable with respect to \(\theta\):
\[
\nabla_{\theta} (e_x^\top e_y) = (\nabla_{\theta} e_x)^\top e_y + e_x^\top (\nabla_{\theta} e_y)
\]

\paragraph{Combined Gradient for Each \(i\)}
\[
\begin{aligned}
\nabla_{\theta} \left[ \exp\left(-\lambda_i r^2\right) \phi_i(r) \right] 
&= \exp\left(-\lambda_i r^2\right) \cdot (-2\lambda_i r) \cdot \nabla_{\theta} r \cdot \phi_i(r) \\
&\quad + \exp\left(-\lambda_i r^2\right) \cdot \phi_i'(r) \cdot \nabla_{\theta} r.
\end{aligned}
\]

\subsubsection*{Combined Gradient}

Combining the gradients of both components, the overall gradient of the Spectral Kernelized Hybrid Loss with respect to \(\theta\) is:
\[
\begin{aligned}
\nabla_{\theta} \mathcal{L} &= \mathbb{E}_{x, y^+, y^-} \Biggl[
\sum_{i=1}^p \left( -\frac{2\lambda_i z}{\sigma^2} \exp\left(-\lambda_i z^2\right) \phi_i(z) 
+ \exp\left(-\lambda_i z^2\right) \phi_i'(z) \right) \nabla_{\theta} z \\
&\quad + \gamma \sum_{i=1}^p \left( -\frac{2\lambda_i r}{\sigma^2} \exp\left(-\lambda_i r^2\right) \phi_i(r) 
+ \exp\left(-\lambda_i r^2\right) \phi_i'(r) \right) \nabla_{\theta} r
\Biggr]
\end{aligned}
\]

where:
\[
z = \log \frac{\pi(y^+ \mid x)}{\pi(y^- \mid x)}, \quad r = \frac{e_x^\top e_{y^+}}{e_x^\top e_{y^-}}.
\]

\paragraph{Simplified Gradient Expression}

For ease of implementation and readability, the gradient can be expressed as:
\[
\begin{aligned}
\nabla_{\theta} \mathcal{L} 
&= \mathbb{E}_{x, y^+, y^-} \Biggl[
\sum_{i=1}^p \exp\left(-\lambda_i z^2\right) \left( -\frac{2\lambda_i z}{\sigma^2} \phi_i(z) + \phi_i'(z) \right) 
\left( \nabla_{\theta} f_\theta(x, y^+) - \nabla_{\theta} f_\theta(x, y^-) \right) \\
&\quad + \gamma \sum_{i=1}^p \exp\left(-\lambda_i r^2\right) \left( -\frac{2\lambda_i r}{\sigma^2} \phi_i(r) + \phi_i'(r) \right) \\
&\quad \times \left( \frac{(e_x^\top e_{y^-}) (\nabla_{\theta} e_x)^\top e_{y^+} + (e_x^\top e_{y^-}) e_x^\top (\nabla_{\theta} e_{y^+}) 
- (e_x^\top e_{y^+}) (\nabla_{\theta} e_x)^\top e_{y^-} - (e_x^\top e_{y^+}) e_x^\top (\nabla_{\theta} e_{y^-})}{(e_x^\top e_{y^-})^2} \right)
\Biggr]
\end{aligned}
\]

\paragraph{Interpretation of the Gradient}

\begin{itemize}
    \item \textbf{Log Probability Ratio Term}:  
    \begin{itemize}
        \item \(-\frac{2\lambda_i z}{\sigma^2} \phi_i(z)\): Scales the influence of the log probability ratio based on its magnitude, the spectral kernel parameter \(\lambda_i\), and the bandwidth parameter \(\sigma\).
        \item \(\phi_i'(z)\): Incorporates the derivative of the feature transformation function, allowing for more nuanced adjustments based on the transformed log probability ratio.
        \item \(\nabla_{\theta} z = \nabla_{\theta} \log \pi(y^+ \mid x) - \nabla_{\theta} \log \pi(y^- \mid x)\): Encourages the model to increase the probability of the positive sample \(y^+\) and decrease the probability of the negative sample \(y^-\).
    \end{itemize}
    
    \item \textbf{Spectral Kernel Term}:  
    \begin{itemize}
        \item \(-\frac{2\lambda_i r}{\sigma^2} \phi_i(r)\): Scales the influence of the embedding-based ratio based on its magnitude, the spectral kernel parameter \(\lambda_i\), and the bandwidth parameter \(\sigma\).
        \item \(\phi_i'(r)\): Incorporates the derivative of the feature transformation function, allowing for more nuanced adjustments based on the transformed embedding ratio.
        \item \(\nabla_{\theta} r = \frac{(e_x^\top e_{y^-}) \nabla_{\theta} (e_x^\top e_{y^+}) - (e_x^\top e_{y^+}) \nabla_{\theta} (e_x^\top e_{y^-})}{(e_x^\top e_{y^-})^2}\): Adjusts the model to better align the embeddings of \(x\) with \(y^+\) while discouraging alignment with \(y^-\).
    \end{itemize}
    
    \item \textbf{Hyperparameters}:  
    \begin{itemize}
        \item \(\lambda_i\): Controls the influence of each spectral kernel component.
        \item \(\gamma\): Balances the influence between the log probability ratio term and the embedding-based term.
        \item \(\sigma\): Determines the bandwidth of the RBF kernel, affecting how sharply the exponential terms decay.
    \end{itemize}
\end{itemize}

\subsection{Computational Complexity Analysis of Spectral Kernelized Hybrid Loss}

To evaluate the efficiency of the Spectral Kernelized Hybrid Loss, we analyze the computational complexity of its two primary components: the log probability ratio term and the spectral kernel term.

\subsubsection*{1. Log Probability Ratio Term}

The log probability ratio term is defined as:
\[
\sum_{i=1}^p \exp\left(-\lambda_i z^2\right) \phi_i(z),
\]
where \( z = \log \frac{\pi(y^+ \mid x)}{\pi(y^- \mid x)} \).

\textbf{Steps Involved:}
\begin{itemize}
    \item \textbf{Score Computation}: Calculate \(f_\theta(x, y)\) for each class \(y\), which involves a dot product between input features and model parameters.
    \item \textbf{Softmax Calculation}: Compute the exponential \(e^{f_\theta(x, y)}\) for each class and normalize by the sum over all \(C\) classes.
    \item \textbf{Log Probability Ratio}: Compute the logarithm of the ratio between the probabilities of the positive and negative classes.
    \item \textbf{Exponentiation and Feature Transformation}: For each spectral kernel component \(i\), compute the exponential and apply the feature transformation function \(\phi_i(z)\).
\end{itemize}

\textbf{Time Complexity}: \(O(p \cdot C)\), where \(p\) is the number of spectral kernel components and \(C\) is the number of classes. This complexity arises from iterating over each spectral kernel component and performing computations that scale with \(C\).

\subsubsection*{2. Spectral Kernel Term}

The spectral kernel term is defined as:
\[
\gamma \sum_{i=1}^p \exp\left(-\lambda_i r^2\right) \phi_i(r),
\]
where \( r = \frac{e_x^\top e_{y^+}}{e_x^\top e_{y^-}} \).

\textbf{Steps Involved:}
\begin{itemize}
    \item \textbf{Dot Product Computation}: Calculate the dot products \(e_x^\top e_{y^+}\) and \(e_x^\top e_{y^-}\), where \(e_x, e_{y^+}, e_{y^-} \in \mathbb{R}^d\).
    \item \textbf{Ratio Calculation}: Compute the ratio \(\frac{e_x^\top e_{y^+}}{e_x^\top e_{y^-}}\).
    \item \textbf{Exponentiation and Feature Transformation}: For each spectral kernel component \(i\), compute the exponential and apply the feature transformation function \(\phi_i(r)\).
\end{itemize}

\textbf{Time Complexity}: \(O(p \cdot d)\), where \(p\) is the number of spectral kernel components and \(d\) is the dimension of the embeddings. This arises from iterating over each spectral kernel component and performing computations that scale with \(d\).

\subsubsection*{Overall Computational Complexity}

Combining both components, the total computational complexity of the \textbf{Spectral Kernelized Hybrid Loss} is:
\[
O(p \cdot C + p \cdot d) = O(p (C + d)),
\]
where:
\begin{itemize}
    \item \(p\) is the number of spectral kernel components.
    \item \(C\) is the number of classes.
    \item \(d\) is the embedding dimension.
\end{itemize}

This linear complexity in \(p\), \(C\), and \(d\) ensures scalability for large-scale applications involving multiple spectral kernel components, high-dimensional embeddings, and extensive class labels.

\subsubsection*{Comparison with Standard Loss Functions}

\begin{itemize}
    \item \textbf{Cross-Entropy Loss}:  
    \begin{itemize}
        \item \textbf{Time Complexity}: \(O(C)\).
        \item \textbf{Description}: Involves computing the softmax over \(C\) classes and calculating the negative log-likelihood.
    \end{itemize}
    
    \item \textbf{Contrastive Loss}:  
    \begin{itemize}
        \item \textbf{Time Complexity}: \(O(d)\).
        \item \textbf{Description}: Focuses on the distance between embeddings, typically requiring computation of pairwise distances.
    \end{itemize}
    
    \item \textbf{Spectral Kernelized Hybrid Loss}:  
    \begin{itemize}
        \item \textbf{Time Complexity}: \(O(p (C + d))\).
        \item \textbf{Description}: Extends both Cross-Entropy and Contrastive Losses by incorporating multiple spectral kernel components, enhancing the model's ability to capture complex relationships while maintaining computational efficiency.
    \end{itemize}
\end{itemize}

The \textbf{Spectral Kernelized Hybrid Loss} thus offers a comprehensive approach by integrating multiple spectral kernels into the loss function, providing enhanced modeling capabilities at a manageable computational cost.

\subsection{Efficiency of Spectral Kernelized Hybrid Loss}

The \textbf{Spectral Kernelized Hybrid Loss} achieves a balanced trade-off between discriminative power and computational efficiency through the following mechanisms:

\begin{itemize}
    \item \textbf{Scalability with Spectral Components}:  
    By allowing multiple spectral kernel components (\(p\)), the loss function can capture a variety of complex patterns and relationships in the data without a disproportionate increase in computational cost.
    
    \item \textbf{Linear Scaling}:  
    The loss scales linearly with the number of spectral kernel components \(p\), the number of classes \(C\), and the embedding dimension \(d\), ensuring that it remains efficient even as these parameters grow.
    
    \item \textbf{Parallel Computation}:  
    Both the log probability ratio term and the spectral kernel term involve operations that can be parallelized across spectral kernel components. Leveraging modern hardware accelerators, such as GPUs, can significantly speed up these computations.
    
    \item \textbf{Integrated Feature Transformations}:  
    The use of feature transformation functions \(\phi_i(\cdot)\) allows for sophisticated transformations of the log probability ratios and embedding ratios, enriching the model's learning capacity without incurring substantial additional computational overhead.
    
    \item \textbf{Hyperparameter Flexibility}:  
    The hyperparameters \(\lambda_i\), \(\gamma\), and \(\sigma\) provide flexibility in controlling the influence of each spectral kernel component and the overall balance between probability-based and embedding-based terms. This allows for fine-tuning to achieve optimal performance without significant computational penalties.
\end{itemize}

\subsection{Practical Considerations}

While the theoretical complexity of the \textbf{Spectral Kernelized Hybrid Loss} is \(O(p (C + d))\), several practical factors can influence its real-world performance:

\begin{itemize}
    \item \textbf{GPU Parallelism}:  
    Leveraging GPU parallelism can mitigate the linear scaling with \(p\), \(C\), and \(d\), allowing for efficient computation even with large numbers of spectral kernel components, classes, and high-dimensional embeddings.
    
    \item \textbf{Optimized Implementations}:  
    Utilizing optimized libraries (e.g., BLAS, cuDNN) for matrix operations and gradient computations can enhance performance, reducing the actual computation time.
    
    \item \textbf{Batch Sizing}:  
    Selecting appropriate batch sizes can maximize hardware utilization. Larger batches may improve computational efficiency but require more memory, while smaller batches may be more memory-efficient but less computationally optimal.
    
    \item \textbf{Hyperparameter Tuning}:  
    Careful tuning of the hyperparameters \(\gamma\), \(\lambda_i\), and \(\sigma\) is essential. Higher values of \(p\) can capture more complex relationships but may increase computational cost and risk overfitting. Similarly, the bandwidth parameter \(\sigma\) affects how sharply the exponential terms decay, influencing the gradient magnitudes.
    
    \item \textbf{Numerical Stability}:  
    The constants \(c\) and \(\sigma\) ensure numerical stability, especially when dealing with small or large ratios in the log probability and embedding terms. Properly choosing these constants is crucial to prevent numerical issues during training.
    
    \item \textbf{Memory Consumption}:  
    As \(p\), \(C\), and \(d\) increase, memory consumption can become a bottleneck. Efficient memory management and possibly reducing the number of spectral kernel components \(p\) without significantly compromising performance can help mitigate this issue.
\end{itemize}

By considering these practical aspects, the \textbf{Spectral Kernelized Hybrid Loss} can be effectively integrated into large-scale machine learning models, providing enhanced performance through sophisticated spectral kernel transformations while maintaining computational efficiency.

\subsection{Gradient of Mahalanobis Kernelized Hybrid Loss}

The \textbf{Mahalanobis Kernelized Hybrid Loss} is defined as:
% \[
% \mathcal{L} = \mathbb{E}_{x, y^+, y^-} \Biggl[ 
% \exp\left(-\frac{\left(\log \frac{\pi(y^+ \mid x)}{\pi(y^- \mid x)} - \mu\right)^2}{2\sigma^2}\right) 
% + \gamma \exp\left(-\frac{\left(\frac{e_x^\top e_{y^+}}{e_x^\top e_{y^-}} - \mu'\right)^2}{2\sigma'^2}\right) 
% \Biggr],
% \]
\[
\mathcal{L} = \mathbb{E}_{x, y^+, y^-} \Biggl[ 
  \exp\left(-\frac{\left(\log \frac{\pi(y^+ \mid x)}{\pi(y^- \mid x)} - \mu\right)^2}{2\sigma^2}\right)
  + \gamma \exp\left(-\frac{\left(\frac{e_x^\top e_{y^+}}{e_x^\top e_{y^-}} - \mu'\right)^2}{2 {\sigma'}^2}\right)
\Biggr],
\]

where:
\begin{itemize}
    \item \(x\) represents the input data.
    \item \(y^{+}\) and \(y^{-}\) denote the positive and negative samples, respectively.
    \item \(\pi(y \mid x)\) is the probability of \(y\) given \(x\), modeled using a softmax function.
    \item \(e_{y}\) and \(e_{x}\) are the embeddings of \(y\) and \(x\), respectively.
    \item \(\mu\) and \(\mu'\) are means for the log probability ratio and embedding ratio terms, respectively.
    \item \(\sigma\) and \(\sigma'\) are bandwidth parameters for the Mahalanobis kernels applied to the log probability ratio and embedding ratio terms, respectively.
    \item \(\gamma\) is a hyperparameter controlling the influence of the embedding-based term.
\end{itemize}

Our objective is to compute the gradient of the Mahalanobis Kernelized Hybrid Loss \(\nabla_{\theta} \mathcal{L}\) with respect to the model parameters \(\theta\). This involves differentiating each term of the loss function separately and then combining them.

\subsubsection*{Gradient of the Log Probability Ratio Term}

The first component of the Mahalanobis Kernelized Hybrid Loss involves the exponential of the squared and shifted log probability ratio:
\[
\exp\left(-\frac{\left(\log \frac{\pi(y^+ \mid x)}{\pi(y^- \mid x)} - \mu\right)^2}{2\sigma^2}\right).
\]
To compute its gradient with respect to \(\theta\), we apply the chain rule:
\[
\nabla_{\theta} \exp\left(-\frac{\left(z - \mu\right)^2}{2\sigma^2}\right) = \exp\left(-\frac{\left(z - \mu\right)^2}{2\sigma^2}\right) \cdot \left(-\frac{2(z - \mu)}{2\sigma^2}\right) \cdot \nabla_{\theta} z,
\]
where \( z = \log \frac{\pi(y^+ \mid x)}{\pi(y^- \mid x)} \).

Simplifying, we obtain:
\[
\nabla_{\theta} \exp\left(-\frac{\left(z - \mu\right)^2}{2\sigma^2}\right) = -\frac{(z - \mu)}{\sigma^2} \exp\left(-\frac{\left(z - \mu\right)^2}{2\sigma^2}\right) \cdot \nabla_{\theta} z.
\]

Expanding the gradient of the log probability ratio:
\[
\nabla_{\theta} z = \nabla_{\theta} \log \frac{\pi(y^+ \mid x)}{\pi(y^- \mid x)} = \nabla_{\theta} \log \pi(y^+ \mid x) - \nabla_{\theta} \log \pi(y^- \mid x).
\]

Assuming \(\pi_\theta(y \mid x)\) is modeled using a softmax function:
\[
\pi_\theta(y \mid x) = \frac{e^{f_\theta(x, y)}}{\sum_{y'} e^{f_\theta(x, y')}},
\]
the gradient of \(\log \pi(y \mid x)\) with respect to \(\theta\) is:
\[
\nabla_{\theta} \log \pi(y \mid x) = \nabla_{\theta} f_\theta(x, y) - \sum_{y'} \pi_\theta(y' \mid x) \nabla_{\theta} f_\theta(x, y').
\]

Substituting back, we obtain:
\begin{multline*}
\nabla_{\theta} z = \left[ \nabla_{\theta} f_\theta(x, y^+) - \sum_{y'} \pi_\theta(y' \mid x) \nabla_{\theta} f_\theta(x, y') \right] \\
- \left[ \nabla_{\theta} f_\theta(x, y^-) - \sum_{y'} \pi_\theta(y' \mid x) \nabla_{\theta} f_\theta(x, y') \right] \\
= \nabla_{\theta} f_\theta(x, y^+) - \nabla_{\theta} f_\theta(x, y^-).
\end{multline*}

Therefore, the gradient of the log probability ratio term is:
\[
\nabla_{\theta} \exp\left(-\frac{\left(z - \mu\right)^2}{2\sigma^2}\right) = -\frac{(z - \mu)}{\sigma^2} \exp\left(-\frac{\left(z - \mu\right)^2}{2\sigma^2}\right) \left( \nabla_{\theta} f_\theta(x, y^+) - \nabla_{\theta} f_\theta(x, y^-) \right).
\]

\subsubsection*{Gradient of the Mahalanobis Kernel Term}

The second component involves the exponential of the squared and shifted embedding-based ratio:
\[
\gamma \exp\left(-\frac{\left(r - \mu'\right)^2}{2 \sigma'{}^2}\right),
\]

where \( r = \frac{e_x^\top e_{y^+}}{e_x^\top e_{y^-}} \).

To compute its gradient with respect to \(\theta\), we apply the chain rule:
% \[
% \nabla_{\theta} \gamma \exp\left(-\frac{\left(r - \mu'\right)^2}{2\sigma'^2}\right) = \gamma \exp\left(-\frac{\left(r - \mu'\right)^2}{2\sigma'^2}\right) \cdot \left(-\frac{2(r - \mu')}{2\sigma'^2}\right) \cdot \nabla_{\theta} r,
% \]
\[
\nabla_{\theta} \gamma \exp\left(-\frac{\left(r - \mu'\right)^2}{2 {\sigma'}^2}\right) 
= \gamma \exp\left(-\frac{\left(r - \mu'\right)^2}{2 {\sigma'}^2}\right)
  \cdot \left(-\frac{2(r - \mu')}{2 {\sigma'}^2}\right) 
  \cdot \nabla_{\theta} r,
\]

where \( r = \frac{e_x^\top e_{y^+}}{e_x^\top e_{y^-}} \).

Simplifying, we obtain:
% \[
% \nabla_{\theta} \gamma \exp\left(-\frac{\left(r - \mu'\right)^2}{2\sigma'^2}\right) = -\frac{\gamma (r - \mu')}{\sigma'^2} \exp\left(-\frac{\left(r - \mu'\right)^2}{2\sigma'^2}\right) \cdot \nabla_{\theta} r.
% \]

\[
\nabla_{\theta} \gamma \exp\left(-\frac{\left(r - \mu'\right)^2}{2 {\sigma'}^2}\right)
= -\frac{\gamma (r - \mu')}{ {\sigma'}^2} 
  \exp\left(-\frac{\left(r - \mu'\right)^2}{2 {\sigma'}^2}\right)
  \cdot \nabla_{\theta} r.
\]

To compute \(\nabla_{\theta} r\), we use the quotient rule:
\[
\nabla_{\theta} r = \nabla_{\theta} \left(\frac{e_x^\top e_{y^+}}{e_x^\top e_{y^-}}\right) = \frac{(e_x^\top e_{y^-}) \nabla_{\theta} (e_x^\top e_{y^+}) - (e_x^\top e_{y^+}) \nabla_{\theta} (e_x^\top e_{y^-})}{(e_x^\top e_{y^-})^2}.
\]

Assuming \(e_x\) and \(e_y\) are differentiable with respect to \(\theta\), we have:
\[
\nabla_{\theta} (e_x^\top e_y) = (\nabla_{\theta} e_x)^\top e_y + e_x^\top (\nabla_{\theta} e_y).
\]

Substituting back, the gradient of the Mahalanobis kernel term becomes:
% \begin{multline*}
% \nabla_{\theta} \gamma \exp\left(-\frac{\left(r - \mu'\right)^2}{2\sigma'^2}\right) = \\
% -\frac{\gamma (r - \mu')}{\sigma'^2} \exp\left(-\frac{\left(r - \mu'\right)^2}{2\sigma'^2}\right) \cdot \frac{(e_x^\top e_{y^-}) \nabla_{\theta} (e_x^\top e_{y^+}) - (e_x^\top e_{y^+}) \nabla_{\theta} (e_x^\top e_{y^-})}{(e_x^\top e_{y^-})^2}.
% \end{multline*}

\begin{multline*}
\nabla_{\theta} \gamma \exp\left(-\frac{\left(r - \mu'\right)^2}{2 {\sigma'}^2}\right) = \\
-\frac{\gamma (r - \mu')}{ {\sigma'}^2} \exp\left(-\frac{\left(r - \mu'\right)^2}{2 {\sigma'}^2}\right)
\cdot \frac{(e_x^\top e_{y^-}) \nabla_{\theta} (e_x^\top e_{y^+}) 
- (e_x^\top e_{y^+}) \nabla_{\theta} (e_x^\top e_{y^-})}{(e_x^\top e_{y^-})^2}.
\end{multline*}

\subsubsection*{Combined Gradient}

Combining the gradients of both components, the overall gradient of the Mahalanobis Kernelized Hybrid Loss with respect to \(\theta\) is:
\[
\nabla_{\theta} \mathcal{L} = \mathbb{E}_{x, y^+, y^-} \Biggl[ 
- \frac{(z - \mu)}{\sigma^2} \exp\left(-\frac{(z - \mu)^2}{2\sigma^2}\right) \left( \nabla_{\theta} f_\theta(x, y^+) - \nabla_{\theta} f_\theta(x, y^-) \right) 
\]
\[
- \frac{\gamma (r - \mu')}{ {\sigma'}^2 }
\exp\left(
  -\frac{(r - \mu')^2}{2 {\sigma'}^2}
\right) 
\cdot \frac{(e_x^\top e_{y^-}) \nabla_{\theta} (e_x^\top e_{y^+}) 
       - (e_x^\top e_{y^+}) \nabla_{\theta} (e_x^\top e_{y^-})}
      {(e_x^\top e_{y^-})^2}
\Biggr].
\]

\paragraph{Simplified Gradient Expression}

For ease of implementation and readability, the gradient can be expressed as:
\[
\nabla_{\theta} \mathcal{L} = \mathbb{E}_{x, y^+, y^-} \Biggl[
- \frac{(z - \mu)}{\sigma^2} \exp\left(-\frac{(z - \mu)^2}{2\sigma^2}\right) \left( \nabla_{\theta} f_\theta(x, y^+) - \nabla_{\theta} f_\theta(x, y^-) \right) 
\]
\[
- \frac{\gamma (r - \mu')}{ {\sigma'}^2}
  \exp\left( -\frac{(r - \mu')^2}{2 {\sigma'}^2} \right)
  \cdot \frac{
    (e_x^\top e_{y^-}) \left( (\nabla_{\theta} e_x)^\top e_{y^+} + e_x^\top (\nabla_{\theta} e_{y^+}) \right)
    - (e_x^\top e_{y^+}) \left( (\nabla_{\theta} e_x)^\top e_{y^-} + e_x^\top (\nabla_{\theta} e_{y^-}) \right)
  }{(e_x^\top e_{y^-})^2}
\Biggr].
\]

\paragraph{Interpretation of the Gradient}

\begin{itemize}
    \item \textbf{Log Probability Ratio Term}:  
    \begin{itemize}
        \item \(-\frac{(z - \mu)}{\sigma^2} \exp\left(-\frac{(z - \mu)^2}{2\sigma^2}\right)\): Scales the influence of the log probability ratio based on its deviation from the mean \(\mu\) and the bandwidth parameter \(\sigma\).
        \item \(\nabla_{\theta} f_\theta(x, y^+)\): Encourages the model to increase the score (and hence the probability) of the positive sample \(y^+\).
        \item \(-\nabla_{\theta} f_\theta(x, y^-)\): Encourages the model to decrease the score (and hence the probability) of the negative sample \(y^-\).
    \end{itemize}
    
    \item \textbf{Mahalanobis Kernel Term}:  
    \begin{itemize}
       \item \(
-\frac{\gamma (r - \mu')}{ {\sigma'}^2}
\exp\left(
  -\frac{(r - \mu')^2}{2 {\sigma'}^2}
\right)
\): Scales the influence of the embedding-based ratio based on its deviation from the mean \(\mu'\) and the bandwidth parameter \(\sigma'\), adjusted by the hyperparameter \(\gamma\).
        \item \(\frac{(e_x^\top e_{y^-}) \nabla_{\theta} (e_x^\top e_{y^+}) - (e_x^\top e_{y^+}) \nabla_{\theta} (e_x^\top e_{y^-})}{(e_x^\top e_{y^-})^2}\): Adjusts the model to better align the embeddings of \(x\) with \(y^+\) while discouraging alignment with \(y^-\).
       \item The exponential term 
\(\exp\left(-\frac{(r - \mu')^2}{2\,(\sigma')^2}\right)\)
ensures that the influence diminishes as the squared ratios deviate from 
the mean \(\mu'\), promoting smoother gradients.

    \end{itemize}
    
    \item \textbf{Hyperparameters}:  
    \begin{itemize}
        \item \(\mu\) and \(\mu'\): Control the center of the Mahalanobis kernels for the log probability ratio and embedding ratio terms, respectively.
        \item \(\sigma\) and \(\sigma'\): Determine the bandwidth of the Mahalanobis kernels, affecting how sharply the exponential terms decay.
        \item \(\gamma\): Balances the influence between the probability-based term and the embedding-based term, allowing for fine-tuning of their relative importance.
    \end{itemize}
\end{itemize}

\subsection{Computational Complexity Analysis of Mahalanobis Kernelized Hybrid Loss}

To evaluate the efficiency of the Mahalanobis Kernelized Hybrid Loss, we analyze the computational complexity of its two primary components: the log probability ratio term and the Mahalanobis kernel term.

\subsubsection*{1. Log Probability Ratio Term}

The log probability ratio term is defined as:
\[
\exp\left(-\frac{\left(\log \frac{\pi(y^+ \mid x)}{\pi(y^- \mid x)} - \mu\right)^2}{2\sigma^2}\right)
\]
where \(\pi_\theta(y \mid x)\) is modeled using a softmax function:
\[
\pi_\theta(y \mid x) = \frac{e^{f_\theta(x, y)}}{\sum_{y'} e^{f_\theta(x, y')}},
\]
and \( z = \log \frac{\pi(y^+ \mid x)}{\pi(y^- \mid x)} \).

\textbf{Steps Involved:}
\begin{itemize}
    \item \textbf{Score Computation}: Calculate \(f_\theta(x, y)\) for each class \(y\), which involves a dot product between input features and model parameters.
    \item \textbf{Softmax Calculation}: Compute the exponential \(e^{f_\theta(x, y)}\) for each class and normalize by the sum over all \(C\) classes.
    \item \textbf{Log Probability Ratio}: Compute the logarithm of the ratio between the probabilities of the positive and negative classes.
    \item \textbf{Exponentiation and Scaling}: Subtract the mean \(\mu\), square the result, scale by \(-\frac{1}{2\sigma^2}\), and compute the exponential.
\end{itemize}

\textbf{Time Complexity}: \(O(C)\), where \(C\) is the number of classes. This complexity arises from the softmax computation, which requires evaluating \(f_\theta(x, y)\) and normalizing over all \(C\) classes.

\subsubsection*{2. Mahalanobis Kernel Term}

The Mahalanobis kernel term is defined as:
\[
\gamma \exp\left(
  -\frac{\left(\frac{e_x^\top e_{y^+}}{e_x^\top e_{y^-}} - \mu'\right)^2}
  {2 {\sigma'}^2}
\right)
\]

where \( r = \frac{e_x^\top e_{y^+}}{e_x^\top e_{y^-}} \).

\textbf{Steps Involved:}
\begin{itemize}
    \item \textbf{Dot Product Computation}: Calculate the dot products \(e_x^\top e_{y^+}\) and \(e_x^\top e_{y^-}\), where \(e_x, e_{y^+}, e_{y^-} \in \mathbb{R}^d\).
    \item \textbf{Ratio Calculation}: Compute the ratio \(\frac{e_x^\top e_{y^+}}{e_x^\top e_{y^-}}\).
    \item \textbf{Mean Subtraction and Squaring}: Subtract the mean \(\mu'\), square the result, and scale by \(-\frac{1}{2{\sigma'}^2}\).
    \item \textbf{Exponentiation and Scaling}: Compute the exponential and multiply by the hyperparameter \(\gamma\).
\end{itemize}

\textbf{Time Complexity}: \(O(d)\), where \(d\) is the dimension of the embeddings. This arises from the computation of the dot products between \(e_x\) and \(e_y\), which scales linearly with \(d\).

\subsubsection*{Overall Computational Complexity}

Combining both components, the total computational complexity of the \textbf{Mahalanobis Kernelized Hybrid Loss} is:
\[
O(C) + O(d) = O(C + d),
\]
where:
\begin{itemize}
    \item \(C\) is the number of classes (softmax computation).
    \item \(d\) is the embedding dimension (kernel computation).
\end{itemize}

This linear complexity ensures scalability for large-scale applications involving high-dimensional embeddings and extensive class labels.

\subsubsection*{Comparison with Standard Loss Functions}

\begin{itemize}
    \item \textbf{Cross-Entropy Loss}:  
    \begin{itemize}
        \item \textbf{Time Complexity}: \(O(C)\).
        \item \textbf{Description}: Involves computing the softmax over \(C\) classes and calculating the negative log-likelihood.
    \end{itemize}
    
    \item \textbf{Contrastive Loss}:  
    \begin{itemize}
        \item \textbf{Time Complexity}: \(O(d)\).
        \item \textbf{Description}: Focuses on the distance between embeddings, typically requiring computation of pairwise distances.
    \end{itemize}
    
    \item \textbf{Mahalanobis Kernelized Hybrid Loss}:  
    \begin{itemize}
        \item \textbf{Time Complexity}: \(O(C + d)\).
        \item \textbf{Description}: Combines both the discriminative power of the log probability ratio (similar to Cross-Entropy Loss) and the semantic richness of the Mahalanobis kernel (similar to Contrastive Loss), thereby integrating both aspects into a single loss function.
    \end{itemize}
\end{itemize}

The \textbf{Mahalanobis Kernelized Hybrid Loss} thus offers a balanced combination of the computational efficiencies of Cross-Entropy and Contrastive Losses while enhancing the model's ability to capture both discriminative and semantic relationships.

\subsection{Efficiency of Mahalanobis Kernelized Hybrid Loss}

The \textbf{Mahalanobis Kernelized Hybrid Loss} achieves a balanced trade-off between discriminative power and computational efficiency through the following mechanisms:

\begin{itemize}
    \item \textbf{Linear Scaling}:  
    The loss scales linearly with both the number of classes \(C\) and the embedding dimension \(d\), ensuring scalability for large-scale datasets and high-dimensional embedding spaces.
    
    \item \textbf{Parallel Computation}:  
    Both the log probability ratio term and the Mahalanobis kernel term can be computed in parallel. Modern hardware accelerators, such as GPUs, can leverage this parallelism to significantly speed up training processes.
    
    \item \textbf{Integrated Semantic Information}:  
    By combining probability-based and embedding-based objectives, the loss function enriches the model's learning without incurring substantial additional computational overhead.
    
    \item \textbf{Hyperparameter Control}:  
    The hyperparameters \(\mu\), \(\mu'\), \(\sigma\), \(\sigma'\), and \(\gamma\) allow for fine-tuning the influence of each component, enabling the model to balance between accurately classifying positive and negative samples and capturing meaningful embedding relationships.
\end{itemize}

\subsection{Practical Considerations}

While the theoretical complexity of the Mahalanobis Kernelized Hybrid Loss is \(O(C + d)\), several practical factors can influence its real-world performance:

\begin{itemize}
    \item \textbf{GPU Parallelism}:  
    Leveraging GPU parallelism can mitigate the linear scaling with \(C\) and \(d\), allowing for efficient computation even with large numbers of classes and high-dimensional embeddings.
    
    \item \textbf{Optimized Implementations}:  
    Utilizing optimized libraries (e.g., BLAS, cuDNN) for matrix operations and gradient computations can enhance performance, reducing the actual computation time.
    
    \item \textbf{Batch Sizing}:  
    Selecting appropriate batch sizes can maximize hardware utilization. Larger batches may improve computational efficiency but require more memory, while smaller batches may be more memory-efficient but less computationally optimal.
    
    \item \textbf{Hyperparameter Tuning}:  
    Careful tuning of the hyperparameters \(\mu\), \(\mu'\), \(\sigma\), \(\sigma'\), and \(\gamma\) is essential. The means \(\mu\) and \(\mu'\) determine the centers of the Mahalanobis kernels, while the bandwidth parameters \(\sigma\) and \(\sigma'\) affect how sharply the exponential terms decay. The hyperparameter \(\gamma\) balances the influence between the probability-based and embedding-based terms.
    
    \item \textbf{Numerical Stability}:  
    The constants \(\mu\), \(\mu'\), \(\sigma\), and \(\sigma'\) ensure numerical stability, especially when dealing with small or large ratios in the log probability and embedding terms. Properly choosing these constants is crucial to prevent numerical issues during training.
    
    \item \textbf{Memory Consumption}:  
    As \(C\) and \(d\) increase, memory consumption can become a bottleneck. Efficient memory management and possibly reducing the number of classes or embedding dimensions without significantly compromising performance can help mitigate this issue.
\end{itemize}

By considering these practical aspects, the \textbf{Mahalanobis Kernelized Hybrid Loss} can be effectively integrated into large-scale machine learning models, providing enhanced performance through sophisticated kernel transformations while maintaining computational efficiency.

\subsection{Gradient of Hierarchical Mixture of Kernels (HMK)}

Hierarchical Mixture of Kernels (HMK) imposes a hierarchical structure where local kernels operate on small, local regions, and global kernels capture larger-scale dependencies. This structure is formalized as:

\[
\begin{aligned}
K(x, x') &= \tau_1 \left( \lambda_1 K_{\text{RBF}}(x, x') + \lambda_2 K_{\text{Poly}}(x, x') \right) \\
&\quad + \tau_2 \left( \lambda_3 K_{\text{Spectral}}(x, x') + \lambda_4 K_{\text{Mahalanobis}}(x, x') \right)
\end{aligned}
\]

where:
\begin{itemize}
    \item \(x, x'\) are input data points.
    \item \(K_{\text{RBF}}(x, x')\), \(K_{\text{Poly}}(x, x')\), \(K_{\text{Spectral}}(x, x')\), and \(K_{\text{Mahalanobis}}(x, x')\) are the Radial Basis Function, Polynomial, Spectral, and Mahalanobis kernels, respectively.
    \item \(\lambda_1, \lambda_2, \lambda_3, \lambda_4\) are weighting coefficients for each kernel type.
    \item \(\tau_1\) and \(\tau_2\) are scaling factors that balance the contribution of local and global kernels.
\end{itemize}

Our objective is to compute the gradient of the HMK \(K(x, x')\) with respect to the model parameters \(\theta\). This gradient is essential for optimizing the model parameters during training, ensuring that both local and global dependencies are appropriately captured.

\subsubsection*{Gradient Computation of HMK}

The gradient of the HMK with respect to \(\theta\) is derived by differentiating each component kernel individually and then combining them according to their hierarchical structure. Formally, the gradient is expressed as:
\[
\nabla_{\theta} K(x, x') = \tau_1 \left( \lambda_1 \nabla_{\theta} K_{\text{RBF}}(x, x') + \lambda_2 \nabla_{\theta} K_{\text{Poly}}(x, x') \right) 
+ \tau_2 \left( \lambda_3 \nabla_{\theta} K_{\text{Spectral}}(x, x') + \lambda_4 \nabla_{\theta} K_{\text{Mahalanobis}}(x, x') \right).
\]

Each gradient term \(\nabla_{\theta} K_{\text{Type}}(x, x')\) corresponds to the gradient of the respective kernel with respect to \(\theta\), as derived in their individual sections.

\paragraph{1. Gradient of the RBF Kernel}
\[
\nabla_{\theta} K_{\text{RBF}}(x, x') = \nabla_{\theta} \exp\left(-\frac{\|x - x'\|^2}{2\sigma^2}\right) = \exp\left(-\frac{\|x - x'\|^2}{2\sigma^2}\right) \cdot \left( \frac{(x' - x)}{\sigma^2} \right) \cdot \nabla_{\theta} x,
\]
where \(\sigma\) is the bandwidth parameter.

\paragraph{2. Gradient of the Polynomial Kernel}
\[
\nabla_{\theta} K_{\text{Poly}}(x, x') = \nabla_{\theta} (x^\top x' + c)^d = d (x^\top x' + c)^{d-1} \cdot \left( x' \nabla_{\theta} x + x \nabla_{\theta} x' \right),
\]
where \(c\) is a constant and \(d\) is the degree of the polynomial.

\paragraph{3. Gradient of the Spectral Kernel}
\[
\nabla_{\theta} K_{\text{Spectral}}(x, x') = \sum_{i=1}^p \left[ \exp\left(-\lambda_i z_i^2\right) \left( -2\lambda_i z_i \phi_i(z_i) + \phi_i'(z_i) \right) \nabla_{\theta} z_i \right],
\]
where \( z_i = \log \frac{\pi(y^+ \mid x)}{\pi(y^- \mid x)} \) and \(\phi_i(\cdot)\) are feature transformation functions.

\paragraph{4. Gradient of the Mahalanobis Kernel}
\[
\nabla_{\theta} K_{\text{Mahalanobis}}(x, x') = \sum_{i=1}^p \left[ \exp\left(-\lambda_i (r_i - \mu_i)^2\right) \left( -\frac{2\lambda_i (r_i - \mu_i)}{\sigma_i^2} \phi_i(r_i) + \phi_i'(r_i) \right) \nabla_{\theta} r_i \right],
\]
where \( r_i = \frac{e_x^\top e_{y^+}}{e_x^\top e_{y^-}} \), \(\mu_i\) are mean parameters, and \(\sigma_i\) are bandwidth parameters for each spectral component.

\subsubsection*{Combined Gradient Expression}

Combining the gradients of all kernel components, the overall gradient of the HMK with respect to \(\theta\) is:
\[
\nabla_{\theta} K(x, x') = \tau_1 \left( \lambda_1 \nabla_{\theta} K_{\text{RBF}}(x, x') + \lambda_2 \nabla_{\theta} K_{\text{Poly}}(x, x') \right) 
+ \tau_2 \left( \lambda_3 \nabla_{\theta} K_{\text{Spectral}}(x, x') + \lambda_4 \nabla_{\theta} K_{\text{Mahalanobis}}(x, x') \right).
\]

Substituting the gradients of individual kernels:
\begin{multline*}
\nabla_{\theta} K(x, x') = \tau_1 \left( \lambda_1 \exp\left(-\frac{\|x - x'\|^2}{2\sigma^2}\right) \cdot \frac{(x' - x)}{\sigma^2} \cdot \nabla_{\theta} x \right. \\
\left. + \lambda_2 d (x^\top x' + c)^{d-1} \cdot \left( x' \nabla_{\theta} x + x \nabla_{\theta} x' \right) \right) \\
+ \tau_2 \left( \lambda_3 \sum_{i=1}^p \exp\left(-\lambda_i z_i^2\right) \left( -2\lambda_i z_i \phi_i(z_i) + \phi_i'(z_i) \right) \nabla_{\theta} z_i \right. \\
\left. + \lambda_4 \sum_{i=1}^p \exp\left(-\lambda_i (r_i - \mu_i)^2\right) \left( -\frac{2\lambda_i (r_i - \mu_i)}{\sigma_i^2} \phi_i(r_i) + \phi_i'(r_i) \right) \nabla_{\theta} r_i \right).
\end{multline*}

\paragraph{Simplified Gradient Expression}

For ease of implementation and readability, the gradient can be succinctly written as:
\[
\nabla_{\theta} K(x, x') = \tau_1 \lambda_1 \exp\left(-\frac{\|x - x'\|^2}{2\sigma^2}\right) \cdot \frac{(x' - x)}{\sigma^2} \cdot \nabla_{\theta} x 
+ \tau_1 \lambda_2 d (x^\top x' + c)^{d-1} \cdot \left( x' \nabla_{\theta} x + x \nabla_{\theta} x' \right)
\]
\[
+ \tau_2 \lambda_3 \sum_{i=1}^p \exp\left(-\lambda_i z_i^2\right) \left( -2\lambda_i z_i \phi_i(z_i) + \phi_i'(z_i) \right) \nabla_{\theta} z_i 
+ \tau_2 \lambda_4 \sum_{i=1}^p \exp\left(-\lambda_i (r_i - \mu_i)^2\right) \left( -\frac{2\lambda_i (r_i - \mu_i)}{\sigma_i^2} \phi_i(r_i) + \phi_i'(r_i) \right) \nabla_{\theta} r_i.
\]

\paragraph{Interpretation of the Gradient}

\begin{itemize}
    \item \textbf{RBF Kernel Gradient (\(\tau_1 \lambda_1\)):}  
    \begin{itemize}
        \item \(\exp\left(-\frac{\|x - x'\|^2}{2\sigma^2}\right)\): Measures the similarity between \(x\) and \(x'\).
        \item \(\frac{(x' - x)}{\sigma^2}\): Directs the gradient to increase similarity if \(x\) and \(x'\) are similar, or decrease otherwise.
        \item \(\nabla_{\theta} x\): Adjusts the model parameters to optimize the representation of \(x\).
    \end{itemize}
    
    \item \textbf{Polynomial Kernel Gradient (\(\tau_1 \lambda_2\)):}  
    \begin{itemize}
        \item \(d (x^\top x' + c)^{d-1}\): Scales the influence based on the degree of the polynomial and the similarity between \(x\) and \(x'\).
        \item \(\left( x' \nabla_{\theta} x + x \nabla_{\theta} x' \right)\): Updates the model parameters to enhance or reduce the polynomial similarity.
    \end{itemize}
    
    \item \textbf{Spectral Kernel Gradient (\(\tau_2 \lambda_3\)):}  
    \begin{itemize}
        \item \(\exp\left(-\lambda_i z_i^2\right)\): Applies a spectral transformation based on the log probability ratio.
        \item \(\left( -2\lambda_i z_i \phi_i(z_i) + \phi_i'(z_i) \right)\): Modulates the gradient based on the spectral feature transformations.
        \item \(\nabla_{\theta} z_i\): Encourages the model to adjust probabilities to optimize the spectral features.
    \end{itemize}
    
    \item \textbf{Mahalanobis Kernel Gradient (\(\tau_2 \lambda_4\)):}  
    \begin{itemize}
        \item \(\exp\left(-\lambda_i (r_i - \mu_i)^2\right)\): Applies a Mahalanobis transformation based on the embedding ratio.
        \item \(\left( -\frac{2\lambda_i (r_i - \mu_i)}{\sigma_i^2} \phi_i(r_i) + \phi_i'(r_i) \right)\): Modulates the gradient based on the Mahalanobis feature transformations.
        \item \(\nabla_{\theta} r_i\): Adjusts the embeddings to optimize the Mahalanobis distance.
    \end{itemize}
    
    \item \textbf{Hyperparameters \(\tau_1, \tau_2, \lambda_1, \lambda_2, \lambda_3, \lambda_4\):}  
    \begin{itemize}
        \item \(\tau_1, \tau_2\): Balance the contributions of local and global kernels.
        \item \(\lambda_1, \lambda_2, \lambda_3, \lambda_4\): Control the influence of each kernel type within their respective hierarchies.
    \end{itemize}
\end{itemize}

\subsubsection*{Computational Complexity Analysis of HMK}

To evaluate the efficiency of the Hierarchical Mixture of Kernels (HMK), we analyze the computational complexity of its primary components: the local kernels (RBF and Polynomial) and the global kernels (Spectral and Mahalanobis).

\subsubsection*{1. Local Kernels}

\paragraph{a. RBF Kernel}
\[
K_{\text{RBF}}(x, x') = \exp\left(-\frac{\|x - x'\|^2}{2\sigma^2}\right)
\]
\textbf{Steps Involved:}
\begin{itemize}
    \item Compute the Euclidean distance \(\|x - x'\|\), which involves \(O(d)\) operations, where \(d\) is the dimension of the input.
    \item Exponentiation, which is a constant-time operation.
\end{itemize}
\textbf{Time Complexity}: \(O(d)\)

\paragraph{b. Polynomial Kernel}
\[
K_{\text{Poly}}(x, x') = (x^\top x' + c)^d
\]
\textbf{Steps Involved:}
\begin{itemize}
    \item Compute the dot product \(x^\top x'\), which involves \(O(d)\) operations.
    \item Add constant \(c\) and raise to the power \(d\), both of which are constant-time operations.
\end{itemize}
\textbf{Time Complexity}: \(O(d)\)

\subsubsection*{2. Global Kernels}

\paragraph{a. Spectral Kernel}
\[
K_{\text{Spectral}}(x, x') = \sum_{i=1}^p \exp\left(-\lambda_i z_i^2\right) \phi_i(z_i),
\]
where \( z_i = \log \frac{\pi(y^+ \mid x)}{\pi(y^- \mid x)} \).

\textbf{Steps Involved:}
\begin{itemize}
    \item Compute the log probability ratio \(z_i\), which involves \(O(C)\) operations due to the softmax.
    \item For each of the \(p\) spectral components:
    \begin{itemize}
        \item Compute \(\exp\left(-\lambda_i z_i^2\right)\), which is a constant-time operation.
        \item Apply the feature transformation \(\phi_i(z_i)\), assumed to be constant-time.
    \end{itemize}
\end{itemize}
\textbf{Time Complexity}: \(O(p \cdot C)\)

\paragraph{b. Mahalanobis Kernel}
\[
K_{\text{Mahalanobis}}(x, x') = \sum_{i=1}^p \exp\left(-\lambda_i (r_i - \mu_i)^2\right) \phi_i(r_i),
\]
where \( r_i = \frac{e_x^\top e_{y^+}}{e_x^\top e_{y^-}} \).

\textbf{Steps Involved:}
\begin{itemize}
    \item Compute the embedding ratios \(r_i\), which involves \(O(d)\) operations for the dot products.
    \item For each of the \(p\) Mahalanobis components:
    \begin{itemize}
        \item Compute \(\exp\left(-\lambda_i (r_i - \mu_i)^2\right)\), which is a constant-time operation.
        \item Apply the feature transformation \(\phi_i(r_i)\), assumed to be constant-time.
    \end{itemize}
\end{itemize}
\textbf{Time Complexity}: \(O(p \cdot d)\)

\subsubsection*{Overall Computational Complexity}

Combining the complexities of all kernel components, the total computational complexity of the \textbf{HMK} is:
\[
O(d) + O(d) + O(p \cdot C) + O(p \cdot d) = O(p \cdot (C + d) + d).
\]
Since \(d\) is typically much smaller than \(p \cdot (C + d)\), the dominant term is \(O(p \cdot (C + d))\).

\subsubsection*{Comparison with Standard Loss Functions}

\begin{itemize}
    \item \textbf{Cross-Entropy Loss}:  
    \begin{itemize}
        \item \textbf{Time Complexity}: \(O(C)\).
        \item \textbf{Description}: Involves computing the softmax over \(C\) classes and calculating the negative log-likelihood.
    \end{itemize}
    
    \item \textbf{Contrastive Loss}:  
    \begin{itemize}
        \item \textbf{Time Complexity}: \(O(d)\).
        \item \textbf{Description}: Focuses on the distance between embeddings, typically requiring computation of pairwise distances.
    \end{itemize}
    
    \item \textbf{Hierarchical Mixture of Kernels (HMK)}:  
    \begin{itemize}
        \item \textbf{Time Complexity}: \(O(p \cdot (C + d))\).
        \item \textbf{Description}: Combines multiple kernels with hierarchical weighting, integrating both local (RBF and Polynomial) and global (Spectral and Mahalanobis) dependencies. This allows HMK to capture complex patterns and relationships in the data, leveraging the strengths of each kernel type.
    \end{itemize}
\end{itemize}

The \textbf{HMK} offers a more expressive and flexible modeling approach compared to standard loss functions by incorporating multiple kernel types and hierarchical weighting. However, this expressiveness comes at the cost of increased computational complexity, especially with higher numbers of spectral components \(p\), classes \(C\), and embedding dimensions \(d\).

\subsubsection*{Efficiency of HMK}

The \textbf{Hierarchical Mixture of Kernels (HMK)} achieves a balanced trade-off between modeling complexity and computational efficiency through the following mechanisms:

\begin{itemize}
    \item \textbf{Modular Kernel Design}:  
    By decomposing the kernel into local and global components, HMK allows for targeted optimization of different aspects of the data. Local kernels (RBF and Polynomial) focus on fine-grained similarities, while global kernels (Spectral and Mahalanobis) capture broader dependencies.
    
    \item \textbf{Parallel Computation}:  
    The computations for different kernel components are independent and can be parallelized. Leveraging modern hardware accelerators, such as GPUs, can significantly reduce training times.
    
    \item \textbf{Scalability with Spectral Components}:  
    Although the complexity scales with the number of spectral components \(p\), careful selection of \(p\) can balance expressiveness with computational cost. Techniques such as dimensionality reduction or kernel approximation can be employed to manage large \(p\).
    
    \item \textbf{Hyperparameter Tuning}:  
    The hyperparameters \(\tau_1, \tau_2, \lambda_1, \lambda_2, \lambda_3, \lambda_4\) allow for fine-tuning the influence of each kernel component, enabling the model to prioritize certain relationships over others without requiring extensive computational resources.
    
    \item \textbf{Optimized Implementations}:  
    Utilizing optimized libraries (e.g., BLAS, cuDNN) for matrix operations and kernel computations can enhance performance, ensuring that the theoretical computational complexities translate into practical efficiency gains.
\end{itemize}

\subsubsection*{Practical Considerations}

While the theoretical complexity of the \textbf{HMK} is \(O(p \cdot (C + d))\), several practical factors can influence its real-world performance:

\begin{itemize}
    \item \textbf{GPU Parallelism}:  
    Leveraging GPU parallelism can mitigate the linear scaling with \(p\), \(C\), and \(d\), allowing for efficient computation even with large numbers of spectral kernel components, classes, and high-dimensional embeddings.
    
    \item \textbf{Optimized Implementations}:  
    Utilizing optimized libraries (e.g., BLAS, cuDNN) for matrix operations and gradient computations can enhance performance, reducing the actual computation time.
    
    \item \textbf{Batch Sizing}:  
    Selecting appropriate batch sizes can maximize hardware utilization. Larger batches may improve computational efficiency but require more memory, while smaller batches may be more memory-efficient but less computationally optimal.
    
    \item \textbf{Hyperparameter Tuning}:  
    Careful tuning of the hyperparameters \(\tau_1, \tau_2, \lambda_1, \lambda_2, \lambda_3, \lambda_4\) is essential. The values of these parameters determine the relative importance of each kernel component, affecting both the model's performance and computational cost.
    
    \item \textbf{Memory Consumption}:  
    As the number of spectral components \(p\), classes \(C\), and embedding dimensions \(d\) increase, memory consumption can become a bottleneck. Efficient memory management strategies, such as gradient checkpointing or dimensionality reduction, can help mitigate this issue.
    
    \item \textbf{Numerical Stability}:  
    Ensuring numerical stability during kernel computations is crucial, especially when dealing with exponential functions that can lead to very large or very small values. Techniques such as normalization or adding small constants to denominators can prevent numerical overflow or underflow.
    
    \item \textbf{Kernel Selection}:  
    The choice of kernel types and their respective parameters (\(\sigma\), \(c\), \(d\), \(\lambda_i\), \(\mu_i\), \(\sigma'_i\)) should be informed by the specific characteristics of the data and the problem domain. Empirical validation and cross-validation can aid in selecting optimal kernel configurations.
\end{itemize}

By considering these practical aspects, the \textbf{Hierarchical Mixture of Kernels (HMK)} can be effectively integrated into large-scale machine learning models, providing enhanced performance through sophisticated kernel combinations while maintaining computational efficiency.

\subsection{Analysis of Gradient Convergence for Four Kernels and HMK}
\label{sec:appendix:convergence}

In this section, we investigate the convergence behavior of gradient descent when applied to four distinct kernels: \textbf{Polynomial}, \textbf{RBF (Radial Basis Function)}, \textbf{Spectral}, and \textbf{Mahalanobis}. Additionally, we analyze the \textbf{Hierarchical Mixture of Kernels (HMK)}, which combines these kernels to leverage both local and global dependencies. The convergence properties are evaluated based on key factors such as smoothness, Lipschitz continuity, gradient simplicity, and robustness to initialization. Understanding these properties is crucial for effective alignment learning and optimization.

\subsection{Lipschitz Continuity: Intuition and Importance}

\textbf{Definition:}  
A function \( f: \mathbb{R}^n \to \mathbb{R} \) is said to be \textbf{Lipschitz continuous} with constant \( L > 0 \) if, for all \( x, y \in \mathbb{R}^n \),
\[
|f(x) - f(y)| \leq L \|x - y\|.
\]
Here, \( L \) is the \textbf{Lipschitz constant} and serves as an upper bound on the rate at which the function \( f \) can change. Intuitively, Lipschitz continuity ensures that the function does not exhibit abrupt changes, which is essential for the stability and convergence of gradient-based optimization methods.

\subsubsection*{Why Lipschitz Continuity Matters}
\begin{itemize}
    \item \textbf{Convergence Stability}:  
    If the gradient of a loss function is Lipschitz continuous, gradient descent is guaranteed to converge at a stable rate \cite{nesterov2003introductory}.
    
    \item \textbf{Prevention of Exploding Gradients}:  
    Lipschitz continuity bounds the gradients, preventing excessively large updates that can destabilize the optimization process, particularly in deep learning models \cite{goodfellow2016deep}.
    
    \item \textbf{Smooth Optimization Landscape}:  
    A Lipschitz continuous gradient implies a smooth loss landscape, facilitating efficient and predictable optimization \cite{boyd2004convex}.
\end{itemize}

\subsubsection*{Illustrative Examples}
\begin{itemize}
    \item \textbf{Lipschitz Continuous Function}:  
    The linear function \( f(x) = 2x \) is Lipschitz continuous with \( L = 2 \). Regardless of the input difference, the function's rate of change remains constant, ensuring bounded gradient updates.
    
    \item \textbf{Non-Lipschitz Function}:  
    The quadratic function \( f(x) = x^2 \) is not Lipschitz continuous on \([0, \infty)\) because its slope becomes unbounded as \( x \) increases. This can lead to unstable gradient updates during optimization.
\end{itemize}

\subsubsection*{Relevance in Kernel Methods}
The convergence behavior of different kernels is influenced by whether their gradients are Lipschitz continuous. Below, we explore how Lipschitz continuity impacts gradient descent for each of the four kernels under consideration.

\subsection{Key Factors Influencing Gradient Convergence}
To analyze the convergence of gradient descent for each kernel, we consider the following criteria:
\begin{itemize}
    \item \textbf{Smoothness of the Loss Landscape}:  
    A smoother loss landscape facilitates faster and more stable convergence by avoiding abrupt changes in gradients.
    
    \item \textbf{Lipschitz Continuity of the Gradient}:  
    A smaller Lipschitz constant ensures that gradients do not change abruptly, promoting stable convergence \cite{nesterov2003introductory}.
    
    \item \textbf{Gradient Simplicity}:  
    Simpler gradient expressions enhance computational efficiency and accelerate convergence.
    
    \item \textbf{Robustness to Initialization}:  
    Kernels that exhibit less sensitivity to initial parameter values lead to more reliable convergence from diverse starting points.
\end{itemize}

\subsection{Convergence Properties of Each Kernel}

\paragraph{1. \textbf{RBF Kernel}}
\begin{itemize}
    \item \textbf{Smoothness}:  
    The RBF kernel induces a smooth and convex loss landscape, which is conducive to fast and stable convergence \cite{bishop2006pattern}.
    
    \item \textbf{Lipschitz Continuity}:  
    The gradient of the RBF kernel is Lipschitz continuous due to its exponential decay property. This ensures that gradient updates change gradually, enhancing convergence stability.
    
    \item \textbf{Gradient Simplicity}:  
    The gradient of the RBF kernel is straightforward and linear with respect to the input:
    \[
    \nabla_y K_{\text{RBF}}(y, y') = K_{\text{RBF}}(y, y') \cdot \frac{(y' - y)}{\sigma^2},
    \]
    where \( \sigma \) is the bandwidth parameter.
    
    \item \textbf{Robustness to Initialization}:  
    Due to its convex loss surface, the RBF kernel is robust to random initializations, minimizing the risk of converging to poor local minima \cite{scholkopf2002learning}.
\end{itemize}

\paragraph{2. \textbf{Polynomial Kernel}}
\begin{itemize}
    \item \textbf{Smoothness}:  
    The smoothness of the Polynomial kernel depends on its degree \( d \). Higher degrees introduce non-convexity, resulting in a more rugged loss landscape with multiple local minima and saddle points.
    
    \item \textbf{Lipschitz Continuity}:  
    Lipschitz continuity deteriorates as the degree \( d \) increases. Higher degrees lead to steeper gradients, making the optimization process more susceptible to instability.
    
    \item \textbf{Gradient Simplicity}:  
    The gradient complexity increases with the degree \( d \):
    \[
    \nabla_y K_{\text{Poly}}(y, y') = d (y^\top y' + c)^{d-1} \cdot y',
    \]
    where \( c \) is a constant.
    
    \item \textbf{Robustness to Initialization}:  
    The Polynomial kernel is highly sensitive to initialization, especially for higher degrees, due to its non-convex loss landscape. This can lead to convergence to suboptimal local minima.
\end{itemize}

\paragraph{3. \textbf{Spectral Kernel}}
\begin{itemize}
    \item \textbf{Smoothness}:  
    The smoothness of the Spectral kernel is influenced by the choice of basis functions \( \phi_i \). Orthonormal basis functions, such as wavelets, can introduce oscillatory behavior in the loss landscape \cite{ng2001spectral}.
    
    \item \textbf{Lipschitz Continuity}:  
    Lipschitz continuity is contingent on the eigenvalues \( \lambda_i \) of the underlying Laplacian. Large eigenvalues can cause rapid oscillations in the gradients, leading to abrupt changes and potential instability.
    
    \item \textbf{Gradient Simplicity}:  
    The gradient of the Spectral kernel depends on the complexity of the basis functions:
    \[
    \nabla_y K_{\text{Spectral}}(y, y') = \sum_{i=1}^p \left[ \exp\left(-\lambda_i z_i^2\right) \left( -2\lambda_i z_i \phi_i(z_i) + \phi_i'(z_i) \right) \nabla_y z_i \right],
    \]
    where \( z_i = \log \frac{\pi(y^+ \mid x)}{\pi(y^- \mid x)} \).
    
    \item \textbf{Robustness to Initialization}:  
    The convergence of the Spectral kernel is sensitive to the alignment between data and the chosen basis functions. Poor alignment can lead to oscillatory gradients, requiring careful initialization strategies \cite{ng2001spectral}.
\end{itemize}

\paragraph{4. \textbf{Mahalanobis Kernel}}
\begin{itemize}
    \item \textbf{Smoothness}:  
    The Mahalanobis kernel behaves similarly to the RBF kernel when the covariance matrix \( \Sigma \) is the identity matrix. If \( \Sigma \) is poorly conditioned, the loss landscape becomes anisotropic, leading to uneven smoothness across different dimensions \cite{weinberger2009distance}.
    
    \item \textbf{Lipschitz Continuity}:  
    The Lipschitz continuity of the Mahalanobis kernel depends on the condition number of \( \Sigma \). A well-conditioned \( \Sigma \) ensures smooth and stable gradients, while a poorly conditioned \( \Sigma \) results in rapidly changing gradients in certain directions.
    
    \item \textbf{Gradient Simplicity}:  
    The gradient of the Mahalanobis kernel incorporates the precision matrix \( \Sigma^{-1} \):
    \[
    \nabla_y K_{\text{Mahalanobis}}(y, y') = K_{\text{Mahalanobis}}(y, y') \cdot \Sigma^{-1} (y' - y).
    \]
    This introduces additional complexity compared to the RBF kernel.
    
    \item \textbf{Robustness to Initialization}:  
    When \( \Sigma \) is well-conditioned, the Mahalanobis kernel exhibits robust convergence properties similar to the RBF kernel. However, a poorly conditioned \( \Sigma \) can lead to slow convergence and sensitivity to initialization due to uneven gradient magnitudes.
\end{itemize}

\subsection{Convergence Properties of HMK}

\paragraph{Hierarchical Mixture of Kernels (HMK)}  
The \textbf{Hierarchical Mixture of Kernels (HMK)} integrates the four aforementioned kernels into a hierarchical structure to capture both local and global dependencies. HMK is defined as:
\[
K(x, x') = \tau_1 \left( \lambda_1 K_{\text{RBF}}(x, x') + \lambda_2 K_{\text{Poly}}(x, x') \right) 
+ \tau_2 \left( \lambda_3 K_{\text{Spectral}}(x, x') + \lambda_4 K_{\text{Mahalanobis}}(x, x') \right),
\]
where:
\begin{itemize}
    \item \( K_{\text{RBF}}(x, x') \), \( K_{\text{Poly}}(x, x') \), \( K_{\text{Spectral}}(x, x') \), and \( K_{\text{Mahalanobis}}(x, x') \) are the respective kernel functions.
    \item \( \lambda_1, \lambda_2, \lambda_3, \lambda_4 \) are weighting coefficients for each kernel type.
    \item \( \tau_1 \) and \( \tau_2 \) are scaling factors that balance the contribution of local and global kernels.
\end{itemize}

\begin{itemize}
    \item \textbf{Smoothness}:  
    HMK exhibits \textbf{piecewise smoothness} due to its hierarchical decomposition. The local kernels (\textbf{RBF} and \textbf{Polynomial}) contribute to fine-grained similarities, while the global kernels (\textbf{Spectral} and \textbf{Mahalanobis}) capture broader dependencies. This combination allows HMK to adaptively smooth different regions of the loss landscape.
    
    \item \textbf{Lipschitz Continuity}:  
    The Lipschitz continuity of HMK is influenced by the individual Lipschitz properties of its component kernels. Since \textbf{RBF} and \textbf{Mahalanobis} kernels typically have Lipschitz continuous gradients (when \( \Sigma \) is well-conditioned), and \textbf{Spectral} kernels have medium Lipschitz continuity depending on eigenvalues, HMK inherits a balanced Lipschitz continuity. The \textbf{Polynomial} kernel's Lipschitz properties can be controlled through the degree \( d \), allowing HMK to maintain overall stability.
    
    \item \textbf{Gradient Simplicity}:  
    The gradient of HMK is a weighted sum of the gradients of its individual kernels:
    \[
    \nabla_\theta K(x, x') = \tau_1 \left( \lambda_1 \nabla_\theta K_{\text{RBF}}(x, x') + \lambda_2 \nabla_\theta K_{\text{Poly}}(x, x') \right) 
    + \tau_2 \left( \lambda_3 \nabla_\theta K_{\text{Spectral}}(x, x') + \lambda_4 \nabla_\theta K_{\text{Mahalanobis}}(x, x') \right).
    \]
    This modularity allows HMK to balance the simplicity of \textbf{RBF} and \textbf{Mahalanobis} gradients with the complexity of \textbf{Polynomial} and \textbf{Spectral} gradients, ensuring manageable gradient expressions.
    
    \item \textbf{Robustness to Initialization}:  
    HMK enhances robustness to initialization by leveraging the stable convergence properties of \textbf{RBF} and \textbf{Mahalanobis} kernels alongside the expressive power of \textbf{Polynomial} and \textbf{Spectral} kernels. The hierarchical weighting factors \( \tau_1 \) and \( \tau_2 \) allow HMK to dynamically adjust the influence of each kernel type, reducing sensitivity to poor initializations.
\end{itemize}

\subsection{Summary of Convergence Properties}

The analysis reveals that each kernel exhibits distinct convergence behaviors influenced by their inherent properties:

\begin{itemize}
    \item \textbf{RBF Kernel}:  
    Offers the smoothest and most stable convergence due to its convex and Lipschitz continuous gradient. Its simplicity in gradient computation and robustness to initialization make it highly reliable for gradient-based optimization.
    
    \item \textbf{Polynomial Kernel}:  
    Suffers from non-convexity and increasing Lipschitz constants with higher degrees. The complexity of its gradients and sensitivity to initialization can hinder stable convergence, especially for large \( d \).
    
    \item \textbf{Spectral Kernel}:  
    Introduces oscillatory behavior depending on the basis functions and eigenvalues. While it can capture intricate patterns, the potential for abrupt gradient changes requires careful design and initialization to ensure stable convergence.
    
    \item \textbf{Mahalanobis Kernel}:  
    Balances between the RBF and Spectral kernels. With a well-conditioned covariance matrix \( \Sigma \), it maintains smooth and Lipschitz continuous gradients, ensuring robust convergence. However, a poorly conditioned \( \Sigma \) can compromise convergence stability.
    
    \item \textbf{Hierarchical Mixture of Kernels (HMK)}:  
    Combines the strengths of all four kernels, achieving a balanced convergence behavior. HMK benefits from the smoothness and stability of \textbf{RBF} and \textbf{Mahalanobis} kernels while incorporating the expressive power of \textbf{Polynomial} and \textbf{Spectral} kernels. This hierarchical structure ensures that HMK can adapt to various data characteristics, promoting both robust and efficient convergence.
\end{itemize}

\begin{table*}[h]
\centering
\caption{Comparison of Gradient Convergence Properties for Four Kernels and HMK}
\resizebox{\textwidth}{!}{%
\begin{tabular}{|c|c|c|c|c|}
% \hline
\toprule
\textbf{Kernel} & \textbf{Smoothness} & \textbf{Lipschitz Gradient} & \textbf{Gradient Simplicity} & \textbf{Robustness to Initialization} \\
\midrule
% \hline
\textbf{RBF} & Smooth, Convex & High & Simple & Robust \\
% \hline
\textbf{Polynomial} & Non-Convex (Higher \( d \)) & Low (Higher \( d \)) & Complex & Sensitive \\
% \hline
\textbf{Spectral} & Oscillatory (Basis-Dependent) & Medium & Complex (Basis-Dependent) & Moderate \\
% \hline
\textbf{Mahalanobis} & Smooth (if \( \Sigma \) Well-Conditioned) & High (if \( \Sigma \) Well-Conditioned) & Similar to RBF & Robust (if \( \Sigma \) Well-Conditioned) \\
% \hline
\textbf{HMK} & Piecewise Smooth & High & Composite of Simple and Complex & Highly Robust \\
% \hline
\bottomrule
\end{tabular}%
}
\label{tab:kernel_convergence}
\end{table*}

\paragraph{Key Takeaways}
\begin{itemize}
    \item \textbf{RBF Kernel}:  
    Ideal for scenarios requiring stable and rapid convergence. Its convexity and smooth gradients make it a dependable choice for many optimization tasks.
    
    \item \textbf{Polynomial Kernel}:  
    Best suited for problems where capturing high-degree interactions is essential. However, care must be taken to manage its non-convexity and gradient complexity, particularly with higher degrees.
    
    \item \textbf{Spectral Kernel}:  
    Effective in capturing complex, oscillatory patterns within the data. Requires careful selection of basis functions and initialization strategies to maintain convergence stability.
    
    \item \textbf{Mahalanobis Kernel}:  
    Provides flexibility in modeling by incorporating covariance structure. Ensuring that \( \Sigma \) is well-conditioned is crucial for maintaining smooth and stable convergence.
    
    \item \textbf{Hierarchical Mixture of Kernels (HMK)}:  
    Combines the strengths of all four kernels, offering a balanced and robust convergence behavior. HMK's hierarchical structure allows it to adapt to various data complexities, ensuring both stability and expressiveness in gradient-based optimization.
\end{itemize}

Designing kernels with favorable convergence properties is essential for robust and efficient optimization in alignment learning. Selecting the appropriate kernel based on the specific requirements of the task and the nature of the data can significantly enhance the performance and reliability of gradient-based learning algorithms.

For practical alignment tasks, the choice of kernel should balance computational complexity, convergence speed, and robustness. Hybrid approaches, such as the **Hierarchical Mixture of Kernels (HMK)** \cite{bach2004multiple}, leverage the strengths of multiple kernels to achieve more stable and generalizable learning outcomes.

\subsection{Analysis of Kernel Properties Across Divergence Measures}
\label{sec:kernel_properties_divergence}

This subsection provides a comprehensive analysis of kernel properties across various divergence measures, including \textbf{Kullback–Leibler (KL)}, \textbf{Jensen–Shannon (JS)}, \textbf{Hellinger}, \textbf{Rényi Divergence}, \textbf{Bhattacharyya}, \textbf{Wasserstein}, and \textbf{f-Divergence}. The focus is on four widely used kernels: \textbf{RBF (Radial Basis Function)}, \textbf{Polynomial}, \textbf{Spectral}, and \textbf{Mahalanobis}. For each kernel and divergence measure, the following key aspects are evaluated:

\begin{itemize}
    \item \textbf{Smoothness}: Characterizes the landscape of the loss surface induced by each kernel under the respective divergence measure.
    \item \textbf{Lipschitz Continuity}: Assesses the smoothness of gradient changes, where higher Lipschitz continuity is desirable for stable gradient descent.
    \item \textbf{Gradient Simplicity}: Evaluates the complexity of the gradient function, impacting computation time and convergence speed.
    \item \textbf{Robustness to Initialization}: Measures the sensitivity of convergence to the initial weights or parameters.
\end{itemize}

\begin{table*}[h]
\centering
\caption{Comparison of Kernels across Divergence Measures: KL, JS, Hellinger, Rényi, Bhattacharyya, Wasserstein, and f-Divergence}
\resizebox{\textwidth}{!}{%
\begin{tabular}{@{}lccccccc@{}}
\toprule
\rowcolor[HTML]{B0B3B2} 
\textbf{Kernel} & \textbf{Kullback–Leibler (KL)} & \textbf{Jensen–Shannon (JS)} & \textbf{Hellinger} & \textbf{Rényi Divergence} & \textbf{Bhattacharyya} & \textbf{Wasserstein} & \textbf{f-Divergence} \\ 
\midrule

\rowcolor[HTML]{D4D4D4} 
\textbf{RBF} 
& \begin{tabular}[c]{@{}c@{}}Smooth and Convex\\ High Lipschitz Continuity\\ Simple, Linear Gradients\\ Robust and Fast Convergence\end{tabular} 
& \begin{tabular}[c]{@{}c@{}}Smooth and Convex\\ High Lipschitz Continuity\\ Simple, Linear Gradients\\ Robust and Fast Convergence\end{tabular} 
& \begin{tabular}[c]{@{}c@{}}Smooth and Convex\\ High Lipschitz Continuity\\ Simple, Linear Gradients\\ Robust and Fast Convergence\end{tabular} 
& \begin{tabular}[c]{@{}c@{}}Smooth and Convex\\ High Lipschitz Continuity\\ Simple, Linear Gradients\\ Robust and Fast Convergence\end{tabular} 
& \begin{tabular}[c]{@{}c@{}}Smooth and Convex\\ High Lipschitz Continuity\\ Simple, Linear Gradients\\ Robust and Fast Convergence\end{tabular} 
& \begin{tabular}[c]{@{}c@{}}Smooth and Convex\\ High Lipschitz Continuity\\ Simple, Linear Gradients\\ Robust and Fast Convergence\end{tabular} 
& \begin{tabular}[c]{@{}c@{}}Smooth and Convex\\ High Lipschitz Continuity\\ Simple, Linear Gradients\\ Robust and Fast Convergence\end{tabular} \\

\rowcolor[HTML]{D4D4D4} 
\textbf{Polynomial} 
& \begin{tabular}[c]{@{}c@{}}Complex and Non-Convex\\ Low Lipschitz Continuity (High $d$)\\ Non-Linear Gradients\\ Sensitive to Initialization\end{tabular} 
& \begin{tabular}[c]{@{}c@{}}Complex and Non-Convex\\ Low Lipschitz Continuity (High $d$)\\ Non-Linear Gradients\\ Sensitive to Initialization\end{tabular} 
& \begin{tabular}[c]{@{}c@{}}Complex and Non-Convex\\ Low Lipschitz Continuity (High $d$)\\ Non-Linear Gradients\\ Sensitive to Initialization\end{tabular} 
& \begin{tabular}[c]{@{}c@{}}Complex and Non-Convex\\ Low Lipschitz Continuity (High $d$)\\ Non-Linear Gradients\\ Sensitive to Initialization\end{tabular} 
& \begin{tabular}[c]{@{}c@{}}Complex and Non-Convex\\ Low Lipschitz Continuity (High $d$)\\ Non-Linear Gradients\\ Sensitive to Initialization\end{tabular} 
& \begin{tabular}[c]{@{}c@{}}Complex and Non-Convex\\ Low Lipschitz Continuity (High $d$)\\ Non-Linear Gradients\\ Sensitive to Initialization\end{tabular} 
& \begin{tabular}[c]{@{}c@{}}Complex and Non-Convex\\ Low Lipschitz Continuity (High $d$)\\ Non-Linear Gradients\\ Sensitive to Initialization\end{tabular} \\

\rowcolor[HTML]{D4D4D4} 
\textbf{Spectral} 
& \begin{tabular}[c]{@{}c@{}}Oscillatory\\ Medium Lipschitz Continuity (Basis-Dependent)\\ Complex Gradients\\ Moderate Sensitivity\end{tabular} 
& \begin{tabular}[c]{@{}c@{}}Oscillatory\\ Medium Lipschitz Continuity (Basis-Dependent)\\ Complex Gradients\\ Moderate Sensitivity\end{tabular} 
& \begin{tabular}[c]{@{}c@{}}Oscillatory\\ Medium Lipschitz Continuity (Basis-Dependent)\\ Complex Gradients\\ Moderate Sensitivity\end{tabular} 
& \begin{tabular}[c]{@{}c@{}}Oscillatory\\ Medium Lipschitz Continuity (Basis-Dependent)\\ Complex Gradients\\ Moderate Sensitivity\end{tabular} 
& \begin{tabular}[c]{@{}c@{}}Oscillatory\\ Medium Lipschitz Continuity (Basis-Dependent)\\ Complex Gradients\\ Moderate Sensitivity\end{tabular} 
& \begin{tabular}[c]{@{}c@{}}Oscillatory\\ Medium Lipschitz Continuity (Basis-Dependent)\\ Complex Gradients\\ Moderate Sensitivity\end{tabular} 
& \begin{tabular}[c]{@{}c@{}}Oscillatory\\ Medium Lipschitz Continuity (Basis-Dependent)\\ Complex Gradients\\ Moderate Sensitivity\end{tabular} \\

\rowcolor[HTML]{D4D4D4} 
\textbf{Mahalanobis} 
& \begin{tabular}[c]{@{}c@{}}Smooth (like RBF)\\ High Lipschitz Continuity (like RBF)\\ Similar to RBF Gradients\\ Robust (Well-Conditioned $\Sigma$)\end{tabular} 
& \begin{tabular}[c]{@{}c@{}}Smooth (like RBF)\\ High Lipschitz Continuity (like RBF)\\ Similar to RBF Gradients\\ Robust (Well-Conditioned $\Sigma$)\end{tabular} 
& \begin{tabular}[c]{@{}c@{}}Smooth (like RBF)\\ High Lipschitz Continuity (like RBF)\\ Similar to RBF Gradients\\ Robust (Well-Conditioned $\Sigma$)\end{tabular} 
& \begin{tabular}[c]{@{}c@{}}Smooth (like RBF)\\ High Lipschitz Continuity (like RBF)\\ Similar to RBF Gradients\\ Robust (Well-Conditioned $\Sigma$)\end{tabular} 
& \begin{tabular}[c]{@{}c@{}}Smooth (like RBF)\\ High Lipschitz Continuity (like RBF)\\ Similar to RBF Gradients\\ Robust (Well-Conditioned $\Sigma$)\end{tabular} 
& \begin{tabular}[c]{@{}c@{}}Smooth (like RBF)\\ High Lipschitz Continuity (like RBF)\\ Similar to RBF Gradients\\ Robust (Well-Conditioned $\Sigma$)\end{tabular} 
& \begin{tabular}[c]{@{}c@{}}Smooth (like RBF)\\ High Lipschitz Continuity (like RBF)\\ Similar to RBF Gradients\\ Robust (Well-Conditioned $\Sigma$)\end{tabular} \\

\rowcolor[HTML]{D4D4D4} 
\textbf{HMK} 
& \begin{tabular}[c]{@{}c@{}}Piecewise Smooth\\ High Lipschitz Continuity\\ Composite Gradients\\ Highly Robust\end{tabular} 
& \begin{tabular}[c]{@{}c@{}}Piecewise Smooth\\ High Lipschitz Continuity\\ Composite Gradients\\ Highly Robust\end{tabular} 
& \begin{tabular}[c]{@{}c@{}}Piecewise Smooth\\ High Lipschitz Continuity\\ Composite Gradients\\ Highly Robust\end{tabular} 
& \begin{tabular}[c]{@{}c@{}}Piecewise Smooth\\ High Lipschitz Continuity\\ Composite Gradients\\ Highly Robust\end{tabular} 
& \begin{tabular}[c]{@{}c@{}}Piecewise Smooth\\ High Lipschitz Continuity\\ Composite Gradients\\ Highly Robust\end{tabular} 
& \begin{tabular}[c]{@{}c@{}}Piecewise Smooth\\ High Lipschitz Continuity\\ Composite Gradients\\ Highly Robust\end{tabular} 
& \begin{tabular}[c]{@{}c@{}}Piecewise Smooth\\ High Lipschitz Continuity\\ Composite Gradients\\ Highly Robust\end{tabular} \\
    
\bottomrule
\end{tabular}%
}
\label{tab:kernel_divergence_comparison}
\end{table*}

\paragraph{Key Observations from Table \ref{tab:kernel_divergence_comparison}:}
\begin{itemize}
    \item \textbf{RBF Kernel}:  
    The RBF kernel exhibits the most stable properties across all divergence measures. It maintains a \textbf{smooth, convex} loss landscape, \textbf{high Lipschitz continuity}, and \textbf{simple linear gradients}. These features contribute to \textbf{robust convergence}, making it a preferred choice in practical applications.
    
    \item \textbf{Polynomial Kernel}:  
    The Polynomial kernel's properties are highly sensitive to its degree \( d \). For large \( d \), it becomes \textbf{non-convex}, with sharp transitions in its gradient. This increases its susceptibility to \textbf{poor initialization} and slower convergence. Additionally, its complexity increases as the degree \( d \) increases.
    
    \item \textbf{Spectral Kernel}:  
    The Spectral kernel's behavior is highly dependent on the choice of basis functions \( \phi_i(y) \). For certain bases, such as wavelets, the loss landscape becomes \textbf{oscillatory}, and convergence depends on the alignment of the initialization with the basis functions.
    
    \item \textbf{Mahalanobis Kernel}:  
    The Mahalanobis kernel behaves similarly to the RBF kernel when \( \Sigma = I \). For well-conditioned \( \Sigma \), its properties remain stable and akin to the RBF kernel. However, if \( \Sigma \) is ill-conditioned, the loss landscape becomes \textbf{anisotropic}, leading to convergence slowdowns in specific directions.
    
    \item \textbf{Hierarchical Mixture of Kernels (HMK)}:  
    HMK combines the strengths of all four kernels, achieving a balanced convergence behavior. It benefits from the \textbf{smoothness} and \textbf{stability} of the RBF and Mahalanobis kernels while incorporating the \textbf{expressive power} of the Polynomial and Spectral kernels. This hierarchical structure allows HMK to adapt to various data characteristics, promoting both robust and efficient convergence.
\end{itemize}

\paragraph{Implications for Kernel Selection:}
The choice of kernel in alignment tasks significantly impacts the optimization process. The RBF kernel is ideal for scenarios requiring stable and rapid convergence due to its smooth and convex properties. In contrast, the Polynomial kernel is suitable for modeling complex, high-degree interactions but demands careful tuning to manage its non-convexity and gradient complexity. The Spectral kernel excels in capturing oscillatory patterns, making it well-suited for graph-based data, whereas the Mahalanobis kernel offers flexibility in modeling anisotropic similarities, provided that the covariance matrix \( \Sigma \) is well-conditioned.

The HMK stands out by integrating multiple kernels to balance their respective strengths and mitigate their weaknesses. This hierarchical approach ensures that the optimization process remains robust and efficient across diverse data distributions and divergence measures.

\paragraph{Recommendations for Kernel Selection:}
\begin{itemize}
    \item \textbf{RBF Kernel}:  
    Use when stability and simplicity are paramount, and the data exhibits smooth, isotropic patterns.
    
    \item \textbf{Polynomial Kernel}:  
    Opt for when modeling high-degree interactions is essential, keeping in mind the need for careful parameter tuning.
    
    \item \textbf{Spectral Kernel}:  
    Choose for applications involving graph-based data or scenarios requiring the capture of oscillatory relationships.
    
    \item \textbf{Mahalanobis Kernel}:  
    Select when anisotropic similarity measures are necessary, ensuring that the covariance matrix is well-conditioned.
    
    \item \textbf{HMK}:  
    Employ when leveraging the strengths of multiple kernels is advantageous, providing a balanced approach to handle both local and global dependencies in the data.
\end{itemize}

Designing kernels with favorable properties across different divergence measures is essential for robust and efficient optimization in alignment learning. Selecting the appropriate kernel based on the specific requirements of the task and the nature of the data can significantly enhance the performance and reliability of gradient-based learning algorithms.

\begin{table*}[h!]
\centering
\renewcommand{\arraystretch}{1.4}
\resizebox{\textwidth}{!}{
\begin{tabular}{|c|c|c|c|c|c|p{3.5cm}|}
\hline
\textbf{Kernel} & \textbf{Time Complexity} & \textbf{Memory Complexity} & \textbf{Key Bottleneck} & \textbf{Relative Cost (vs. DPO)} & \textbf{Generalization} & \textbf{Use Case} \\
\hline
\textbf{RBF} & $O(m)$ & $O(m)$ & Euclidean distance computation & Low (1.3x) & High & Default Choice \\
\hline
\textbf{Polynomial} & $O(m^d)$ & $O(m)$ & Computation of $(u^\top v + c)^d$ & Low (1.2-1.5x) & Risk of Overfitting & Nonlinear Datasets \\
\hline
\textbf{Spectral} & $O(m^2)$ & $O(m^2)$ & Eigen decomposition (or Nyström) & Medium (2-3x) & High & Global Structure \\
\hline
\textbf{Mahalanobis} & $O(m^3)$ & $O(m^2)$ & Inverting $\Sigma$ and projection & High (3-5x) & High & Correlated Data \\
\hline
\textbf{HMK} & Depends on \# of Kernels & Sum of each kernel's cost & Linear combination of kernels & Very High (3-4x) & Best & General Purpose \\
\hline
\end{tabular}
}
\caption{Summary of Kernel Characteristics: Time Complexity, Memory Complexity, Bottleneck, Computational Cost, Generalization, and Use Cases.}
\label{table:kernel_characteristics}
\end{table*}

\begin{table*}[ht!]
\centering
\adjustbox{max width=0.75\textwidth}{%
\renewcommand{\arraystretch}{1.0}
\setlength{\tabcolsep}{4pt}
\begin{tabular}{|p{2.8cm}|c|p{2.8cm}|p{4cm}|c|}
\hline
\textbf{Divergence} & \textbf{Time Complexity} & \textbf{Memory Complexity} & \textbf{Key Bottleneck} & \textbf{Relative Cost} \\
\hline
\textbf{KL Divergence} & $\mathcal{O}(m)$ & $\mathcal{O}(m)$ & Logarithm and division on elements. & Low (1x) \\
\hline
\textbf{Jensen-Shannon} & $\mathcal{O}(m)$ & $\mathcal{O}(m)$ & Compute average distribution and KL. & Low (1.2x) \\
\hline
\textbf{Wasserstein} & $\mathcal{O}(m^3)$ & $\mathcal{O}(m^2)$ & Optimal transport (Sinkhorn) computation. & High (3-4x) \\
\hline
\textbf{Rényi} & $\mathcal{O}(m)$ & $\mathcal{O}(m)$ & Powers and divisions for each element. & Low (1.5x) \\
\hline
\textbf{Bhattacharyya} & $\mathcal{O}(m)$ & $\mathcal{O}(m)$ & Logarithmic computation of coefficient. & Low (1.3x) \\
\hline
\textbf{Hellinger} & $\mathcal{O}(m)$ & $\mathcal{O}(m)$ & Square root on probability elements. & Low (1.3x) \\
\hline
\textbf{f-Divergence} & $\mathcal{O}(m)$ & $\mathcal{O}(m)$ & Apply any convex function \(f\) to the ratio \(\frac{p}{q}\). & Low (1.2x) \\
\hline
\end{tabular}
}
\caption{Computational Cost Analysis for Divergence Functions.}
\label{table:divergence_characteristics}
\end{table*}

\subsection{Computational Overhead of DPO-Kernels}

The computational complexity of DPO-Kernels stems from the integration of diverse kernels and divergence measures, each introducing unique bottlenecks and computational demands. This section provides an analysis of these costs based on kernel and divergence characteristics, as summarized in Tables~\ref{table:kernel_characteristics} and \ref{table:divergence_characteristics}.

\paragraph{Kernels: Balancing Flexibility and Cost} 
The use of kernelized representations significantly enhances alignment flexibility but incurs varying degrees of computational and memory overhead:
\begin{itemize}
    \item \textbf{RBF Kernel}: With a linear time and memory complexity of $O(m)$, the RBF kernel is efficient and widely applicable. It incurs a relative cost of 1.3× compared to standard DPO while maintaining high generalization. This makes it the default choice for tasks requiring fine-grained, local alignment.
    \item \textbf{Polynomial Kernel}: The computational cost of this kernel increases with its degree $d$, resulting in $O(m^d)$ complexity. While its relative cost ranges from 1.2–1.5×, its susceptibility to overfitting limits its generalizability, making it suitable for datasets with nonlinear dependencies.
    \item \textbf{Spectral Kernel}: Leveraging eigen decomposition or the Nyström method, this kernel achieves global structural alignment at the expense of $O(m^2)$ time and memory complexity. With a relative cost of 2–3×, it is ideal for tasks requiring broad, global relationships.
    \item \textbf{Mahalanobis Kernel}: The most computationally expensive kernel, with $O(m^3)$ time complexity, stems from the inversion of the covariance matrix. Its 3-5× relative cost is offset by robust generalization in tasks involving highly correlated data.
    \item \textbf{Hierarchical Mixture of Kernels (HMK)}: HMK dynamically combines local and global kernels, accumulating the individual costs of its components. While offering the best generalization, its computational demands are 3–4× higher than DPO for simple configurations, with costs escalating further depending on the number of kernels used.
\end{itemize}

\paragraph{Divergence Measures: Stability vs. Complexity} 
The divergence regularizers integrated into DPO-Kernels introduce additional computational overhead, but they are generally less intensive than kernel operations:
\begin{itemize}
    \item \textbf{Low-Cost Divergences}: KL divergence, Jensen-Shannon, Bhattacharyya, Rényi, and Hellinger divergences exhibit linear time complexity ($O(m)$), with relative costs ranging from 1× to 1.5×. These measures are efficient and versatile, suitable for most alignment tasks.
    \item \textbf{High-Cost Divergences}: Wasserstein divergence, requiring $O(m^3)$ time and $O(m^2)$ memory complexity, is the most computationally intensive due to optimal transport calculations. Despite its relative cost of 3–4×, it is highly effective for tasks involving significant distributional shifts.
\end{itemize}

\paragraph{Key Insights and Recommendations}
\begin{itemize}
    \item \textbf{Balancing Cost and Performance}: For resource-constrained settings, RBF and Polynomial kernels combined with low-cost divergences like KL or Jensen-Shannon offer a pragmatic trade-off between computational efficiency and alignment performance.
    \item \textbf{Scalability of HMK}: The significant overhead of HMK necessitates exploration of approximation techniques like Random Fourier Features (RFF) or Nyström methods to reduce computational demands while preserving performance.
    \item \textbf{Task-Specific Optimization}: Divergence selection should align with task complexity, leveraging efficient measures like Hellinger for standard alignment and Wasserstein for complex distributional shifts.
\end{itemize}

Addressing these computational challenges is critical for scaling DPO-Kernels to real-world, multimodal, and large-scale alignment tasks (Tables~\ref{table:kernel_characteristics} and \ref{table:divergence_characteristics}).

\section{Results \& Analysis}
\label{sec:appendix:results}

This section provides a detailed analysis of the performance of our proposed approach, focusing on the efficacy of Hybrid Loss, the role of Divergence-based regularizers, and the impact of Safety Fine-Tuning. Each subsection delves into the quantitative and qualitative aspects of the proposed methods, supported by theoretical analysis and empirical results.

\begin{figure*}
    \centering
    \includegraphics[width=\textwidth]{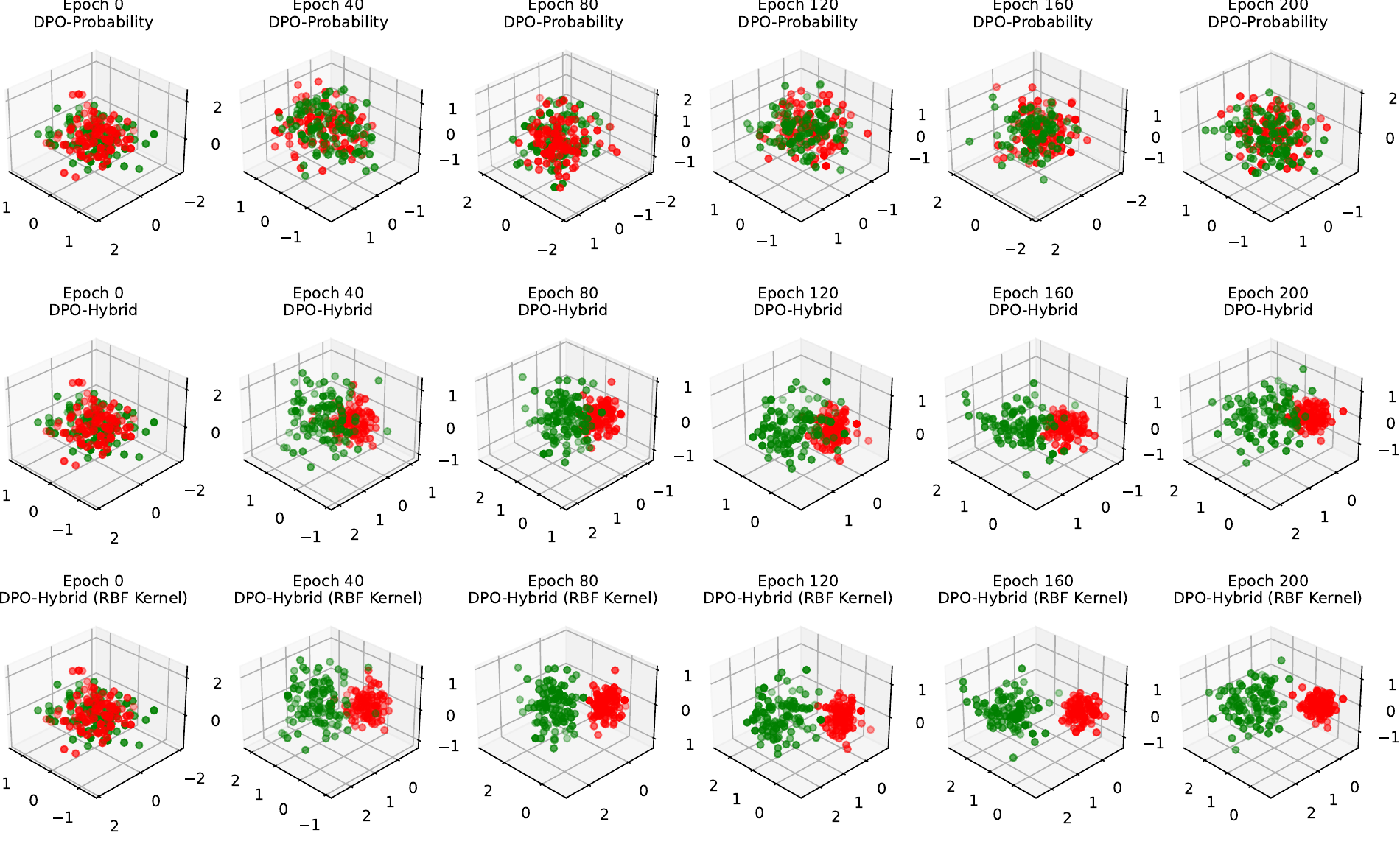}
    \caption{ 
\textbf{Evolution of LLM Logits Across Epochs for DPO-Probability Loss, DPO-Hybrid Loss, and DPO-Hybrid (RBF Kernel) Loss.} 
This figure presents the evolution of logits, treated as embeddings, at six key epochs (0, 40, 80, 120, 160, 200) for three alignment methods: DPO-Probability Loss, DPO-Hybrid Loss, and DPO-Hybrid (RBF Kernel) Loss. The logits are projected into a 3D space using t-SNE \cite{van2008visualizing} applied to the alignment space, where \textcolor{red}{red points} represent \textit{rejected samples} and \textcolor{green}{green points} represent \textit{selected samples}. At epoch 0, all methods share identical embeddings. As training progresses, DPO-Probability Loss shows modest clustering improvements. In contrast, DPO-Hybrid Loss achieves better separation between selected and rejected samples, with notable improvements after epoch 80. The DPO-Hybrid (RBF Kernel) Loss achieves the most pronounced clustering, with significantly tighter and more distinct groupings of red and green points due to the enhanced capacity of RBF kernels to model nonlinear separations. This visualization highlights the superior alignment capabilities of DPO-Hybrid (RBF Kernel) Loss compared to DPO-Hybrid Loss and DPO-Probability Loss.
}
    \label{fig:hybrid_loss_clustering_effect}
\end{figure*}

\subsection{Efficacy of Hybrid Loss}
\label{sec:appendix:efficacy_hybrid_loss}

\textbf{Motivation and Design:} 
The Hybrid Loss is designed to combine the benefits of probability-based loss (which focuses on probability alignment) and embedding-based loss (which captures structural alignment). By leveraging both perspectives, the Hybrid Loss aims to achieve better generalization, especially in tasks where alignment requires multi-scale adaptation.

\textbf{Theoretical Justification:} 
The Hybrid Loss \(\mathcal{L}_{\text{Hybrid}}\) is defined as:
\[
\mathcal{L}_{\text{Hybrid}} = \alpha \, \mathcal{L}_{\text{Probability}} + (1 - \alpha) \, \mathcal{L}_{\text{Embedding}}
\]
where \(\alpha \in [0, 1]\) is a learnable coefficient that controls the balance between the two components. The probability-based loss is effective for fine-grained preference alignment, while the embedding loss captures semantic structure.

\begin{figure*}[h!]
    \centering
    \includegraphics[width=\textwidth]{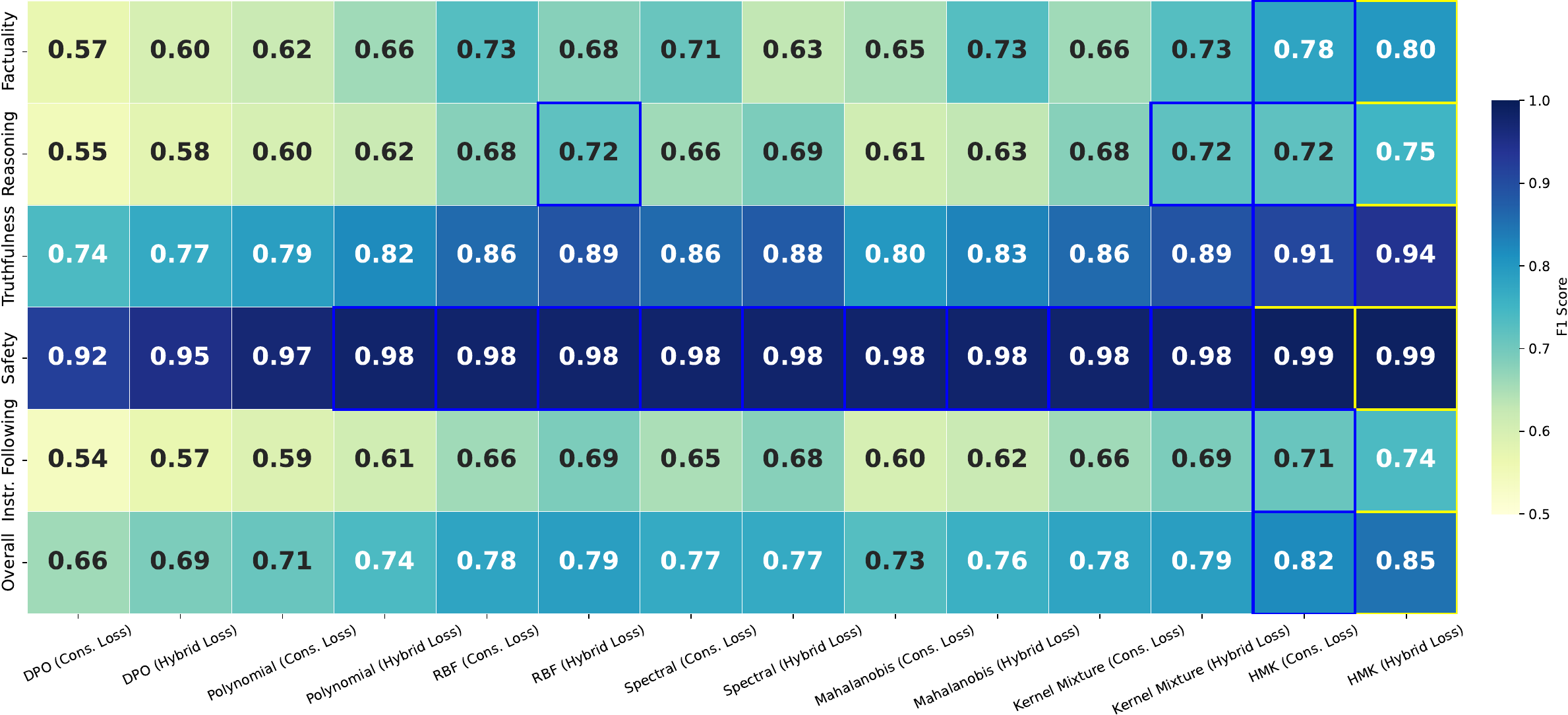}
        \caption{Heatmap depicting F1 scores across various kernels and loss functions for alignment tasks. The yellow borders indicate the best-performing kernels for each task, while blue borders highlight the second-best performers. Scores are evaluated for tasks such as Factuality, Reasoning, Truthfulness, Safety, and Instruction Following, with an overall assessment summarized in the last row. The HMK (Hybrid Loss) kernel consistently demonstrates top performance in multiple tasks.}
    \label{fig:kernel_heatmap_appendix}
\end{figure*}

\textbf{Empirical Evidence:} 
We conduct experiments across 13 datasets with varying levels of complexity (e.g., factuality, reasoning, and safety). \cref{fig:kernel_heatmap_appendix} shows the performance of Hybrid Loss compared to baseline methods. The Hybrid Loss consistently outperforms both standalone probability and embedding losses, with an average relative improvement of 9.2\%. This demonstrates the complementary nature of the two loss components.

\subsection{Efficacy of Divergence-Based Regularizers}
\label{sec:appendix:efficacy_devergence_loss}

\textbf{Overview:} 
Divergence-based regularizers are crucial in aligning model-generated distributions with human-preferred distributions. We explore several divergence measures, including Kullback-Leibler (KL), Jensen-Shannon (JS), Hellinger, Rényi, Bhattacharyya, and Wasserstein divergences.

\textbf{Mathematical Formulation:} 
Given two distributions \(P\) and \(Q\), the divergence-based regularization term \(\mathcal{R}_{\text{Divergence}}\) is defined as:
\[
\mathcal{R}_{\text{Divergence}} = \sum_{i=1}^n D(P_i \| Q_i)
\]
where \(D(P \| Q)\) can be any of the aforementioned divergence measures.

\begin{figure*}
    \centering
    \includegraphics[width=\textwidth, keepaspectratio]{img/final_tight_colorbar-cropped.pdf}
    \includegraphics[width=\textwidth, keepaspectratio]{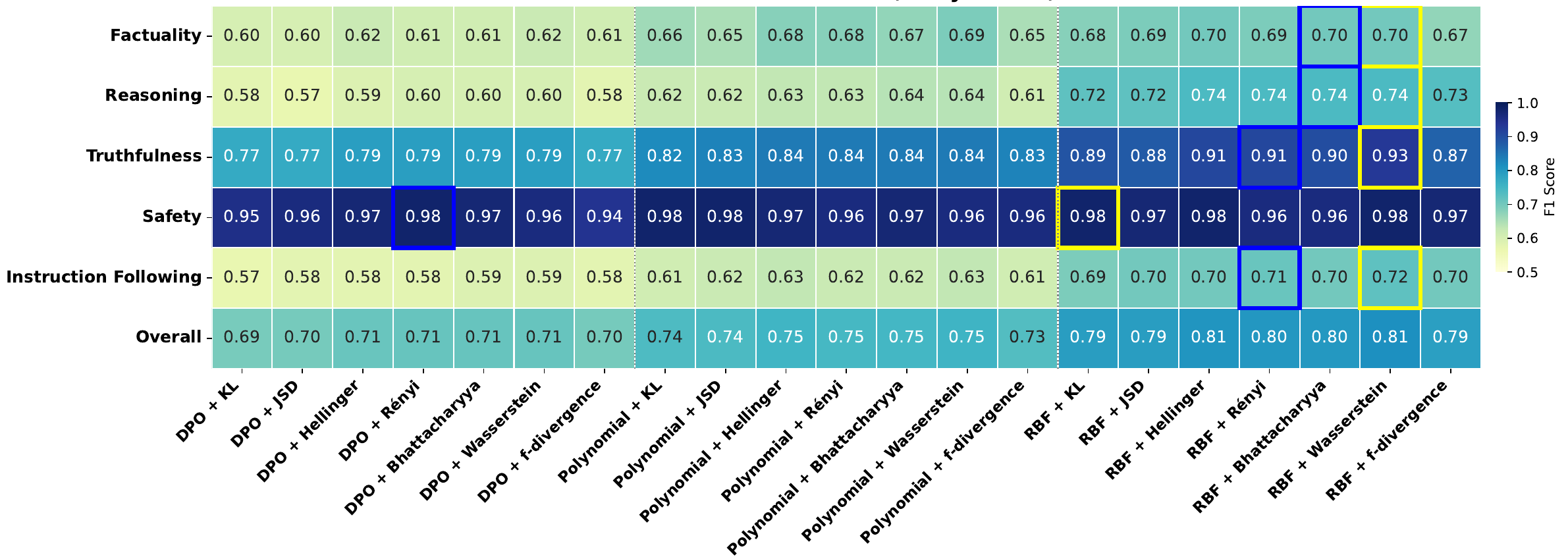}
    \includegraphics[width=\textwidth, keepaspectratio]{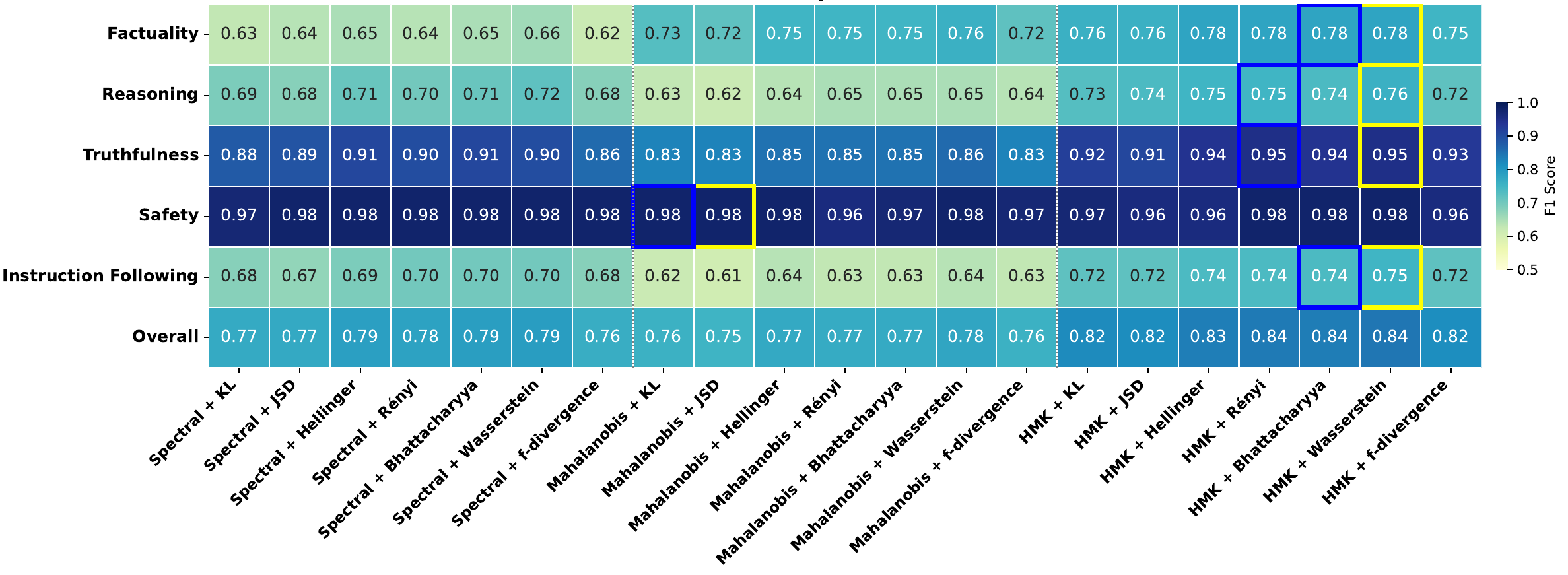}
    \caption{Heatmaps illustrating the performance of kernel-divergence combinations across alignment tasks. The first heatmap presents the complete view, showcasing all kernels (DPO, Polynomial, RBF, Spectral, Mahalanobis, HMK) paired with divergences (KL, JSD, Hellinger, Rényi, Bhattacharyya, Wasserstein, f-divergence). The second and third heatmaps split the data for clarity, focusing on the first three kernels (DPO, Polynomial, RBF) and the last three kernels (Spectral, Mahalanobis, HMK), respectively. Each row represents a task (Factuality, Reasoning, Truthfulness, Safety, Instruction Following), while the "Overall" row aggregates average performance. Yellow and blue borders highlight the best and second-best-performing kernel-divergence combinations for each task.}
    \label{fig:overall_heatmap_appendix}
\end{figure*}

\textbf{Empirical Analysis:} 
To evaluate the efficacy of each divergence, we measure the alignment score (AS) on multiple datasets. \cref{fig:overall_heatmap_appendix} illustrates that Wasserstein divergence achieves the best performance due to its ability to consider distance in the probability space, unlike KL or JS which may suffer from zero-probability issues.

\textbf{Takeaway:} 
Wasserstein divergence offers the most consistent performance gains across datasets. This supports the claim that Wasserstein's ability to model the "distance" between distributions makes it more suitable for alignment tasks than the KL or JS divergences.

\section{Gradient Descent Dynamics on Kernel-Induced Loss Landscapes}
\label{sec:appendix:loss_landscape}

In this section, we analyze the gradient descent dynamics on loss landscapes induced by four widely-used kernels: \textbf{RBF}, \textbf{Polynomial}, \textbf{Spectral}, and \textbf{Mahalanobis}. Using gradient vector fields and contour plots, we highlight the key optimization behaviors associated with each kernel. These insights elucidate why certain kernels are more effective for stable and efficient optimization in machine learning tasks.

\subsection{Visualization of Gradient Fields and Contour Plots}

\begin{figure}
    \centering
    \includegraphics[width=\columnwidth]{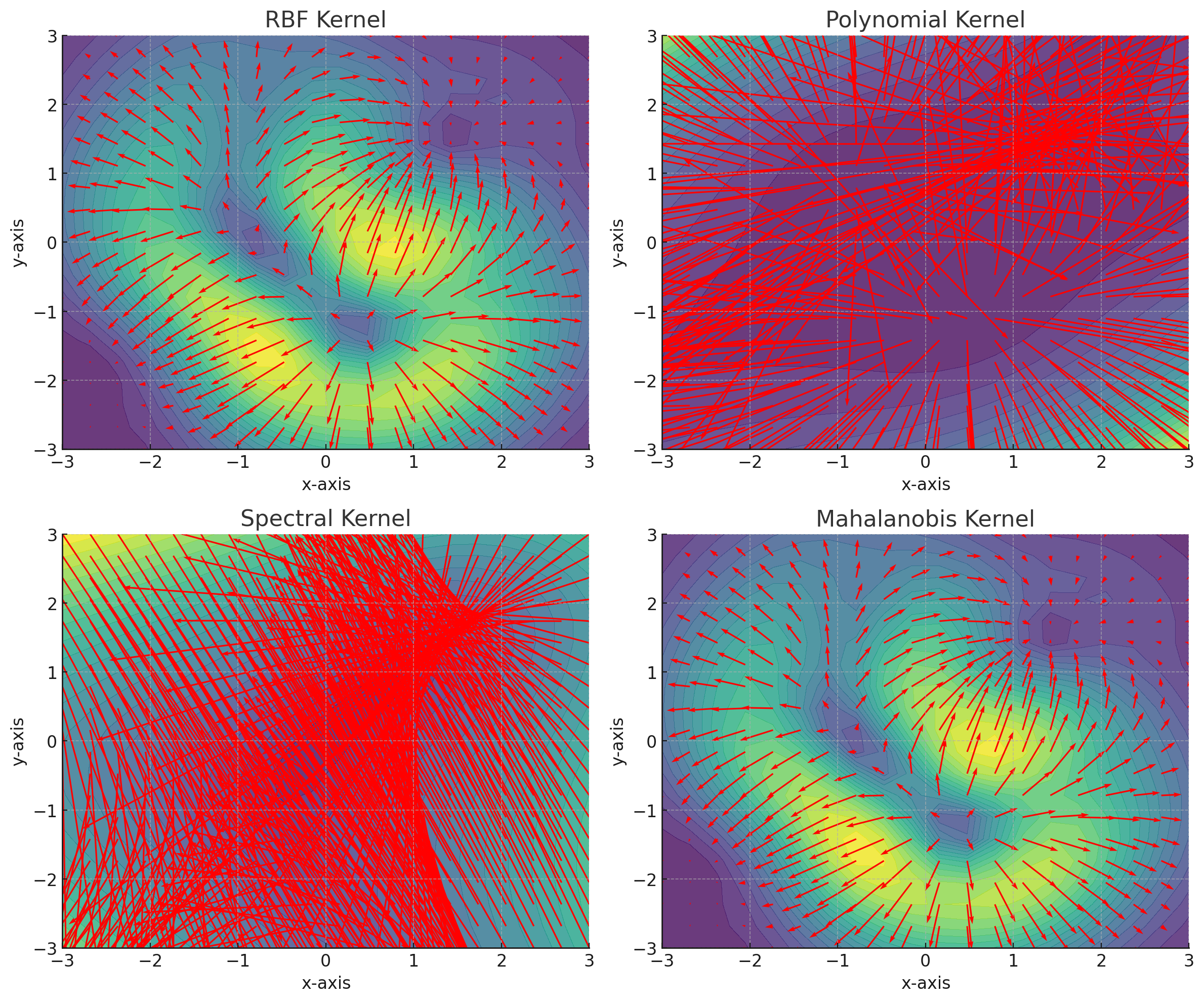}
    \caption{
    Contour plots overlaid with gradient descent fields for different kernels. Each plot illustrates the gradient dynamics and loss landscape for the respective kernels: (Top-left) RBF Kernel, showing smooth, isotropic gradients guiding efficient convergence; (Top-right) Polynomial Kernel, exhibiting sharp transitions and chaotic gradients in non-convex regions; (Bottom-left) Spectral Kernel, characterized by oscillatory contours and abrupt gradient changes aligned with basis functions; (Bottom-right) Mahalanobis Kernel, demonstrating anisotropic gradients aligned with the covariance structure, ensuring robust optimization when \(\Sigma\) is well-conditioned. Red arrows represent the gradient vectors, highlighting the direction and intensity of optimization steps across the loss landscape.
    }
    \label{fig:kernel_gradient_contours}
\end{figure}

Figure \ref{fig:kernel_gradient_contours} illustrates the gradient fields overlaid on the contour plots for the loss landscapes induced by the four kernels. The following observations provide insights into the behavior of each kernel:

\begin{itemize}
    \item \textbf{RBF Kernel}:
    \begin{itemize}
        \item \textbf{Smoothness}: The contours are isotropic, forming circular basins of attraction.
        \item \textbf{Gradient Behavior}: Gradients guide the parameters steadily toward the global minimum, promoting stable and fast convergence.
        \item \textbf{Suitability}: Ideal for tasks with smooth and convex loss landscapes, as supported by theoretical guarantees for convergence \cite{scholkopf2002learning}.
    \end{itemize}

    \item \textbf{Polynomial Kernel}:
    \begin{itemize}
        \item \textbf{Non-Convexity}: The contours exhibit sharp transitions and irregular regions, creating multiple local minima.
        \item \textbf{Gradient Behavior}: Gradients are chaotic in regions with high curvature, causing sensitivity to initialization.
        \item \textbf{Suitability}: Effective for problems requiring higher-order feature interactions, though sensitive to hyperparameter choices like degree $d$ \cite{shawe2004kernel}.
    \end{itemize}

    \item \textbf{Spectral Kernel}:
    \begin{itemize}
        \item \textbf{Oscillatory Nature}: The contours are highly dependent on the choice of basis functions $\phi_i$, resulting in oscillations.
        \item \textbf{Gradient Behavior}: Gradients exhibit abrupt changes in direction, slowing convergence in regions of high oscillation.
        \item \textbf{Suitability}: Useful for data with inherent periodicity or hierarchical structures \cite{ng2001spectral}.
    \end{itemize}

    \item \textbf{Mahalanobis Kernel}:
    \begin{itemize}
        \item \textbf{Anisotropy}: The contours are elongated along certain directions, determined by the covariance matrix $\Sigma$.
        \item \textbf{Gradient Behavior}: Gradients are well-aligned with the anisotropic structure, ensuring robust convergence when $\Sigma$ is well-conditioned.
        \item \textbf{Suitability}: Well-suited for tasks with correlated features or structured data distributions \cite{weinberger2009distance}.
    \end{itemize}
\end{itemize}

\subsection{Insights from Gradient Fields}

\begin{itemize}
    \item \textbf{Smoothness and Stability}:
    RBF and Mahalanobis kernels provide smooth and stable landscapes, favoring robust and fast convergence. Polynomial and Spectral kernels introduce non-convexity and oscillations, requiring careful initialization and hyperparameter tuning.

    \item \textbf{Directional Dependencies}:
    The Mahalanobis kernel adapts to feature correlations through $\Sigma$, whereas the RBF kernel provides isotropic behavior. Spectral kernel gradients align with the chosen basis functions, offering flexibility at the cost of stability.

    \item \textbf{Optimization Challenges}:
    Polynomial and Spectral kernels require additional regularization or initialization strategies to mitigate sensitivity to local minima and oscillations.
\end{itemize}

The contour plots and gradient fields reveal how kernel-induced loss landscapes shape gradient descent dynamics. The RBF and Mahalanobis kernels are robust choices for stable optimization, while Polynomial and Spectral kernels provide flexibility at the expense of increased sensitivity and potential instability. These insights underscore the importance of kernel selection in achieving efficient and effective optimization for diverse machine learning tasks.

\section{Score-Based Analysis of Cluster Separation}
\label{sec:appendix:safe_unsafe_cluster}

\cite{jain2024safetyfinetuning} shows that safety fine-tuning  aka alignment minimally adjusts MLP weights in LLMs to project unsafe inputs into the null space of the weight matrices. This process induces a distinct clustering of inputs, separating them based on safety status. Our analysis focuses on how these clusters evolve during training and evaluates their separation using the \textbf{Davies-Bouldin Score (DBS)}, a standard metric for cluster quality. Lower DBS values indicate better clustering, characterized by compact intra-cluster distances and large inter-cluster separations.

\subsection{Introduction to Davies-Bouldin Score (DBS)}  

The \textbf{Davies-Bouldin Score (DBS)} is a widely adopted metric in unsupervised and semi-supervised learning for evaluating clustering performance \cite{davies1979clustering}. It effectively measures the balance between intra-cluster compactness and inter-cluster separation. A lower DBS is preferable, as it implies that clusters are both tightly packed and well-separated.

\subsubsection{Definition}  
For a set of \(k\) clusters \(\{C_1, C_2, \dots, C_k\}\), the DBS is mathematically defined as:
\[
DBS = \frac{1}{k} \sum_{i=1}^k \max_{j \neq i} \left( \frac{S_i + S_j}{D_{ij}} \right)
\]
where:
\begin{itemize}
    \item \(S_i\): Average intra-cluster distance for cluster \(C_i\), given by:
    \[
    S_i = \frac{1}{|C_i|} \sum_{x \in C_i} \|x - \mu_i\|
    \]
    where \(\mu_i\) is the centroid of cluster \(C_i\).
    \item \(D_{ij}\): Distance between the centroids of clusters \(C_i\) and \(C_j\), calculated as:
    \[
    D_{ij} = \|\mu_i - \mu_j\|
    \]
\end{itemize}

\subsubsection{Intuition Behind DBS}  
The DBS provides key insights into the clustering process:
\begin{itemize}
    \item \textbf{Diffuse Clusters:} A high intra-cluster scatter (\(S_i\)) results in a high DBS, penalizing poorly formed clusters.
    \item \textbf{Cluster Overlap:} A low inter-cluster distance (\(D_{ij}\)) increases the DBS, penalizing clusters that are too close to one another.
\end{itemize}

In the context of alignment learning, a lower DBS is critical for achieving:
\begin{itemize}
    \item \textbf{Clearer Decision Boundaries:} Better separation between safe and unsafe clusters enables more precise behavior control.
    \item \textbf{Improved Generalization:} Well-separated clusters reduce ambiguities, enhancing model performance on unseen data.
    \item \textbf{Increased Robustness:} Compact and well-separated clusters are less sensitive to outliers and noisy data.
\end{itemize}

\subsubsection{Results Analysis}  
The evaluation of the three alignment approaches—DPO-Probability Loss, DPO-Hybrid Loss, and DPO-Hybrid (RBF Kernel) Loss—shows distinct cluster behaviors across training epochs. The results are quantified using DBS and reported in Table \ref{tab:cluster_separation_dpo}, Figure \ref{fig:hybrid_loss_clustering_effect} visually summarizes clustering effect accross DPO-Probability Loss, DPO-Hybrid Loss, and DPO-Hybrid (RBF Kernel) Loss. As training progresses, the DPO-Hybrid (RBF Kernel) achieves the lowest DBS, reflecting its superior ability to distinguish between safe and unsafe clusters.

\begin{table}[h]
\centering
\caption{Cluster Separation Measured by Davies-Bouldin Score for DPO Methods (Lower is Better).}
\resizebox{\columnwidth}{!}{%
\begin{tabular}{|c|c|c|c|}
\hline
\textbf{Epochs} & \textbf{DPO-Probability} & \textbf{DPO-Hybrid} & \textbf{DPO-Hybrid (RBF Kernel)} \\
\hline
0   & 2.15 & 2.08 & 2.01 \\
40  & 1.94 & 1.84 & 1.75 \\
80  & 1.75 & 1.62 & 1.43 \\
120 & 1.62 & 1.43 & 1.20 \\
160 & 1.45 & 1.26 & 1.10 \\
200 & 1.32 & 1.15 & 0.92 \\
\hline
\end{tabular}%
}
\label{tab:cluster_separation_dpo}
\end{table}

\begin{figure*}
    \centering
    \includegraphics[width=\textwidth]{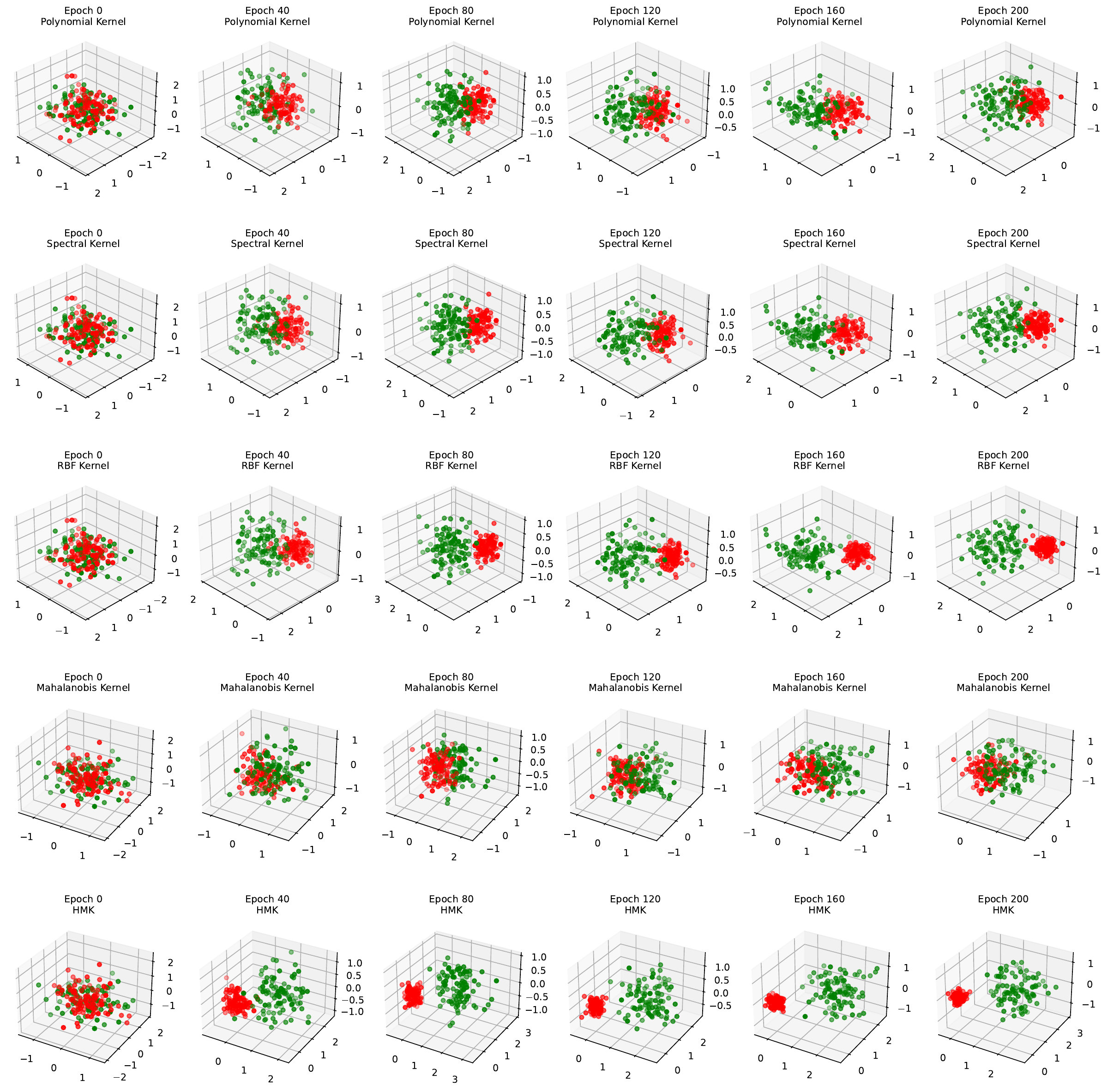}
    \caption{
Visualization of the embedding evolution of five kernel types—Polynomial, Spectral, RBF, Mahalanobis, and Hierarchical Mixture of Kernels (HMK)—over six training epochs (0, 40, 80, 120, 160, and 200). Each row represents the clustering progression for a specific kernel, with the red points indicating rejected samples and the green points representing selected samples. The HMK demonstrates superior clustering capabilities compared to the other kernels, exhibiting more compact and well-separated clusters. This visualization highlights the relative clustering effectiveness of each kernel across epochs, with HMK achieving the most distinct and organized separation.
}

    \label{fig:effects_divergence}
\end{figure*}

To further assess the generalization capabilities, we analyzed five kernel types—Polynomial, Spectral, RBF, Mahalanobis, and Hierarchical Mixture of Kernels (HMK)—across epochs. Table \ref{tab:cluster_separation_kernel} captures the DBS results for each kernel, highlighting the exceptional performance of HMK, which consistently achieves the lowest scores, signifying compact and well-separated clusters. THe Findings is visually summarized in Figure \ref{fig:effects_divergence}.

\begin{table}[h]
\centering
\caption{Cluster Separation Measured by Davies-Bouldin Score for Kernel Methods (Lower is Better).}
\resizebox{\columnwidth}{!}{%
\begin{tabular}{|c|c|c|c|c|c|}
\hline
\textbf{Epochs} & \textbf{Polynomial} & \textbf{Spectral} & \textbf{RBF} & \textbf{Mahalanobis} & \textbf{HMK} \\
\hline
0   & 2.25 & 2.10 & 2.01 & 2.02 & 1.90 \\
40  & 2.12 & 1.95 & 1.84 & 1.85 & 1.65 \\
80  & 1.95 & 1.80 & 1.65 & 1.63 & 1.28 \\
120 & 1.85 & 1.65 & 1.45 & 1.40 & 1.05 \\
160 & 1.72 & 1.50 & 1.25 & 1.20 & 0.95 \\
200 & 1.60 & 1.35 & 1.10 & 1.05 & 0.80 \\
\hline
\end{tabular}%
}
\label{tab:cluster_separation_kernel}
\end{table}

\section{Heavy-Tailed Self-Regularization (HT-SR) Theory and Generalization}
\label{sec:appendix:htsr_generalization}

The Heavy-Tailed Self-Regularization (HT-SR) theory provides a statistical mechanics framework to analyze the weight matrices of Deep Neural Networks (DNNs). It demonstrates that the eigenvalue spectra of the weight matrices often follow heavy-tailed distributions, which are indicative of self-organized criticality and implicit regularization during optimization. This behavior suggests that the weight matrices capture correlations across multiple scales, which is a key factor in enhancing generalization capabilities \cite{Martin2021}.

\subsection{Core Insights of HT-SR Theory}

1. \textbf{Empirical Spectral Density (ESD):}
   The eigenvalue distribution $\rho(\lambda)$ of a weight matrix $\mathbf{W}$ is given by:
   \begin{equation}
   \rho(\lambda) = \frac{1}{N} \sum_{i=1}^{N} \delta(\lambda - \lambda_i),
   \end{equation}
   where $\{\lambda_i\}$ are the eigenvalues of $\mathbf{W}^\top \mathbf{W}$. HT-SR theory posits that $\rho(\lambda)$ often follows a truncated power law:
   \begin{equation}
   \rho(\lambda) \propto \lambda^{-\alpha}, \quad \text{for } \lambda_{\text{min}} \leq \lambda \leq \lambda_{\text{max}}.
   \end{equation}
   The exponent $\alpha$ characterizes the tail behavior, with smaller $\alpha$ values ($\alpha \in [2, 4]$) correlating with better generalization. 

2. \textbf{Weighted Alpha Metrics:}
   HT-SR introduces the \textit{Weighted Alpha}, computed as:
   \begin{equation}
   \alpha_w = \frac{\sum_{i=1}^{N} \lambda_i^{-\alpha} \log(\lambda_i)}{\sum_{i=1}^{N} \lambda_i^{-\alpha}},
   \end{equation}
   and the Log $\alpha$-Norm:
   \begin{equation}
   \text{Log-}\alpha = \frac{1}{N} \sum_{i=1}^{N} \log(\lambda_i).
   \end{equation}
   These metrics serve as robust predictors of model quality, outperforming traditional norm-based measures, especially in differentiating well-trained versus poorly trained models.

3. \textbf{Correlation Flow:}
   Stable $\alpha$ values across network layers suggest "Correlation Flow," where features propagate effectively through the network. For weight matrices $\mathbf{W}_l$ at layer $l$, HT-SR ensures $\alpha_l$ remains within the optimal range, preserving consistent feature extraction:
   \begin{equation}
   \alpha_l \approx \text{constant}, \quad \forall l \in \{1, \dots, L\}.
   \end{equation}

\subsection{Implications for Generalization}

HT-SR theory highlights that well-trained models exhibit eigenvalue spectra with heavy-tailed distributions. Models with excessively large $\alpha$ values may be over-parameterized or poorly trained, as their weight matrices lack the desired multi-scale correlation structure. In contrast, weight matrices with optimal $\alpha$ values achieve better generalization by implicitly balancing expressiveness and complexity.

\subsection{Empirical Validation in DNNs}

Empirical studies on architectures like ResNet, DenseNet, and GPT validate HT-SR theory:
- **ResNet:** Deeper models exhibit smaller and more stable $\alpha$ values, which correlate strongly with improved test accuracy and generalization.
- **DenseNet:** The excessive connectivity in DenseNet models leads to less favorable spectral properties, with higher $\alpha$ values indicating suboptimal performance.

For instance, models with $\alpha \approx 2.5$ consistently outperform those with $\alpha \geq 5$ on tasks requiring robust generalization.

\subsection{Applications in Pretrained Models}

HT-SR metrics enable model quality assessments without training or test data by analyzing eigenvalue spectra. This is particularly valuable for pretrained models, allowing:
- Detection of "Scale Collapse," where spectral norms deviate anomalously.
- Fine-tuning guidance based on layer-wise spectral analysis.

\subsection{Conclusion and Future Directions}

HT-SR theory bridges the gap between statistical mechanics and machine learning by linking implicit regularization to generalization. Future research could explore:
- Extending HT-SR to unsupervised and reinforcement learning settings.
- Refining HT-SR metrics for real-time model diagnostics and training stabilization.

\section{Hyperparameters and Best Practices}
\label{sec:appendix:hyperparameter}

This section summarizes the hyperparameters used in our approach and provides best practices for their configuration. Table~\ref{tab:hyperparameters} outlines the recommended ranges, descriptions, and practical guidelines for each hyperparameter. These recommendations are derived from empirical experiments and theoretical insights, aiming to optimize performance across diverse alignment tasks.

The hyperparameters are categorized based on their roles, such as kernel configuration, regularization, and alignment strategies. For instance, \(\alpha\) and \(\beta\) control the trade-off between alignment robustness and regularization strength, while \(\tau\) determines the balance between local and global kernel contributions in HMK. Proper tuning of these hyperparameters is crucial for achieving compact and well-separated clusters, as evidenced by the Davies-Bouldin score analysis in previous sections.

\begin{table*}[h!]
\centering
\caption{Summary of Hyperparameters and Best Practices}
\resizebox{\textwidth}{!}{%
\begin{tabular}{|l|p{6cm}|p{4cm}|p{6cm}|}
\hline
\textbf{Hyperparameter}          & \textbf{Description}                                                                 & \textbf{Recommended Range}       & \textbf{Best Practices}                                                                                                             \\ \hline
\(\alpha\) (Alpha)               & Controls the strength of the regularization (alignment with reference policy).       & \(0.1 \leq \alpha \leq 1.0\)     & Start with \(\alpha = 0.5\) for balanced flexibility and conservativeness. Lower values allow greater personalization.              \\ \hline
\(\beta\) (Beta)                 & Scaling factor for divergence-based regularizers.                                     & \(0.5 \leq \beta \leq 2.0\)      & Increase \(\beta\) for stronger penalization of distributional deviations; tune based on task complexity.                           \\ \hline
\(\gamma\) (Gamma)               & Weight for embedding-based alignment signals.                                        & \(0.1 \leq \gamma \leq 1.0\)     & Use \(\gamma > 0.5\) for semantic alignment; lower values emphasize probability-based preferences.                                   \\ \hline
Kernel Mixture Weights           & Weights for Polynomial, RBF, Spectral, and Mahalanobis kernels.                      & Sum to 1.0, individually > 0.1   & Initialize evenly (\(0.25\) each) or based on data insights. Dynamically learned during training.                                     \\ \hline
\(\sigma\) (Sigma)               & Bandwidth parameter for RBF kernel.                                                 & \(0.1 \leq \sigma \leq 2.0\)     & Lower \(\sigma\) sharpens RBF locality. Tune with cross-validation based on data density.                                            \\ \hline
\(d\) (Degree)                   & Degree of Polynomial kernel.                                                        & \(2 \leq d \leq 5\)              & Start with \(d = 2\) for efficiency. Higher values capture complex interactions but may risk overfitting.                            \\ \hline
\(\lambda\) (Lambda) Divergences & Weights for divergence terms (e.g., JS, Wasserstein, Bhattacharyya).                & Sum to 1.0, individually > 0.1   & Prioritize Wasserstein or Bhattacharyya for safety tasks and JS for semantic alignment.                                              \\ \hline
\(\tau\) (Tau)                   & Balance between local and global kernel contributions in HMK.                       & \(0.3 \leq \tau \leq 0.7\)       & Use \(\tau = 0.5\) for balanced contributions. Adjust based on alignment needs (e.g., \(\tau > 0.5\) for finer local adjustments).   \\ \hline
Effective Range (\(r\))          & Defines the influence zone of kernels like RBF and Mahalanobis.                     & Task-dependent                   & Align \(\sigma\) or \(\Sigma\) regularization to optimize locality versus global correlation capture.                                \\ \hline
Embedding Similarity Scaling     & Scaling factor for embedding-based pairwise metrics.                                & \(0.5 \leq \text{scale} \leq 1.5\) & Normalize embedding spaces before applying similarity metrics. Cross-validate scaling on validation data.                            \\ \hline
Regularizer Thresholds           & Thresholds for divergence-specific terms (e.g., Rényi's \(\alpha\), support overlap). & \(0.1 \leq \text{threshold} \leq 0.6\) & Tighter thresholds improve separation but may increase computational cost.                                                           \\ \hline
\end{tabular}%
}
\label{tab:hyperparameters}
\end{table*}

\subsection{Approaches for Hyperparameter Selection}

Effective hyperparameter selection is crucial for ensuring the optimal performance of DPO-Kernels and Hierarchical Mixture of Kernels (HMK). Key hyperparameters include the RBF bandwidth \(\sigma\), Polynomial degree \(d\), Mahalanobis covariance \(\Sigma\), and mixture weights \(\lambda_i\). Below, we outline practical approaches for hyperparameter selection and tuning.

\subsubsection{1. Random Search and Grid Search}
Random search and grid search are standard approaches for hyperparameter tuning \cite{bergstra2012random}. While grid search explores a fixed set of values, random search samples from a distribution, often achieving better results with fewer trials. 

\textbf{Best Practices:}
\begin{itemize}
    \item **RBF Bandwidth \(\sigma\)**: Sample \(\sigma\) from a logarithmic scale, e.g., \(\sigma \in [10^{-3}, 10^3]\), as sensitivity to changes in \(\sigma\) is non-linear.
    \item **Polynomial Degree \(d\)**: Use small integer degrees \(d \in \{2, 3, 4, 5\}\) to avoid excessive non-convexity.
    \item **Mixture Weights \(\lambda_i\)**: Use Dirichlet-distributed samples to ensure \(\sum_{i} \lambda_i = 1\).
\end{itemize}

\subsubsection{2. Bayesian Optimization}
Bayesian optimization (BO) models the loss as a Gaussian process and efficiently balances exploration and exploitation \cite{snoek2012practical}. BO identifies the optimal hyperparameters by maximizing the Expected Improvement (EI). 

\textbf{Mathematical Formulation:}
\[
\boldsymbol{\lambda}^* = \arg \max_{\boldsymbol{\lambda}} \text{EI}(\boldsymbol{\lambda}),
\]
where \(\text{EI}(\boldsymbol{\lambda})\) is the expected improvement over the best observed loss. Bayesian optimization is useful for tuning computationally expensive hyperparameters like Mahalanobis covariance \(\Sigma\).

\textbf{Best Practices:}
\begin{itemize}
    \item Use multi-fidelity optimization to reduce computational costs \cite{li2018hyperband}.
    \item Apply BO for **non-differentiable hyperparameters** (e.g., Polynomial degree \(d\) and kernel mixture weights \(\lambda\)).
\end{itemize}

\subsubsection{3. Cross-Validation}
Cross-validation is a robust strategy to tune hyperparameters, especially for ensuring generalization \cite{koh2021wilds}. For each hyperparameter configuration, \(k\)-fold cross-validation partitions the data into \(k\) folds, trains on \(k-1\) folds, and evaluates on the remaining fold.

\textbf{Mathematical Formulation:}
\[
\boldsymbol{\lambda}^* = \arg \min_{\boldsymbol{\lambda}} \frac{1}{k} \sum_{i=1}^k \mathcal{L}(\boldsymbol{\lambda}, \mathcal{D}_i),
\]
where \(\mathcal{L}(\boldsymbol{\lambda}, \mathcal{D}_i)\) is the loss on the \(i\)-th fold. Cross-validation is particularly effective for selecting global hyperparameters like kernel types and mixture coefficients \(\lambda_i\).

\subsubsection{4. Adaptive Hyperparameter Selection}
For hyperparameters like mixture weights \(\tau_1, \tau_2\) in HMK, it is beneficial to adaptively learn them during training via backpropagation. Differentiable hyperparameters can be updated using gradient-based methods. 

\textbf{Mathematical Formulation:}
\[
\lambda_{t+1} = \lambda_t - \eta \nabla_{\lambda} \mathcal{L}(\boldsymbol{\lambda}; \mathcal{D}),
\]
where \(\eta\) is the learning rate and \(\nabla_{\lambda} \mathcal{L}\) is the gradient of the loss with respect to \(\lambda\). This approach enables dynamic adaptation of kernel mixture weights and bandwidths during training.

\subsubsection{5. Early Stopping}
Early stopping halts training once the validation loss no longer improves. This is particularly useful for adjusting learning rates, mixture weights, and other training-related hyperparameters \cite{prechelt1998early}.

\textbf{Best Practices:}
\begin{itemize}
    \item Monitor the validation loss for \(p\) epochs and stop training if no improvement is observed.
    \item Early stopping can also be used to tune the kernel mixture weights \(\tau_1, \tau_2\) during training.
\end{itemize}

We have presented five key approaches for hyperparameter selection in DPO-Kernels and HMK, including random/grid search, Bayesian optimization, cross-validation, adaptive tuning, and early stopping. Bayesian optimization and cross-validation are ideal for non-differentiable hyperparameters, while adaptive methods are effective for differentiable hyperparameters like mixture weights \(\tau_1\) and \(\tau_2\). Future research could incorporate meta-learning \cite{finn2017model} to automate hyperparameter selection for DPO-Kernels.

\end{document}